\documentclass[10pt,journal,compsoc,x11names]{IEEEtran}
\usepackage[switch]{lineno}
\usepackage{amsfonts}
\usepackage{amsmath, bm}
\usepackage{xspace}
\usepackage{enumitem}
\usepackage{amssymb}
\usepackage{booktabs}
\usepackage[hyphens]{url}

\usepackage{ragged2e}
\usepackage{hyperref}
\usepackage{multirow}
\usepackage{graphicx}
\usepackage{xcolor}
\usepackage{color}
\usepackage{colortbl}
\usepackage{tablefootnote}
\usepackage{pifont}
\usepackage{makecell}
\usepackage[most]{tcolorbox}
\usepackage{framed}
\usepackage{mdframed}
\usepackage{subfigure}
\usepackage{caption}
\usepackage{longtable}
\usepackage{float}
\usepackage{booktabs}

\newcommand{\paratitle}[1]{\vspace{1.5ex}\noindent\textbf{#1}}
\newcolumntype{H}{>{\setbox0=\hbox\bgroup}c<{\egroup}@{}}

\newcommand{\ie}{\emph{i.e.,}\xspace}
\newcommand{\aka}{\emph{a.k.a.,}\xspace}
\newcommand{\eg}{\emph{e.g.,}\xspace}

\newcommand{\etc}{\emph{etc}}
\newcommand{\ignore}[1]{}

\definecolor{gold}{RGB}{205,133,63}
\definecolor{fGreen}{RGB}{34,139,34}
\definecolor{tOrange}{RGB}{255,215,0}
\definecolor{tBlue}{RGB}{135,206,250}
\definecolor{tPink}{RGB}{255,204,204}
\definecolor{tGreen}{RGB}{205,230,199}
\definecolor{tGold}{RGB}{255,215,0}

\usepackage{cite}
\usepackage[numbers,sort&compress]{natbib}

\ifCLASSINFOpdf
\else
\fi

\begin{document}
\title{A Survey of Large Language Models}

\author{Wayne Xin Zhao, Kun Zhou*, Junyi Li*, Tianyi Tang, Xiaolei Wang, Yupeng Hou, Yingqian Min, Beichen Zhang, Junjie Zhang, Zican Dong, Yifan Du, Chen Yang, Yushuo Chen, Zhipeng Chen, Jinhao Jiang, Ruiyang Ren, Yifan Li, Xinyu Tang, Zikang Liu, Peiyu Liu, Jian-Yun Nie and Ji-Rong Wen$^\dagger$
\IEEEcompsocitemizethanks{
\IEEEcompsocthanksitem Version: v18 (major update on March 7, 2026). 
\IEEEcompsocthanksitem GitHub link: \url{https://github.com/RUCAIBox/LLMSurvey} 
\IEEEcompsocthanksitem Chinese book link: \url{lmbook-zh.github.io} 
\IEEEcompsocthanksitem $\dagger$ Corresponding Author. 
\IEEEcompsocthanksitem * K. Zhou and J. Li contribute equally to this work. 
\IEEEcompsocthanksitem 
The authors are mainly with Gaoling School of Artificial Intelligence and School of Information, Renmin University of China, Beijing, China;  Jian-Yun Nie is with DIRO, Universit\'{e} de Montr\'{e}al, Canada. \\
Contact e-mail: batmanfly@gmail.com
\IEEEcompsocthanksitem \textcolor{red}{The authors of this survey paper reserve all the copyrights of the figures/tables, and any use of these materials for publication purpose must be officially granted by the survey authors.} 
}%
}

\markboth{}%
{Shell \MakeLowercase{\textit{et al.}}: Bare Advanced Demo of IEEEtran.cls for IEEE Computer Society Journals}

\IEEEtitleabstractindextext{%
\begin{abstract}
\justifying
Ever since the Turing Test was proposed in the 1950s,  humans have explored the mastering of  language intelligence by machine.  
Language is essentially a complex, intricate system of human expressions governed by grammatical rules.
 It poses  a significant challenge to develop capable artificial intelligence~(AI) algorithms  for comprehending and grasping a language. 
 As a major approach, 
 \emph{language modeling}
 has been widely studied  for language understanding and generation in the past two decades, evolving from statistical language models to neural language models. 
Recently, pre-trained language models~(PLMs) have been proposed by pre-training Transformer models over large-scale corpora, showing strong capabilities in solving various natural language processing~(NLP) tasks. 
Since the researchers have found that model scaling can lead to an improved model capacity, they further investigate the scaling effect by increasing the parameter scale to an even larger size.  
Interestingly, when the parameter scale exceeds a certain level, these enlarged language models not only achieve a significant performance improvement, but also exhibit some special abilities (\eg in-context learning) that are not present in small-scale  language models (\eg BERT). 
To discriminate the language models in different parameter scales,   
the research community has coined the term \emph{large language models~(LLM)} for the PLMs of significant  size  {(\eg containing tens or hundreds of billions of parameters)}. 
Recently, the research on LLMs has been largely advanced by both academia and industry, and a remarkable progress 
is the launch of ChatGPT (a powerful AI chatbot developed based on LLMs), which has attracted   widespread attention from society. 
The technical evolution of LLMs has been making an important impact on the entire AI community, which would  revolutionize the way how we develop and use AI algorithms. Considering this rapid technical progress, in this survey, we review the recent advances of LLMs by introducing the background, key findings, and mainstream techniques. In particular, we focus on four major aspects of LLMs, namely pre-training, adaptation tuning, utilization, and capacity evaluation. Furthermore, we also summarize the available resources for developing LLMs and discuss the remaining issues for future directions. This survey provides an up-to-date review of the literature on LLMs, which can be a useful resource for both researchers and engineers. 
\end{abstract}

\begin{IEEEkeywords}
Large Language Models; Emergent Abilities; Adaptation Tuning;  
{Utilization; Alignment; Capacity Evaluation}
\end{IEEEkeywords}}

\maketitle

\IEEEdisplaynontitleabstractindextext

\IEEEpeerreviewmaketitle

\IEEEraisesectionheading{\section{Introduction}\label{sec:introduction}}

\begin{flushright}
\rightskip=0.8cm\textit{``The limits of my language mean the limits of my world.''} \\
\vspace{.2em}
\rightskip=.8cm---\emph{Ludwig Wittgenstein}
\end{flushright}
\vspace{1em}

\ignore{
\begin{figure*}[h]
    \centering
    \begin{minipage}{0.45\textwidth}
           \includegraphics[width=\textwidth]{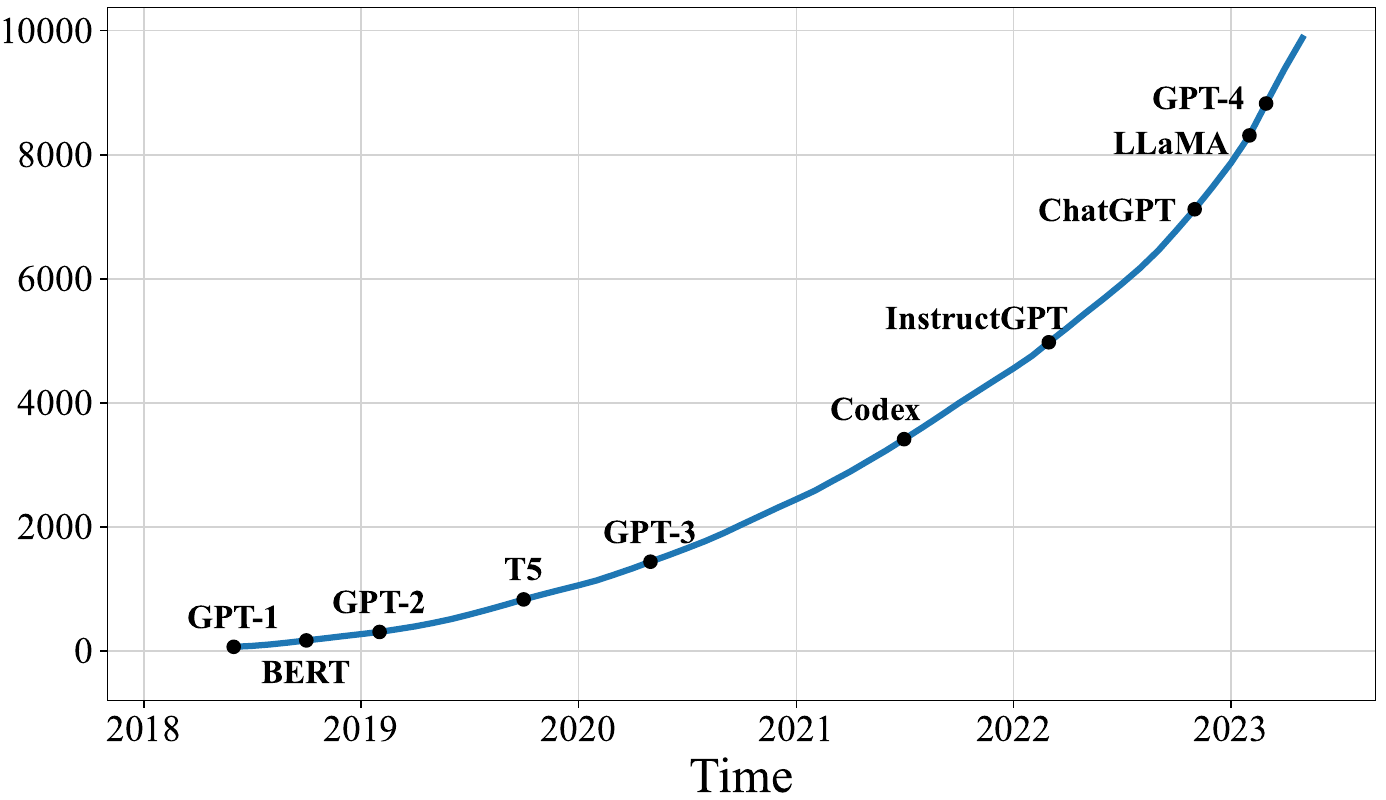}
           \captionof*{figure}{(a) Query="Language Model"}
        \end{minipage}
        \qquad
        \begin{minipage}{0.45\textwidth}
           \includegraphics[width=\textwidth]{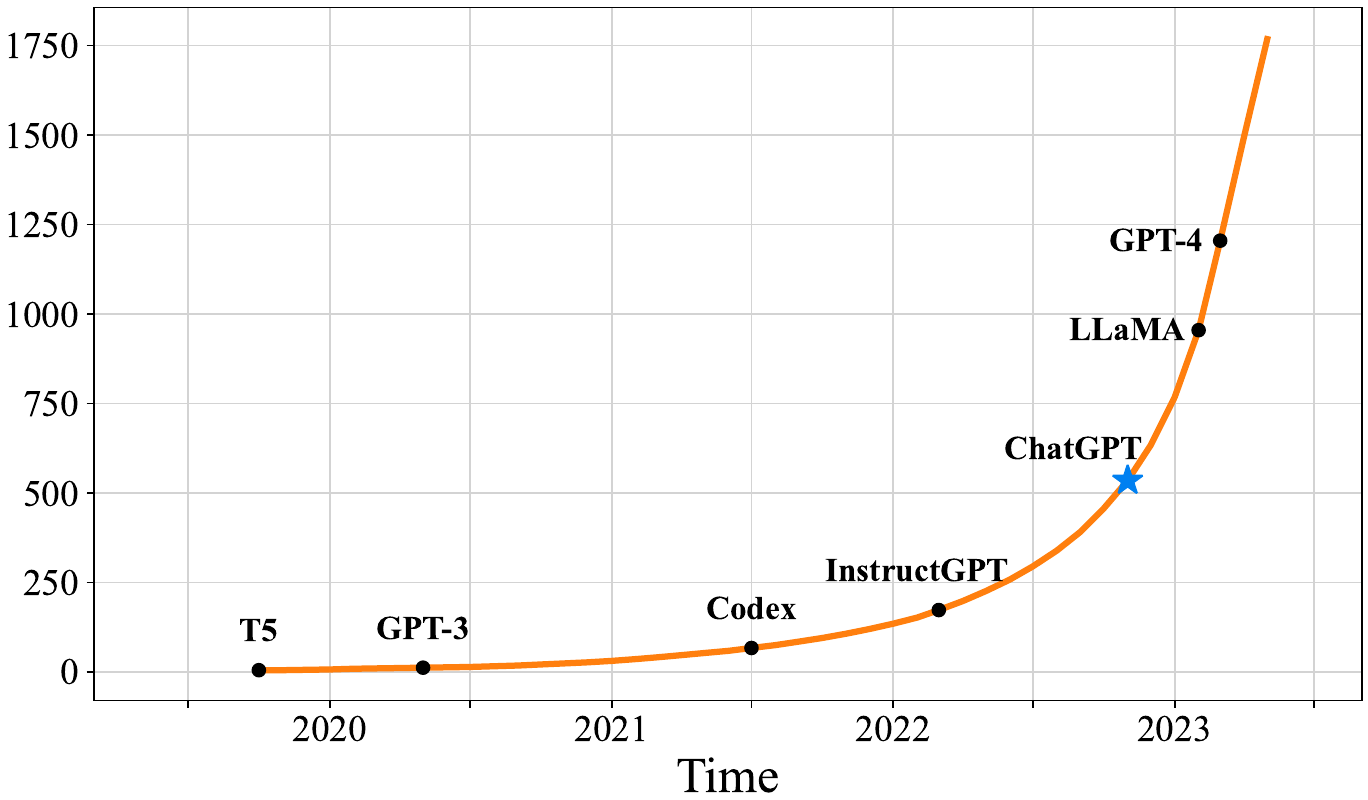}
           \captionof*{figure}{(b) Query="Large Language Model"}
    \end{minipage}
    \caption{{The trends of the  cumulative numbers  of arXiv papers that contain  the keyphrases  ``\emph{language model}" (since June 2018)  and ``\emph{large language model}'' (since October 2019), respectively. The statistics are calculated using   exact match by querying the keyphrases in title or abstract by months. We set different x-axis ranges for the two keyphrases, because ``language models'' have been explored at an earlier time.  We label the points corresponding to  important landmarks in the research progress of LLMs. A sharp increase occurs after the release of ChatGPT: the average number of published  arXiv papers that contain ``\emph{large language model}'' in title or abstract goes from 0.40 per day to 8.58 per day (Figure~\ref{fig:paper_number}(b)). }}
    \label{fig:paper_number}
\end{figure*}}

\begin{figure*}[h]
    \centering
    \begin{minipage}{0.45\textwidth}
           \includegraphics[width=\textwidth]{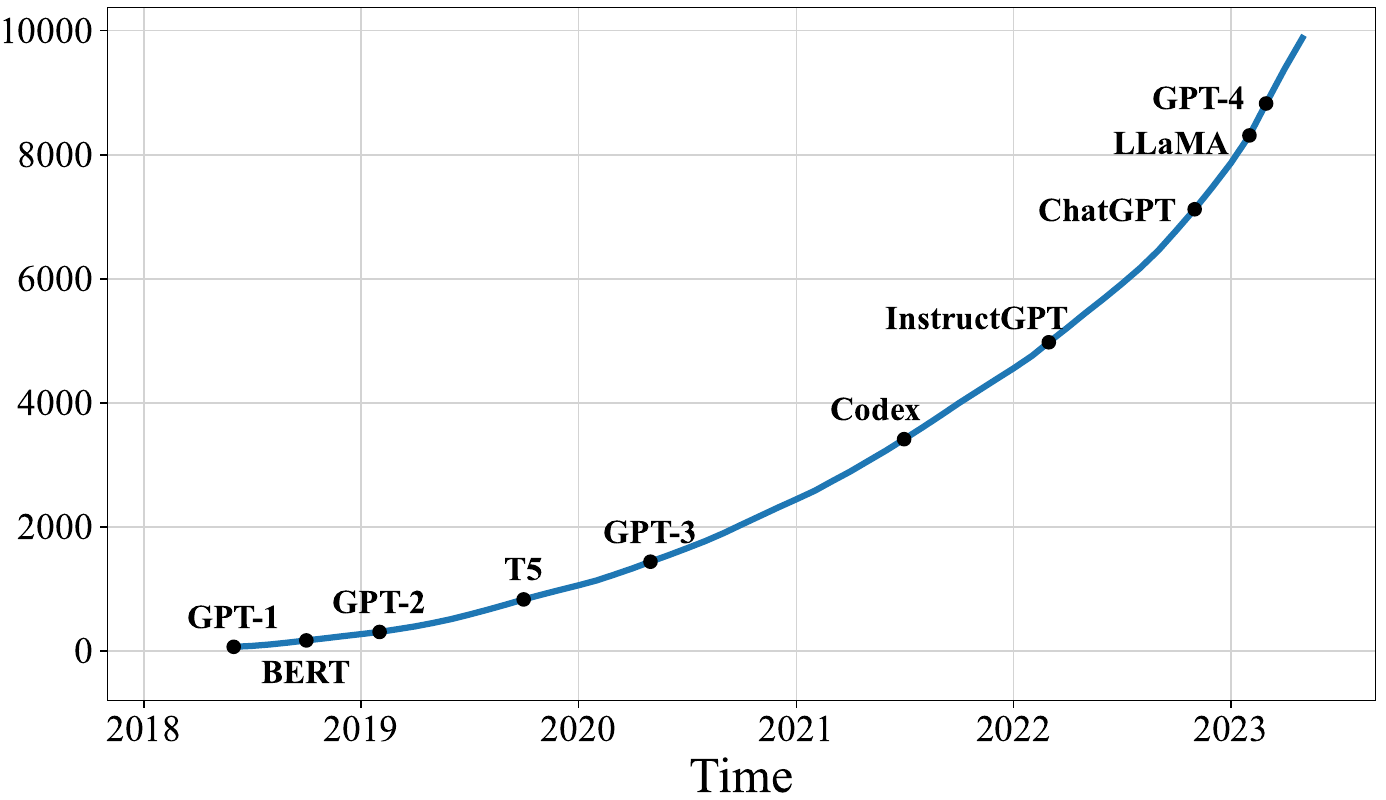}
           \captionof*{figure}{(a) Query="Language Model"}
        \end{minipage}
        \qquad
        \begin{minipage}{0.45\textwidth}
           \includegraphics[width=\textwidth]{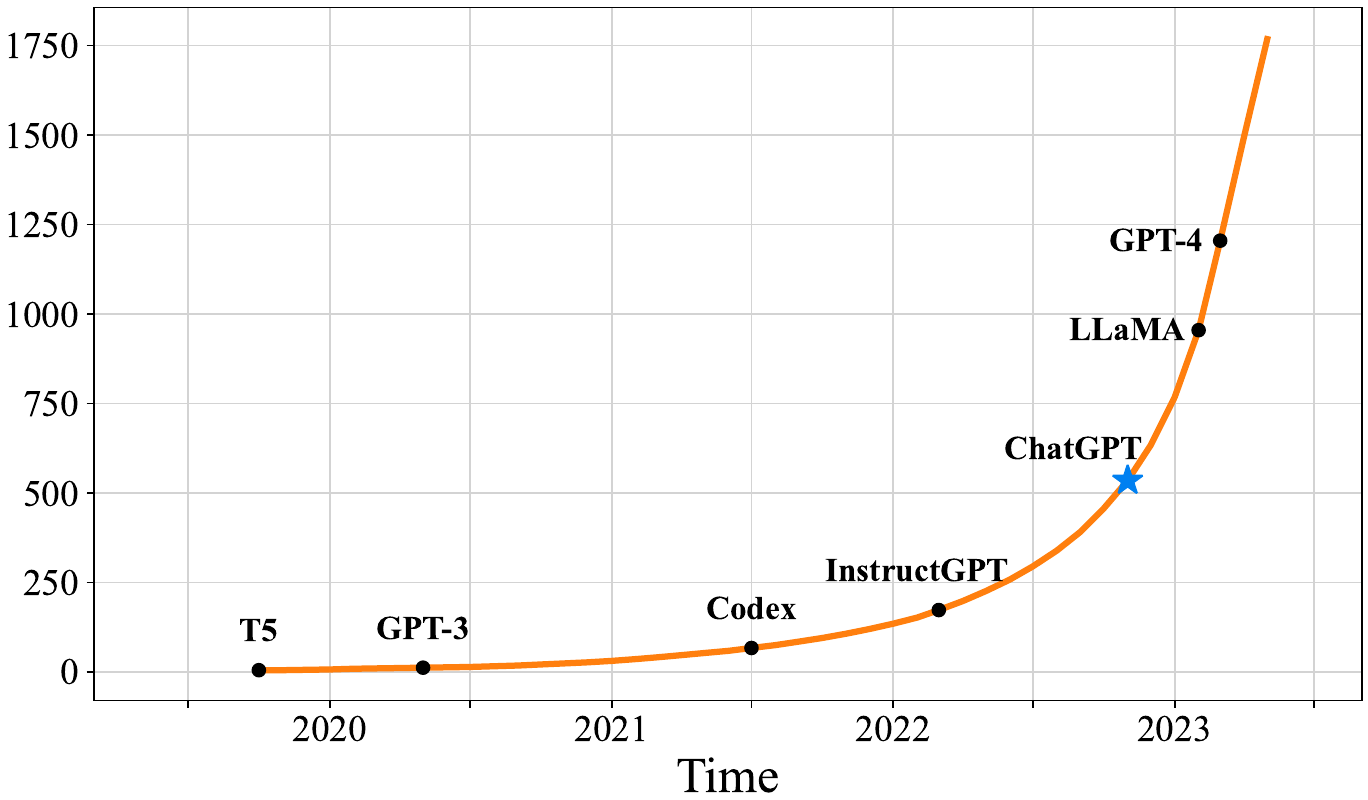}
           \captionof*{figure}{(b) Query="Large Language Model"}
    \end{minipage}
    \caption{{The trends of the  cumulative numbers  of arXiv papers that contain  the keyphrases  ``\emph{language model}" (since June 2018)  and ``\emph{large language model}'' (since October 2019), respectively. The statistics are calculated using   exact match by querying the keyphrases in title or abstract by months. We set different x-axis ranges for the two keyphrases, because ``language models'' have been explored at an earlier time.  We label the points corresponding to  important landmarks in the research progress of LLMs. A sharp increase occurs after the release of ChatGPT: the average number of published  arXiv papers that contain ``\emph{large language model}'' in title or abstract goes from 0.40 per day to 8.58 per day (Figure~\ref{fig:paper_number}(b)). }}
    \label{fig:paper_number}
\end{figure*}

\begin{figure*}
    \centering
\includegraphics[width=.9\textwidth]{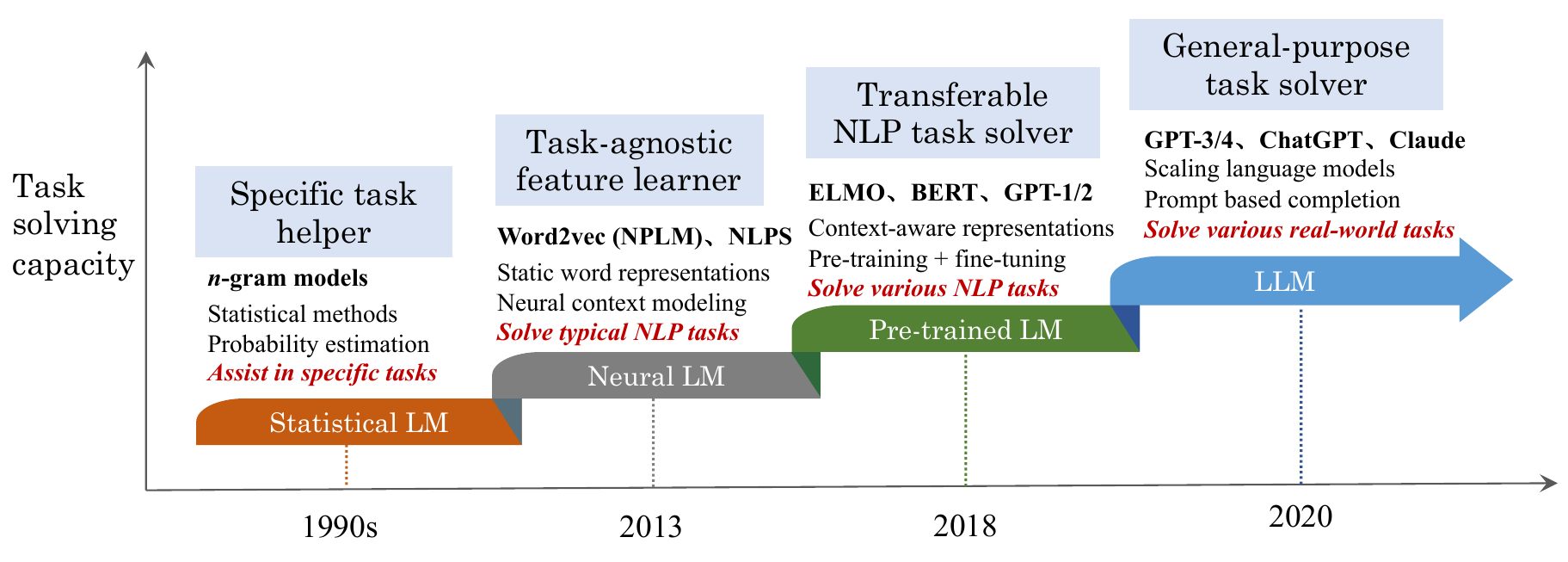}
    \caption{An  evolution process of the four generations of language models~(LM) from the perspective of task solving capacity. Note that the time period for each stage may not be very accurate, and we set the time mainly according to the publish date of the most  representative  studies at each stage. For neural language models, we abbreviate the paper titles of two representative studies to name the two approaches: NPLM~\cite{Bengio-JMLR-2003-A}   (``\emph{A neural probabilistic language model}'') and NLPS~\cite{Collobert-JMLR-2011}    (``\emph{Natural language processing (almost) from scratch}''). Due to the space limitation, we don't list all representative studies in this figure. } \label{fig:task_solvers}
\end{figure*}

\IEEEPARstart{L}{anguage}  is a prominent ability in human beings to express and communicate, which develops in early childhood and evolves over a lifetime~\cite{instinct-book,hauser-science-2002-faculty}.  Machines, however, cannot naturally grasp the abilities of understanding and communicating in the form of human language, unless equipped with powerful artificial intelligence~(AI) algorithms. It has been a longstanding research challenge to achieve this goal, to enable machines to read, write, and communicate like humans~\cite{turing-test}.

Technically, \emph{language modeling~(LM)} is one of the major approaches to advancing language intelligence of machines.  
In general, LM aims to model  the generative likelihood of word sequences, so as to predict the probabilities of future (or missing) tokens. 
The research of LM has received extensive attention in the literature, which can be divided into four major development stages:

$\bullet$ \emph{Statistical language models~(SLM)}.  SLMs~\cite{NLP-speech-book,SLM-2004,rosenfeld2000two,stolcke2002srilm} are developed based on  \emph{statistical learning} methods that rose in the 1990s. The basic idea is to build the word prediction model based on the Markov assumption, \eg predicting the next word based on the most recent context. The SLMs with a fixed context length $n$ are also called $n$-gram language models, \eg   bigram and trigram language models. SLMs have been widely applied to enhance task performance in information retrieval~(IR)~\cite{SLM-IR1,SLM-IR2} and natural language processing~(NLP)~\cite{Thede-acl-1999-a,bahl1989tree,Brants-emnlp-2007-large}.  
However, they often suffer from the curse of dimensionality: it is difficult to accurately estimate high-order language models since an exponential number of transition probabilities need to be  estimated.
{Thus}, specially designed smoothing strategies  such as back-off estimation~\cite{Katz-IEEE-1987-estimation} and Good–Turing estimation~\cite{Gale-JQL-1995-good} have been introduced {to alleviate the data sparsity problem. }

$\bullet$ \emph{Neural language models~(NLM)}. NLMs~\cite{Bengio-JMLR-2003-A,Mikolov-INTERSPEECH-2010,Kombrink-INTERSPEECH-2011} characterize the probability of  word sequences by neural networks, \eg multi-layer perceptron~(MLP) and recurrent neural networks~(RNNs).   
As a remarkable contribution, the work in \cite{Bengio-JMLR-2003-A} introduced the concept of \emph{distributed representation} of words and built the word prediction function conditioned on the aggregated context features (\ie the distributed word vectors).  
By extending the idea of learning effective features for text data, a general neural network approach was developed to build a unified, end-to-end  solution for various NLP tasks~\cite{Collobert-JMLR-2011}. Furthermore, word2vec~\cite{Mikolov-NIPS-2013,Mikolov-ICLR-2013} was proposed to build a simplified shallow neural network for learning distributed word representations, which were demonstrated to be very effective across a variety of NLP tasks. 
These studies have initiated the use of language models for representation learning (beyond word sequence modeling),  having  an important impact on the field of NLP.

$\bullet$ \emph{Pre-trained language models~(PLM)}. As an early attempt,  ELMo~\cite{Peters-NAACL-2018} was proposed to  capture  context-aware word representations by first pre-training a bidirectional LSTM~(biLSTM)  network ({instead of learning fixed word representations}) and then fine-tuning the biLSTM network according to specific downstream tasks.  Furthermore, based on the highly parallelizable Transformer 
architecture~\cite{Vaswani-NIPS-2017-Attention} with self-attention mechanisms, 
BERT~\cite{Devlin-NAACL-2019-BERT} was proposed by  pre-training  bidirectional language models with specially designed pre-training tasks on large-scale unlabeled corpora.  These pre-trained context-aware word representations are very effective as general-purpose semantic features, which have largely raised the performance bar of NLP tasks. This study has inspired a large number of follow-up work, which sets the ``\emph{pre-training} and \emph{fine-tuning}'' learning paradigm. 
Following this paradigm, a great number of studies on PLMs have been developed, introducing either different architectures~\cite{Lewis-ACL-2020-BART,Fedus-JMLR-2021-Switch} (\eg GPT-2~\cite{radford-blog-2019-language} and BART~\cite{Lewis-ACL-2020-BART}) or improved pre-training strategies~\cite{Liu-CoRR-2019-RoBERTa,Sanh-ICLR-2022-Multitask,Wang-ICML-2022-What}. In this paradigm, it often requires fine-tuning the PLM for adapting to different downstream tasks.  

$\bullet$ \emph{Large language models~(LLM)}. Researchers find that scaling PLM (\eg scaling model size or data size)  often leads to an improved model capacity on downstream tasks (\ie following the scaling law~\cite{Kaplan-arxiv-2020-Scaling}). A number of studies have explored the performance limit by training an ever larger PLM (\eg the 175B-parameter GPT-3 and the 540B-parameter PaLM). Although scaling is mainly conducted in model size (with similar architectures and pre-training tasks), these large-sized PLMs display different behaviors from smaller PLMs (\eg 330M-parameter BERT and 1.5B-parameter GPT-2) and show surprising abilities (called \emph{emergent abilities}~\cite{Wei-arxiv-2022-Emergent}) in solving a series of complex tasks. 
For example, GPT-3 can  solve few-shot tasks through \emph{in-context learning}, whereas GPT-2 cannot do well. 
Thus, the research community coins the  term  ``\emph{large language models~(LLM)}''\footnote{Note that a LLM is not necessarily more capable than a small PLM, and  emergent abilities may not  occur in some LLMs. } for these large-sized PLMs~\cite{Shanahan-arxiv-2022-Talking,Wei-arxiv-2022-chain,Hoffmann-arxiv-2022-Training,Taylor-arxiv-2022-Galactica}, which attract  increasing   research attention (See Figure~\ref{fig:paper_number}). 
A remarkable application of LLMs is \emph{ChatGPT}\footnote{https://openai.com/blog/chatgpt/}  that adapts the LLMs from the GPT series for dialogue, which presents an amazing conversation ability with humans. We can observe a sharp increase of the  arXiv papers that are related to LLMs after the release of ChatGPT in Figure~\ref{fig:paper_number}.

{
As discussed before, language model is not a new technical concept specially for LLMs, but has evolved with the advance of artificial intelligence over the decades. Early language models mainly aim to model and generate text data, while latest language models (\eg GPT-4) focus on  {complex task solving}. From \emph{language modeling} to \emph{task solving}, it is an important leap in scientific thinking, which is the key to understand the development of   language models in the research history. 
From the perspective of task solving, the four generations of language models have exhibited different levels of model capacities.  
  In Figure~\ref{fig:task_solvers}, we describe the  evolution  process  of language models in terms of the task solving capacity.
At first, statistical language models mainly assisted in some specific tasks (\eg retrieval or speech tasks), in which the predicted or estimated probabilities can enhance the performance of task-specific approaches.
Subsequently, neural language models focused on learning task-agnostic representations (\eg features), aiming  to reduce the efforts for human feature engineering. Furthermore, pre-trained language models  learned context-aware representations that can be optimized according to downstream  tasks. For the latest generation of language model, LLMs are enhanced by exploring the scaling effect on model capacity, which can be considered as   general-purpose task solvers. To summarize, in the evolution process, the task scope that can be solved by language models have been greatly extended, and the task performance attained by language models have been significantly enhanced. 
}

In the existing literature, PLMs have been widely discussed and surveyed~\cite{Liu-survey-2023-Pre-train,Zhou-arxiv-2023-A,Han-AIopen-2021-PTM,qiu-CoRR-2020-PTM}, while LLMs are seldom reviewed in a systematic way. To  motivate our survey, we first highlight three major differences between LLMs and PLMs. 
First, LLMs display some surprising emergent abilities that may not be observed in previous smaller  PLMs. These abilities are key to the performance of language models on complex tasks, making AI algorithms unprecedently powerful and effective.   
Second, LLMs would  revolutionize  the way that humans  develop and use AI algorithms.    
Unlike small PLMs, the major approach to accessing LLMs is  through the prompting interface (\eg GPT-4 API). Humans have to understand how LLMs work and format their tasks in a way that LLMs can follow.   
Third, the development of LLMs no longer draws a clear distinction between research and engineering. The training of LLMs requires extensive  practical experiences in large-scale data processing and distributed parallel training.  
To develop capable LLMs, researchers have to solve complicated engineering issues, working with engineers or being engineers. 

Nowadays,  LLMs are posing a significant impact on the AI community, and the advent of ChatGPT and GPT-4 leads to the rethinking of the possibilities of artificial general intelligence~(AGI). OpenAI has published a technical article entitled ``\emph{Planning for AGI and beyond}'', which discusses the short-term and long-term plans to approach  AGI~\cite{OpenAI-blog-2023-Planning}, and a more recent paper has argued that GPT-4 might be considered as an early version of an AGI system~\cite{Bubeck-arxiv-2023-Sparks}. 
The research areas of AI are being revolutionized by the rapid progress of LLMs. 
In the field of NLP,  LLMs can serve as a general-purpose  language task solver (to some extent), and  the research paradigm has been shifting towards the use of LLMs. 
In the field of IR, traditional search engines are challenged by the new information seeking way through AI chatbots (\ie ChatGPT), and  \emph{New Bing}\footnote{https://www.bing.com/new} presents an initial attempt that enhances  the search results based on LLMs. 
In the field of  CV, the researchers  try to develop ChatGPT-like vision-language models that can better serve multimodal dialogues~\cite{Huang-CoRR-2023,Cao-arxiv-2023-comprehensive, driess-arxiv-2023-palm,wu-arxiv-2023-visual}, and  GPT-4~\cite{OpenAI-OpenAI-2023-GPT-4} has  supported multimodal input by integrating the visual information.  
This new wave of technology would potentially lead to a prosperous
 ecosystem of real-world applications based on LLMs. 
For instance, Microsoft 365  is being empowered by  LLMs (\ie Copilot) to automate the office work, and OpenAI supports the use of   plugins in ChatGPT for implementing special functions.

Despite the  progress and impact,  the underlying principles of LLMs are still not well explored. Firstly, it is  mysterious   why emergent abilities occur in  LLMs,   instead of smaller PLMs. As a more general issue, there  lacks a deep, detailed investigation of the key factors that contribute to the superior abilities of LLMs. 
It is important to study when and how LLMs obtain such abilities~\cite{FU-blog-2022-how}. Although there are some meaningful discussions about this problem~\cite{Wei-arxiv-2022-Emergent,FU-blog-2022-how}, more principled investigations are needed to uncover the ``\emph{secrets}`` of LLMs. 
Secondly, it is difficult for the research community to train capable LLMs.  
Due to the huge demand of computation resources, it is very costly to carry out repetitive, ablating studies for investigating the effect of various strategies for training LLMs. 
Indeed,  LLMs are mainly trained by industry, where many important training details (\eg data collection and cleaning) are not revealed to the public. 
Thirdly, it is  challenging to align LLMs with human values or preferences. Despite the capacities, LLMs are also likely to produce toxic, fictitious, or harmful contents. It requires effective and efficient control approaches  to  eliminating the potential risk of the use of LLMs~\cite{OpenAI-OpenAI-2023-GPT-4}.

Faced with both opportunities and challenges, it needs more  attention on the research and development of LLMs. 
In order to provide a basic understanding of LLMs,  
this survey conducts a literature review of the recent advances in LLMs 
from four major aspects, including \emph{pre-training} (how to pre-train a capable LLM), \emph{adaptation} (how to effectively adapt pre-trained LLMs for better use), \emph{utilization} (how to use LLMs for solving various downstream tasks) and   
\emph{capability evaluation} (how to evaluate the abilities of LLMs and existing empirical findings).  
We thoroughly comb the literature and summarize the key findings, techniques, and methods of LLMs.  
For this survey, we also 
create a GitHub project website by collecting the supporting resources for LLMs, at the link \url{https://github.com/RUCAIBox/LLMSurvey}.  
We are also aware of several related review articles on PLMs or LLMs~\cite{Han-AIopen-2021-PTM,qiu-CoRR-2020-PTM,Li-IJCAI-2021-Pretrained,Liu-survey-2023-Pre-train,Lu-arxiv-2022-Survey,Dong-arxiv-2023-A,Shanahan-arxiv-2022-Talking,Huang-arxiv-2022-Towards,Qiao-arxiv-2022-Reasoning,Cao-arxiv-2023-comprehensive,Zhou-FITEE-2023-ChatGPT,Zhao-arxiv-2022-Dense}. These papers either  discuss PLMs or some specific (or general) aspects of LLMs. 
Compared with them, we focus on the techniques and methods to develop and use LLMs and provide a relatively comprehensive reference to important aspects  of LLMs.

The remainder of this survey is organized as follows: Section 2 introduces the background for LLMs and the evolution of GPT-series models, followed by the summarization of available resources for developing LLMs in Section 3. Sections 4, 5, 6, and 7 review and summarize the recent progress from the four aspects of pre-training, adaptation, utilization, and capacity evaluation, respectively. 
Then, Section 8 discusses the practical guide for prompt design, and Section 9 reviews the applications of LLMs in several representative domains. 
Finally, we conclude the survey in Section 10 by summarizing the major findings and discuss the remaining issues for future work. 

\section{Overview}
\label{sec-overview}
In this section, we present an overview  about the background of LLMs and then summarize the 
technical evolution of the GPT-series models.

\subsection{Background for LLMs}\label{sec-background}

Typically, \emph{large language models}~(LLMs) refer to Transformer language models 
that contain hundreds of billions~(or more) of parameters\footnote{In existing literature, there is no formal consensus on the minimum parameter scale for LLMs, since the model capacity is also related to  data size and total compute. In this survey, we take a slightly loose definition of LLMs, and mainly focus on discussing language models  with a model size larger than 10B. }, which are trained on massive text  data~\cite{Shanahan-arxiv-2022-Talking}, such as GPT-3~\cite{Brown-NeurIPS-2020-Language}, PaLM~\cite{Chowdhery-arxiv-2022-PaLM},  Galactica~\cite{Taylor-arxiv-2022-Galactica}, and LLaMA~\cite{Touvron-arxiv-2023-LLaMA}.  LLMs exhibit strong capacities to understand natural language and solve complex tasks (via text generation). To have a quick understanding of how LLMs work, this part introduces the basic background for LLMs, including scaling laws, emergent abilities and key techniques.

\paratitle{Formulation of Scaling Laws for LLMs}. Currently, LLMs are mainly built upon the Transformer architecture~\cite{Vaswani-NIPS-2017-Attention}, where  multi-head attention layers are stacked in a very deep neural network. 
Existing LLMs  adopt   similar Transformer architectures and pre-training objectives (\eg language modeling) as  small language models. 
However, LLMs significantly extend the model size, data size, and total compute (orders of magnification). Extensive research has shown that scaling can largely improve the model capacity of LLMs~\cite{radford-blog-2019-language,Brown-NeurIPS-2020-Language,Chowdhery-arxiv-2022-PaLM}. Thus, it is useful to establish a quantitative approach to characterizing the scaling effect. 
Next, we introduce two representative scaling laws for Transformer language models~\cite{Kaplan-arxiv-2020-Scaling,Hoffmann-arxiv-2022-Training}.

$\bullet$ \emph{KM scaling law}\footnote{Since there was not a model trained following this law in the original paper, we took the last names of the two co-first authors to name this scaling law. 
}. In 2020, Kaplan et al.~\cite{Kaplan-arxiv-2020-Scaling} (the OpenAI team) firstly proposed to model the power-law relationship of model performance with respective to three major factors, namely model size ($N$), dataset size ($D$), and the amount of training compute ($C$), for neural language models. Given a compute budget $c$, they empirically presented three basic formulas for the scaling law\footnote{Here, $N_c$, $D_c$ and $C_c$ are measured in the number of non-embedding parameters, the number of training tokens and the number of FP-days, respectively. According to the original paper~\cite{Kaplan-arxiv-2020-Scaling}, $C_c$ and $C$ should be denoted by $C_c^{min}$ and $C_{min}$, corresponding to the optimal use of compute. We use the simplified notations for ease of discussions. }: 

\begin{small}
\begin{eqnarray}
L(N) &=& \bigg(\frac{N_c}{N}\bigg)^{\alpha_N}, \text{~~~} \alpha_N \sim 0.076, N_c \sim 8.8\times 10^{13} \\\nonumber
L(D) &=& \bigg(\frac{D_c
}{D}\bigg)^{\alpha_D},  \text{~~~} \alpha_D \sim 0.095, D_c \sim 5.4\times 10^{13} \\\nonumber
L(C) &=& \bigg(\frac{C_c}{C}\bigg)^{\alpha_C},  \text{~~~} \alpha_C \sim 0.050, C_c \sim 3.1\times 10^{8}\nonumber
\end{eqnarray}
\end{small}

\noindent where $L(\cdot)$ denotes  the cross entropy loss in nats, and a follow-up study~\cite{Henighan-2020-scalinglaw} from OpenAI has shown that the language modeling loss can be decomposed into two parts, namely \emph{irreducible loss} (the entropy of the true data distribution) and \emph{reducible loss} (an estimate of the KL divergence between the true and model distributions). The three laws were derived by fitting the model performance with varied data sizes (22M to 23B tokens), model sizes  (768 to 1.5B non-embedding parameters) and training compute, under some assumptions (\eg the analysis of one factor should be not bottlenecked by the other two factors). They showed that the model performance has a strong dependence relation on the three factors. 

$\bullet$ \emph{Chinchilla scaling law}. As another representative study,  Hoffmann et al.~\cite{Hoffmann-arxiv-2022-Training} (the Google DeepMind team) 
proposed an alternative form for scaling laws to instruct the compute-optimal training for LLMs. They conducted rigorous  experiments by varying a larger range of model sizes (70M to 16B) and data sizes (5B to 500B tokens), and fitted a similar scaling law yet with different coefficients  as below~\cite{Hoffmann-arxiv-2022-Training}:
 \begin{equation}
L(N, D) = E + \frac{A}{N^\alpha} + \frac{B}{D^{\beta}},
\end{equation}
where $E = 1.69, A = 406.4, B = 410.7$, $\alpha=0.34$ and $\beta=0.28$. By optimizing the loss $L(N, D)$ under the constraint $C\approx 6ND$, they showed that  the optimal  allocation of compute budget to model size and data size can be  derived as follows: 

\begin{small}
\begin{eqnarray}\label{eq-CSL}
N_{opt}(C)=G \bigg(\frac{C}{6}\bigg)^a, \text{~~~}
D_{opt}(C)=G^{-1} \bigg(\frac{C}{6}\bigg)^b, 
\end{eqnarray}\nonumber
\end{small}

\noindent where $a=\frac{\alpha}{\alpha+\beta}$,  $b=\frac{\beta}{\alpha+\beta}$ and $G$ is a scaling coefficient that can be computed by $A$, $B$, $\alpha$ and $\beta$. As analyzed in \cite{Hoffmann-arxiv-2022-Training}, given an increase in compute budget, the KM scaling law favors a larger budget allocation in model size than the data size, while the Chinchilla scaling law argues that the two sizes should be increased in equal scales, \ie having similar values for $a$ and $b$ in Equation~\eqref{eq-CSL}. 

\paratitle{Discussion on Scaling Laws}. After introducing the formulations, we continue to discuss scaling law in the following two aspects, to enhance its understanding: 

{
$\bullet$ \emph{Predictable scaling}. 
In practice, scaling law can be used to instruct the training of LLMs, and 
it has been proven   feasible to reliably estimate the performance of larger models based on that of smaller models,   called \emph{predictable scaling}~\cite{OpenAI-OpenAI-2023-GPT-4}. 
The benefits of predictable scaling for training LLMs are mainly twofold.  
Firstly, for large models, it is infeasible to rigorously examine various training tricks or variants, and it would be very  helpful if experiences gained from small models could also apply to large models. 
For instance, small proxy models can be trained to find the optimal schedule of the data mixture for large models~\cite{Xie-arxiv-2023-doremi}. 
Secondly, the training of large-scale models takes a long time,  often suffering from issues such as training loss spike, and scaling law can be employed to monitor the training status of LLMs, \eg identifying abnormal performance at an early time. Despite that scaling law characterizes a smooth trend of performance increase (or loss decrease), it also indicates that  \emph{diminishing returns}\footnote{https://en.wikipedia.org/wiki/Diminishing\_returns} might occur as model scaling. 
An empirical study~\cite{Henighan-2020-scalinglaw}  from the OpenAI team has shown that 
representation quality or semantic content can still effectively improve even if approaching the point of diminishing returns (\ie approaching the irreducible loss)~\cite{Henighan-2020-scalinglaw}. This finding suggests  that training large models are  promising for improving the  performance of downstream tasks. 
To further explore scaling effect, a potential issue is that the amount of available data for training LLMs is actually limited. With the ever-increasing model scale, the public text data would be soon ``exhausted'' for LLMs~\cite{Villalobos-arXiv-2023-runout}.  Thus, it will be meaningful to study how scaling laws apply to  a data-constrained regime~\cite{Muennighoff-arXiv-2023-dataconstrained}, where data  repetition or augmentation might be useful to alleviate data scarcity.   }    

{
$\bullet$ \emph{Task-level  predictability}.  
Existing research of scaling laws are mostly conducted in terms of language modeling loss (\eg per-token cross-entropy loss in nats~\cite{Kaplan-arxiv-2020-Scaling}), while in practice we are more concerned about the performance of LLMs on actual tasks. 
Thus, a basic problem is that how the decrease of language modeling loss translates into the improvement of task performance~\cite{Henighan-2020-scalinglaw}. 
Intuitively, a model with a smaller language modeling  loss tends to yield a better performance on downstream tasks, since language modeling loss can be considered as a general measure of the overall model capacity. 
 GPT-4~\cite{OpenAI-OpenAI-2023-GPT-4} has reported that some capabilities (\eg coding ability) can be accurately predicted via scaling law.  Despite that, 
readers should be aware that a direct decrease in language modeling loss does not always indicate an improvement of model performance on downstream  tasks. Specially, the phenomenon of \emph{inverse scaling} would occur for some  tasks, where task performance surprisingly  becomes worse as the language modeling loss decreases~\cite{McKenzie-2022-inverse}.  Overall, it is more difficult to explore and characterize task-level scaling laws, since it might be also dependent on task-related information (task metric, task difficulty, etc.). Furthermore, some capacities (\eg in-context learning~\cite{Brown-NeurIPS-2020-Language}) are unpredictable according to the scaling law, which can be observed only when the model size exceeds a certain level (as discussed below).  }

\paratitle{Emergent Abilities of LLMs}.  In the literature~\cite{Wei-arxiv-2022-Emergent}, \emph{emergent abilities} of LLMs are formally defined as ``the abilities that are not present in small models but arise in large models'',  which is one of the most prominent features that distinguish LLMs from previous PLMs. 
It further introduces a notable characteristic when emergent abilities occur~\cite{Wei-arxiv-2022-Emergent}:  performance 
rises significantly above random when the scale reaches a certain level. By analogy, such an emergent pattern has close connections with the phenomenon of  \emph{phase transition} in physics~\cite{Wei-arxiv-2022-Emergent,Huberman-AI-1987-phase}. In principle, emergent abilities can be  defined in relation to some complex tasks~\cite{Rae-arxiv-2021-Scaling,Wei-arxiv-2022-Emergent}, while we are more concerned with general abilities that can be applied to solve a variety of  tasks. Here, we briefly introduce three typical emergent abilities for LLMs and  representative models that possess such an ability\footnote{It is difficult to accurately examine the critical size for emergent abilities of LLMs (\ie the minimum size to possess an ability), since it might vary for different models or tasks. Also, existing studies often test emergent abilities on very limited model sizes for a specific LLM. For example, PaLM is often tested with three sizes of 8B, 62B and 540B. It is unclear about the model performance of the untested  sizes.}.

$\bullet$ \emph{In-context learning.}  
{The in-context learning~(ICL) ability is formally introduced by GPT-3~\cite{Brown-NeurIPS-2020-Language}: assuming that the language model has been provided with a natural language instruction and/or  several task demonstrations, it can generate the expected output for the test instances by completing the word sequence of input text, without requiring additional training or gradient update\footnote{In a recent study~\cite{Dai-arxiv-2022-Why}, it also shows that in-context learning implicitly performs meta-optimization through the attention mechanism.}.} 
Among the GPT-series models, the 175B GPT-3 model exhibited a strong ICL ability in general, but not the GPT-1 and GPT-2 models. 
Such an ability also depends on the specific downstream task. For example, the ICL ability can emerge on the arithmetic tasks (\eg the 3-digit addition and subtraction) for the 13B GPT-3, but 175B GPT-3 even cannot work well on the Persian QA task~\cite{Wei-arxiv-2022-Emergent}.

$\bullet$ \emph{Instruction following.} 
By fine-tuning with a mixture of multi-task  datasets formatted via natural language descriptions (called  \emph{instruction tuning}),  
LLMs are shown to  perform well on unseen tasks that are also described in the form of  instructions~\cite{Ouyang-arxiv-2022-Training,Wei-ICLR-2022-Finetuned,Sanh-ICLR-2022-Multitask}. 
{With instruction tuning,  LLMs are enabled to follow  
the task instructions for new tasks without using explicit examples, thus having an improved generalization ability.} 
According to the experiments in~\cite{Wei-ICLR-2022-Finetuned}, instruction-tuned LaMDA-PT~\cite{Thoppilan-CoRR-2022-LaMDA} started to significantly outperform the untuned one on unseen tasks when the model size reached 68B, but not for 8B or smaller model sizes. A recent study~\cite{Chung-arxiv-2022-Scaling} found that a model size of 62B is at least required for PaLM to perform  well on various tasks in four evaluation benchmarks (\ie MMLU, BBH, TyDiQA and MGSM), though a much smaller size might suffice for some specific tasks (\eg MMLU).

$\bullet$ \emph{Step-by-step reasoning.} 
For small language models, it is usually difficult to solve complex tasks that involve multiple reasoning steps, \eg mathematical word problems.
{In contrast, with the chain-of-thought~(CoT) prompting strategy~\cite{Wei-arxiv-2022-chain}, LLMs can solve such tasks by utilizing the prompting mechanism that involves intermediate reasoning steps for deriving the final answer.} 
This ability is speculated to be potentially obtained by training on code~\cite{FU-blog-2022-how,Wei-arxiv-2022-chain}. 
An empirical study~\cite{Wei-arxiv-2022-chain} has shown that CoT prompting can bring performance gains (on arithmetic reasoning benchmarks) when applied to PaLM and LaMDA variants with a model size larger than 60B, while its advantage over the standard prompting becomes more evident when the model size exceeds 100B.  
Furthermore, the performance improvement with CoT prompting seems to be also varied for different tasks, \eg GSM8K $>$ MAWPS $>$ SWAMP for PaLM~\cite{Wei-arxiv-2022-chain}. 

{
\paratitle{How Emergent Abilities Relate to Scaling Laws}. In existing literature~\cite{Kaplan-arxiv-2020-Scaling,Hoffmann-arxiv-2022-Training,Wei-arxiv-2022-Emergent}, scaling laws and emergent abilities  provide two perspectives to understand the advantage of large models over small models. In general, scaling law (often measured by \emph{language modeling loss}) describes predictable performance relation with the potential effect of diminishing returns, while emergent abilities (often measured by \emph{task performance}) are unpredictable but very profitable once such abilities actually emerge.  
Since the two perspectives reflect different performance trends (continuous improvement \emph{v.s.} sharp performance leap), they might lead to misaligned findings or observations. 
There are also extensive debates on the rationality of emergent abilities. 
A popular speculation is that  emergent abilities might be partially attributed to the evaluation setting for special tasks (\eg the discontinuous evaluation metrics)~\cite{Srivastava-arxiv-2022-Beyond,Schaeffer-arXiv-2023-mirage}: when evaluation metrics are altered accordingly, the sharpness of the emergent ability curve would disappear. 
However, the performance of LLMs on most tasks are perceived by users naturally in a discontinuous way.  
For instance, end users prefer a reliable code generated by LLMs that can successfully pass  the test case, but are less interested in selecting a better code with fewer errors between two failed ones.     
More recently, a study~\cite{Hu-arXiv-2023-unlock} proposes a new evaluation setting that can enlarge the resolution of task metrics, making task performance more predictable. Despite these efforts, more fundamental research (\eg grokking\footnote{Grokking refers that ``a pattern in the data, improving generalization performance from random chance level to perfect generalization'', quoted from the original paper~\cite{Power-arxiv-2022-grokking}.}) about the working mechanism of LLMs is still in need to understand the emergence of certain abilities. The subtle relation  between scaling law and emergent abilities can be explained by analogy with the ability acquisition of human\footnote{This explanation is only for ease of understanding, and there is not direct evidence to connect the two points. }.
Take the speaking ability as an example. For children,   language development  (especially infants) can be also considered as a multi-level  process where ``emergent abilities'' occur. Specially, the language ability would relatively stable within a time interval, but qualitative change only occurs when evolving into another ability level (\eg from speaking simple words to speaking simple sentences). Such a learning process is essentially not  \emph{smooth} and \emph{stable} (\ie language ability does not develop at a constant rate over time), though a child actually grows every day. It is interesting that young parents would be often surprised by unexpected progress of the speaking ability exhibited by their babies.
}


\paratitle{Key Techniques for LLMs}. It has been a long way that LLMs evolve into the current state:  \emph{general} and \emph{capable} learners.  
In the development process, a number of important techniques are proposed, which largely improve the capacity of LLMs.   
Here, we briefly list several important techniques that (potentially)  lead to the success of LLMs, as follows. 

$\bullet$ \emph{Scaling}. As discussed in previous parts, there exists an evident scaling effect in Transformer language models:   larger model/data sizes and more training compute typically lead to an improved model capacity~\cite{Kaplan-arxiv-2020-Scaling,Hoffmann-arxiv-2022-Training}. 
As two representative models,  GPT-3 and PaLM explored the scaling limits  by increasing the model size to 175B and 540B, respectively. Since compute budget is usually limited,  scaling laws can be further  employed to conduct a more compute-efficient allocation of the compute resources.  
For example,  Chinchilla (with more training tokens) outperforms its 
counterpart model Gopher (with a larger model size) by increasing the data scale with the same compute budget~\cite{Hoffmann-arxiv-2022-Training}. 
In addition,  data scaling should be with careful cleaning process,  since the quality of pre-training data plays a key role in the model capacity.

$\bullet$ \emph{Training}. Due to the huge model size, it is very challenging to successfully train a capable LLM. Distributed training algorithms are needed to learn
the network parameters of LLMs, in which various parallel strategies  are often jointly utilized. To support distributed training, several optimization frameworks have been released to facilitate the implementation and deployment of parallel algorithms, such as DeepSpeed~\cite{Rasley-KDD-2020-DeepSpeed} and Megatron-LM~\cite{Shoeybi-arXiv-2019-Megatron, Narayanan-ACM-2021-Efficient, Korthikanti-arxiv-2022-reducing}. Also, optimization tricks are also important for training stability and model performance, \eg  restart to overcome  training loss spike~\cite{Chowdhery-arxiv-2022-PaLM} and mixed precision training~\cite{Scao-arxiv-2022-BLOOM}. 
More recently, GPT-4~\cite{OpenAI-OpenAI-2023-GPT-4} proposes to develop special infrastructure and optimization methods that reliably predict the performance of large models with much smaller models.

$\bullet$ \emph{Ability eliciting}. After being pre-trained on large-scale corpora, LLMs are endowed with potential abilities as general-purpose task solvers. 
These abilities might not be explicitly exhibited when LLMs perform some specific  tasks. As the technical  approach, it is useful to design suitable task instructions or specific  in-context learning strategies to elicit such abilities.
For instance, chain-of-thought prompting has been shown to be useful to solve complex reasoning tasks by including intermediate reasoning steps.  
Furthermore, we can  perform instruction tuning on LLMs with task descriptions expressed in natural language, for improving  the generalizability  of LLMs on unseen tasks. 
These eliciting  techniques  mainly correspond to the emergent abilities of LLMs, which may not show the same effect on small language models.

$\bullet$ \emph{Alignment tuning}. Since  LLMs are trained to capture the data characteristics of pre-training corpora (including both high-quality and low-quality data),  they are likely to generate toxic, biased, or even harmful content for humans. It is necessary to align LLMs with human values, \eg  \emph{helpful}, \emph{honest}, and \emph{harmless}. For this purpose, InstructGPT~\cite{Ouyang-arxiv-2022-Training} designs an effective tuning approach that enables LLMs to follow the expected instructions, which utilizes the technique of  \emph{reinforcement learning with human feedback}~\cite{Christiano-NeurIPS-2017-Deep,Ouyang-arxiv-2022-Training}.  It incorporates human in the training  loop with elaborately designed labeling strategies. 
ChatGPT is indeed developed on a similar technique to InstructGPT,  which shows a strong alignment capacity in producing high-quality, harmless responses, \eg rejecting to answer insulting questions.

$\bullet$ \emph{Tools manipulation}. In essence, LLMs are trained as text generators over massive plain text corpora, thus performing less well on the tasks that are not best expressed in the form of text (\eg numerical computation). In addition, their capacities are also limited to the pre-training data, \eg the inability to capture up-to-date information. To tackle these issues, a recently proposed technique is to employ external tools to compensate for the deficiencies of LLMs~\cite{Schick-arxiv-2023-Toolformer,Nakano-arxiv-2021-WebGPT}. For example, LLMs can utilize the calculator for accurate computation~\cite{Schick-arxiv-2023-Toolformer} and employ search engines to retrieve unknown information~\cite{Nakano-arxiv-2021-WebGPT}. 
More recently, ChatGPT has enabled the mechanism of using external plugins (existing or newly created apps)\footnote{https://openai.com/blog/chatgpt-plugins}, which are by analogy with the ``\emph{eyes and ears}'' of LLMs.  Such a mechanism can broadly expand the scope of capacities for LLMs.

In addition, many other factors (\eg the upgrade of hardware) also contribute to the success of LLMs. Currently, we limit our discussion to the major technical approaches and key findings for developing LLMs.

\begin{figure*}
    \centering
    \includegraphics[width=\textwidth]{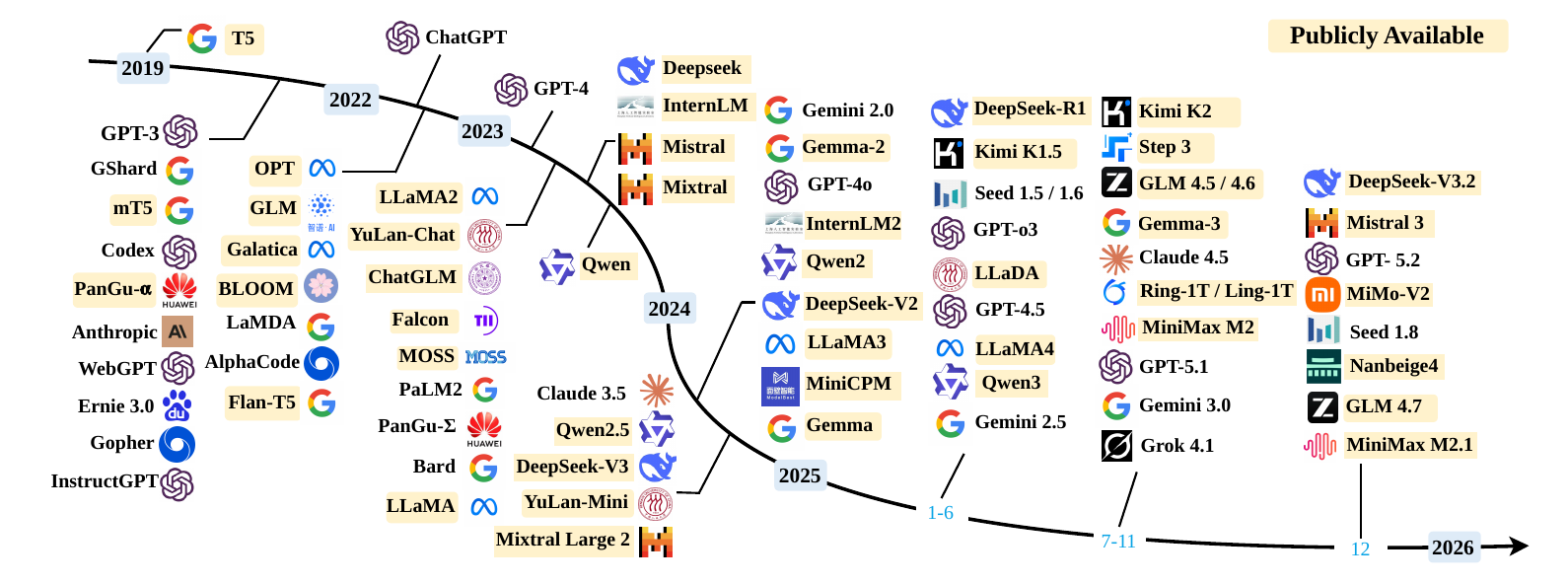}
    \caption{A timeline of representative LLMs released in recent years. Models with publicly available checkpoints are highlighted in yellow.}
    \label{fig:llms_timeline}
\end{figure*}

\begin{table*}[htbp]
\centering
\caption{Statistics  of  large language models (having a size larger than 10B in this survey) in recent years, including the capacity evaluation, pre-training data scale (either in the number of tokens or storage size)  and hardware resource costs. In this table, we only include LLMs with a public paper about the technical details.
Here, ``Release Time"   indicates the date when the corresponding paper was officially released. ``Publicly Available" means  that the model checkpoints can be publicly accessible while ``Closed Source"  means the opposite.
``Adaptation'' indicates whether the model has been with subsequent fine-tuning: IT denotes instruction tuning and RLHF denotes reinforcement learning with human feedback.
``Evaluation'' indicates whether the model has been evaluated with  corresponding abilities in their original paper: ICL denotes in-context learning and CoT denotes chain-of-thought. ``*'' denotes the largest publicly available version. 
}
\label{tab:resource_model}
\footnotesize
\renewcommand\tabcolsep{2.5pt}
\begin{tabular}{llcrccccccccc}
\toprule
  &  &   & \multicolumn{1}{c}{}  & & \multicolumn{2}{c}{\textbf{Adaptation}} &   &    & &    & \multicolumn{2}{c}{\textbf{Evaluation}}   \\
\multirow{-2}{*}{}    & \multirow{-2}{*}{\textbf{Model}} & \multirow{-2}{*}{\textbf{\begin{tabular}[c]{@{}c@{}}Release\\ Time\end{tabular}}} & \multicolumn{1}{c}{\multirow{-2}{*}{\textbf{\begin{tabular}[c]{@{}c@{}}Size\\ (B)\end{tabular}}}} & \multirow{-2}{*}{\textbf{\begin{tabular}[c]{@{}c@{}}Base\\ Model\end{tabular}}} & \textbf{IT}   & \textbf{RLHF} & \multirow{-2}{*}{\textbf{\begin{tabular}[c]{@{}c@{}}Pre-train\\ Data Scale\end{tabular}}} & \multirow{-2}{*}{\textbf{\begin{tabular}[c]{@{}c@{}}Latest Data\\ Timestamp\end{tabular}}} & \multirow{-2}{*}{\textbf{\begin{tabular}[c]{@{}c@{}}Hardware\\ (GPUs / TPUs)\end{tabular}}} & \multirow{-2}{*}{\textbf{\begin{tabular}[c]{@{}c@{}}Training\\ Time\end{tabular}}} & \textbf{ICL} & \textbf{CoT} \\
\midrule
  & T5~\cite{Raffel-JMLR-2020-Exploring}   & Oct-2019    & 11    & -   & - & - & {1T tokens}    & Apr-2019 & 1024 TPU v3 & -  & $\checkmark$ & -    \\
  & mT5~\cite{Xue-NAACL-2021-mT5}  & Oct-2020    & 13    & -   & - & - & 1T tokens & - & -   & -  & $\checkmark$ & - \\
  &  {PanGu-$\alpha$}~\cite{Zeng-arxiv-2021-PanGualpha}   & Apr-2021    & 13*    & -  &  - & - & 1.1TB   & -  & 2048 Ascend 910 &  -  & $\checkmark$ & - \\
  & CPM-2~\cite{Zhang-arXiv-2021-CPM-2}    & Jun-2021    & 198   & -   & - & - & {2.6TB}  & -  & - & - & -       & -    \\
  & T0~\cite{Sanh-ICLR-2022-Multitask}   & Oct-2021    & 11    & T5  & $\checkmark$  & - & -   & -  & 512 TPU v3  & 27 h   & $\checkmark$ & - \\
  & CodeGen~\cite{nijkamp-arxiv-2022-Codegen}  & Mar-2022    & 16    & -   & - & - & 577B tokens & -  & -   & -  & $\checkmark$ & -  \\
  & GPT-NeoX-20B~\cite{Black-CoRR-2022-GPT} & Apr-2022    & 20    & -   & - & - & 825GB & - & 96 40G A100 & -  & $\checkmark$ & -      \\
  & Tk-Instruct~\cite{Wang-EMNLP-2022-Super}  & Apr-2022    & 11    & T5  & $\checkmark$  & - & - & -  & 256 TPU v3   & 4 h  & $\checkmark$    & -    \\
  & UL2~\cite{Tay-arxiv-2022-UL2}  & May-2022    & 20    & -   & - & - & 1T tokens & Apr-2019 & 512 TPU v4  & -  & $\checkmark$ & $\checkmark$    \\
  & OPT~\cite{Zhang-arxiv-2022-OPT}  & May-2022    & 175   & -   & - & - & 180B tokens   & -  & 992 80G A100    & -  & $\checkmark$  & -    \\
  & NLLB~\cite{Marta-arxiv-2022-NLLB}  & Jul-2022    & 54.5   & -   & - & - & -  & -  & -   & -  & $\checkmark$  & -    \\
  & CodeGeeX~\cite{Zheng-arXiv-2023-CodeGeex}    & Sep-2022    & 13    & -   & - & - & 850B tokens   & - & 1536 Ascend 910 & 60 d   & $\checkmark$ & - \\
  & GLM~\cite{Zeng-arxiv-2022-GLM}  & Oct-2022    & 130   & -   & - & - & 400B tokens   & -  & 768 40G A100    & 60 d  & $\checkmark$ & -    \\
  & Flan-T5~\cite{Chung-arxiv-2022-Scaling}  & Oct-2022    & 11    & T5  & $\checkmark$  & - & - & -  & -   & -  & $\checkmark$ & $\checkmark$  \\
  & BLOOM~\cite{Scao-arxiv-2022-BLOOM}    & Nov-2022    & 176   & -   & - & - & 366B tokens & -  & 384 80G A100    & 105 d  &  $\checkmark$    & -    \\
  & mT0~\cite{Muennighoff-2022-arxiv-Crosslingual}  & Nov-2022    & 13    & mT5 & $\checkmark$  & - & - & -  & -   & -  & $\checkmark$ & - \\
  & Galactica~\cite{Taylor-arxiv-2022-Galactica} & Nov-2022    & 120   & -   & - & - & 106B tokens & -  & -   & -  & $\checkmark$ & $\checkmark$    \\
  & BLOOMZ~\cite{Muennighoff-2022-arxiv-Crosslingual}   & Nov-2022    & 176   & BLOOM   & $\checkmark$  & - & - & -  & -   & -  & $\checkmark$ & - \\
  & OPT-IML~\cite{Iyer-arxiv-2022-OPT}  & Dec-2022    & 175   & OPT & $\checkmark$  & - & - & -  & 128 40G A100   & -  & $\checkmark$ & $\checkmark$    \\
  & LLaMA~\cite{Touvron-arxiv-2023-LLaMA}    & Feb-2023    & 65    & -   & - & - & 1.4T tokens   & - & 2048 80G A100 & 21 d   & $\checkmark$ & -    \\
  & Pythia~\cite{Biderman-arxiv-2023-Pythia}    & Apr-2023    & 12    & -   & - & - & 300B tokens   & - & 256 40G A100 & -  & $\checkmark$ & - \\
  & CodeGen2~\cite{Nijkamp-2023-codegen2-arxiv}    & May-2023    & 16    & -   & - & - & 400B tokens   & - & - & -  & $\checkmark$ & - \\
  & StarCoder~\cite{Li-2023-arxiv-Starcoder}    & May-2023    & 15.5    & -   & - & - & 1T tokens   & - & 512 40G A100 & -  & $\checkmark$ & $\checkmark$ \\
   & LLaMA2~\cite{Touvron-2023-llama2-arxiv}  & Jul-2023    & 70   & - & $\checkmark$   & $\checkmark$ & 2T tokens & -  & 2000 80G A100   & -  & $\checkmark$ & - \\
    & Baichuan2~\cite{yang-2023-baichuan2}  & Sep-2023    & 13   & - & $\checkmark$   & $\checkmark$ & 2.6T tokens & -  & 1024 A800   & -  & $\checkmark$ & - \\
    & QWEN~\cite{bai-2023-qwen}  & Sep-2023    & 14   & - & $\checkmark$   & $\checkmark$ & 3T tokens & -  & -   & -  & $\checkmark$ & - \\
    & FLM~\cite{Li-arxiv-2023-FLM}  & Sep-2023    & 101   & - & $\checkmark$  & - & 311B tokens & -  & 192 A800   & 22 d  & $\checkmark$ & - \\
\multirow{-18}{*}{\begin{tabular}[c]{@{}c@{}}Publicly\\ Available\end{tabular}}  & Skywork~\cite{wei-2023-skywork}  & Oct-2023    & 13   & - & -    & - & 3.2T tokens & -  & 512 80G A800   & -  & $\checkmark$ & - \\

\midrule
\midrule
  & GPT-3~\cite{Brown-NeurIPS-2020-Language}    & May-2020    & 175   & -   & - & - & {300B tokens} & -  & -   & -  & $\checkmark$ & -       \\
  & GShard~\cite{Lepikhin-ILR-2021-GShard}   & Jun-2020    & 600   & -   & - & - & 1T tokens & -  & 2048 TPU v3 & 4 d & -    & -       \\
  & Codex~\cite{Chen-arxiv-2021-evaluating}    & Jul-2021    & 12    & GPT-3   & - & - & 100B tokens & May-2020 & -   & -  & $\checkmark$    & - \\
  & ERNIE 3.0~\cite{Sun-arXiv-2021-ERNIE3.0}    & Jul-2021    & 10    & -   & - & - & 375B tokens & - & 384 V100   & -  & $\checkmark$ & -    \\
  & Jurassic-1~\cite{lieber-2021-jurassic}   & Aug-2021    & 178   & -   & - & - & 300B tokens   & -  & 800 GPU & -  & $\checkmark$    & -       \\
  & HyperCLOVA~\cite{Kim-EMNLP-2021-HyperCLOVA}   & Sep-2021    & 82    & - & -  & - &  300B tokens  & - & 1024 A100   & 13.4 d  & $\checkmark$ & -      \\
  & FLAN~\cite{Wei-ICLR-2022-Finetuned} & Sep-2021    & 137   & LaMDA-PT   & $\checkmark$  & - & -  & -  & 128 TPU v3  & 60 h   & $\checkmark$  & -    \\
  & Yuan 1.0~\cite{Wu-arxiv-2021-Yuan}   & Oct-2021    & 245    &  - & -  & - &  180B tokens  &  -  & 2128 GPU  &  -  & $\checkmark$ & - \\
  & Anthropic~\cite{Askell-arxiv-2021-Anthropic}   & Dec-2021    & 52   & -   & - & - & {400B tokens}   & - & -   & -  & $\checkmark$    & -        \\
  & WebGPT~\cite{Nakano-arxiv-2021-WebGPT}   & Dec-2021    & 175   & GPT-3   & - & $\checkmark$  & - & -  & -   & -  & $\checkmark$ &  -    \\
  & Gopher~\cite{Rae-arxiv-2021-Scaling}   & Dec-2021    & 280   & -   & - & - & 300B tokens   & -  & 4096 TPU v3 & 920 h  & $\checkmark$    & -        \\
  & ERNIE 3.0 Titan~\cite{Wang-arxiv-2021-ERNIE}   & Dec-2021    & 260    &  - & -  & - &  -  & - & -  & -  & $\checkmark$ & -  \\
  & GLaM~\cite{Du-ICML-2022-GLaM} & Dec-2021    & 1200  & -   & - & - & 280B tokens   & -  & 1024 TPU v4 & 574 h  & $\checkmark$ & -       \\
  & LaMDA~\cite{Thoppilan-CoRR-2022-LaMDA}    & Jan-2022    & 137   & -   & - & - & 768B tokens  & -  & 1024 TPU v3 & 57.7 d & -    & -       \\
  & MT-NLG~\cite{Smith-CoRR-2022-Using}   & Jan-2022    & 530   & -   & - & - & {270B tokens}   & - & 4480 80G A100   & -  & $\checkmark$    & -        \\
  & AlphaCode~\cite{Li-Science-2022-AlphaCode}    & Feb-2022    & 41    & -   & - & - & 967B tokens & Jul-2021 & -   & -  & -    & -        \\
  & InstructGPT~\cite{Ouyang-arxiv-2022-Training}  & Mar-2022    & 175   & GPT-3   & $\checkmark$  & $\checkmark$  & -  & -  & -   & -  & $\checkmark$  & -    \\
  & Chinchilla~\cite{Hoffmann-arxiv-2022-Training}   & Mar-2022    & 70    & -   & - & - & 1.4T tokens   & -  & -   & -  & $\checkmark$    & -        \\
  & PaLM~\cite{Chowdhery-arxiv-2022-PaLM} & Apr-2022    & 540   & -   & - & - & 780B tokens   & -  & 6144 TPU v4 & -  & $\checkmark$       & $\checkmark$ \\
  & AlexaTM~\cite{Soltan-arxiv-2022-AlexaTM20B}  & Aug-2022    & 20    & -   & - & - & 1.3T tokens   & -  & 128 A100    & 120 d  & $\checkmark$ & $\checkmark$       \\
  & Sparrow~\cite{Glaese-arxiv-2022-Improving}  & Sep-2022    & 70    & -   & - & $\checkmark$  & - & -  & 64 TPU v3   & -  & $\checkmark$    & -       \\
  & WeLM~\cite{Su-arxiv-2022-WeLM}  & Sep-2022    & 10    & -   & - & -  & 300B tokens & -  & 128 A100 40G   & 24 d  & $\checkmark$    & -       \\
  & U-PaLM~\cite{Tay-arxiv-2022-Transcending}   & Oct-2022    & 540   & PaLM & - & - & - & -  & 512 TPU v4  & 5 d    & $\checkmark$       & $\checkmark$ \\
  & Flan-PaLM~\cite{Chung-arxiv-2022-Scaling}    & Oct-2022    & 540   & PaLM    & $\checkmark$  & - & - & -  & 512 TPU v4  & 37 h   & $\checkmark$ & $\checkmark$  \\
  & Flan-U-PaLM~\cite{Chung-arxiv-2022-Scaling}  & Oct-2022    & 540   & U-PaLM  & $\checkmark$  & - & - & -  & -   & -  & $\checkmark$ & $\checkmark$  \\
  & GPT-4~\cite{OpenAI-OpenAI-2023-GPT-4}    & Mar-2023    & - & -   & $\checkmark$  & $\checkmark$  & - & -  & -   & -  & $\checkmark$ & $\checkmark$  \\
  & {PanGu-$\Sigma$}~\cite{Ren-arXiv-2023-PanGusigma}  & Mar-2023  & 1085 & {PanGu-$\alpha$}   & -  & -  & 329B tokens & -  & 512 Ascend 910   & 100 d  & $\checkmark$  & -   \\
\multirow{-25}{*}{\begin{tabular}[c]{@{}c@{}}Closed\\ Source\end{tabular}} & PaLM2~\cite{Anil-arxiv-2023-palm2}    & May-2023    & 16 & -   & $\checkmark$  & -  & 100B tokens & -  & -   & -  & $\checkmark$ & $\checkmark$  \\
\bottomrule
\end{tabular}
\end{table*}

\begin{figure*}[h]
    \centering
    \includegraphics[width=\textwidth]{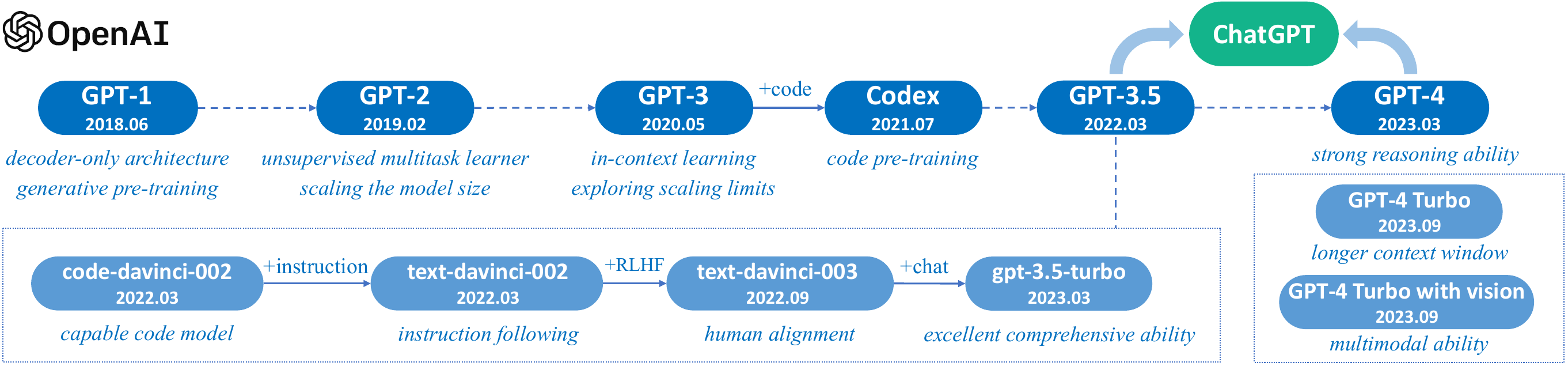}
    \caption{A brief  illustration for the technical evolution of GPT-series models. We plot this figure mainly based on the papers, blog articles and official APIs from OpenAI. Here,  \emph{solid lines}  denote that there exists an explicit evidence (\eg the official statement that a new model is developed based on a base model) on the evolution path between two models, while \emph{dashed lines} denote a relatively weaker evolution relation. }
    \label{fig:openai}
\end{figure*}

\subsection{Technical Evolution  of GPT-series Models}\label{sec-GPT-series}

Due to the  excellent capacity in communicating  with humans,  ChatGPT has ignited the excitement of the AI community since its release.   
ChatGPT is developed based on the powerful GPT model with specially optimized conversation  capacities.   
Considering the ever-growing interest in ChatGPT and GPT models, we add a special discussion about the technical evolution of the GPT-series models, to briefly summarize the progress how they have been developed in the past years. 
{Meanwhile, we drew a schematic diagram depicting the technological evolution of the GPT-series models in Figure~\ref{fig:openai}.} 
The basic principle underlying GPT models is to compress the world knowledge into the decoder-only Transformer model by language modeling,  such that it can  recover (or memorize) the semantics of world knowledge and serve as a general-purpose task solver. Two key points to the success are  (I)   training decoder-only Transformer language models that can \emph{accurately predict the next word}  and (II) \emph{scaling  up the size of  language models}.   
Overall, the research of OpenAI on LLMs can be roughly divided into the following stages\footnote{Note that the discussion of this part can be somewhat subjective. The overall viewpoints and summaries are made based on the understanding of the survey authors by reading  the papers, blog articles, interview reports and APIs released by OpenAI.  }.  

\paratitle{Early Explorations}. According to one interview with Ilya Sutskever\footnote{\url{https://hackernoon.com/an-interview-with-ilya-sutskever-co-founder-of-openai}} (a co-founder and chief scientist of OpenAI), the idea of approaching  intelligent systems  with language models was already explored in the early days of OpenAI, while it was attempted with recurrent neural networks~(RNN)~\cite{Radford-CoRR-2017-Learning}. With the advent of Transformer, OpenAI developed two initial GPT models, namely GPT-1~\cite{radford-openai-2018-improving} and GPT-2~\cite{radford-blog-2019-language}, which can be considered as the foundation to  more powerful models subsequently \ie GPT-3 and GPT-4.   

$\bullet$ \emph{GPT-1}. In 2017, the Transformer model~\cite{Vaswani-NIPS-2017-Attention} was introduced by Google, and the OpenAI team quickly adapted their language modeling work to this new neural network architecture. They released the first GPT model in 2018, \ie GPT-1~\cite{radford-openai-2018-improving}, and coined the abbreviation term \emph{GPT} as the model name,  standing for \emph{Generative Pre-Training}. GPT-1 was developed based on a generative, decoder-only Transformer architecture, and adopted a hybrid  approach of  unsupervised pre-training and supervised fine-tuning. 
GPT-1 has set up the core architecture for the GPT-series models and established the underlying principle to model natural language text, \ie predicting the next word.  

$\bullet$ \emph{GPT-2}. Following a similar architecture of GPT-1,  GPT-2~\cite{radford-blog-2019-language} increased the parameter scale to 1.5B, which was trained with a large webpage dataset WebText. As claimed in the paper of GPT-2, it sought to perform tasks via unsupervised language modeling, without  explicit fine-tuning using labeled data. To motivate the approach, they  introduced a   probabilistic form for  multi-task solving, \ie  $p(output|input, task)$ (similar approaches have been adopted in \cite{McCann-CoRR-2018-The}), which predicts the output conditioned on the input and task information.  To model this conditional probability,  language text can be naturally employed as a unified way to  format  input, output and task information.  In this way, the process of solving a task  can be cast as a word prediction problem for generating the solution text. Further, they introduced  a more formal claim for this idea: ``Since the (task-specific) supervised objective is the same as the unsupervised (language modeling) objective but only evaluated on a subset of the sequence, the global minimum of the unsupervised objective is also the global minimum of the supervised objective (for various tasks)''~\cite{radford-blog-2019-language}\footnote{To better understand this sentence, we put some explanation words in parentheses.}. 
A basic understanding of this claim is that each (NLP) task can be considered as the word prediction problem based on a subset of the world text. Thus,  unsupervised language modeling could be capable in solving various tasks, if it was trained to have sufficient capacity in recovering the world text.   
These early discussion in GPT-2's paper echoed in the interview of Ilya Sutskever by Jensen Huang: ``What the neural network learns is some representation of the process that produced the text. This text is actually a projection of the world...the more accurate you are in predicting the next word, the higher the fidelity, the more resolution you get in this process...''\footnote{\url{https://lifearchitect.ai/ilya/}}.

\paratitle{Capacity Leap}. Although GPT-2 is intended to be an  ``unsupervised multitask learner'', it overall has an inferior performance compared with supervised fine-tuning state-of-the-art methods.  
Because it has a relatively small model size, it has been widely fine-tuned in downstream tasks, especially the dialog tasks~\cite{Zhang-ACL-2020-DIALOGPT,Ham-ACL-2020-End}. Based on GPT-2, GPT-3 demonstrates a  key capacity leap by scaling of the (nearly same) generative pre-training architecture.  
 
$\bullet$ \emph{GPT-3}. GPT-3~\cite{Brown-NeurIPS-2020-Language} was released in 2020, which scaled the model parameters to an ever larger size of 175B. In the GPT-3's paper, it formally introduced the concept of in-context learning~(ICL)\footnote{GPT-2 essentially used ICL for unsupervised task learning, though it wasn't called ICL at that time.  }, which utilizes LLMs in a few-shot or zero-shot way. ICL can  teach (or instruct) LLMs to understand the tasks  in the form of natural language text. 
With ICL, the pre-training and utilization of LLMs converge to the same language modeling paradigm: pre-training predicts the following text sequence conditioned on the context, while ICL predicts the correct task solution, which can be  also formatted as a text sequence, given the task description and demonstrations. 
GPT-3 not only demonstrates very excellent performance in a variety of NLP tasks, but also on a number of specially designed tasks that require the abilities of reasoning or domain adaptation. Although the GPT-3's paper does not explicitly discuss the emergent abilities of LLMs, we can observe  large performance leap that might transcend the basic scaling law~\cite{Kaplan-arxiv-2020-Scaling}, \eg  larger models have significantly stronger ICL ability (illustrated in the original Figure~1.2 of the GPT-3's  paper~\cite{Brown-NeurIPS-2020-Language}).  Overall, GPT-3 can be viewed as a remarkable landmark in the journey evolving  from PLMs to LLMs.  It has empirically proved that scaling the neural networks to a  significant size can lead to a huge increase in model capacity.

\paratitle{Capacity Enhancement}. Due to the strong capacities,  GPT-3  has been the base model to develop even more capable LLMs for OpenAI. Overall, OpenAI has explored  two major approaches to further improving the GPT-3 model, \ie training on code data and alignment with human preference, which are detailed as follows. 

$\bullet$ \emph{Training on code data}. A major limitation of the original  GPT-3 model  (pre-trained on plain text) lies in the lack of the reasoning ability on complex tasks, \eg completing the code and solving math problems. To enhance this ability,  Codex~\cite{Chen-arxiv-2021-evaluating} was introduced by OpenAI in July 2021, which was a GPT model fine-tuned on 
a large corpus of GitHub code. It demonstrated that Codex can solve very difficult programming problems, and also lead to a significant performance  improvement in solving math problems~\cite{Drori-CoRR-2021-A}. Further, a contrastive approach~\cite{Neelakantan-CoRR-2022-Text} to training text and code embedding was reported in January 2022, which was shown to improve a series of related tasks (\ie  linear-probe classification, text search and code search). Actually, the GPT-3.5 models are developed based on a code-based GPT model (\ie \texttt{code-davinci-002}), which indicates that training on code data is a very useful practice to improve the model capacity  of GPT models, especially the reasoning ability.  
Furthermore, there is  also a speculation  that training on code data can greatly increase the chain-of-thought prompting abilities of LLMs~\cite{FU-blog-2022-how}, while it is still worth further investigation with more thorough verification.

$\bullet$ \emph{Human alignment}. The related research of human alignment can be dated back to the year 2017 (or earlier) for OpenAI: a blog article entitled ``learning from human preferences''\footnote{\url{https://openai.com/research/learning-from-human-preferences}} was posted on the OpenAI blog describing a work that applied   reinforcement learning~(RL) to learn from the \emph{preference comparisons}  annotated by  humans~\cite{Christiano-NeurIPS-2017-Deep} (similar to the \emph{reward training}  step in the aligning algorithm of InstructGPT in Figure~\ref{fig:RLHF}). 
Shortly after the release of this RL paper~\cite{Christiano-NeurIPS-2017-Deep}, the paper of the Proximal Policy Optimization~(PPO)~\cite{schulman-arxiv-2017-proximal} was published in July 2017, which now has been the foundational RL algorithm for learning from human preferences~\cite{Ouyang-arxiv-2022-Training}. 
Later in January 2020, GPT-2 was fine-tuned using the aforementioned RL algorithms~\cite{Christiano-NeurIPS-2017-Deep,schulman-arxiv-2017-proximal}, which leveraged human preferences to improve the capacities of GPT-2 on NLP tasks. In the same year, another work~\cite{Stiennon-arxiv-2020-learning} trained a summarization  model for optimizing  human preferences in a similar way. 
Based on these prior work, InstructGPT~\cite{Ouyang-arxiv-2022-Training} was proposed in January 2022 to improve the GPT-3 model for human alignment, which formally established a three-stage    \emph{reinforcement learning from human feedback~(RLHF)} algorithm. 
Note that it seems that the wording of ``\emph{instruction tuning}'' has seldom been used in OpenAI's paper and documentation, which is substituted by \emph{supervised fine-tuning on human demonstrations} (\ie the first step of the RLHF algorithm~\cite{Ouyang-arxiv-2022-Training}). 
In addition to improving the instruction following capacity, the RLHF algorithm is particularly useful to mitigate the issues of generating harm or toxic content for LLMs, which is key to the safe deployment of LLMs in practice. 
 OpenAI describes  their approach to alignment research in a technical article~\cite{OpenAI-blog-2022-alignment},  which has summarized three promising directions: ``training AI systems to use human feedback, to  assist human evaluation and to do  alignment research''.  

These enhancement techniques lead to the improved GPT-3 models with stronger capacities, which are called GPT-3.5 models  by OpenAI (see the discussion about the OpenAI API in Section~\ref{subsec-models-apis}).

\ignore{
\begin{figure*}[h]
    \centering
    \includegraphics[width=1\textwidth]{images/LLM pipeline.pdf}
    \caption{A brief summary of important aspects and techniques to develop and use large language models.}
    \label{fig:pipeline}
\end{figure*}
}

\paratitle{The Milestones of Language Models}. 
Based on all the exploration efforts, two major milestones have been achieved by OpenAI, namely  ChatGPT~\cite{OpenAI-blog-2022-ChatGPT} and GPT-4~\cite{OpenAI-OpenAI-2023-GPT-4}, which have largely raised the capacity bar of existing AI systems. 

$\bullet$ \emph{ChatGPT}. In November 2022, OpenAI released the conversation model  ChatGPT, based on the GPT models (GPT-3.5 and GPT-4). As the official blog article introduced~\cite{OpenAI-blog-2022-ChatGPT}, ChatGPT was trained in a similar way as InstructGPT (called ``a sibling model to InstructGPT'' in the original post), while specially optimized for dialogue. 
They reported a difference between the training of ChatGPT and InstructGPT in  the data collection setup: human-generated conversations (playing both the roles of user and AI) are combined with the InstructGPT dataset in a dialogue format for training ChatGPT.  
ChatGPT exhibited superior capacities in communicating with humans: possessing  a vast store of knowledge,  skill at reasoning on mathematical problems, tracing the context accurately in multi-turn dialogues, and aligning well with human values for safe use. Later on, the plugin mechanism has been supported in ChatGPT, which further  extends the capacities of ChatGPT with existing tools or apps.
So far, it seems to be the ever most powerful chatbot in the AI history. The launch of ChatGPT has a significant impact on the AI research in the future, which sheds light on the exploration of  human-like AI systems. 

$\bullet$ \emph{GPT-4}. As another remarkable progress,  GPT-4~\cite{OpenAI-OpenAI-2023-GPT-4} was released in March 2023, which extended the text input to multimodal signals. 
Overall, GPT-4 has  stronger capacities  in solving complex tasks than GPT-3.5, showing a large performance improvement on many evaluation tasks. 
A recent study~\cite{Bubeck-arxiv-2023-Sparks}  investigated the capacities of GPT-4 by conducting qualitative tests with human-generated  problems, spanning a diverse range of difficult  tasks, and showed  that GPT-4 can achieve more superior performance than prior GPT models. 
Furthermore, GPT-4  responds more safely to  malicious or provocative queries, due to a six-month iterative alignment (with an additional  safety reward signal in the RLHF training). 
In the technical report, OpenAI has emphasized how to safely develop GPT-4 and applied  a number of  intervention strategies  to mitigate the possible issues of LLMs, such as hallucinations, privacy and overreliance. For example, they introduced the mechanism called \emph{red teaming}~\cite{Ganguli-arxiv-2022-Red}  to reduce the harm or toxic content generation. 
As another important aspect, GPT-4 has been developed on a well-established deep learning infrastructure with improved optimization methods. They  introduced a new mechanism called \emph{predictable scaling} that can  accurately predict the final performance with a small proportion of compute during model training. 

{
$\bullet$ \emph{GPT-4V, GPT-4 turbo, and beyond}. Based on the work done for GPT-4~\cite{OpenAI-OpenAI-2023-GPT-4}, OpenAI further released GPT-4V in September 2023, 
which focused on the safe deployment of the vision capabilities of GPT-4. In the GPT-4V's  system card~\cite{OpenAI-OpenAI-2023-GPT-4v}, it has extensively discussed the  assessment and mitigation of risks related to visually augmented inputs.
Specially, GPT-4V  exhibited strong vision capacities in various application scenarios, showing the great potential as a powerful multimodal learning  system. 
More recently, in November 2023, OpenAI 
 released an upgraded generation of GPT-4 model at DevDay, named \emph{GPT-4 Turbo}, with a series of  technical improvements. 
 GPT-4 Turbo is featured by the improved model capacity (more capable than GPT-4), the extended knowledge source (up to April 2023), long context window (up to 128k tokens), optimized model performance (cheaper price), and other useful functionality updates (function call, reproducible outputs, etc.). 
 At the same time,  Assistants API was  launched to ease the rapid development of  agent-like assistants. With this API, developers can easily create goal-oriented assistants within their applications, by leveraging specific instruction,  extra knowledge and tool use. Furthermore, multimodal capacities (see, hear, and speak) were also enhanced in this new release, supported by  GPT-4 Turbo with vision, DALL·E 3, Text-to-speech~(TTS), and Listen to voice samples. 
 These improvements  have greatly extended the capacity scope and enhanced the task  performance of GPT models. 
 More importantly, the application ecosystem will be greatly strengthened with the technology upgrade in improved models, APIs, and functionalities.  
 }
 





Despite the huge progress, there are still limitations with these superior LLMs, \eg generating hallucinations with factual errors or potentially risky response within some specific context~\cite{OpenAI-OpenAI-2023-GPT-4}. More limitations or issues of LLMs will be discussed in Section~\ref{sec-evaluation}. 
It poses  long-standing research challenges to develop more capable, safer LLMs. 
From the perspective of engineering, OpenAI has adopted an iterative deployment strategy~\cite{OpenAI-blog-2022-lessons}  to develop the models and products by following a five-stage development and deployment life-cycle, which aims to effectively reduce the potential risks of using the models. 
In the following, we will dive into the technical details in order to have a specific understanding of how they have been  developed.

\begin{figure*}
    \centering
    \includegraphics[width=\textwidth]{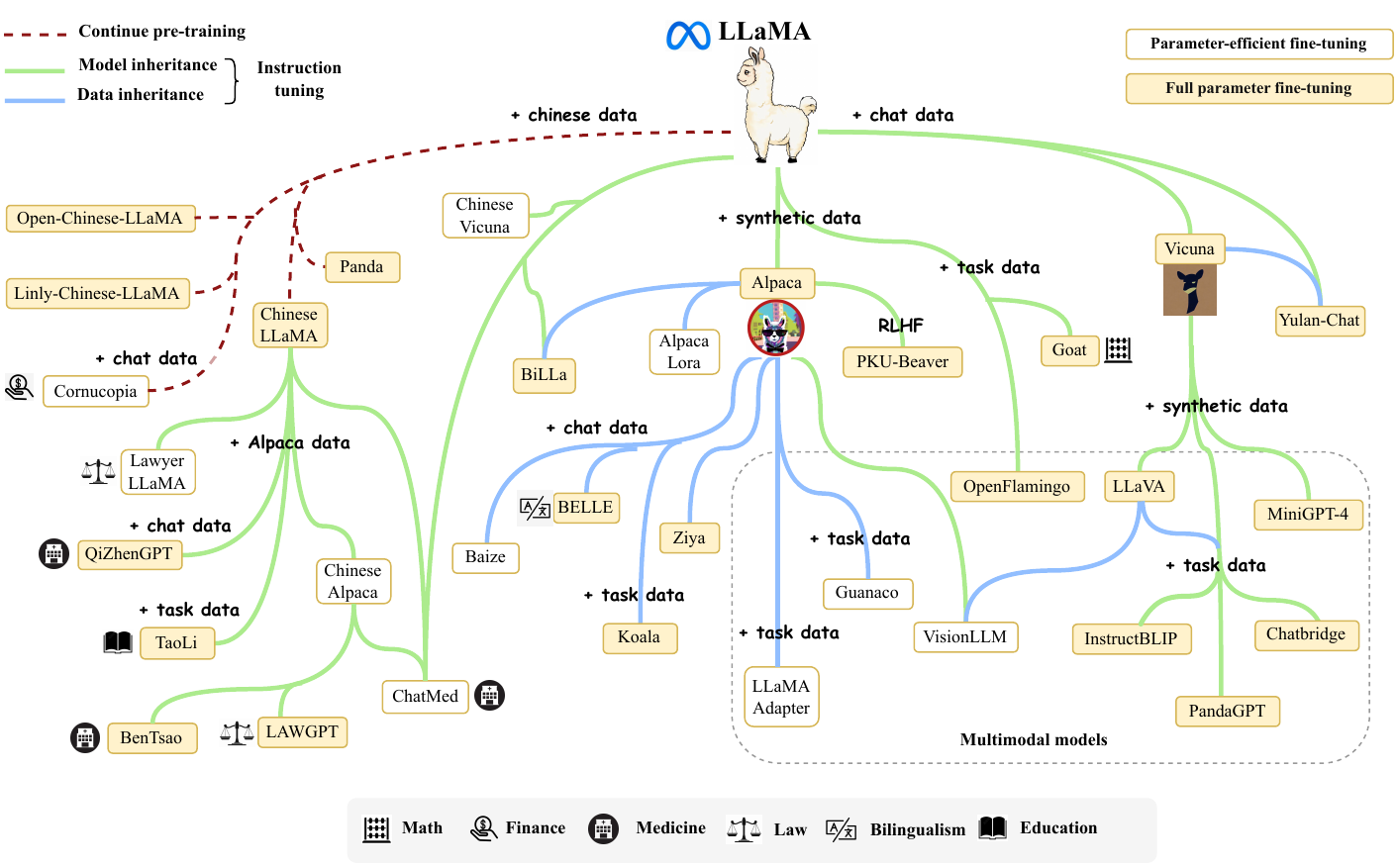}
    \caption{An evolutionary graph of the research  work conducted on LLaMA. Due to the huge number, we cannot include  all the LLaMA variants in this figure, even much excellent work. To support incremental update, we share the source file of this figure, and welcome the readers to include the desired models by submitting the pull requests on our GitHub page. }
    \label{fig:llama_family}
    
\end{figure*}

\section{Resources of LLMs}
\label{sec-resource}
It is by no means an easy job to develop or reproduce LLMs, considering the challenging technical issues and huge demands of computation resources.  
A feasible way is to learn experiences from existing LLMs and reuse publicly available resources for incremental development or experimental study.  In this section, we briefly summarize the publicly available resources for developing LLMs, including model checkpoints (or APIs), corpora and libraries.

\subsection{Publicly Available Model Checkpoints or APIs}\label{subsec-models-apis}
Given the huge cost of model pre-training, well-trained model checkpoints are critical to the study and development of LLMs for the research community. 
Due to space limitation, we can only selectively discuss several representative LLMs. 
In addition, for inference, we can directly employ public APIs to perform our tasks, without running the model 
{locally}.  Next, we introduce the publicly available model checkpoints and APIs. 

\paratitle{Publicly Available Model Checkpoints.} To assist researchers in selecting a suitable model based on the resource budget and usage needs, we focus on discussing the model's parameter size, data and computational resources required for training, the relevant technologies employed by the model, and its performance evaluation in downstream tasks. For more details of LLMs, see Table ~\ref{tab:resource_model}.

{
$\bullet$ \textit{LLaMA.} 
The LLaMA series of models has gained immense popularity and widespread attention due to its openness and effectiveness. From LLaMA~\cite{Touvron-arxiv-2023-LLaMA}, LLaMA-2~\cite{Touvron-2023-llama2-arxiv}, LLaMA-3~\cite{LLaMa3} to LLaMA-3.1~\cite{LLaMA-3.1}, continuous updates have been made and the development is still ongoing.
With increased parameters (the largest version has 405B), more pre-training tokens (15T tokens), and an extended context window (128K), LLaMA-3.1 has significantly enhanced its capabilities, and it also integrates additional components that work in synergy with the model, including new security and safety tools.
In evaluation, LLaMa-3.1~(405B version) achieves competitive performance against prominent closed-source LLMs, such as GPT-4, GPT-4o, and Claude 3.5 Sonnet in various benchmarks~(\eg MMLU, GSM8k, and HumanEval). 
The pre-training of LLaMA~(65B version) involves 2,048 A100-80G GPUs, whereas LLaMA-3.1~(405B version) involves more than 16,000 H100 GPUs.
}

$\bullet$ \textit{Mistral.} 
The Mistral series~\cite{jiang-2023-arxiv-mistral, Jiang-2024-arxiv-Mixtral} consist of Mistral~(7B), Mistral NeMo~(12B), Mistral Large 2~(123B), and Mixtral~(8$\times$7B and 8$\times$22B), which have been widely known for their strong performance on various mainstream benchmarks~(\eg MMLU and GSM8k). 
Mistral NeMo is featured with a long context window of 128K at the parameter scale of 12B.
Although Mistral NeMo is trained with quantization awareness, it enables FP8 inference without sacrificing performance.
Mistral Large 2 is the largest and most powerful model of the Mistral series, which supports 11 natural languages and more than 80 programming languages. Mixtral is a kind of sparse Mixture-of-Experts (SMoE) model that activates only part of the parameters during inference, making it more efficient compared to dense models of the same size.

{
$\bullet$ \textit{Gemma.} Gemma~\cite{Mesnard-2024-arxiv-Gemma, Morgane-2024-arxiv-Gemma2} is a series of lightweight, strong, and open models, consisting of Gemma-1~(2B and 7B) and Gemma-2~(2B, 9B, and 27B). During the pre-training stage, Gemma-2 2B, 9B, and 27B versions are trained on 2T, 8T, and 13T primarily English tokens, respectively.
The largest version of Gemma-2 is trained on 6144 TPUv5p chips.
Gemma-2 has achieved excellent performance in multiple benchmarks~(\eg ARC-c, MMLU, and GSM8k). 
}

{
$\bullet$ \textit{Qwen.} Qwen~\cite{qwen2, qwen2.5} is an open-source large model series consisting of Qwen~(raging from 7B to 72B), Qwen1.5~(raging from 0.5B to 110B), Qwen2~(ranging from 0.5B to 72B), and Qwen2.5~(ranging from 0.5B to 72B).
Qwen2.5 is the newest LLM collection of Qwen, which is pre-trained on up to 18T tokens. Compared to Qwen2, 
Qwen2.5 demonstrates a significant increase in knowledge retention, as well as notable advancements in coding and mathematical abilities.
Qwen2.5 has also shown large improvements in instruction following,  long texts generation (over 8K tokens),  structured data understanding and generation (\eg JSON).
}

{
$\bullet$ \textit{GLM.} GLM~\cite{glm-2024-arxiv-ChatGLM} is a series of LLMs featuring comprehensive capabilities in both English and Chinese. GLM has been upgraded to its fourth-generation model, GLM-4, with a parameter scale of up to 9B, possesses excellent conversational abilities. It has achieved excellent performance in evaluations from multiple perspectives including semantics, mathematics, reasoning, code, and knowledge. In addition to the base model GLM-4-9B, it has open-sourced human preference-aligned model GLM-4-9B-Chat, and long context conversational model GLM-4-9B-Chat-1M.
}

{$\bullet$ \emph{Baichuan}.} Baichuan is a series of open-source bilingual LLMs and the latest version is Baichuan-2. Both Baichuan and Baichuan-2 have two available parameter sizes~(7B and 13B). Baichuan supports both Chinese and English, with pre-training data reaching 1.2 trillion tokens. Furthermore, Baichuan-2 expands its pre-training data to 2.6 trillion tokens. 
Baichuan-2 surpasses Baichuan in all evaluation benchmarks, demonstrating excellent multilingual capabilities and showing potential for vertical applications in the domains such as law and healthcare~(\eg JEC-QA~\cite{Zhong-2024-AAAI-JECQA} and MedQA~\cite{Jin-2021-MedQA}).

\paratitle{LLaMA Model Family}. The collection of LLaMA models~\cite{Touvron-arxiv-2023-LLaMA}  
were introduced by Meta AI in February, 2023, consisting of four sizes (7B, 13B, 30B and 65B).  
Since released, LLaMA has attracted extensive attention from both research and industry communities. 
LLaMA models have achieved very excellent performance on various open benchmarks, which have become the most popular open language models thus far. 
A large number of researchers have extended LLaMA models by either instruction tuning or continual pre-training. In particular, instruction tuning LLaMA has become a major approach to developing customized or specialized models, due to the relatively low computational costs.  
To effectively adapt LLaMA models in non-English languages, it often needs to extend the original vocabulary (trained mainly  on English corpus) or fine-tune it with instructions or data in the target language. 
Among these extended models,  Stanford Alpaca~\cite{Taori-github-2023-Stanford} is the first open instruct-following model fine-tuned based on LLaMA~(7B). It is trained by 52K instruction-following demonstrations generated via self-instruct~\cite{Wang-arXiv-2022-Self} using \texttt{text-davinci-003}. 
The instruction data, named \emph{Alpaca-52K}, and training code have been extensively adopted  in subsequent work, such as Alpaca-LoRA~\cite{Alpaca-LoRA} (a reproduction of Stanford Alpaca using LoRA~\cite{Hu-ICLR-2022-LoRA}), Koala~\cite{koala_blogpost_2023}, and BELLE~\cite{BELLE}. In addition, Vicuna~\cite{vicuna2023} is another popular LLaMA variant,   trained upon user-shared conversations collected from ShareGPT~\cite{ShareGPT}. 
Due to the excellent  performance and availability of the LLaMA model family, many multimodal models incorporate them as the base language models, to achieve strong language understanding and generation abilities. 
Compared with other variants, Vicuna is more preferred in multimodal language models, which have led to the emergence of a variety of popular models, including  LLaVA~\cite{Liu-arxiv-2023-Visual}, MiniGPT-4~\cite{Zhu-arxiv-2023-MiniGPT-4}, InstructBLIP~\cite{Dai-2023-arxiv-InstructBLIP}, and PandaGPT~\cite{su-2023-arxiv-pandagpt}. 
The release of LLaMA has greatly advanced the research progress of LLMs. 
{To  summarize the research work conducted on  LLaMA, we present a brief evolutionary graph in Figure~\ref{fig:llama_family}. 
}

\ignore{
The openness and effectiveness of LLaMA~\cite{Touvron-arxiv-2023-LLaMA} have generated considerable interest within the research community. With the aim of implementing novel models or tools, multiple endeavors have been directed towards fine-tuning and continuously pre-training its diverse model versions (7B, 13B, 30B, and 65B). The most common training method is instructing tuning, which enables the model to better follow instructions. Stanford Alpaca~\cite{Taori-github-2023-Stanford} is the first open instruct-following model fine-tuned from LLaMA~(7B). It is trained on 52K instruction-following demonstrations generated in the style of self-instruct~\cite{Wang-arXiv-2022-Self} using text-davinci-003. The instruction data, named Alpaca-52k, and training code have been extensively applied in other subsequent models, such as Alpaca-LoRA~\cite{Alpaca-LoRA} (a reproduction of Stanford Alpaca using LoRA~\cite{Hu-ICLR-2022-LoRA}), Koala~\cite{koala_blogpost_2023}, and BELLE~\cite{BELLE}. In addition, Vicuna~\cite{vicuna2023} is trained upon user-shared conversations collected from ShareGPT~\cite{ShareGPT}. Due to the outstanding performance of the LLaMA family models, many multimodal models also them as their base models for training. Actually, when it comes to selecting a base model for multimodal models, Vicuna is frequently preferred. LLaVA~\cite{Liu-arxiv-2023-Visual}, MiniGPT-4~\cite{Zhu-arxiv-2023-MiniGPT-4}, InstructBLIP~\cite{Dai-2023-arxiv-InstructBLIP}, and PandaGPT~\cite{su-2023-arxiv-pandagpt}) are all trained upon it. }

\paratitle{Public API of LLMs}.
\label{sec:apis_for_llms}
Instead of directly using the model copies, APIs provide a more convenient way for common users to use LLMs, without the need of running the model locally. As a representative interface for using LLMs, the APIs for the GPT-series models~\cite{Brown-NeurIPS-2020-Language, Chen-arxiv-2021-evaluating, Ouyang-arxiv-2022-Training, OpenAI-OpenAI-2023-GPT-4} have been widely used for both academia and industry\footnote{{https://platform.openai.com/docs/api-reference/introduction}}. OpenAI has provided  {seven major interfaces to the models in GPT-3 series: \texttt{ada}, \texttt{babbage}, \texttt{curie}, \texttt{davinci} (the most powerful version in GPT-3 series), \texttt{text-ada-001}, \texttt{text-babbage-001}, and \texttt{text-curie-001}.}  Among them, the first four interfaces can be further fine-tuned on the host server of OpenAI.  
In particular, \texttt{babbage}, \texttt{curie}, and \texttt{davinci} correspond to the GPT-3~(1B), GPT-3~(6.7B), and GPT-3~(175B) models, respectively~\cite{Brown-NeurIPS-2020-Language}. 
In addition, there are also two APIs related to Codex~\cite{Chen-arxiv-2021-evaluating}, called \texttt{code-cushman-001} (a powerful and multilingual version of the Codex~(12B)~\cite{Chen-arxiv-2021-evaluating}) and \texttt{code-davinci-002}.
Further, GPT-3.5 series include one base model \texttt{code-davinci-002} and  three enhanced versions, namely \texttt{text-davinci-002}, \texttt{text-davinci-003}, and \texttt{gpt-3.5-turbo}.
As more powerful alternatives, in this year, OpenAI has released the model interfaces for GPT-4 series, including \texttt{gpt-4}, \texttt{gpt-4-32k}, \texttt{gpt-4-1106-preview}~(\ie GPT-4 Turbo) and \texttt{gpt-4-vision-preview}~(\ie GPT-4 Turbo with vision, a multimodal model). 
It is worth noting that OpenAI has been  maintaining and upgrading these model interfaces (\texttt{gpt-3.5-turbo}, \texttt{gpt-4}, \texttt{gpt-4-32k}), so the API name will actually point to the latest version. 
Currently, ChatGPT can be  powered by either GPT-3.5 or GPT-4 models.  Overall, one select the suitable model interface based on the specific application scenarios and response requirements.
The detailed usage can be found on their project websites\footnote{https://platform.openai.com/docs/models/overview}.

\begin{table}[htbp]
    \centering
    \caption{Statistics of commonly-used data sources. }
    \label{tab:corpora}
    \footnotesize
    \renewcommand\tabcolsep{2.5pt}
    \begin{tabular}{llcc}
    \toprule
    {\textbf{Corpora}} & {\textbf{Size}} &  {\textbf{Source}} & {\textbf{Latest Update Time}} \\
    \midrule
    BookCorpus~\cite{Zhu-ICCV-2015-Aligning} & 5GB   & Books    & Dec-2015 \\
    Gutenberg~\cite{Gutenberg}   & -    & Books    & Dec-2021 \\ 
    C4~\cite{Raffel-JMLR-2020-Exploring}  & 800GB    & CommonCrawl  & Apr-2019 \\
    CC-Stories-R~\cite{Trinh-CoRR-2018-A}  & 31GB    & CommonCrawl  & Sep-2019 \\
    CC-NEWS~\cite{Liu-CoRR-2019-RoBERTa} & 78GB  & CommonCrawl  & Feb-2019 \\
    REALNEWs~\cite{Zellers-NeurIPS-2019-Defending}    & 120GB & CommonCrawl  & Apr-2019 \\
    OpenWebText~\cite{Gokaslan2019OpenWeb} & 38GB  & Reddit links & Mar-2023 \\
    Pushift.io~\cite{Baumgartner-AAAI-2020-The}  & 2TB    & Reddit links & Mar-2023 \\
    Wikipedia~\cite{Wikipedia}   & 21GB    & Wikipedia    & Mar-2023 \\
    BigQuery~\cite{bigquery-google}    & -    & Codes    & Mar-2023 \\
    the Pile~\cite{Gao-arxiv-2021-Pile} & 800GB & Other    & Dec-2020 \\
    ROOTS~\cite{Laurencon-NIPS-2022-The}   & 1.6TB & Other    & Jun-2022 \\ 
    \bottomrule
    \end{tabular}
    \label{tab:methods}
\end{table}

\subsection{Commonly Used Corpora for Pre-training}
\label{sec:commonly_used_corpora}
{In contrast to earlier PLMs, LLMs which consist of a significantly larger number of parameters require a higher volume of training data that covers a broad range of content. For this need, there are increasingly more accessible training datasets that have been released for research.   
In this section, we will briefly summarize several widely used corpora for training LLMs. } 
Based on their content types, we categorize these corpora into five groups: 
{web pages, books, Wikipedia, code, and others.}

\paratitle{Web pages.} Web pages are a primary data source for training language models. 

$\bullet$ \textit{CommonCrawl.} CommonCrawl~\cite{commoncrawl} is one of the largest open-source web crawling databases, containing a petabyte-scale data volume, which has been widely used as training data for existing LLMs.
As the whole dataset is very large, existing studies mainly extract subsets of web pages from it within a specific period or specific needs (\eg extracting mathematical texts).
However, due to the widespread existence of noisy and low-quality information in web page data, it is necessary to perform data preprocessing before usage.
{One commonly used toolkit for cleaning CommonCrawl is CC-Net~\cite{Wenzek-2020-ccnet}, which is 
developed by Facebook and has been used in processing datasets like RedPajama-Data~\cite{Together-2023-Redpajama}.}

$\bullet$ \textit{C4.} {The Colossal Clean Crawled Corpus (C4) includes five {variants}\footnote{https://www.tensorflow.org/datasets/catalog/c4 }, namely en~(806G), en.noclean~(6T), realnewslike~(36G), webtextlike~(17G), and multilingual~(38T). The \emph{en} version has been utilized for pre-training T5~\cite{Raffel-JMLR-2020-Exploring}, LaMDA~\cite{Thoppilan-CoRR-2022-LaMDA}, Gopher~\cite{Rae-arxiv-2021-Scaling}, and UL2~\cite{Tay-arxiv-2022-UL2}. The multilingual C4, also called mC4, has been used in mT5~\cite{Xue-NAACL-2021-mT5}.}

$\bullet$ \textit{RedPajama-Data.} {RedPajama-Data~\cite{Together-2023-Redpajama}} is a publicly available comprehensive web dataset, comprising 100 billion documents from Common Crawl. It has been cleaned, filtered, and deduplicated using the CCNet tool, resulting in approximately 30T tokens, which is available for download on Hugging Face. RedPajama-Data is a multilingual dataset that includes five languages: English, French, Spanish, German, and Italian. Additionally, it offers over 40 quality labels, making it feasible to filter or reweight the dataset according to specific criteria. 
The dataset is continuously updated and maintained, with all data processing scripts open-sourced on GitHub for  convenient use.

$\bullet$ \textit{RefinedWeb.} 
{RefinedWeb~\cite{Penedo-2023-arxiv-Refinedweb}} is a web dataset obtained through rigorous selection and deduplication based on data from Common Crawl, encompassing all Common Crawl web records from 2008 to June 2023, totaling around 5T tokens. The open-source portion consists of 600B tokens, with a data size of approximately 500GB. After decompression, it requires 2.8TB of local storage space and is available for download on Hugging Face. This dataset serves as the primary training dataset for the open-source large language model Falcon.

$\bullet$ \textit{WebText.} {WebText~\cite{radford-blog-2019-language} is a well-known corpus composed of highly upvoted links from Reddit, a social media platform that enables users to submit links and text posts, but it is not publicly available.
As a surrogate, there is a readily accessible open-source alternative called OpenWebText~\cite{Gokaslan2019OpenWeb}.}

\paratitle{Books \& Academic Data.} Books and academic data contains a wealth of world knowledge and linguistic information, serving as a high-quality corpus for model learning. 

$\bullet \textit{ Book Data.}$ BookCorpus~\cite{Zhu-ICCV-2015-Aligning} is a commonly used dataset in previous small-scale models (\eg GPT~\cite{radford-openai-2018-improving} and GPT-2~\cite{radford-blog-2019-language}), consisting of over 11,000 books covering a wide range of topics and genres (\eg novels and biographies).
Another large-scale book corpus is Project Gutenberg~\cite{Gutenberg}, consisting of over 70,000 literary books including novels, essays, poetry, drama, history, science, philosophy, and other types of works in the public domain. It is currently one of the largest open-source book collections, which is used in training of MT-NLG~\cite{Smith-CoRR-2022-Using} and LLaMA~\cite{Touvron-arxiv-2023-LLaMA}. As for Books1~\cite{Brown-NeurIPS-2020-Language} and Books2~\cite{Brown-NeurIPS-2020-Language} used in GPT-3~\cite{Brown-NeurIPS-2020-Language}, they are much larger than BookCorpus but have not been publicly released so far. 

$\bullet$ \textit{ Academic Data. }
In addition to book data, scientific publication data such as paper is also important for model pre-training.  arXiv Dataset~\cite{clement-arxiv-2019-ontheuse} is a corpus of 1.7 million academic papers, covering a wide range of papers in the fields of physics, mathematics, and computer science.
S2ORC~\cite{lo-2020-ACL-s2orc} is a corpora that consists of 136M academic papers {collected by 
Semantic Scholar. It also releases a derivative dataset peS2o~\cite{peS2o}, which contains about 42B tokens. 
}

\paratitle{Wikipedia.} 
Wikipedia~\cite{Wikipedia} is an online encyclopedia containing a large volume of high-quality articles on diverse topics.
{Most of these articles are composed in an expository style of writing (with supporting references), covering a wide range of languages and fields. }
Typically, the English-only filtered versions of Wikipedia are widely used in most LLMs (\eg GPT-3~\cite{Brown-NeurIPS-2020-Language}, LaMDA~\cite{Thoppilan-CoRR-2022-LaMDA}, and LLaMA~\cite{Touvron-arxiv-2023-LLaMA}). Wikipedia is available in multiple languages,  so it can be used in multilingual settings.

\paratitle{Code.} To collect code data, existing work mainly crawls open-source licensed codes from the Internet.
Two major sources are public code repositories  under open-source licenses (\eg  GitHub) and code-related question-answering platforms (\eg StackOverflow).
{Google has publicly released the BigQuery dataset~\cite{bigquery-google}, which includes a substantial number of open-source licensed code snippets in various programming languages, serving as a representative code dataset. CodeGen has utilized BIGQUERY~\cite{nijkamp-arxiv-2022-Codegen}, a subset of the BigQuery dataset, for training the multilingual version of CodeGen (CodeGen-Multi).}
{In addition, Hugging Face has collected and released a code dataset named The Stack~\cite{kocetkov-arxiv-2022-thestack}, covering more than 30 programming languages. The Stack is continuously updated, and the v1.2 version has expanded to 358 programming languages. Based on this dataset, BigCode further { processed it} and released StarCoder~\cite{Li-2023-arxiv-Starcoder}, which is also the pre-training data of the model StarCoder.
}

\paratitle{Mixed Data.} In addition to the aforementioned specific types of datasets, different types of data have been combined to facilitate usage by researchers. 
The Pile~\cite{Gao-arxiv-2021-Pile} is a large-scale, diverse, and open-source text dataset consisting of over 800GB of data from multiple sources, including books, websites, codes, scientific papers, and social media platforms. It is constructed from 22 diverse high-quality subsets.
The Pile dataset is widely used in models with different parameter scales, such as GPT-J~(6B)~\cite{Wang-GitHub-2021-GPT-J}, CodeGen~(16B)~\cite{nijkamp-arxiv-2022-Codegen}, and Megatron-Turing NLG~(530B)~\cite{Smith-CoRR-2022-Using}. 
ROOTS~\cite{Laurencon-NIPS-2022-The} is composed of various smaller datasets (totally 1.61 TB of text) and covers 59 different languages (containing natural languages and programming languages), which have been used for training BLOOM~\cite{Scao-arxiv-2022-BLOOM}.
{Another mixture dataset is Dolma~\cite{Soldani-arxiv-2024-dolma}, which includes web text from Common Crawl, academic papers from Semantic Scholar, GitHub code, books, social media from Reddit, and Wikipedia data. Dolma consisting of 3T tokens of approximately 200TB of raw text and has been used to train OLMo~\cite{groeneveld-2024-arxiv-olmo}.
}

In practice, it commonly requires a mixture of different data sources for pre-training LLMs (see Figure~\ref{fig:source-ratio}), instead of a single corpus.
{Therefore, existing studies commonly mix several ready-made datasets (\eg C4, OpenWebText, and the Pile), and then perform further processing to obtain the pre-training corpus.
Furthermore, to train the LLMs that are adaptive to specific applications, it is also important to extract  data from  relevant sources (\eg Wikipedia and BigQuery) for enriching the corresponding information in pre-training data.} 




\begin{table}[h]
    \centering
    \caption{A detailed list of available collections for instruction tuning. %
    }
    \footnotesize
    \renewcommand\tabcolsep{2.5pt}
    \begin{tabular}{clcccc}
    \toprule
    
    \textbf{Categories} & \textbf{Collections} & \textbf{Time} & \textbf{\#Examples} \\
    \midrule
    \multirow{7}{*}{Task}
    & Nat. Inst.~\cite{Mishra-ACL-2022-Cross} & Apr-2021 & 193K \\
    & FLAN~\cite{Wei-ICLR-2022-Finetuned} & Sep-2021 & 4.4M \\
    & P3~\cite{Bach-ACL-2022-PromptSource} & Oct-2021 & 12.1M \\
    & Super Nat. Inst.~\cite{Wang-EMNLP-2022-Super} & Apr-2022 & 5M \\
    & MVPCorpus~\cite{Tang-arxiv-2022-MVP} & Jun-2022 & 41M \\
    & xP3~\cite{Muennighoff-2022-arxiv-Crosslingual} & Nov-2022 & 81M \\
    & OIG\cite{Nguyen-laion-2023-The} & Mar-2023 & 43M \\
    \midrule
    \multirow{5}{*}{Chat}
    & HH-RLHF~\cite{Bai-arxiv-2022-Training} & Apr-2022 & 160K \\
    & HC3~\cite{guo-arxiv-2023-how} & Jan-2023 & 87K \\
    & ShareGPT~\cite{ShareGPT} & Mar-2023 & 90K \\
    & Dolly~\cite{Conover-2023-arxiv-Dolly} & Apr-2023 & 15K \\
    & OpenAssistant~\cite{kopf-arxiv-2023-openassistant} & Apr-2023 & 161K \\
    \midrule
    \multirow{5}{*}{Synthetic}
    & Self-Instruct~\cite{Wang-arXiv-2022-Self} & Dec-2022 & 82K \\
    & Alpaca~\cite{alpaca} & Mar-2023 & 52K \\
    & Guanaco~\cite{Cheung-2023-Guanaco} & Mar-2023 & 535K \\
    & Baize~\cite{xu-arxiv-2023-baize} & Apr-2023 & 158K \\
    & BELLE~\cite{ji-arxiv-2023-towards} & Apr-2023 & 1.5M \\
    \bottomrule
    \end{tabular}
    \label{tab:instruction-collection}
\end{table}

\begin{table}[h]
    \centering
    \caption{A list of available collections for alignment. %
    }\label{tab:rlhf-datasets}
    \footnotesize
    \renewcommand\tabcolsep{2.5pt}
    \begin{tabular}{lccc}
    \toprule

    \textbf{Dataset} & \textbf{Release Time} & \textbf{\#Examples} \\
    \midrule
Summarize from Feedback~\cite{Stiennon-arxiv-2020-learning}  & Sep-2020 & 193K \\
SHP~\cite{Ethayarajh-ICLM-2022-Understanding}  & Oct-2021 & 385K \\
WebGPT Comparisons~\cite{Nakano-arxiv-2021-WebGPT}    & Dec-2021 & 19K  \\
Stack Exchange Preferences~\cite{Lambert-2023-StackH4}  & Dec-2021 & 10M  \\
HH-RLHF~\cite{Bai-arxiv-2022-Training}         & Apr-2022 & 169K \\
Sandbox Alignment Data~\cite{Liu-arxiv-2023-training}      & May-2023 & 169K \\
CValues~\cite{Xu-2023-arxiv-CValues}  & Jul-2023 & 145K \\
PKU-SafeRLHF~\cite{Dai-arxiv-2023-SafeRLHF}    & Oct-2023 & 330K \\  
    \bottomrule
    \end{tabular}
    \label{tab:alignment-collection}
\end{table}

\subsection{Commonly Used Datasets for Fine-tuning}
\label{sec:commonly_used_fituning}

{After pre-training, 
it requires further fine-tuning LLMs to enhance the model capacity, which often involve two major steps, namely instruction tuning (supervised fine-tuning) and alignment tuning. In this section, we mainly focus on discussing the related available  datasets for the two kinds of tuning approaches, and more algorithm details can be found in Section~\ref{sec-adaptation}.  

\subsubsection{Instruction Tuning Datasets}
\label{sec:it-dataset}

{
After pre-training, instruction tuning (\aka supervised fine-tuning) is an important method to enhance or unlock specific abilities of LLMs (\eg instruction following). In this part, we introduce several widely used datasets for instruction tuning, and categorize them into three main types based on the construction method of formatted instruction instances, namely NLP task datasets, daily chat datasets and synthetic datasets. 
We show their details in Table~\ref{tab:instruction-collection}.}

\paratitle{NLP Task Datasets.}  
{This kind of datasets are formatted based on collected NLP task datasets~(\eg text classification and summarization) with corresponding natural language task descriptions. In this category, P3~\cite{Sanh-2022-ICLR-P3} and FLAN~\cite{Wei-ICLR-2022-Finetuned, Longpre-2023-arxiv-Flan_v2} are two widely used datasets for instruction tuning. }

$\bullet$ \emph{P3}~\cite{Sanh-2022-ICLR-P3} is composed of 170 English NLP datasets and 2,052 English prompt templates, where the input and output of each data example have been formatted with specific prompt templates for composing the training instance.

$\bullet$ \emph{FLAN}{~\cite{Wei-ICLR-2022-Finetuned} consists of 62 widely used NLP benchmarks in its original version. Recently, FLAN-v2~\cite{Longpre-2023-arxiv-Flan_v2} is also proposed, which expands FLAN by mixing additional instruction datasets, including  Muffin~\cite{Wei-ICLR-2022-Finetuned}, NIV2~\cite{Wang-EMNLP-2022-Super}, T0-SF~\cite{Sanh-ICLR-2022-Multitask}, and CoT~\cite{Cobbe-arxiv-2021-Training, Geva-tacl-2021-Did, Camburu-2020-ACL-Make}. Muffin contains 62 tasks from the original FLAN and additional 26 tasks, including conversation and code synthesis tasks. T0-SF is extracted from T0~\cite{Sanh-ICLR-2022-Multitask} while ensuring no overlap with Muffin. NIV2 refers to the Natural-Instructions v2 dataset~\cite{Wang-EMNLP-2022-Super}, and CoT~\cite{Cobbe-arxiv-2021-Training, Geva-tacl-2021-Did, Camburu-2020-ACL-Make} is a combination of nine reasoning tasks with corresponding chain-of-thought prompts and outputs. }

\paratitle{Daily Chat Datasets.} 
{This kind of datasets are constructed based on real user conversations where queries are posed by humans and responses are mainly  generated by human labelers or LLMs~(\eg ChatGPT, GPT-4). The conversation types include open-ended generation, question answering, brainstorming, and chatting. In this category,  ShareGPT~\cite{ShareGPT}, OpenAssistant~\cite{kopf-arxiv-2023-openassistant} and Dolly~\cite{Conover-2023-arxiv-Dolly} are three commonly used datasets for LLM fine-tuning.}

$\bullet$ \emph{ShareGPT}{~\cite{ShareGPT} is collected from a data collection platform where users can upload their conversations with ChatGPT or GPT-4 through the ShareGPT API. Currently, this dataset consists of approximately 90,000 conversations, including real instructions or inquiries from human and responses from ChatGPT.} 

$\bullet$ \emph{OpenAssistant}{~\cite{kopf-arxiv-2023-openassistant} is a multilingual corpus containing 66,497 real-world conversation trees between human and AI assistant. Each conversation tree consists of multiple nodes, and each node represents the information generated by a role in the dialogue.
It spans 35 languages and includes 461,292 manually annotated quality ratings of responses. }

$\bullet$ \emph{Dolly}{~\cite{Conover-2023-arxiv-Dolly} is an English dataset comprising 15,000 human-generated data instances (prompt-response pairs) from Databricks.  
This dataset covers seven domains outlined in the InstructGPT~\cite{Ouyang-arxiv-2022-Training}, including brainstorming, classification, closed-book quality assurance, generation, information extraction, open-book quality assurance, and summarization. }

\paratitle{Synthetic Datasets.} {This kind of datasets are typically constructed  by instructing  LLMs, based on pre-defined guidance rules or methods.
In this category, 
Self-Instruct-52K~\cite{Wang-arXiv-2022-Self}, Alpaca~\cite{Taori-github-2023-Stanford} and Baize~\cite{xu-arxiv-2023-baize} are three commonly used synthetic datasets for LLMs.}

$\bullet$ \emph{Self-Instruct-52K}~\cite{Wang-arXiv-2022-Self} is an instruction dataset generated through the self-instruct~\cite{Wang-arXiv-2022-Self} method, consisting of 82,000 instances with 52,000 instructions. Concretely, the authors construct 175 seed instances, and then iteratively prompt the LLM~\cite{Brown-NeurIPS-2020-Language} to synthesize additional instructions based on randomly selected 8 instructions as reference. 
Subsequently, the LLM is further instructed  to generate instance inputs and their corresponding outputs based on the synthetic instructions, and finally obtain the Self-Instruct-52K dataset. 

$\bullet$ \emph{Alpaca}~\cite{Taori-github-2023-Stanford} is also a synthetic dataset based on the self-instruct~\cite{Wang-arXiv-2022-Self} method. It utilizes the \texttt{text-davinci-003} model on the 175 seed datasets from Self-Instruct-52K to obtain 52,000 new  instructions and corresponding inputs and outputs. Moreover,  
60\% of the examples are pure instructions without the input part in the final dataset.

$\bullet$ \emph{Baize}~\cite{xu-arxiv-2023-baize} is an English multi-turn conversation corpus constructed using ChatGPT, comprising 111.5K instances. To create Baize, a method called ``self-chat"~\cite{xu-arxiv-2023-baize} is purposed, where ChatGPT takes on the roles of both the user and the AI assistant in turns, generating information in a conversational format. 

\subsubsection{Alignment Datasets}
\label{sec:commonly_used_aligntuning}

Apart from instruction tuning, it is important to construct high-quality datasets for aligning LLMs with   human values and preferences~(\eg helpfulness, honesty, and harmlessness). In this section, we introduce several widely used datasets for alignment tuning, including  HH-RLHF~\cite{Bai-arxiv-2022-Training}, SHP~\cite{Ethayarajh-ICLM-2022-Understanding}, PKU-SafeRLHF~\cite{Dai-arxiv-2023-SafeRLHF}, Stack Exchange Preferences~\cite{Lambert-2023-StackH4} and Sandbox Alignment Data~\cite{Liu-arxiv-2023-training}. We show their details in Table~\ref{tab:rlhf-datasets}.

$\bullet$  \textbf{HH-RLHF}~\cite{Bai-arxiv-2022-Training} consists of around 169K instances, and can be divided into two parts that focus on the helpfulness and harmlessness of LLMs, respectively.
Each instance is an open-ended conversation between a crowdworker and a chat model, about seeking assistance, advice, or task completion.
The chat model provides two responses  to each user query, and the more helpful or harmful responses will be chosen as the annotations.

$\bullet$  {\textbf{SHP}}{~\cite{Ethayarajh-ICLM-2022-Understanding} focuses on the helpfulness of responses. It comprises 385K collective human preferences over responses to questions/instructions across 18 diverse subject areas, spanning topics from cooking to legal advice. Each instance is a Reddit post containing a question or instruction and a pair of top-level comments, one of which is deemed as more preferable by Reddit users and the other one is deemed as less helpful. 
Different from HH-RLHF~\cite{Bai-arxiv-2022-Training}, the data in SHP consists of naturally occurring and human-written responses.
 }

$\bullet$  {\textbf{PKU-SafeRLHF}}{~\cite{Dai-arxiv-2023-SafeRLHF} encompasses more than 330K instances of expert comparison data, concentrating on the helpfulness and harmlessness. Each instance in the dataset includes a question and two responses, accompanied by safety labels for each response and two preference annotations between the two responses according to helpfulness and harmlessness.
The harmlessness of a response indicates its classification as risk-neutral across all 14 harm categories, while the helpfulness of a response is evaluated based on its effectiveness in addressing the question. 
 }

$\bullet$  {\textbf{Stack Exchange Preferences}}{~\cite{Lambert-2023-StackH4} focuses on the helpfulness of answers. It comprises about 10M questions and answers from Stack Overflow. Each instance consists of a question and more than two corresponding answers. Each answer is annotated with a score calculated based on its votes and a label denoting whether it is selected. 
 }


$\bullet$  {\textbf{Sandbox Alignment Data}}{~\cite{Liu-arxiv-2023-training} is an alignment dataset containing feedback from LLMs rather than human. It comes from a virtual interaction environment called SANDBOX, where the model simulates social interactions with other models and revise responses according to the feedback from other models. The dataset contains 169K instances, and each instance consists of a societal query, several responses, and corresponding ratings from other models.
 }

 }

\subsection{Library Resource}
\label{sec:library}

In this part, we briefly introduce a series of available libraries for developing LLMs. 

$\bullet$ \textbf{Transformers}~\cite{Wolf-EMNLP-2020-Transformers} is an open-source Python library for building models using the Transformer architecture,  which is developed and maintained by Hugging Face. 
It has a simple and user-friendly API, {making it easy to use and customize various pre-trained models.} It is a powerful library with a large and active community of users and developers who regularly update and improve the models and algorithms.

{$\bullet$ \textbf{DeepSpeed}~\cite{Rasley-KDD-2020-DeepSpeed} is a deep learning optimization library (compatible with PyTorch) developed by Microsoft, which has been used to train a number of LLMs, such as MT-NLG~\cite{Smith-CoRR-2022-Using} and BLOOM~\cite{Scao-arxiv-2022-BLOOM}.} It provides the support of various  optimization techniques for distributed training, such as memory optimization (ZeRO technique,  gradient checkpointing), and pipeline parallelism.

$\bullet$ \textbf{Megatron-LM}~\cite{Shoeybi-arXiv-2019-Megatron, Narayanan-ACM-2021-Efficient, Korthikanti-arxiv-2022-reducing} is a deep learning library developed by NVIDIA for training large-scale language models. It also provides rich optimization techniques for distributed training, {including  model and data parallelism, mixed-precision training, and FlashAttention.} These optimization techniques can largely improve the training efficiency and speed,  enabling efficient distributed training across GPUs.

$\bullet$ \textbf{JAX}~\cite{Bradbury-github-2018-jax} is a Python library for high-performance machine learning algorithms developed by Google, allowing users to easily perform computations on arrays with hardware acceleration (\eg GPU or TPU).   
{It enables  efficient computation on various devices and also supports  several featured functions, such as automatic differentiation and just-in-time compilation. 
}

{$\bullet$ \textbf{Colossal-AI}~\cite{Bian-CoRR-2021-Colossal-AI} is a deep learning library developed by HPC-AI Tech for training large-scale AI models. It is implemented based on PyTorch and supports a rich collection of  parallel training  strategies.
Furthermore, it can also optimize heterogeneous memory management with methods proposed by PatrickStar~\cite{Fang-arxiv-2021-PatrickStar}.
{Recently, a ChatGPT-like model called ColossalChat~\cite{ColossalChat} has been publicly released with two versions (7B and 13B), which are developed 
using Colossal-AI based on LLaMA~\cite{Touvron-arxiv-2023-LLaMA}.}}

$\bullet$  {\textbf{BMTrain}~\cite{BMTrain} is an efficient library developed by OpenBMB for training models with large-scale parameters in a distributed manner, which emphasizes code simplicity, low resource, and high availability. BMTrain has already incorporated several common LLMs (\eg Flan-T5~\cite{Chung-arxiv-2022-Scaling} and GLM~\cite{Zeng-arxiv-2022-GLM}) into its ModelCenter, where developers can use these models directly.}

$\bullet$ {\textbf{FastMoE}~\cite{He-arXiv-2021-FastMoE} is a specialized training library for MoE (\ie mixture-of-experts) models. It is developed {based on} PyTorch, prioritizing both efficiency and user-friendliness in its design. FastMoE simplifies the process of transferring Transformer models to MoE models and supports both data parallelism and model parallelism during training. }

{$\bullet$ {\textbf{vLLM}~\cite{kwon-2023-SIGOPS-efficient} is a fast, memory efficient, and easy-to-use library for LLM inference and serving. 
To enable fast  inference, it is specially optimized with high serving throughput, effective attention memory management using PagedAttention~\cite{kwon-2023-SIGOPS-efficient}, continuous batching, and optimized CUDA kernels.
Furthermore, vLLM also supports  various decoding algorithms, tensor parallelism and streaming outputs.
To ease the integration with other systems, vLLM is friendly to the use of HuggingFace models, and also provide OpenAI-compatible API servers. }
}

{$\bullet$ {\textbf{DeepSpeed-MII}~\cite{DeepSpeed-MII} is also a memory efficient Python library developed by DeepSpeed~\cite{Rasley-KDD-2020-DeepSpeed}. It aims to democratize LLMs inference by prioritizing high throughput, low latency, and cost-effectiveness. DeepSpeed-MII achieves accelerated text generation inference by leveraging four essential technologies: blocked KV caching, continuous batching, dynamic SplitFuse, and high-performance CUDA Kernels. It currently supports over 13,000 models across three popular model architectures, such as LLaMA~\cite{Touvron-arxiv-2023-LLaMA}, Mistral~\cite{jiang-2023-arxiv-mistral}, and OPT~\cite{Zhang-arxiv-2022-OPT}. }
}

{$\bullet$ {\textbf{DeepSpeed-Chat}~\cite{yao-2023-arxiv-dschat}} is a fast, cost-effective, and easy-to-use system framework that enables the integration of the complete RLHF process during model training. It is featured by three major  functionalities: (1) it simplifies the training and inference process for ChatGPT-like models, enabling using a simple script to implement multiple training or inference steps;
(2) it replicates the training mode of InstructGPT~\cite{Ouyang-arxiv-2022-Training} and provides a complete pipeline for three training steps~(\ie SFT, reward model fine-tuning, and RLHF); (3) it integrates the training engine and inference engine of Deepspeed into a unified hybrid engine~(Deepspeed HE) for RLHF training, which enables seamless switch between training and inference modes, and leveraging various optimizations from DeepSpeed Inference. 
}

{
In addition to the above library resources,  existing  deep learning  frameworks {(\eg PyTorch~\cite{Paszke-NeurIPS-2019-Pytorch}, TensorFlow~\cite{Abadi-OSDI-2016-TensorFlow}, MXNet~\cite{Chen-arxiv-2015-MXNet}, PaddlePaddle~\cite{Ma-fodc-2019-PaddlePaddle}, MindSpore~\cite{Huawei-Springer-2022-MindSpore}  and OneFlow~\cite{Yuan-arXiv-2021-OneFlow}) have also provided the support for parallel algorithms, which are commonly used for training large-scale models.  
}
}
\begin{figure*}
    \centering
    \includegraphics[width=\textwidth]{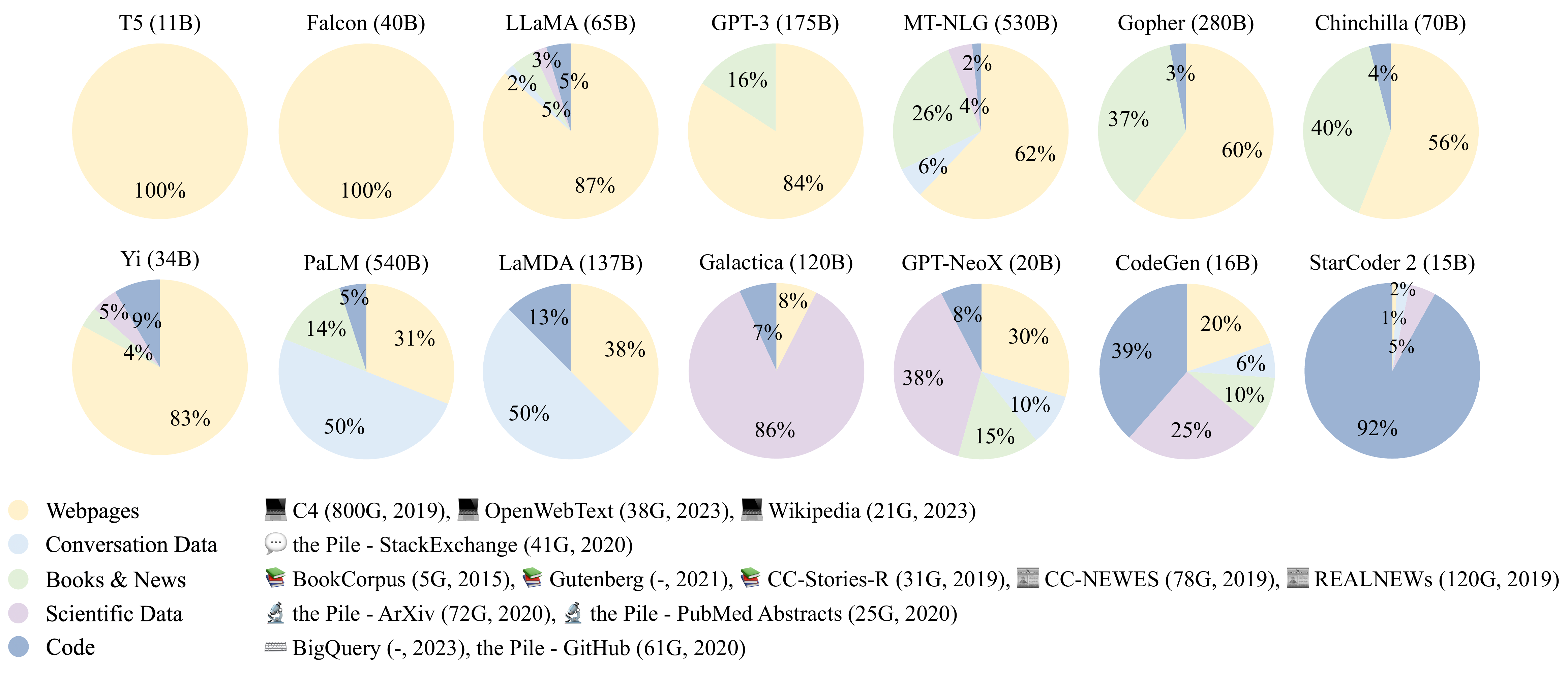}
    \caption{Ratios of various data sources in the pre-training data for existing  LLMs.}
    \label{fig:source-ratio}
\end{figure*}

\section{Pre-training}
\label{sec-pretraining}

Pre-training establishes the basis of the abilities of LLMs. By pre-training on large-scale corpora, LLMs can acquire essential language understanding and %
{generation} skills~\cite{Brown-NeurIPS-2020-Language,Chowdhery-arxiv-2022-PaLM}.
In this process, the scale and quality of the pre-training corpus are critical for LLMs to attain powerful capabilities.
Furthermore, to effectively pre-train LLMs,  model architectures, acceleration methods, and optimization techniques need to be well designed. In what follows, we  first discuss the data collection and processing in Section~\ref{sec:data_collection}, then introduce the commonly used model architectures in Section~\ref{sec:architecture}, and finally present the training techniques to stably and efficiently  optimize LLMs in Section~\ref{sec:training_settings}.

\subsection{Data Collection and Preparation}
\label{sec:data_collection}
Compared with small-scale language models, LLMs have a stronger demand for high-quality data for model pre-training, and their model capacities  largely rely on the pre-training corpus and how it has been preprocessed. In this part, we discuss the collection and processing of pre-training data, including data sources, preprocessing methods, and important analysis of how pre-training data affects the performance of LLMs.

\subsubsection{Data Source}\label{sec-source}
To develop a capable LLM, it is key to collect a large amount of natural language corpus from various data sources.
Existing LLMs mainly leverage a mixture of diverse public textual datasets as the pre-training corpus. 
Figure~\ref{fig:source-ratio} shows  the distribution of the sources of pre-training data for a number of representative LLMs. 

The source of pre-training corpus can be broadly categorized into two types: general data and specialized data. General data, such as webpages, books, and conversational text, is utilized by most LLMs~\cite{Chowdhery-arxiv-2022-PaLM,Brown-NeurIPS-2020-Language,Zhang-arxiv-2022-OPT} due to its large, diverse, and accessible nature, which can enhance the language modeling and generalization abilities of LLMs. In light of the impressive generalization capabilities exhibited by LLMs, there are also studies that extend their pre-training corpus to more specialized datasets, such as multilingual data, scientific data, and code, endowing LLMs with specific task-solving capabilities~\cite{Chowdhery-arxiv-2022-PaLM,Taylor-arxiv-2022-Galactica, nijkamp-arxiv-2022-Codegen}. In what follows, we describe these two types of pre-training data sources and their effects on LLMs.  {For a detailed introduction to the commonly used corpus, one can refer to Section~\ref{sec:commonly_used_corpora}.} 

\paratitle{General Text Data.}
As we can see in Figure~\ref{fig:source-ratio}, the vast majority of LLMs adopt general-purpose pre-training data, such as webpages, books, and conversational text, which provides rich text sources on a variety of topics. 
Next, we briefly summarize three important kinds of general data.

$\bullet$ \emph{Webpages.} Owing to the proliferation of the Internet, various types of data have been created, which enables LLMs to gain diverse linguistic knowledge and enhance their generalization capabilities~\cite{radford-blog-2019-language,Raffel-JMLR-2020-Exploring}. For convenient use of these data resources, a large amount of data is crawled from the web in previous work, such as CommonCrawl~\cite{commoncrawl}. However, the crawled web data tends to contain both high-quality text, such as Wikipedia and low-quality text, like spam mail, thus it is important to filter and process webpages for improving the data quality. 

$\bullet$ \emph{Conversation text.} %
Conversation data can enhance the conversational competence of LLMs~\cite{Zhang-arxiv-2022-OPT} and potentially improve their performance on a range of question-answering tasks~\cite{Chowdhery-arxiv-2022-PaLM}. 
Researchers can utilize  subsets of public conversation corpus (\eg PushShift.io Reddit corpus)~\cite{Roller-ACL-2021-Recipes,Baumgartner-AAAI-2020-The} or collect conversation data from online social media.  
{Since online conversational data often involves discussions among multiple participants, an effective processing way  is to transform a conversation into a tree structure, where the utterance is linked to the one it responds to.
In this way, the multi-party conversation tree can be divided into multiple sub-conversations, which can be collected in the pre-training corpus.}
Furthermore, a potential risk is that the excessive integration of dialogue data into LLMs may result in a side effect~\cite{Zhang-arxiv-2022-OPT}:  declarative instructions and direct interrogatives are erroneously perceived as the beginning of conversations, thus leading to a decline in the efficacy of the instructions. 

$\bullet$ \emph{Books.} Compared to other corpus, books provide an important source of  formal long texts, which are potentially   beneficial for %
{LLMs to learn  linguistic knowledge, model long-term dependency, and generate narrative and coherent texts.}
To obtain open-source book data, existing studies usually adopt the Books3 and Bookcorpus2 datasets, which are available in the Pile dataset~\cite{Gao-arxiv-2021-Pile}. 

\paratitle{Specialized Text Data.} Specialized  datasets are  useful to improve the specific capabilities of LLMs on downstream tasks.  
Next, we introduce three kinds of specialized data.

$\bullet$ \emph{Multilingual text.}  
In addition to the text in the target language, 
integrating a multilingual corpus can enhance the multilingual abilities of language understanding and generation. 
 For example, BLOOM~\cite{Scao-arxiv-2022-BLOOM} and PaLM~\cite{Chowdhery-arxiv-2022-PaLM} have curated multilingual data covering 46 and 122 languages, respectively, within their pre-training corpora. FLM~\cite{Li-arxiv-2023-FLM} mixes Chinese and English corpora in nearly equal proportions. These models demonstrate impressive performance in multilingual tasks, such as translation, multilingual summarization, and multilingual question answering, and achieve comparable or superior performance to the state-of-the-art models that are fine-tuned on the corpus in the target language(s).

$\bullet$ \emph{Scientific text.}
The exploration of science by humans has been witnessed by the increasing growth  of scientific publications. 
In order to enhance the understanding of scientific knowledge for LLMs~\cite{Taylor-arxiv-2022-Galactica,Lewkowycz-arxiv-2022-Solving}, 
it is useful to incorporate a scientific corpus for model pre-training~\cite{Taylor-arxiv-2022-Galactica,Lewkowycz-arxiv-2022-Solving}. %
By pre-training on a vast amount of scientific text, LLMs can achieve impressive performance in scientific and reasoning tasks~\cite{Saier-arxiv-2023-unarXive}.
To construct the scientific corpus, existing efforts mainly collect arXiv papers, {scientific textbooks}, math webpages, and other related scientific resources.  
 Due to the complex nature of data in scientific fields, such as mathematical symbols and protein sequences, specific tokenization and preprocessing techniques are usually required to transform these different formats of data into a unified form that can be processed by language models.

$\bullet$ \emph{Code.} Program synthesis has been widely studied in the research community~\cite{Simon-JACM-1963-Experiments,Manna-CommunACM-1971-Toward,Feng-EMNLPFindings-2020-CodeBERT,Chen-arxiv-2021-evaluating,Austin-arxiv-2021-Program}, especially the use of PLMs trained on code~\cite{Black-GitHub-2021-GPT-Neo,Wang-GitHub-2021-GPT-J}.   
However, it remains challenging for these PLMs (\eg GPT-J~\cite{Wang-GitHub-2021-GPT-J}) to generate high-quality and accurate programs.
Recent studies~\cite{Chen-arxiv-2021-evaluating,Austin-arxiv-2021-Program} have found that training LLMs  on a vast code corpus can lead to a substantial improvement in the quality of the synthesized programs. The generated programs can successfully pass expert-designed unit-test cases~\cite{Chen-arxiv-2021-evaluating} or solve competitive programming questions~\cite{Li-Science-2022-AlphaCode}. %
{In general, two types of code corpora are commonly used for pre-training LLMs. The first source is  from programming question answering  communities like Stack Exchange~\cite{Xu-SIGPLAN-2022-Systematic}. The second source is  from public software repositories such as GitHub~\cite{Chen-arxiv-2021-evaluating,Austin-arxiv-2021-Program,nijkamp-arxiv-2022-Codegen}, where code data (including comments and docstrings)  are collected for utilization.}
Compared to natural language text, code is  in the format of a programming language, corresponding to long-range  dependencies and accurate execution logic~\cite{Madaan-emnlp-2022-Language}. 
{A recent study~\cite{FU-blog-2022-how}} also speculates that training on code might be a source of complex reasoning abilities (\eg chain-of-thought ability~\cite{Wei-arxiv-2022-chain}).  
{Furthermore, it has been shown that  formatting reasoning tasks into code can help  LLMs generate more accurate results~\cite{Madaan-emnlp-2022-Language}.}

\subsubsection{Data Preprocessing}
\label{sec:data_pre_processing}
\begin{figure*}
    \centering
    \includegraphics[width=1\textwidth]{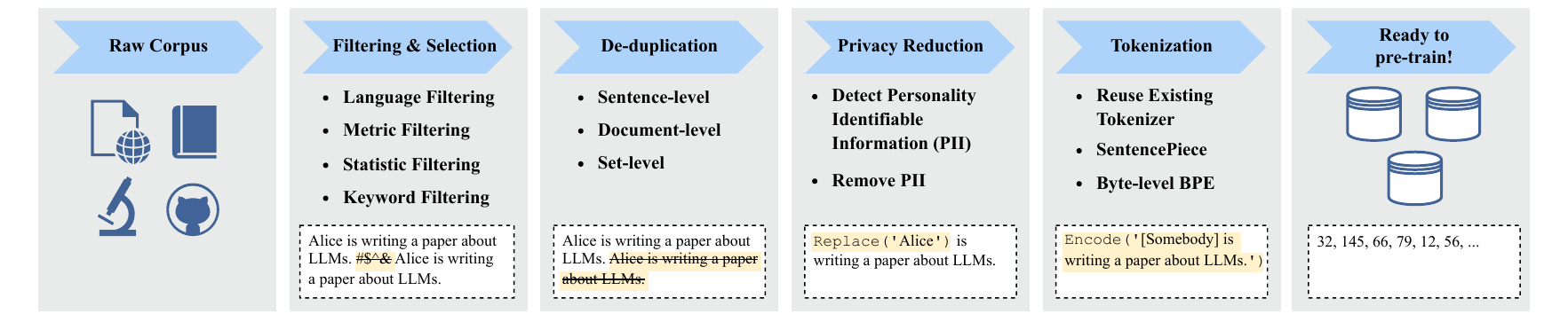}
    \caption{An illustration of a typical data preprocessing pipeline for pre-training large language models. }
    \label{fig:processing-pipeline}
\end{figure*}

After collecting a large amount of text data, it is essential to preprocess the data for constructing  the pre-training corpus, especially removing  noisy, redundant, irrelevant, and potentially toxic data~\cite{Rae-arxiv-2021-Scaling, Chowdhery-arxiv-2022-PaLM, Longpre-arxiv-2023-pretrainer}, which may largely affect the  capacity and performance of LLMs.  
{To facilitate the data processing, 
a recent study~\cite{Chen-2023-arxiv-Data} proposes a useful data processing system for LLMs, named Data-Juicer, which provides over 50 processing operators and tools.}
In this part, we review the detailed data preprocessing strategies to improve the quality of the collected data~\cite{Rae-arxiv-2021-Scaling,Du-ICML-2022-GLaM,Scao-arxiv-2022-BLOOM}. 
{A typical pipeline of preprocessing the pre-training data for LLMs has been illustrated in Figure~\ref{fig:processing-pipeline}.}

\paratitle{Filtering and Selection.}
To remove low-quality data from the collected corpus, existing work generally adopts two approaches, namely  classifier-based and heuristic-based. 
The former approach trains a selection classifier based on high-quality texts and leverages it to identify and filter out low-quality data. 
{Typically, these methods train a binary classifier using positive instances that are: well-curated data (\eg Wikipedia pages)~\cite{Brown-NeurIPS-2020-Language,Du-ICML-2022-GLaM,Chowdhery-arxiv-2022-PaLM},   high-quality synthesized data~\cite{LLaMa3,Abdin-arxiv-2024-phi3,Penedo-arxiv-2024-fineweb,Maini-ICLRworkshop-2024-rephrasing}, or a combination of both.}
They sample candidate data as negative instances and predict the score that measures the quality of each data example. 
However, several studies~\cite{Du-ICML-2022-GLaM, Rae-arxiv-2021-Scaling} find that a classifier-based approach may result in the unintentional removal of high-quality texts in dialectal, colloquial, and sociolectal languages, which potentially leads to bias in the pre-training corpus and diminishes the corpus diversity.
As the second approach, several studies, such as BLOOM~\cite{Scao-arxiv-2022-BLOOM} and Gopher~\cite{Rae-arxiv-2021-Scaling}, employ heuristic-based approaches to eliminate low-quality texts through a set of well-designed rules, which can be summarized as follows:

\begin{itemize}[leftmargin=0.5cm, itemsep=5pt]
\item \emph{Language based filtering.} If a LLM would  be mainly used  in the tasks of certain languages, the text in other languages can be filtered.

\item \emph{Metric based filtering.} Evaluation metrics about the generated texts, \eg perplexity, can be employed to detect and remove unnatural sentences.

\item \emph{Statistic based  filtering.} Statistical features of a corpus, \eg the punctuation distribution, symbol-to-word ratio, and sentence length, can be utilized to measure the text quality and filter the low-quality data.

\item \emph{Keyword based filtering.} Based on specific keyword set, the noisy or unuseful elements in the text, such as HTML tags, hyperlinks, boilerplates, and offensive words, can be identified and removed. 
\end{itemize}

In addition to the above methods, LLMs (especially relatively small models) can be also employed for data selection, either by computing perplexity~\cite{Marion-arxiv-2023-data_pruning} or directly prompting LLMs~\cite{Sachdeva-arxiv-2024-askllm} for measuring the sample importance. However,  using LLMs is unavoidably computationally intensive for large-scale data selection.


\paratitle{De-duplication.}
Existing work~\cite{Hernandez-arxiv-2022-Scaling} has found that duplicate data in a corpus would reduce the diversity of language models, which may  cause the training process to become unstable and thus affect the model performance.
Therefore, it is necessary to de-duplicate the pre-training corpus. 
Specially, de-duplication can be performed at different granularities, including sentence-level, document-level, and dataset-level de-duplication. 
First,  low-quality sentences that contain repeated words and phrases should be removed,  as they may introduce repetitive patterns in language modeling~\cite{Holtzman-2019-ICLR-The}. At the document level, existing studies mostly rely on the overlap ratio of surface features (\eg words and $n$-grams overlap) between documents to detect and remove duplicate documents containing similar contents~\cite{Rae-arxiv-2021-Scaling,Touvron-arxiv-2023-LLaMA,Scao-arxiv-2022-BLOOM,Lee-ACL-2022-Deduplicating}. 
Furthermore, to avoid the dataset contamination problem, it is also crucial to prevent the overlap between the training and evaluation sets~\cite{Chowdhery-arxiv-2022-PaLM}, by removing the possible duplicate texts  from the training set.
It has been shown that the three levels of de-duplication are useful to improve the training of LLMs~\cite{Carlini-arxiv-2022-Quantifying,Chowdhery-arxiv-2022-PaLM}, which should be jointly used in practice. 

\paratitle{Privacy Reduction.} 
Thus, it is necessary to remove the \emph{personally identifiable information~(PII)}  from the pre-training corpus. One direct and effective approach is to employ rule-based methods, such as keyword spotting, to detect and remove PII such as names, addresses, and phone numbers~\cite{Laurencon-NIPS-2022-The}. Furthermore, researchers also find that the vulnerability of LLMs under privacy attacks can be attributed to the presence of {duplicate PII data} in the pre-training corpus~\cite{Kandpal-ICML-2022-Deduplicating}. Therefore, de-duplication can also reduce privacy risks to some extent.

\paratitle{Tokenization.}
Tokenization is also a crucial step for data preprocessing.  
It aims to segment raw text into sequences of individual tokens, which are subsequently used as the inputs of LLMs. In traditional NLP research (\eg sequence labeling with conditional random fields~\cite{Lafferty-ICML-2001}), word-based tokenization is the predominant approach, which is more aligned with human's language cognition. However, word-based tokenization can yield different segmentation results for the same input in some languages (\eg Chinese word segmentation), generate  a huge word vocabulary containing many low-frequency words, and also suffer from the ``\emph{out-of-vocabulary}'' issue. Thus, several  neural network models employ \emph{character} as the minimum unit to derive the word representation (\eg a CNN word encoder in ELMo~\cite{Peters-NAACL-2018}). Recently, \emph{subword tokenizers} have been widely used in Transformer based language models, typically including Byte-Pair Encoding tokenization, WordPiece   tokenization and Unigram tokenization. HuggingFace has maintained an excellent online NLP course  on tokenizer\footnote{https://huggingface.co/learn/nlp-course/chapter6} with running examples, and we refer to the beginners to this course. Next, we briefly describe the three representative  tokenization methods. 

$\bullet$ \emph{Byte-Pair Encoding~(BPE) tokenization}. BPE was originally proposed as a general data compression algorithm in 1994~\cite{Philip-1994-BPE}, and then adapted to NLP for tokenization~\cite{Sennrich-ACL-2016-nueral}. It starts with a set of basic symbols (\eg the alphabets and boundary characters), and iteratively combine frequent pairs of two consecutive tokens in the corpus as new tokens (called \emph{merge}). For each merge, the selection criterion is based on the co-occurrence frequency of two   contiguous tokens: the top frequent pair would be selected. The merge process continues until it reaches the predefined size.  
Further, Byte-level BPE has been used  to improve the tokenization quality for multilingual corpus (\eg the text containing non-ASCII characters) by considering \emph{bytes} as the basic symbols for merge. 
Representative language models with this tokenization approach include GPT-2, BART, and LLaMA.

$\bullet$ \emph{WordPiece tokenization}.  WordPiece was a Google internal subword tokenization algorithm. 
It was originally proposed by Google in developing voice search systems~\cite{Mike-ICASSP-2012-Japanese}. Then, it was  used in the neural machine translation system in 2016~\cite{Wu-CoRR-2016}, and  was adopted as the word tokenizer for BERT in 2018~\cite{Devlin-NAACL-2019-BERT}.  WordPiece has a very similar idea with BPE by iteratively merging consecutive tokens, whereas taking a slightly different  selection criterion for the merge. To conduct the merge, it first trains a language model and  employs it to score all possible pairs. Then, at each merge, it selects the pair that leads to the most increase in the likelihood of training data. 
Since Google has't released the official implementation of the WordPiece algorithm, HuggingFace gives a more intuitive selection measure  in its online NLP course: a pair is scored by dividing the co-occurrence count by the product of the occurrence counts of two tokens in the pair based on training corpus.

$\bullet$ \emph{Unigram tokenization}.
Unlike BPE and WordPiece, Unigram tokenization~\cite{Kudo-ACL-2018-Subword} starts with a sufficiently large set of  possible substrings or subtokens for a corpus, and iteratively removes the tokens in the current vocabulary until the expected vocabulary size is reached. 
As the selection criterion, it calculates  the yielded increase in the likelihood of training corpus by assuming that some token was removed from current vocabulary.  
This step is conducted based on a trained unigram language model. To estimate the unigram language model, it adopts 
an expectation–maximization~(EM) algorithm: at each iteration, we first find the currently optimal tokenization of words based on the old language model, and then re-estimate the probabilities of unigrams to update the language model. During this procedure, dynamic programming algorithms (\ie the Viterbi algorithm) are  used to efficiently find the optimal decomposition way of a word given the language model. 
Representative models that adopt this tokenization approach include T5 and mBART.

Although it is expedient to leverage an existing tokenizer (\eg OPT~\cite{Zhang-arxiv-2022-OPT} and GPT-3~\cite{Brown-NeurIPS-2020-Language} utilize the tokenizer of GPT-2~\cite{radford-blog-2019-language}), using a  tokenizer specially designed for the pre-training corpus can be highly beneficial~\cite{Scao-arxiv-2022-BLOOM}, especially for the corpus that consists of diverse domains, languages, and {formats}. 
Therefore, recent LLMs often train the customized tokenizers specially for the  pre-training corpus with the SentencePiece library~\cite{Kudo-EMNLP-2018-SentencePiece}, which includes Byte-level BPE and Unigram tokenization. 
A note is that normalization techniques in BPE, such as NFKC~\cite{Davis-arxiv-2001-Unicode}, may  degrade the tokenization performance~\cite{Hoffmann-arxiv-2022-Training,Rae-arxiv-2021-Scaling,Scao-arxiv-2022-BLOOM}. 
When extending existing LLMs  (\ie continual pre-training or instruction tuning), we should be also aware of the potential side effect with customized tokenizers.   
For example, LLaMA trains  
the BPE tokenizer  based on a pre-training corpus mainly consisting of English texts, and the derived vocabulary might be less capable in processing non-English data, \eg taking longer inference latency to generate Chinese texts. 

\paratitle{Discussion on Effect of Data Quality.} 
For pre-training, the quality of pre-training data is vital to the  model capacities of LLMs.
Existing work has shown that pre-training on the low-quality corpus, such as noisy, toxic, and duplicate data, would largely hurt the performance of models~\cite{Rae-arxiv-2021-Scaling, Lee-ACL-2022-Deduplicating, Kandpal-ICML-2022-Deduplicating, Hernandez-arxiv-2022-Scaling}.
Recent studies, such as T5~\cite{Raffel-JMLR-2020-Exploring}, GLaM~\cite{Du-ICML-2022-GLaM}, and Gopher~\cite{Rae-arxiv-2021-Scaling}, have investigated the influence of data quality on the LLMs' capacities.
By comparing the performance of models trained on the filtered and unfiltered corpus, they have reached the similar conclusion that pre-training LLMs on cleaned data can improve the model performance.  %
More specifically, the duplication of data may result in ``\emph{double descent}'' (referring to the phenomenon of performance initially deteriorating and subsequently improving)~\cite{Hernandez-arxiv-2022-Scaling,Nakkiran-ICLR-2020-Deep}, or even overwhelm the training process~\cite{Hernandez-arxiv-2022-Scaling}. 
In addition, it has been shown that duplicate data degrades the ability of LLMs to copy from the context, which might further affect the generalization capacity of LLMs using in-context learning~\cite{Hernandez-arxiv-2022-Scaling}.
Therefore, as suggested in~\cite{Rae-arxiv-2021-Scaling, Scao-arxiv-2022-BLOOM, Chowdhery-arxiv-2022-PaLM, Longpre-arxiv-2023-pretrainer}, it is essential to utilize  preprocessing methods like quality filtering, toxic filtering and deduplication to carefully clean  the pre-training corpus  (as illustrated in Section~\ref{sec:data_pre_processing}), to improve stability of the training process and avoid affecting the model performance.

\begin{figure}
    \centering
    \includegraphics[width=0.48\textwidth]{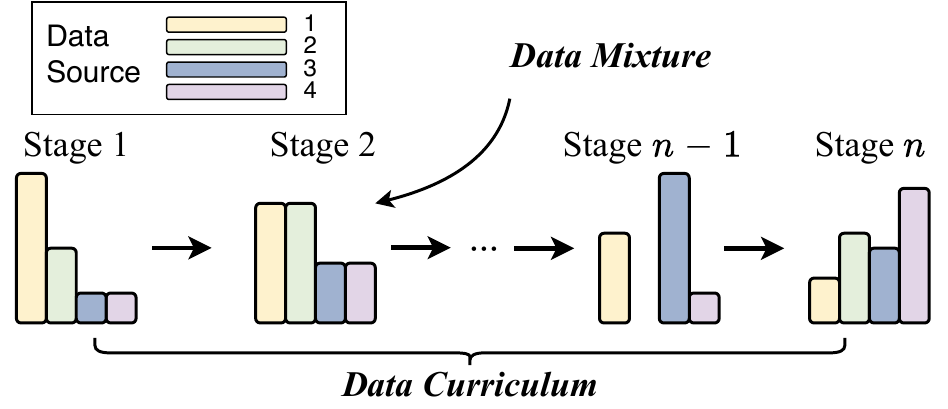}
    \caption{An illustration of  data scheduling for pre-training LLMs. }
    \label{fig:data_schedule}
\end{figure}

\subsubsection{Data Scheduling}
\label{sec:data_scheduling}
After data preprocessing, it is essential to design suitable strategies to schedule these multi-source data  for pre-training a capable LLM. Generally, two key aspects should be paid close attention for data scheduling: the proportion of each data source (\emph{data mixture}), and the order in which each data source is scheduled for training (\emph{data curriculum}).
Next, we discuss the two aspects in detail. An illustration of data scheduling has been presented in Figure~\ref{fig:data_schedule}.

\paratitle{Data Mixture.}
Since each kind of data source is closely related to the development of certain capacities for  LLMs (referring to the discussions in Section~\ref{sec:data_collection}), it is important to set a suitable  distribution to mix these data. 
The data mixture is generally set in a global level (\ie the distribution of the entire pre-training data), and can be also locally set to varied  proportions at different training stages. 
During pre-training, data samples from different sources would be selected according to the mixture proportions: more data will be sampled from a data source with a larger weight.  
Typically, existing LLMs such as LLaMA~\cite{Touvron-arxiv-2023-LLaMA} may employ upsampling or downsampling on the full data of each source to create specific data mixtures as pre-training data.
As Figure~\ref{fig:source-ratio} illustrates, existing LLMs use different data mixtures to construct the pre-training data.  
As a representative model, the pre-training data of LLaMA~\cite{Touvron-arxiv-2023-LLaMA} mainly consists of webpages (over 80\%),  alongside 6.5\% of code-heavy data from GitHub and StackExchange, 4.5\% from books, and 2.5\% of scientific data sourced from arXiv, which has become an important reference for training  general-purpose LLMs. 
Furthermore, special data mixtures can be used to facilitate different purposes. For example, Falcon~\cite{Penedo-2023-arxiv-Refinedweb} is trained on pure webpages, and CodeGen~\cite{nijkamp-arxiv-2022-Codegen} largely increases the amount of code data.
In practice, data mixture is often determined empirically, and we summarize several common strategies for finding an effective data mixture as follows: 

$\bullet$ \emph{Increasing the diversity of data sources.}
Recent studies have empirically shown that training on excessive data about a certain domain would degrade the generalization capability of LLMs on other domains~\cite{Taylor-arxiv-2022-Galactica,Rae-arxiv-2021-Scaling}. In contrast, increasing the data source heterogeneity (\eg  including diverse data sources) is critical for improving the downstream performance of LLMs~\cite{Longpre-arxiv-2023-pretrainer,Tirumala-2023-arXiv-D4,Shen-2023-arXiv-SlimPajamaDC}. 
To further examine the effect of different data sources, some studies  have conducted ablation experiments by removing  each data source one by one,  and pre-train LLMs with specially curated datasets~\cite{Longpre-arxiv-2023-pretrainer}. It has been shown  that dropping data sources with high heterogeneity (\eg webpages) impacts LLM's abilities more severely than dropping sources with low heterogeneity (\eg academic corpus).

$\bullet$ \emph{Optimizing data mixtures.}
In addition to manually setting the data mixtures, several studies have proposed to optimize the data mixtures for improving the model pre-training~\cite{Xie-arxiv-2023-DSIR,Xie-arxiv-2023-doremi}. Given the target downstream tasks, one can select pre-training data with either higher proximity in the feature space~\cite{Xie-arxiv-2023-DSIR} or those that provide positive influences on downstream  task performance~\cite{Wang-2023-arXiv-farewell}. 
Further, to reduce the reliance of target tasks, DoReMi~\cite{Xie-arxiv-2023-doremi} first trains a small reference model using   given initial domain weights, and then trains another small proxy model, upweighting the domains   on which the greatest discrepancies in  likelihood between the two models are observed. Finally, the learned  domain weights of the proxy model are applied to train a much larger LLM.
In a more simple way, one can train several small  language models with different data mixtures, and select the data mixture that leads to  the most desirable performance.
However, an assumption made in this  approach is,   when trained in a similar way,  small models would resemble with large models in model abilities or behaviors, which may not always hold in practice. 

$\bullet$ \emph{Specializing the targeted abilities.} 
The model capacities of LLMs heavily rely on data selection and mixture, and one can boost the proportions of specific data sources to enhance certain model abilities~\cite{Rae-arxiv-2021-Scaling,Longpre-arxiv-2023-pretrainer}.  
For example, the mathematical reasoning and coding abilities can be specially enhanced by training with more mathematical texts and code data, respectively. 
Furthermore, experimental results on the LAMBADA dataset~\cite{Paperno-ACL-2016-LAMBADA} show that increasing the proportion of books data can improve the model capacity  in capturing long-term dependencies from text, and increasing the proportion of the C4 dataset~\cite{Raffel-JMLR-2020-Exploring} leads to performance improvement on the C4 validation dataset~\cite{Rae-arxiv-2021-Scaling}.  
Generally, it is important to identify more implicit relations between data sources and  model abilities. {
To enhance specific skills such as mathematics and coding in LLMs, or to develop specialized LLMs, a practical way is to employ a multi-stage training approach, \eg general and skill-specific data can be  scheduled at two consecutive stages. This approach of training LLMs on varying sources or proportions of data across multiple stages is also known as ``data curriculum'', which will be introduced below.

\paratitle{Data Curriculum.} 
After preparing the data mixture, it is important to schedule the order that specific data is presented to LLMs for pre-training. 
It has been shown that, in some cases, to learn a certain skill, learning in a skill-set sequence (\eg basic skills $\rightarrow$ target skill) outperforms direct learning from a corpus focused solely on the target skill~\cite{Chen-2023-arXiv-skill,Roziere-arxiv-2023-codellama}. 
Following the idea of curriculum learning~\cite{Bengio-2009-arXiv-curriculum}, \emph{data curriculum} has been proposed and widely used in model pre-training~\cite{Chen-2023-arXiv-skill,Xu-2023-arXiv-contrastive,Roziere-arxiv-2023-codellama,Tworkowski-arxiv-2023-Focused}. It aims to  organize different parts of pre-training data for LLMs in a specific order, \eg  starting  with easy/general examples and progressively introducing  more challenging/specialized ones. 
More generally, 
it can broadly refer to the adaptive adjustment of data proportions for different sources during pre-training. 
Existing work about  data curriculum mainly focuses on continual pre-training, such as specialized coding LLMs (\eg CodeLLaMA~\cite{Roziere-arxiv-2023-codellama}) or long context LLMs (\eg LongLLaMA~\cite{Tworkowski-arxiv-2023-Focused}).
However, it still lacks of more detailed report about data curriculum for general-purpose LLMs (\eg LLaMA) in the literature. 
To determine data curriculum, a practical approach is to monitor the development of key abilities of LLMs based on specially constructed evaluation benchmarks, and then adaptively adjust the data mixture during pre-training.
Next, we take  three common abilities as examples to introduce how the concept of data curriculum\footnote{We utilize the symbol ``$\rightarrow$'' to represent the data order  in  data curriculum. For example, ``2T webpage  tokens $\rightarrow$ 500B code tokens'' means that the LLM is firstly trained with 2T webpage tokens and subsequently with 500B code data tokens. } applies in continual pre-training.

$\bullet$ \emph{Coding}. To improve the coding ability of LLMs, CodeLLaMA~\cite{Roziere-arxiv-2023-codellama} is developed based on LLaMA 2~\cite{Touvron-2023-llama2-arxiv}
(2T general tokens $\rightarrow$ 500B code-heavy tokens), aiming to improve the code generation ability and retain natural language understanding skills.   CodeLLaMA also provides a version that is further specialized to a certain programming language, namely CodeLLaMA-Python (2T general tokens $\rightarrow$ 500B code-heavy tokens $\rightarrow$ 100B Python-heavy tokens).

$\bullet$ \emph{Mathematics.} Llemma~\cite{Azerbayev-arxiv-2023-llemma} is proposed to enhance the mathematical capacities of general-purpose LLMs.
It is developed based on CodeLLaMA. 
Although CodeLLaMA~\cite{Roziere-arxiv-2023-codellama} mainly focuses on the coding ability, experiments have shown that it performs better than its base model LLaMA 2 on mathematics benchmarks~\cite{Azerbayev-arxiv-2023-llemma}. 
Based on CodeLLaMA, Llemma is continually trained on mixtures of scientific papers, web data containing mathematical text and  code (2T general tokens $\rightarrow$ 500B code-heavy tokens $\rightarrow$ 50$\sim$200B math-heavy tokens). Note that the pre-training data of Llemma also contains 5\% general domain data as a form of regularization.

$\bullet$ \emph{Long context}. Long context modeling is an important ability for LLMs, and many studies have explored extending the context windows of LLMs via continually training~\cite{Roziere-arxiv-2023-codellama,Tworkowski-arxiv-2023-Focused}. With modifications on position embeddings (\ie  position interpolation) of RoPE-based LLMs~\cite{Touvron-2023-llama2-arxiv,Touvron-arxiv-2023-LLaMA,Chen-arxiv-2023-Extending}, 
CodeLLaMA further extends the context window of LLaMA 2 (2.5T tokens with 4K context window $\rightarrow$ 20B tokens with 16K context window).
LongLLaMA~\cite{Tworkowski-arxiv-2023-Focused} also achieves longer context window 
with the help of external memory and a unique training objective (1T tokens with 2K context window $\rightarrow$ 10B tokens with 8K context window).

{
\subsubsection{Summary of Data Preparation}
\label{sec:data_prepare_sug}
In this part, we summarize the general procedure and key points  to prepare pre-training data for LLMs, which are  detailed in the following three aspects.    
}

{
$\bullet$ \emph{Data collection.} It is suggested to include  diverse data sources in  the pre-training data.  Although Falcon~\cite{Penedo-2023-arxiv-Refinedweb} shows that webpages alone can be employed to train powerful LLMs, a more typical approach is to also incorporate  diverse high-quality text like code, books, scientific papers, \etc.
If a LLM is specialized with a certain skill, the proportion of corresponding data source should be increased accordingly.  
For example, Gopher~\cite{Rae-arxiv-2021-Scaling} and Chinchilla~\cite{Hoffmann-arxiv-2022-Training} are trained with approximately 40\% of data from books. PaLM~\cite{driess-arxiv-2023-palm} and LaMDA~\cite{Thoppilan-CoRR-2022-LaMDA} use  approximately 50\% conversational data.
}

{
$\bullet$ \emph{Data cleaning.} After data collection, it is crucial to clean the raw corpus to enhance its quality as possible. First, deduplication is commonly  used in existing work~\cite{Touvron-2023-llama2-arxiv,Tirumala-2023-arXiv-D4,Penedo-2023-arxiv-Refinedweb}.
Second,  low-quality text, toxic content, and data with privacy concerns should be removed at different granularities (\eg document, passage or sentence). In practice,  both heuristic and classifier-based methods can be employed for quality and toxicity filtering (\eg CCNet~\cite{Wenzek-2020-LREC-CCNet}, fastText~\cite{Joulin-2017-EACL-fasttext}, and Data-Juicer~\cite{chen-2023-arXiv-DataJuicer}). Third,  with the cleaned data, one can further unify or specify the  format for pre-training data, and perform the tokenization by training the tokenizer on the filtered and deduplicated corpus with libraries like SentencePiece~\cite{Kudo-EMNLP-2018-SentencePiece}. 
}

{
$\bullet$ \emph{Data scheduling.} 
With the preprocessed data, the next step is to determine the data mixture and the specific order of data for pre-training LLMs.  
To determine both settings, a practical way is to first train several small  language models with multiple candidate plans and then select a good plan among them~\cite{Xie-arxiv-2023-doremi}.
Overall, it is more difficult to find a  suitable data curriculum. 
In practice, one can monitor the performance of intermediate model checkpoints on specific evaluation benchmarks, and dynamically tune the data mixture and distribution during  pre-training. In this process, it is also useful to explore the potential relations between data sources and model abilities to instruct the design of data curriculum. 
}

\subsection{Architecture}
\label{sec:architecture}
 In this section, we review the architecture design  of LLMs, \ie mainstream architecture, pre-training objective, and detailed configuration. Table~\ref{model_card}  presents the model cards of several representative LLMs with public details.

\begin{table*}[htb]
    \centering
    \caption{Model cards of several selected  LLMs with public configuration details. Here, PE denotes position embedding, \#L denotes the number of layers, \#H denotes the number of attention heads, $d_{model}$ denotes the size of hidden states, and MCL denotes the maximum context length during training.}
    \begin{tabular}{lcrccccrrrr}
    \toprule
         \textbf{Model}&\textbf{Category}&\textbf{Size}&\textbf{Normalization}&\textbf{PE}&\textbf{Activation}&\textbf{Bias}&\textbf{\#L}&\textbf{\#H}&\textbf{$d_{model}$}&\textbf{MCL} \\
\midrule
GPT3~\cite{Brown-NeurIPS-2020-Language}&Causal decoder&175B&Pre LayerNorm&Learned&GeLU&\checkmark&96&96&12288&2048\\
PanGU-~$\alpha$~\cite{Zeng-arxiv-2021-PanGualpha}&Causal decoder&207B&Pre LayerNorm&Learned&GeLU&\checkmark&64&128&16384&1024\\
OPT~\cite{Zhang-arxiv-2022-OPT}&Causal decoder&175B&Pre LayerNorm&Learned&ReLU&\checkmark&96&96&12288&2048\\
PaLM~\cite{Chowdhery-arxiv-2022-PaLM}&Causal decoder&540B&Pre LayerNorm&RoPE&SwiGLU&$\times$&118&48&18432&2048\\
BLOOM~\cite{Scao-arxiv-2022-BLOOM}&Causal decoder&176B&Pre LayerNorm&ALiBi&GeLU&\checkmark&70&112&14336& 2048\\
MT-NLG~\cite{Smith-CoRR-2022-Using}&Causal decoder&530B&-&-&-&-&105&128&20480& 2048\\
Gopher~\cite{Rae-arxiv-2021-Scaling}&Causal decoder&280B&Pre RMSNorm&Relative&-&-& 80&128&16384&2048 \\
Chinchilla~\cite{Hoffmann-arxiv-2022-Training}&Causal decoder&70B&Pre RMSNorm&Relative&-&-&80&64&8192&-\\
Galactica~\cite{Taylor-arxiv-2022-Galactica}&Causal decoder&120B&Pre LayerNorm&Learned&GeLU&$\times$&96&80&10240 &2048\\
LaMDA~\cite{Thoppilan-CoRR-2022-LaMDA}&Causal decoder&137B&-&Relative&GeGLU&-&64&128&8192&-\\
Jurassic-1~\cite{lieber-2021-jurassic}&Causal decoder&178B&Pre LayerNorm&Learned&GeLU&\checkmark&76 &96&13824& 2048 \\
LLaMA~\cite{Touvron-arxiv-2023-LLaMA}&Causal decoder&65B&Pre RMSNorm&RoPE&SwiGLU&$\times$&80&64&8192&2048\\
LLaMA 2~\cite{Touvron-2023-llama2-arxiv} &Causal decoder&70B&Pre RMSNorm&RoPE&SwiGLU&$\times$&80&64&8192& 4096 \\
Falcon~\cite{Penedo-2023-arxiv-Refinedweb}&Causal decoder&40B&Pre LayerNorm&RoPE&GeLU&$\times$&60&64&8192&2048\\
GLM-130B~\cite{Zeng-arxiv-2022-GLM}&Prefix decoder&130B&Post DeepNorm&RoPE&GeGLU&\checkmark&70&96&12288&2048\\
T5~\cite{Raffel-JMLR-2020-Exploring}&Encoder-decoder&11B&Pre RMSNorm&Relative&ReLU&$\times$&24&128&1024&512\\

\bottomrule
    \end{tabular}
    
    \label{model_card}
\end{table*}

\subsubsection{Typical Architectures}\label{sec:archs}

\begin{figure*}[htb]
    \includegraphics[width=\textwidth]{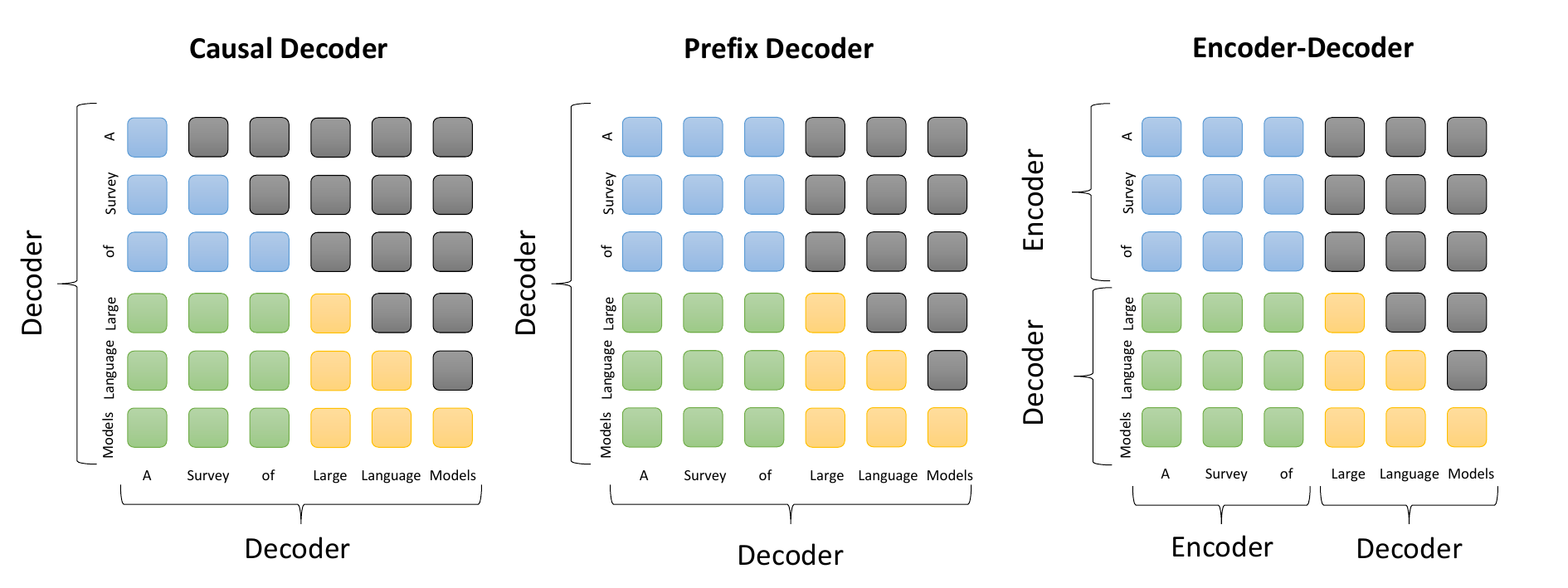}
    \caption{A comparison of the attention patterns in  three mainstream architectures. Here, the blue, green, yellow and grey rounded rectangles indicate the attention between prefix tokens, attention between prefix and target tokens, attention between target tokens, and masked attention respectively.}
    \label{fig:architectures}
\end{figure*}


Due to the excellent parallelizability and capacity, the Transformer architecture~\cite{Vaswani-NIPS-2017-Attention} has become the de facto backbone to develop various LLMs, making it possible to scale language models to hundreds or thousands of billions of parameters.    
In general, the mainstream architectures of existing LLMs can be roughly categorized into three major types, namely encoder-decoder, causal decoder, and prefix decoder, 
as shown in Figure~\ref{fig:architectures}. 

\paratitle{Encoder-decoder Architecture.}
The vanilla Transformer model is built on the encoder-decoder architecture~\cite{Vaswani-NIPS-2017-Attention}, which consists of two stacks of Transformer blocks as the encoder and decoder, respectively. 
The encoder adopts stacked multi-head self-attention layers to encode the input sequence for generating its latent representations, while the decoder performs cross-attention on these representations and autoregressively generates the target sequence. %
Encoder-decoder PLMs (\eg T5~\cite{Raffel-JMLR-2020-Exploring} and BART~\cite{Lewis-ACL-2020-BART}) have shown effectiveness on a variety of NLP tasks.
{
So far, there are only a small number of LLMs that are built based on the encoder-decoder architecture, \eg Flan-T5~\cite{Chung-arxiv-2022-Scaling}. We leave a detailed discussion about the architecture selection  in Section~\ref{sec-summary-arc}.  
}

\paratitle{Causal Decoder Architecture.} 
The causal decoder architecture incorporates the unidirectional attention mask, to guarantee that each input token can only attend to the past tokens and itself. 
The input and output tokens are processed in the same fashion through the decoder. As representative language models of this architecture, the GPT-series models~\cite{radford-openai-2018-improving,radford-blog-2019-language,Brown-NeurIPS-2020-Language} are developed based on the causal-decoder architecture. 
In particular, GPT-3~\cite{Brown-NeurIPS-2020-Language} has successfully demonstrated the effectiveness of this architecture, also showing an amazing in-context learning capability of LLMs.
Interestingly, GPT-1~\cite{radford-openai-2018-improving} and GPT-2~\cite{radford-blog-2019-language} do not exhibit such superior abilities as those in GPT-3, and it seems that scaling plays an important role in increasing the model capacity of this model architecture. 
So far, 
the causal decoders have been widely adopted as the architecture of LLMs by various existing LLMs, such as OPT~\cite{Zhang-arxiv-2022-OPT}, BLOOM~\cite{Scao-arxiv-2022-BLOOM}, and Gopher~\cite{Rae-arxiv-2021-Scaling}. 
{
Note that both the causal decoder and prefix decoder discussed next belong to decoder-only architectures. When  mentioning  ``{decoder-only architecture}'', it mainly refers to the causal decoder architecture in existing literature, unless specified. }

\paratitle{Prefix Decoder Architecture.} 
The prefix decoder architecture {(\aka non-causal decoder~\cite{Zhang-ICML-2022-Examining})}  revises the masking mechanism of causal decoders, to enable performing bidirectional attention over the prefix tokens~\cite{Dong-NIPS-2019-Unified} and unidirectional attention only on generated tokens. 
In this way, like the encoder-decoder architecture, the prefix decoders can bidirectionally encode the prefix sequence and autoregressively predict the output tokens one by one, where the same parameters are shared during encoding and decoding. 
Instead of pre-training from scratch, a practical suggestion is to continually train causal decoders and then convert them into prefix decoders for accelerating convergence~\cite{Wang-ICML-2022-What}, \eg U-PaLM~\cite{Tay-arxiv-2022-Transcending} is derived from PaLM~\cite{Chowdhery-arxiv-2022-PaLM}. Existing representative LLMs based on prefix decoders include GLM-130B~\cite{Zeng-arxiv-2022-GLM} and U-PaLM~\cite{Tay-arxiv-2022-Transcending}.

\paratitle{Mixture-of-Experts.} For the above three types of architectures, we can  further extend  them via  the mixture-of-experts (MoE) scaling, in which a subset of neural network weights for each input are sparsely activated, \eg Switch Transformer~\cite{Fedus-JMLR-2021-Switch} and GLaM~\cite{Du-ICML-2022-GLaM}.  
{The major merit is that MoE is a flexible way to scale up the model parameter while maintaining a constant computational cost~\cite{Fedus-JMLR-2021-Switch}.} It has been shown that substantial  performance improvement can be observed by increasing either the number of experts or the total parameter size~\cite{Clark-ICML-2022-Unified}.  {Despite the merits, training large MoE models may suffer from instability issues due to the complex, hard-switching nature of the routing operation. 
To enhance the training stability of MoE-based language models, techniques such as selectively using high-precision tensors in the routing module or initializing the model with a smaller range have been introduced~\cite{Fedus-JMLR-2021-Switch}.
More recently, there is widespread speculation that GPT-4 has been developed based on the MoE architecture, but without official verification. 
} 

\textcolor{blue}{
\begin{table}[htb]
    \centering  
    \caption{{Comparison of parallelism and complexity of different models.} $T$ represents sequence length, $H$ represents the dimension of the input representation, $N$ represents the dimension after compression in SSMs, and $M$ represents the number of layers in each Hyena module.}
    \label{tab-new-architectures}
    \small
    \resizebox{\linewidth}{!}{
    \begin{tabular}{lcccc}
         \toprule
         Model&Decoding Complexity&Training Complexity  \\
         \midrule
         Transformer  & $O(H(T+H))$   & $O(TH(T+H))$           \\
         SSM      & $O(H(N^2+H))$ & $O(TH(\log T+N^2+H))$ \\
         Mamba       & $O(H(N^2+H))$ & $O(TH(N^2+H))$          \\
         RWKV        & $O(H^2)$      & $O(TH^2)$                \\
         RetNet   & $O(H^2)$      & $O(TH^2)$                \\
         Hyena    & $O(MH(T+H))$ & $O(TMH(\log T+H))$     \\
         \bottomrule
    \end{tabular}}
\end{table}
}

\paratitle{Emergent Architectures.} 
The conventional Transformer architecture typically suffers from quadratic computational complexity with respect to sequence length, resulting in a high processing cost for dealing with long inputs. 
To improve efficiency, recent studies aim to devise new architectures for language modeling, most based on parameterized state space models {(SSM)~\cite{gu-2022-iclr-efficiently}}, which can be viewed as a combination of RNN and CNN. On the one hand, SSM can generate outputs recursively like RNN, meaning that they only need to refer to the single previous state during decoding. It makes the decoding process more efficient as it eliminates the need to revisit all previous states as in conventional Transformers. On the other hand, these models have the capability to encode an entire sequence in parallel like Transformers via convolution computation. Thus, they can benefit from the parallelism of GPUs with techniques such as Parallel Scan~\cite{smith-2023-iclr-s5,orvieto-2023-icml-lru}, FFT~\cite{poli-2023-icml-hyena,peng-2023-arxiv-rwkv}, and Chunkwise Recurrent~\cite{sun-2023-arxiv-retnet}. Despite the high computation efficiency of SSMs, their performance still lags behind Transformer. Thus, several variants of SSM have been proposed, including Mamba~\cite{gu-arxiv-2023-mamba}, RetNet~\cite{sun-2023-arxiv-retnet}, RWKV~\cite{peng-23-arxiv-rwkv}, and Hyena~\cite{poli-2023-icml-hyena}.

$\bullet$ \emph{Mamba.} Mamba~\cite{gu-arxiv-2023-mamba} aims to selectively {filter out or remember} information during  state update. {It replaces the original fixed parameters of SSM layers with functions of the input}, selectively {filtering out} information of the previous state and {the current input depending on the current input}. Compared with traditional SSMs, Mamba has demonstrated improved text modeling capacities.

$\bullet$ \emph{RWKV.} RWKV~\cite{peng-23-arxiv-rwkv} combines the advantages of Transformer and RNN. It employs time-mixing modules, \ie RNN with gating, and channel-mixing modules that are special {feedforward neural networks}~\cite{peng-23-arxiv-rwkv}. Within these modules, token shift, a linear combination of the current and previous token, is used instead of the token representation as the input}. 

{$\bullet$ \emph{RetNet.} RetNet~\cite{sun-2023-arxiv-retnet} proposes multi-scale retention (MSR) to replace the attention module in Transformer. Similar to linear attention, in the MSR module, the input is first mapped into query, key, and value, and the product of key and value is employed to update the state. Then, the query is used to project the state into the output. Similar to traditional SSMs, RetNet keeps the parallel and recurrent computation capacity at the same time.}

{$\bullet$ \emph{Hyena.} Hyena employs long convolution to replace the attention module. In the long convolution module, the filters based on relative positions are used to aggregate information at different positions into the middle representations, and gating functions are employed to further project intermediate representations into the final output. However, due to the long convolution, Hyena can not infer like RNN and must explicitly access all previous states.}

\subsubsection{Detailed Configuration}
\label{sec:configuration}

\begin{table*}[htb]
    \centering
   \caption{Detailed formulations for the network configurations. Here, Sublayer denotes a FFN or a self-attention module in a Transformer layer, $d$ denotes the size of hidden states, $\mathbf{p}_i$ denotes position embedding at position $i$, $A_{ij}$ denotes the attention score between a query and a key, $r_{i-j}$ denotes a learnable scalar based on the offset between the query and the key, and $\mathbf{R}_{\Theta, t}$ denotes a rotary matrix with rotation degree $t\cdot\Theta$.}
    \begin{tabular}{c|c|l}
    \toprule
    \textbf{Configuration}& \textbf{Method }& \textbf{Equation}\\\midrule
    \multirow{3}{*}{Normalization position}     &Post Norm~\cite{Vaswani-NIPS-2017-Attention} & $\mathrm{Norm(}\mathbf{x} \mathrm{+ Sublayer(}\mathbf{x}\mathrm{))}$  \\
         &Pre Norm~\cite{radford-blog-2019-language} &$\mathbf{x}+\mathrm{Sublayer}(\mathrm{Norm}(\mathbf{x}))$ \\
         &Sandwich Norm~\cite{Ding-NIPS-2021-CogView} & $\mathbf{x}+\mathrm{Norm}(\mathrm{Sublayer}(\mathrm{Norm}(\mathbf{x})))$\\
         \midrule
         \multirow{3}{*}{Normalization method}& LayerNorm~\cite{Jimmy-arxiv-2016-Layer}&$ \frac{\mathbf {x} -\mathbf \mu}{\mathbf \sigma}\cdot \gamma +\beta, \text{~~~} \mathbf \mu=\frac 1 d \sum_{i=1}^d  x_i, \text{~~~} \mathbf \sigma=\sqrt{\frac 1 d \sum_{i=1}^d ( x_i-\mathbf \mu))^2}$         \\
         & RMSNorm~\cite{Zhang-NIPS-2019-Root}&  $ \frac {\mathbf x}{\mathrm{RMS}(\mathbf x)} \cdot \gamma, \text{~~~} \mathrm{RMS}(\mathbf x)=\sqrt {\frac 1 d \sum_{i=1}^d  x_i^2}$\\
         & DeepNorm~\cite{Wang-arxiv-2022-DeepNet}& $\mathrm {LayerNorm}(\alpha \cdot \mathbf x + \mathrm{Sublayer}(\mathbf x))$\\
         \midrule
    \multirow{5}{*}{Activation function}& ReLU~\cite{Vinod-ICML-2010-Rectified}& $\mathrm {ReLU}(\mathbf{x}) = \max(\mathbf{x},\mathbf{0})$\\
    & GeLU~\cite{Wang-EMNLP-2018-GLUE}&$\mathrm {GeLU}(\mathbf{x}) = \mathrm{0.5}\mathbf{x} \otimes [1+\mathrm{erf}(\mathbf{x}/\sqrt{2})], \text{~~~} \mathrm{erf}(x)=\frac 2 {\sqrt \pi}\int_0^x e^{-t^2} dt  $\\
    &Swish~\cite{Ramachandran-arXiv-2017-searching} & $\mathrm{Swish}(\mathbf{x}) = \mathbf x \otimes \mathrm{sigmoid}(\mathbf{x}) $\\
    & SwiGLU~\cite{Shazeer-arxiv-2020-GLU}&$\mathrm{SwiGLU}(\mathbf{x}_1,\mathbf{x}_2) = \mathrm{Swish}(\mathbf{x_1})\otimes \mathbf{x_2}$\\
    & GeGLU~\cite{Shazeer-arxiv-2020-GLU}& $\mathrm{GeGLU}(\mathbf{x}_1,\mathbf{x}_2) = \mathrm{GeLU}(\mathbf{x_1})\otimes \mathbf{x_2}$ \\
    \midrule
    \multirow{4}{*}{Position embedding}& Absolute~\cite{Vaswani-NIPS-2017-Attention}&$\mathbf x_i = \mathbf x_i + \mathbf p_i$ \\
    &Relative~\cite{Raffel-JMLR-2020-Exploring}& $A_{ij} = \mathbf W_q\mathbf x_i \mathbf x_j^T \mathbf W_k^T + r_{i-j}$\\
    &RoPE~\cite{Su-arxiv-2021-Roformer}& $A_{ij} = \mathbf W_q\mathbf x_i \mathbf R_{\Theta, i-j}\mathbf x_j^T\mathbf W_k^T =  (\mathbf W_q\mathbf x_i \mathbf R_{\Theta, i})(\mathbf W_k \mathbf x_j R_{\Theta,  j })^T$\\
    &ALiBi~\cite{Press-ICLR-2022-Train}& 
    $A_{ij} = \mathbf W_q \mathbf x_i \mathbf x_j^T\mathbf W_k^T - m (i-j)$\\  
    \bottomrule
    \end{tabular} 
    \label{tab:detailed_configuration}
\end{table*}

Since the launch of Transformer~\cite{Vaswani-NIPS-2017-Attention}, various improvements have been proposed to enhance its training stability, performance, and computational efficiency. In this part, we will discuss the corresponding 
configurations for 
four major parts of the Transformer, including normalization, position embeddings, activation functions, and attention and bias. 
To make this survey more self-contained, we present the detailed formulations for these configurations in Table~\ref{tab:detailed_configuration}. 

\paratitle{Normalization Methods.} {Training instability is a  challenging issue for pre-training  LLMs. To alleviate this issue, normalization is a widely adopted strategy to stabilize the training of neural networks. In the vanilla Transformer~\cite{Vaswani-NIPS-2017-Attention}, LayerNorm~\cite{Jimmy-arxiv-2016-Layer} is employed. Recently, several advanced normalization techniques have been proposed as alternatives to LayerNorm, \eg RMSNorm, and DeepNorm. }

{
    $\bullet$ \emph{LayerNorm.} In the early research, BatchNorm~\cite{Ioffe-2015-ICML-Batch} is a commonly used normalization method. However, it is difficult to deal with sequence data of variable lengths and small-batch data. Thus, LayerNorm~\cite{Jimmy-arxiv-2016-Layer} is introduced to conduct layerwise normalization.  Specifically, the mean and variance over all activations per layer are calculated to re-center and re-scale the activations. }

{
    $\bullet$ \emph{RMSNorm.} To improve the training speed of LayerNorm (LN), RMSNorm~\cite{Zhang-NIPS-2019-Root} is proposed by re-scaling the activations with only the root mean square (RMS) of the summed activations,  instead of the mean and variance. Related research has demonstrated its superiority in training speed and performance on Transformer~\cite{Narang-EMNLP-2021-Do}. Representative models that adopt RMSNorm include Gopher~\cite{Rae-arxiv-2021-Scaling} and Chinchilla~\cite{Hoffmann-arxiv-2022-Training}.}

{
    $\bullet$ \emph{DeepNorm.} DeepNorm is proposed by Microsoft~\cite{Wang-arxiv-2022-DeepNet} to stabilize the training of deep Transformers. 
    With DeepNorm as residual connections, Transformers can be scaled up to 1,000 layers~\cite{Wang-arxiv-2022-DeepNet}, which has shown  the advantages of stability and good performance. It has been adopted by GLM-130B~\cite{Zeng-arxiv-2022-GLM}.
    }
    
\paratitle{Normalization Position.} {In addition to the normalization method, normalization position also plays a  crucial role in the LLMs. There are generally three choices for the normalization position, \ie post-LN, pre-LN, and sandwich-LN. } 

{
    $\bullet$ \emph{Post-LN.} Post-LN is used in the vanilla  Transformer~\cite{Vaswani-NIPS-2017-Attention}, which is placed between residual blocks. However, existing work has found that the training of Transformers with post-LN tends  to be  instable due to the large gradients near the output layer~\cite{Xiong-ICML-2020-On}. Thus, post-LN is rarely employed in existing LLMs except combined with other strategies (\eg combining post-LN with pre-LN in GLM-130B~\cite{Zeng-arxiv-2022-GLM}).  }

{
    $\bullet$ \emph{Pre-LN.} Different from post-LN, pre-LN~\cite{Baevski-2019-ICLR-Adaptive} is applied before each sub-layer, and an additional LN is placed before the final prediction. Compared with post-LN, the Transformers with pre-LN are more stable in training. However, it performs worse than the variants with post-LN~\cite{liu-2020-EMNLP-Understanding}. Despite the decreasing performance, most LLMs still adopt pre-LN due to the training stability. } 
    {
    However, one exception is that pre-LN  has been found   unstable in GLM when training models more than 100B parameters~\cite{Zeng-arxiv-2022-GLM}.
    }

    $\bullet$ \emph{Sandwich-LN.} Based on pre-LN, Sandwich-LN~\cite{Ding-NIPS-2021-CogView} adds extra LN before the residual connections to avoid the  {value explosion issues in Transformer layer outputs.} However, it has been found that Sandwich-LN sometimes fails to stabilize the training of LLMs and may lead to the collapse of training~\cite{Zeng-arxiv-2022-GLM}.
    

\paratitle{Activation Functions.}
To obtain good performance, activation functions also need to be properly set in feed-forward networks. 
In existing LLMs, GeLU activations~\cite{Dan-arxiv-2016-Gaussian} are widely used. %
Specially, in the latest LLMs (\eg PaLM and LaMDA), variants of GLU activation~\cite{Dauphin-ICML-2017-Language,Shazeer-arxiv-2020-GLU} have also been utilized, especially the SwiGLU and GeGLU variants, which 
often achieve better performance in practice~\cite{Narang-EMNLP-2021-Do}. However, compared with GeLU, they require extra parameters (about 50\%) in the feed-forward networks~\cite{Le-EMNLP-2022-What}.

\paratitle{Position Embeddings.} 
Since the self-attention modules
in Transformer are permutation equivariant, position embeddings~(PE) are employed to inject absolute or relative position information for modeling sequences. 

{
    $\bullet$ \emph{Absolute position embedding.} In the vanilla Transformer~\cite{Vaswani-NIPS-2017-Attention}, absolute position embeddings are employed. At the bottoms of the encoder and the decoder, the absolute positional embeddings are added to the input embeddings. 
    There are two variants of absolute position embeddings proposed in the vanilla Transformer~\cite{Vaswani-NIPS-2017-Attention}, \ie sinusoidal and learned position embeddings, where the latter is commonly used in existing pre-trained language models. } 

    $\bullet$ \emph{Relative position embedding.} Unlike absolute position embeddings, relative positional embeddings are generated according to the offsets between keys and queries~\cite{shaw-2018-acl-self}.  
    {A popular variant of relative PE was introduced in Transformer-XL~\cite{dai-2019-acl-transformer,Yang-NeurIPS-2019-xlnet}. The calculation of attention scores between keys and queries has been modified to introduce learnable embeddings corresponding to relative positions.}
 {T5~\cite{Raffel-JMLR-2020-Exploring} further simplified relative positional embeddings, which was subsequently adopted by Gopher~\cite{Rae-arxiv-2021-Scaling}.}
    Specifically, it adds learnable scalars to the attention scores, where the scalars are calculated based on the distances between the positions of the query and the key. Compared with the absolute PE, Transformers with relative position embedding can generalize to sequences longer than those sequences for training, \ie extrapolation~\cite{Press-ICLR-2022-Train}. 

    $\bullet$ \emph{Rotary position embedding.} 
{Rotary position embedding (RoPE)~\cite{Su-arxiv-2021-Roformer} 
    sets specific rotatory matrices based on the absolute position of each key or query.
    The scores between keys and queries can be computed with relative position information (Table~\ref{tab:detailed_configuration}).} 
    {RoPE combines each consecutive pair of elements in query and key vectors as \emph{a dimension},   so there are $d/2$  dimensions for an original $d$-length embedding. For each dimension $i \in \{1, \dots, d/2\}$, the  pair of involved elements will rotate based on the rotation angle $t\cdot \theta_i$, where $t$ denotes  the position index and $\theta_i$ is the basis in the dimension. Following sinusoidal position embeddings~\cite{Vaswani-NIPS-2017-Attention}, RoPE defines the \emph{basis}  $\theta_i$ as an exponentiation of the \emph{base} $b$ (set to $10000$ by  default):}
\begin{equation}
    \Theta = \{\theta_i = b^{-2(i-1)/d} | i \in \{1, 2, \dots , d/2 \}\}.\label{eq:basis}
\end{equation}
{Furthermore, a recent study~\cite{Peng-arxiv-2023-Yarn} defines the distance required to rotate one cycle ($2\pi$) for each dimension as wavelength:}
\begin{equation}
      \lambda_i = 2\pi b^{2(i-1)/d}=2 \pi / \theta_i.\label{eq:wavelength}
\end{equation}       
Due to the excellent performance and the long-term decay property, RoPE is widely adopted in the latest LLMs, \eg PaLM~\cite{Chowdhery-arxiv-2022-PaLM} and LLaMA~\cite{Touvron-arxiv-2023-LLaMA}. Based on RoPE,  xPos~\cite{Sun-2022-arxiv-Length} further improves the translation invariance and length extrapolation of Transformer. At each dimension of the rotation angle vector, xPos adds a special exponential decay that is smaller when the basis is larger. It can alleviate the unstable phenomenon {during training} as the distance increases.

    $\bullet$ \emph{ALiBi.} ALiBi~\cite{Press-ICLR-2022-Train} is proposed to improve the extrapolation of Transformer. Similar to relative position embedding, it biases attention scores with a penalty based on the distances between keys and queries. 
    {Different from the relative positional embedding methods like T5~\cite{Raffel-JMLR-2020-Exploring}, the penalty scores in ALiBi are pre-defined without any trainable parameters.}
    Empirical results in~\cite{Press-ICLR-2022-Train} have shown that ALiBi has {a better extrapolation performance on sequences that are longer than those for training than several popular 
    position embedding methods such as  sinusoidal PE~\cite{Vaswani-NIPS-2017-Attention}, RoPE~\cite{Su-arxiv-2021-Roformer}, and T5 bias~\cite{Raffel-JMLR-2020-Exploring}. }
    In addition, it has been shown that ALiBi can also improve training stability in BLOOM~\cite{Scao-arxiv-2022-BLOOM}.

%

\paratitle{Attention.} 
Attention mechanism is a critical component of Transformer. It allows the tokens across the sequence to interact with each other and compute the representations of the input and output sequence. 

{
    $\bullet$ \emph{Full attention}. In the vanilla  Transformer~\cite{Vaswani-NIPS-2017-Attention}, the attention mechanism is conducted in a pairwise way, considering the relations between all token pairs in a sequence. It 
    adopts scaled dot-product attention, in which the hidden states are mapped into queries, keys, and values. 
    Additionally, Transformer uses multi-head attention instead of single attention, projecting the queries, keys, and values with different projections in different heads. The concatenation of the output of each head is taken as the final output.}

{
    $\bullet$ \emph{Sparse attention}. A crucial challenge of full attention is the quadratic computational complexity, which becomes a burden when dealing with long sequences. Therefore, various efficient Transformer variants are proposed to reduce the computational complexity of the attention mechanism~\cite{Peng-ICLR-2021-Random, Zaheer-NIPS-2020-Big}. For instance, locally banded sparse attention (\ie Factorized Attention~\cite{Child-arxiv-2019-Generating} has been adopted  in GPT-3~\cite{Brown-NeurIPS-2020-Language}. Instead of the whole sequence, each query can only attend to a subset of tokens based on the positions. }

    $\bullet$ \emph{Multi-query/grouped-query attention}. Multi-query attention refers to the attention variant
    where different heads share  {the same  linear transformation matrices on the keys and values~\cite{Shazeer-2019-arxiv-Fast}.} 
    It achieves higher inference speed {with only a minor sacrifice in model quality.}
    Representative models with multi-query attention include PaLM~\cite{Chowdhery-arxiv-2022-PaLM} and StarCoder~\cite{Li-2023-arxiv-Starcoder}.  
    {To make a trade-off between multi-query attention and multi-head attention, grouped-query attention (GQA)~\cite{Ainslie-2023-arxiv-gqa} has been explored. In GQA, heads are assigned into different groups, and those heads that belong to the same group will share the same transformation matrices. Specially, GQA has been adopted and empirically tested in the recently released LLaMA 2 model~\cite{Touvron-2023-llama2-arxiv}.}

    $\bullet$ \emph{FlashAttention}. Different from most existing approximate attention methods that trade-off model quality to improve the computing efficiency, FlashAttention~\cite{Dao-NeurIPS-2020-FLASH} proposes to optimize the speed and memory consumption of attention modules on GPUs from an IO-aware perspective. There exist different levels of memory on modern GPUs, \eg SRAM with a fast IO and HBM with a relatively slow IO. FlashAttention organizes the input into blocks and introduces necessary recomputation, both to make better use of the fast memory SRAM.  Implemented as a fused kernel in CUDA, FlashAttention has been integrated into PyTorch~\cite{Paszke-NeurIPS-2019-Pytorch}, DeepSpeed~\cite{Rasley-KDD-2020-DeepSpeed}, and Megatron-LM~\cite{Shoeybi-arXiv-2019-Megatron}. {The updated version FlashAttention-2~\cite{Dao-2023-arxiv-flashattention2} further optimizes the work partitioning of GPU thread blocks and warps, leading to around 2$\times$ speedup when compared to the original FlashAttention.}

    $\bullet$ \emph{PagedAttention}.  
    {It has been observed when LLM are deployed on servers, GPU memory is largely occupied by cached attention key and value tensors (called \emph{KV cache}). The major reason is that the input lengths are often varied, leading to fragmentation and over-reservation issues. Inspired by the classic paging technique in operating systems, PagedAttention has been proposed to improve the memory efficiency and throughput of deployed LLMs~\cite{vllm-pagedattention}. In detail, PagedAttention partitions each sequence into subsequences, and the corresponding KV caches of these subsequences are allocated into non-contiguous physical blocks. The paging technique increases the GPU utilization  and enables efficient memory sharing in parallel sampling.}


To put all these discussions together, we summarize the suggestions from existing literature for detailed configuration.  
For stronger generalization and training stability, it is suggested to choose the pre RMSNorm for layer normalization, and SwiGLU or GeGLU as the activation function. %
{In addition,  LN may not be used immediately after embedding layers, which is likely to incur  performance degradation.} As for position embeddings, {RoPE} or ALiBi is a better choice since it performs better on long sequences. %

\subsubsection{Pre-training Tasks}
Pre-training plays a key role that encodes general knowledge from large-scale corpus into the massive model parameters. 
For training LLMs, there are two commonly used pre-training tasks, namely language modeling and denoising autoencoding. %

\paratitle{Language Modeling.} 
The language modeling task (LM) is the most commonly used objective to pre-train decoder-only LLMs, \eg GPT3~\cite{Brown-NeurIPS-2020-Language} and PaLM~\cite{Chowdhery-arxiv-2022-PaLM}. Given a sequence of tokens $\mathbf{x}=\{x_1,\dots,x_n\}$, the LM task aims to autoregressively predict the target tokens  %
{$x_i$ based on the preceding tokens $x_{<i}$ in a sequence}. A general  training objective is to maximize the following likelihood: %
\begin{equation}
    \mathcal{L}_{LM}(\mathbf{x})=\sum_{i=1}^n \log P(x_i|\mathbf{x}_{<i}).\label{eq:lm}
\end{equation}

Since most language tasks can be cast as the prediction problem based on the input,  %
{these decoder-only} LLMs might be potentially advantageous to implicitly learn how to accomplish these tasks in a unified LM way. 
Some studies have also revealed that  %
{decoder-only} LLMs can be naturally transferred to certain tasks by autoregressively predicting the next tokens~\cite{radford-blog-2019-language,Brown-NeurIPS-2020-Language}, without fine-tuning.  
An important variant of LM is the \emph{prefix language modeling} task,  %
{which is designed for pre-training models with the prefix decoder architecture.
The tokens within a randomly selected prefix would not be used in computing the loss of prefix language modeling.} 
{
With the same amount of tokens seen during pre-training, prefix language modeling  performs slightly worse than language modeling, since fewer tokens in the sequence are involved for model pre-training~\cite{Wang-ICML-2022-What}.}

\paratitle{Denoising Autoencoding.} 
{
In addition to conventional LM, the denoising autoencoding task (DAE) has also been widely used to pre-train language models~\cite{Lewis-ACL-2020-BART,Raffel-JMLR-2020-Exploring}.
The inputs $\mathbf{x}_{\backslash \Tilde{\mathbf{x}}}$ for DAE task are corrupted text with randomly replaced spans. Then, the language models are trained to recover the replaced tokens $\Tilde{\mathbf{x}}$. Formally, the training objective of DAE is  denoted as follows:}
\begin{equation}
    \mathcal{L}_{DAE}(\mathbf{x})= \log P(\Tilde{\mathbf{x}}|\mathbf{x}_{\backslash \Tilde{\mathbf{x}}}).\label{eq:dae}
\end{equation}

However, the DAE task seems to be more complicated in implementation than LM task. As a result, it has not been widely used to pre-train  large language models. 
Existing LLMs that take DAE as pre-training objectives include T5~\cite{Raffel-JMLR-2020-Exploring} and GLM-130B~\cite{Zeng-arxiv-2022-GLM}. These models are mainly trained to recover the replaced spans in an autoregressive way.

\paratitle{Mixture-of-Denoisers.} Mixture-of-Denoisers (MoD)~\cite{Tay-arxiv-2022-UL2}, also  known as  UL2 loss, was introduced as a unified objective for pre-training language models. MoD regards both LM and DAE objectives as different types of denoising tasks, namely  
{S-denoiser (LM), R-denoiser (DAE, short span and low corruption), and X-denoiser (DAE, long span or high corruption).}
Among the three denoising tasks, S-denoiser is similar to the conventional LM objective (Equation~\eqref{eq:lm}), while R-denoiser and X-denoiser are similar to DAE objectives (Equation~\eqref{eq:dae}) but differ from each other in the lengths of spans and ratio of corrupted text.  
{For input sentences started with different special tokens (\ie \{\texttt{[R]}, \texttt{[S]}, \texttt{[X]}\}), the model will be optimized using the corresponding denoisers.} 
MoD has been  
applied in the latest PaLM 2 model~\cite{Anil-arxiv-2023-palm2}.

\begin{figure}
    \centering
    \begin{tabular}{lc|lc|lc}
    \multicolumn{6}{c}{I am sleepy. I start a pot of \_\_\_\_\_} \\
    \midrule[0.3pt]
    coffee & 0.661 & strong & 0.008 & soup & 0.005 \\
    water & 0.119 & black & 0.008 & \dots & \dots \\
    tea & 0.057 & hot & 0.007 & happy & 4.3e-6 \\
    rice & 0.017 & oat & 0.006 & Boh & 4.3e-6 \\
    chai & 0.012 & beans & 0.006 & \dots & \dots \\
    \end{tabular}
    \caption{The probability distribution over the vocabulary in descending order for the next token of the context ``\emph{I am sleepy. I start a pot of}''. For ease of discussion, this example is given in word units instead of subword units. }
    \label{fig:decoding-example}
\end{figure}

\subsubsection{Decoding Strategy}
\label{sec-decoding}
After the LLMs have been pre-trained, it is essential to employ a specific decoding strategy to generate the appropriate output from the LLMs. 

\paratitle{Background.} We start the discussion with the prevalent decoder-only architecture, and introduce the auto-regressive decoding mechanism.  Since such LLMs are pre-trained based on the  language modeling task (Equation~\ref{eq:lm}), a basic decoding method is \emph{greedy search} that predicts the most likely token at each step based on the previously generated tokens, formally modeled as:
\begin{equation}
{ x_i = \underset{x}{\arg\max} P(x |\mathbf{x}_{<i}),}
\end{equation}
where $x_i$ is the token with the highest probability at $i$-th step of generation conditioned on the context $\mathbf{x}_{<i}$. For instance in Figure~\ref{fig:decoding-example}, when predicting the next token of the sentence \emph{``I am sleepy. I start a pot of''}, greedy search selects the token ``coffee'' which has the highest probability at the current step. Greedy search can achieve satisfactory results in  text generation tasks (\eg machine translation and text summarization), in which  the output is highly dependent on the input~\cite{Murray-WMT-2018-Correcting}. However, in terms of open-ended generation tasks (\eg story generation and dialog), greedy search sometimes tends to generate awkward and repetitive sentences~\cite{Holtzman-2020-ICLR-The}.

As another alternative decoding strategy, sampling-based methods are proposed to randomly select the next token based on the probability distribution to enhance the randomness and diversity during generation:
\begin{equation}
    x_i \sim P(x|\mathbf{x}_{<i}).
\end{equation} 
For the example in  Figure~\ref{fig:decoding-example}, sampling-based methods will sample the word ``coffee'' with higher probability while also retaining the possibilities of selecting the rest words, ``water'', ``tea'', ``rice'', \etc.


Not limited to the decoder-only architecture, these two decoding methods can be generally applied to encoder-decoder models and prefix decoder models in a similar way.   

\paratitle{Improvement for Greedy Search.}
Selecting the token with the highest probability at each step may result in overlooking a sentence with a higher overall probability but a lower local estimation. Next, we introduce several improvement strategies  to alleviate this issue.

$\bullet$ \emph{Beam search.} 
Beam search~\cite{CMU-book-1977-speech} retains the sentences with the $n$ (beam size) highest probabilities at each step during the decoding process, and finally selects the generated response with the top probability. Typically, the beam size is configured within the range of 3 to 6. However, opting for a larger beam size might result in a decline in performance~\cite{Koehn-ACL-2017-Six}.

$\bullet$ \emph{Length penalty.} 
Since beam search favours shorter sentences, imposing length penalty (\aka length normalization) is a commonly used technique~\cite{Wu-arxiv-2016-Google} to overcome this issue, which normalizes the sentence probability according to the sentence length (divided by an exponential power $\alpha$ of the length).

Besides, some researchers~\cite{Paulus-iclr-2018-A} propose to penalize the generation of previously  generated tokens or $n$-grams to alleviate the issue of repetitive generation. In addition, diverse beam search~\cite{Vijayakumar-arxiv-2016-Diverse} can be leveraged to produce a set of diverse outputs based on the same input.

\paratitle{Improvement for Random Sampling.}
Sampling-based methods sample the token over the whole vocabulary, which may select wrong or irrelevant tokens (\eg ``happy'' and ``Boh'' in Figure~\ref{fig:decoding-example}) based on the context. To improve the generation quality, several strategies have been proposed for mitigating or preventing the selection of words with exceedingly low probabilities.

$\bullet$ \emph{Temperature sampling.}
To modulate the randomness of sampling, a practical method is to adjust the temperature coefficient of the softmax function for computing the probability of the $j$-th token over the vocabulary:
\begin{equation}\label{eqn:temperature}
    P(x_j|\mathbf{x}_{<i}) = \frac{\exp{(l_j/t)}}{\sum_{j'} \exp{(l_{j'}/t)}},
\end{equation}
where $l_{j'}$ is the logits of each word and $t$ is the temperature coefficient.  Reducing the temperature $t$ increases the chance of selecting words with high probabilities while decreases the chances of selecting words with low probabilities. 
When $t$ is set to 1, it becomes the default random sampling; when $t$ is approaching 0, it is equivalent to greedy search. 
In addition, when $t$ goes to infinity, it degenerates to uniform sampling. 

$\bullet$ \emph{Top-$k$ sampling.} 
Different from temperature sampling, top-$k$ sampling directly truncates the tokens with lower probability and only samples from the tokens with the top $k$ highest probabilities~\cite{Fan-2018-ACL-Hierarchical}. For example in Figure~\ref{fig:decoding-example}, top-$5$ sampling will sample from the words ``coffee'', ``water'', ``tea'', ``rice'', and ``chai'' from their re-scaled probabilities.

$\bullet$ \emph{Top-$p$ sampling.} 
Since top-$k$ sampling does not consider the overall  possibility distribution, a constant value of  $k$  may be not be suitable for different contexts. Therefore, top-$p$ sampling (\aka nucleus sampling) is proposed by sampling from the smallest set having a cumulative probability above (or equal to) $p$~\cite{Holtzman-2020-ICLR-The}. In practice, the smallest set can be constructed by gradually adding tokens from the vocabulary sorted in descending order of generative probability, until their cumulative value exceeds $p$.

Recently, researchers have also explored other sampling strategies for LLMs. For instance, \emph{$\eta$-sampling}~\cite{Hewitt-emnlp-2022-Truncation} further improves top-$p$ sampling by introducing a dynamic threshold based on the probability distribution. Furthermore, \emph{contrastive search}~\cite{Su-2022-NIPS-A} and \emph{typical sampling}~\cite{Meister-TACL-2023-Locally} can be utilized to improve the generation coherence during decoding. 
{Since it has been found that large models tend to assign higher probability to important tokens compared to small models, \emph{contrastive decoding}~\cite{Li-ACL-2023-Contrastive} utilizes a larger LM (\eg OPT-13B) and a smaller LM (\eg OPT-125M) to measure their log-likelihood differences. Subsequently, tokens are sampled based on the delta value of the probability distribution, thereby amplifying the impact of important tokens.}
{Based on this contrastive idea, DoLa~\cite{Chuang-arxiv-2023-DoLa} further extends this approach to contrasting the logits across different layers of a single LLM, as higher layers tend to assign more weight to important tokens.}

\paratitle{Practical Settings.} 
In practice, existing libraries (\eg Transformers~\cite{Wolf-EMNLP-2020-Transformers}) and public APIs of LLMs (\eg OpenAI) have supported various decoding strategies to serve different scenarios of text generation. Next, we present the decoding settings of several representative LLMs: 

$\bullet$ \emph{T5}~\cite{Raffel-JMLR-2020-Exploring} utilizes greedy search as the default setting and applies beam search (beam size of 4) with a length penalty {of 0.6} for translation and summarization tasks.

$\bullet$ \emph{GPT-3}~\cite{Brown-NeurIPS-2020-Language} employs beam search with a beam size of 4 and a length penalty of 0.6 for all generation tasks.

$\bullet$ \emph{Alpaca}~\cite{Taori-github-2023-Stanford} utilizes sampling-based strategies 
{with top-$k$ ($k=50$), top-$p$  ($p=0.9$)}, and temperature of 0.7 for open-ended generation.

$\bullet$ \emph{LLaMA}~\cite{Touvron-arxiv-2023-LLaMA} applies diverse decoding strategies tailored to specific tasks. For instance, it employs the greedy search for question answering tasks while utilizes a sampling strategy with the temperature settings of 0.1 (pass@1) and 0.8 (pass@100) for code generation.

$\bullet$ \emph{OpenAI API}  
{supports several basic decoding strategies, including greedy search (by setting \texttt{temperature} to 0), beam search (with the setting \texttt{best\_of}), temperature sampling (with the setting \texttt{temperature}), nucleus sampling (with the setting \texttt{top\_p}). It also introduce parameters \texttt{presence\_penalty} and \texttt{frequency\_penalty} to control the repetition degree of generation.} 
{According to the OpenAI's document, their APIs would produce different outputs even if the input and the hyper-parameters are the same. Setting temperature to 0 can yield more deterministic outputs, albeit with a slight chance of variability.}

\subsubsection{Summary and Discussion}\label{sec-summary-arc}

The choice of architecture and pre-training tasks may incur different inductive biases for LLMs, which would lead to different model capacities. %
In this part, we discuss one open issue about the  architecture choice for LLMs.

\begin{center}
\begin{tcolorbox}[colback=blue!5!white,colframe=blue!55!black,width=0.48\textwidth,title={Why does Predicting the Next Word Works?}]
{
The essence of decoder-only architecture is to \emph{accurately predict the next word} for reconstructing the pre-training data. Till now, there has been no formal study that theoretically  demonstrates its advantage over other architectures. An interesting explanation was from  Ilya Sutskever during the interview held by Jensen Huang\footnote{https://www.nvidia.com/en-us/on-demand/session/gtcspring23-S52092/}. 
The original transcript from the interview was copied below\footnote{https://lifearchitect.ai/ilya/}:  
\\

\texttt{Say you read a detective novel. It’s like complicated plot, a storyline, different characters, lots of events, mysteries like clues, it’s unclear. Then, let’s say that at the last page of the book, the detective has gathered all the clues, gathered all the people and saying, "okay, I’m going to reveal the identity of whoever committed the crime and that person’s name is". Predict that word. ...\\
Now, there are many different words. But predicting those words better and better, the understanding of the text keeps on increasing. GPT-4 predicts the next word better. 
}
}
\end{tcolorbox}
\end{center}

\paratitle{Architecture Choice}.  {In earlier literature of pre-trained language models, there are lots of discussions on the effects of different architectures~\cite{Wang-ICML-2022-What,Tay-arxiv-2022-UL2}}.
However, most LLMs are developed based on the causal decoder architecture, and there  still lacks a theoretical analysis on its advantage over the other alternatives.  Next, we briefly summarize existing  discussions on this issue. 

$\bullet$ By pre-training with the LM objective, it seems that causal decoder architecture can achieve a superior zero-shot and few-shot generalization capacity. Existing research has shown that without multi-task fine-tuning, the causal decoder has better zero-shot performance than other architectures~\cite{Wang-ICML-2022-What}. The success of GPT-3~\cite{Brown-NeurIPS-2020-Language} has demonstrates that the large causal decoder model can be a good few-shot learner. In addition, instruction tuning and alignment tuning discussed in Section~\ref{sec-adaptation} have been proven to further enhance the capability of large causal decoder models~\cite{Wei-ICLR-2022-Finetuned,Ouyang-arxiv-2022-Training,Chung-arxiv-2022-Scaling}. 
    
$\bullet$  Scaling law has been widely observed in causal decoders. By scaling the model size, the dataset size, and the total computation, the performance of causal decoders can be substantially   improved~\cite{Kaplan-arxiv-2020-Scaling,Brown-NeurIPS-2020-Language}.
Thus, it has become an important strategy to increase the model capacity of the causal decoder via  scaling.  %
However, more detailed investigation on encoder-decoder models is still lacking, and more efforts are needed to investigate the performance of  encoder-decoder models at a large scale.

More research efforts about the discussions on architectures and pre-training objectives are in need to analyze how the choices of the architecture and pre-training tasks affect the capacity of LLMs, especially for encoder-decoder architectures.  
Despite the effectiveness of decoder-only architecture, it is also suggested to make more diverse exploration on architecture design. 
Besides the major architecture, the detailed configuration of LLM is also worth attention, which has been discussed in Section~\ref{sec:configuration}.

\begin{table*}[htb]
    \centering
    \caption{Detailed optimization settings of several existing  LLMs. 
    }
\resizebox{2.05\columnwidth}{!}{
\begin{tabular}{lrrccccccc}
\toprule
\textbf{Model}          & \begin{tabular}[r]{@{}r@{}}\textbf{Batch Size}\\ \textbf{(\#tokens)}\end{tabular} & \begin{tabular}[r]{@{}r@{}}\textbf{Learning}\\ \textbf{Rate}\end{tabular}   & \textbf{Warmup} & \textbf{Decay Method}   & \textbf{Optimizer}                              & \begin{tabular}[l]{@{}l@{}}\textbf{Precision}\\ \textbf{Type}\end{tabular} & \begin{tabular}[l]{@{}l@{}}\textbf{Weight}\\ \textbf{Decay}\end{tabular}                & \begin{tabular}[l]{@{}l@{}}\textbf{Grad}\\ \textbf{Clip}\end{tabular} & \textbf{Dropout} \\ \midrule
GPT3~(175B)             & 32K→3.2M                                                                          & $6 \times 10^{-5}$                                                          & yes             & cosine decay to 10\%    & Adam                                            & FP16                                                                       & 0.1                                                                                     & 1.0                                                                   & -                \\
PanGu-$\alpha$~(200B)   & -                                                                                 & $2 \times 10^{-5}$                                                          & -               & -                       & Adam                                            & -                                                                          & 0.1                                                                                     & -                                                                     & -                \\
OPT~(175B)              & 2M                                                                                & $1.2 \times 10^{-4}$                                                        & yes             & manual decay            & AdamW                                           & FP16                                                                       & 0.1                                                                                     & -                                                                     & 0.1              \\
PaLM~(540B)             & 1M→4M                                                                             & $1 \times 10^{-2}$                                                          & no              & inverse square root     & Adafactor                                       & BF16                                                                       & $lr^2$                                                                                  & 1.0                                                                   & 0.1              \\
BLOOM~(176B)            & 4M                                                                                & $6 \times 10^{-5}$                                                          & yes             & cosine decay to 10\%    & Adam                                            & BF16                                                                       & 0.1                                                                                     & 1.0                                                                   & 0.0              \\
MT-NLG~(530B)           & 64 K→3.75M                                                                        & $5 \times 10^{-5}$                                                          & yes             & cosine decay to 10\%    & Adam                                            & BF16                                                                       & 0.1                                                                                     & 1.0                                                                   & -                \\
Gopher~(280B)           & 3M→6M                                                                             & $4 \times 10^{-5}$                                                          & yes             & cosine decay to 10\%    & Adam                                            & BF16                                                                       & -                                                                                       & 1.0                                                                   & -                \\
Chinchilla~(70B)        & 1.5M→3M                                                                           & $1 \times 10^{-4}$                                                          & yes             & cosine decay to 10\%    & AdamW                                           & BF16                                                                       & -                                                                                       & -                                                                     & -                \\
Galactica~(120B)        & 2M                                                                                & $7 \times 10^{-6}$                                                          & yes             & linear decay to 10\%    & AdamW                                           & -                                                                          & 0.1                                                                                     & 1.0                                                                   & 0.1              \\
LaMDA~(137B)            & 256K                                                                              & -                                                                           & -               & -                       & -                                               & BF16                                                                       & -                                                                                       & -                                                                     & -                \\
Jurassic-1~(178B)       & 32 K→3.2M                                                                         & $6 \times 10^{-5}$                                                          & yes             & -                       & -                                               & -                                                                          & -                                                                                       & -                                                                     & -                \\
LLaMA~(65B)             & 4M                                                                                & $1.5 \times 10^{-4}$                                                        & yes             & cosine decay to 10\%    & AdamW                                           & -                                                                          & 0.1                                                                                     & 1.0                                                                   & -                \\
LLaMA 2~(70B) & 4M & $1.5 \times 10^{-4}$ & yes & cosine decay to 10\% & AdamW & - & 0.1 & 1.0 & - \\
Falcon~(40B) & 2M & $1.85 \times 10^{-4}$ & yes & cosine decay to 10\% & AdamW & BF16 & 0.1 & - & - \\
GLM~(130B)              & 0.4M→8.25M                                                                        & $8 \times 10^{-5}$                                                          & yes             & cosine decay to 10\%    & AdamW                                           & FP16                                                                       & 0.1                                                                                     & 1.0                                                                   & 0.1              \\
T5~(11B)                & 64K                                                                               & $1 \times 10^{-2}$                                                          & no              & inverse square root     & AdaFactor                                       & -                                                                          & -                                                                                       & -                                                                     & 0.1              \\
ERNIE 3.0 Titan~(260B)  & -                                                                                 & $1 \times 10^{-4}$                                                          & -               & -                       & Adam                                            & FP16                                                                       & 0.1                                                                                     & 1.0                                                                   & -                \\
PanGu-$\Sigma$~(1.085T) & 0.5M                                                                              & $2 \times 10^{-5}$                                                          & yes             & -                       & Adam                                            & FP16                                                                       & -                                                                                       & -                                                                     & -                \\
\bottomrule\end{tabular}
}
\end{table*}

\subsection{Model Training}
\label{sec:training_settings}
In this part, we review the important settings, techniques, or tricks for training LLMs. 

\subsubsection{Optimization Setting}
 For parameter optimization of LLMs, we present the commonly used settings for batch training, learning rate, optimizer, and training stability.

\paratitle{Batch Training.}
For language model pre-training, existing work generally sets the batch size to a large number (\eg 2,048 examples or 4M tokens) to improve the training stability and {throughput}.
For LLMs such as GPT-3 and PaLM, they have introduced  a new strategy that dynamically increases the batch size during training, ultimately reaching a million scale. 
Specifically, the batch size of GPT-3 is gradually increasing from 32K to 3.2M tokens.
Empirical results have demonstrated that the dynamic  schedule of batch size can effectively stabilize the training process of LLMs~\cite{Chowdhery-arxiv-2022-PaLM}.

\paratitle{Learning Rate.}
Existing LLMs usually adopt a similar learning rate schedule with the warm-up and decay strategies during pre-training.
Specifically, in the initial 0.1\% to 0.5\% of the training steps, a linear warm-up schedule is employed for gradually increasing the learning rate to the maximum value that ranges from approximately $5 \times 10^{-5}$ to $1 \times 10^{-4}$ (\eg $6 \times 10^{-5}$ for GPT-3).
Then, a cosine decay strategy is adopted in the subsequent steps,  gradually reducing the learning rate to approximately 10\% of its maximum value, until the convergence of the training loss.

\paratitle{Optimizer.}
The Adam optimizer~\cite{Kingma-arXiv-2015-Adam} and  {AdamW optimizer~\cite{Loshchilov-arxiv-2017-Fixing}} are widely utilized for training LLMs (\eg GPT-3), which are based on adaptive estimates of lower-order moments for first-order gradient-based optimization.
Commonly, its hyper-parameters are set as follows: $\beta_1 = 0.9$, $\beta_2 = 0.95$ and $\epsilon = 10^{-8}$.
Meanwhile, the Adafactor optimizer~\cite{Shazeer-ICML-2018-Adafactor} has also been utilized in training LLMs (\eg PaLM and T5), which is a variant of the Adam optimizer specially designed for conserving GPU memory during training.
The hyper-parameters of the Adafactor optimizer are set as: $\beta_1 = 0.9$ and $\beta_2 = 1.0 - k^{-0.8}$, where $k$ denotes the number of training steps.

\paratitle{Stabilizing the Training.}
During the pre-training of LLMs, it often suffers from the  training instability issue, which may cause the model collapse. 
To address this issue, weight decay and gradient clipping have been widely utilized, where existing studies~\cite{Brown-NeurIPS-2020-Language,Zhang-arxiv-2022-OPT,Scao-arxiv-2022-BLOOM,Smith-CoRR-2022-Using,Zeng-arxiv-2022-GLM}  commonly set the threshold of gradient clipping to 1.0 and weight decay rate to 0.1.  
However, with the scaling of LLMs, the training loss spike is also more likely to occur, leading to  unstable training. 
To mitigate this problem, PaLM~\cite{Chowdhery-arxiv-2022-PaLM} and OPT~\cite{Zhang-arxiv-2022-OPT} use a simple strategy that restarts the training process from an earlier checkpoint before the occurrence of the spike and skips over the data that may have caused the problem.
Further, GLM~\cite{Zeng-arxiv-2022-GLM} finds that the abnormal gradients of the embedding layer usually lead to spikes, and proposes  to shrink the embedding layer gradients to alleviate it.

\subsubsection{Scalable Training Techniques} \label{subsub:scalable}
As  the model and data sizes increase, it has become challenging to efficiently train LLMs under a limited computational resource.  
Especially, two primary technical issues are required to be resolved, \ie increasing training throughput and loading larger models into GPU memory.
In this part, we review several widely used approaches in existing work to address the above two challenges, namely 
3D parallelism~\cite{Huang-NeurIPS-2019-GPipe,Harlap-arXiv-2018-PipeDream,Shoeybi-arXiv-2019-Megatron} and mixed precision training~\cite{Micikevicius-arXiv-2017-Mixed}, and also give general suggestions about how to utilize them for training. 

\paratitle{3D Parallelism.} %
{3D parallelism is actually a combination of three commonly used parallel training techniques, namely data parallelism, pipeline parallelism~\cite{Huang-NeurIPS-2019-GPipe,Harlap-arXiv-2018-PipeDream}, and tensor parallelism~\cite{Shoeybi-arXiv-2019-Megatron}\footnote{Model parallelism is a more broader term that includes tensor parallelism and pipeline parallelism in some work~\cite{Shoeybi-arXiv-2019-Megatron}.}.} We next introduce the three parallel training techniques.

$\bullet$ \emph{Data parallelism.}
Data parallelism is one of the most fundamental approaches to improving the training throughput. 
It  replicates the model parameters and optimizer states across multiple GPUs and then distributes the whole training corpus into these GPUs.
In this way, each GPU only needs to process the assigned data for it, and performs the forward and backward propagation to obtain the gradients.
The computed gradients on different GPUs will be further aggregated to obtain the gradients of the entire batch for updating the models in all GPUs.
In this way, as the calculations of gradients are independently performed on different GPUs, the data parallelism mechanism is highly scalable,  %
{enabling the way that increases the number of GPUs to improve training throughput.} 
Furthermore,  this technique is simple in implementation, and most of existing popular deep learning libraries have already implemented data parallelism, such as TensorFlow and PyTorch.

$\bullet$ \emph{Pipeline parallelism.}
Pipeline parallelism aims to distribute the different layers of a LLM into multiple GPUs.
Especially, in the case of a Transformer model, pipeline parallelism loads consecutive layers onto the same GPU, to reduce the cost of transmitting the computed hidden states or gradients between GPUs.
However, a naive implementation of pipeline parallelism may result in a lower GPU utilization rate as each GPU has to wait for the previous one to complete the computation, leading to the unnecessary cost of \emph{bubbles overhead}{~\cite{Huang-NeurIPS-2019-GPipe}}.
To reduce these bubbles in pipeline parallelism, GPipe~\cite{Huang-NeurIPS-2019-GPipe} and PipeDream~\cite{Harlap-arXiv-2018-PipeDream} propose the techniques of padding multiple batches of data and asynchronous gradient update to improve the pipeline efficiency.

$\bullet$ \emph{Tensor parallelism.}
Tensor parallelism is also a commonly used technique that aims to decompose the LLM for multi-GPU loading. 
Unlike pipeline parallelism, tensor parallelism focuses on decomposing the tensors (the parameter matrices) of LLMs.
For a matrix multiplication operation $Y = XA$ in the LLM, the parameter matrix $A$ can be split into two submatrices, $A_1$ and $A_2$, by column, which can be expressed as $Y = [X A_1, X A_2]$.
By placing matrices $A_1$ and $A_2$ on different GPUs, the matrix multiplication operation would be invoked at two GPUs in parallel, and the final result can be obtained by combining the outputs from the two GPUs through across-GPU communication.
Currently, tensor parallelism has been supported in several open-source libraries, \eg Megatron-LM~\cite{Shoeybi-arXiv-2019-Megatron}, and can be extended to higher-dimensional tensors. 
Also, Colossal-AI has  implemented tensor parallelism for higher-dimensional tensors~\cite{Xu-arXiv-2021-An,Wang-ICPP-2022-Tesseract,Bian-arXiv-2021-Maximizing} and proposed sequence parallelism~\cite{Li-arXiv-2021-Sequence}  especially for sequence data, which can further decompose the attention operation of the Transformer model.


\paratitle{Mixed Precision Training.}
In previous PLMs (\eg BERT~\cite{Devlin-NAACL-2019-BERT}), 32-bit floating-point numbers, also known as FP32, have been predominantly used for pre-training.
In recent years, to pre-train extremely large language models, some studies~\cite{Micikevicius-arXiv-2017-Mixed} have started to utilize 16-bit floating-point numbers (FP16), which reduces memory usage and communication overhead. 
Additionally, as popular NVIDIA GPUs (\eg A100) have twice the amount of FP16 computation units as FP32, the computational efficiency of FP16 can be further improved.
However, existing work has  found that FP16 may lead to the loss of {computational accuracy~\cite{Scao-arxiv-2022-BLOOM,Rae-arxiv-2021-Scaling}}, which affects the final model performance.
To alleviate it,  an alternative called \emph{Brain Floating Point (BF16)} has been used for training, which
{allocates more exponent bits and fewer significant bits than FP16.}
For pre-training,  
BF16 generally performs better than FP16 on representation accuracy~\cite{Scao-arxiv-2022-BLOOM}. 

\paratitle{Overall Training Suggestion.} 
In practice, the above training techniques, especially 3D parallelism, are often jointly used to improve the training throughput and  large model loading. 
For instance, researchers have incorporated 8{-way} data parallelism, 4{-way} tensor parallelism, and 12{-way} pipeline parallelism, enabling the training of BLOOM~\cite{Scao-arxiv-2022-BLOOM} on 384 A100 GPUs. %
Currently, open-source libraries like DeepSpeed~\cite{Rasley-KDD-2020-DeepSpeed}, Colossal-AI~\cite{Bian-CoRR-2021-Colossal-AI},
and Alpa~\cite{Zheng-OSDI-2022-Alpa}
can well support the three parallel training methods. 
{To reduce the memory redundancy, ZeRO, FSDP, and activation recomputation techniques~\cite{Chen-arxiv-2016-training,Korthikanti-arxiv-2022-reducing} can be also employed for training LLMs,  which have already been integrated into DeepSpeed, PyTorch, and Megatron-LM. }
In addition, the mixed precision training technique such as BF16 can be also leveraged to improve the training efficiency and reduce GPU memory usage, while it  requires necessary support on hardware (\eg A100 GPU). 
Because training large models is a time-intensive process,
it would be useful to forecast the model performance and detect abnormal issues at an early stage. 
For this purpose, GPT-4~\cite{OpenAI-OpenAI-2023-GPT-4} has recently introduced a new mechanism called \emph{predictable scaling} built on a deep learning stack, enabling the performance prediction of large models with a much smaller model, which might be quite useful for developing LLMs.  
In practice, one can further leverage the supporting training techniques of 
mainstream deep learning  frameworks. 
For instance, PyTorch supports the data parallel training algorithm  FSDP~\cite{FairScale2021} (\ie fully sharded data parallel), which  allows for  partial offloading of training computations to CPUs if desired. 

\section{Post-training of LLMs}
\label{sec-adaptation}

After pre-training, LLMs can acquire the general abilities for  solving various tasks. However, an increasing number of studies have shown  that LLM's abilities can be further adapted according to specific goals.
In this section, we introduce two major approaches to adapting pre-trained LLMs, namely instruction tuning and alignment tuning. The former approach  mainly aims to enhance (or unlock) the abilities of LLMs, while the latter approach aims to align the behaviors of LLMs with human values or preferences. 
Further, we will also discuss  efficient tuning and quantization for  model adaptation in resource-limited settings.  
In what follows, we will introduce the four parts in detail. 

\begin{figure*}[h]
    \centering
    \includegraphics[width=1\textwidth]{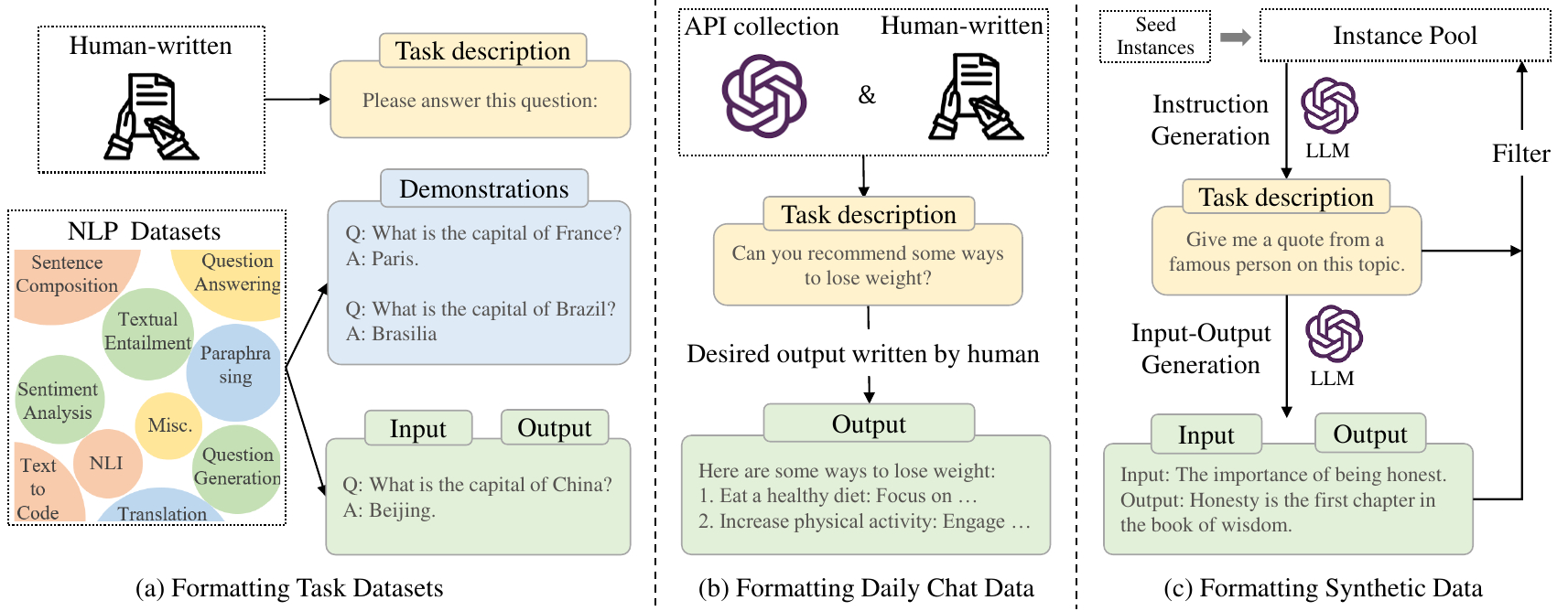}
    \caption{An illustration of instance formatting and three different methods for constructing the instruction-formatted instances. }
    \label{fig:instruction-tuning}
\end{figure*}

\subsection{Instruction Tuning}
\label{sec-instruction}
In essence, instruction tuning is the approach to fine-tuning  pre-trained  LLMs on a collection of formatted instances in the form of natural language~\cite{Wei-ICLR-2022-Finetuned}, which is highly related to supervised fine-tuning~\cite{Ouyang-arxiv-2022-Training} and multi-task prompted training~\cite{Sanh-ICLR-2022-Multitask}. %
In order to perform instruction tuning, we first need to collect or construct instruction-formatted instances. 
Then, we employ these formatted instances to fine-tune LLMs in a supervised learning way (\eg training with the sequence-to-sequence loss). 
After instruction tuning, LLMs can demonstrate superior abilities to generalize to unseen tasks~\cite{Wei-ICLR-2022-Finetuned,Sanh-ICLR-2022-Multitask,Chung-arxiv-2022-Scaling}, even in a multilingual setting~\cite{Muennighoff-2022-arxiv-Crosslingual}.

A recent survey~\cite{Lou-arXiv-2023-Is} presents a systematic overview of the research on instruction tuning. In comparison to that, we mainly focus on the effect of instruction tuning on LLMs and provide detailed guidelines or strategies for instance collection and tuning. In addition, we also discuss the use of instruction tuning for satisfying the real needs of users, which has been widely applied in existing LLMs, \eg InstructGPT~\cite{Ouyang-arxiv-2022-Training} and GPT-4~\cite{OpenAI-OpenAI-2023-GPT-4}.

\subsubsection{Formatted Instance Construction}\label{sec-instruction-formatted} 
Generally, an instruction-formatted instance consists of a task description (called an \emph{instruction}), an optional input, the corresponding output, and a small number of demonstrations (optional).
As important public resources, existing studies have released a large number of labeled data formatted in natural language (see the list of available resources in Table~\ref{tab:instruction-collection}) as introduced in Section~\ref{sec:it-dataset}. 
Next, we introduce four major methods for constructing formatted instances (see an illustration in Figure~\ref{fig:instruction-tuning}) and then discuss several key factors for instance construction.  

\paratitle{Formatting NLP Task Datasets.} 
Before instruction tuning was proposed, several early  studies~\cite{Liu-ACL-2019-Multi,Aghajanyan-EMNLP-2021-Muppet,Tang-arxiv-2022-MVP} collected the instances from a diverse range of traditional NLP tasks (\eg text summarization, text classification, and translation) to create supervised multi-task training datasets. 
As a major source of instruction tuning instances, it is convenient to format these multi-task training  datasets with natural language task descriptions. 
Specifically, recent work~\cite{Wei-ICLR-2022-Finetuned,Sanh-ICLR-2022-Multitask,Ouyang-arxiv-2022-Training,Wang-EMNLP-2022-Super}  augments the labeled datasets with human-written task descriptions, which instructs LLMs to understand the tasks by explaining the task goal.  %
For example, in Figure~\ref{fig:instruction-tuning}(a),  a task description ``\emph{Please answer this question}'' is added for each example in the question-answering task. 
After instruction tuning, LLMs can generalize well to other unseen tasks by following their task descriptions~\cite{Wei-ICLR-2022-Finetuned,Sanh-ICLR-2022-Multitask,Chung-arxiv-2022-Scaling}. 
In particular, it has been shown that instructions are the crucial factor in task generalization ability for LLMs~\cite{Wei-ICLR-2022-Finetuned}:  
by fine-tuning the model on labeled datasets with the task descriptions removed, it results in a dramatic drop in model performance.
To better generate labeled instances for instruction tuning, a crowd-sourcing platform, PromptSource~\cite{Bach-ACL-2022-PromptSource} has been proposed to effectively create, share, and verify the task descriptions for different datasets.  
To enrich the training instances, several studies~\cite{Sanh-ICLR-2022-Multitask,Tang-arxiv-2022-MVP,Longpre-arxiv-2023-The} also try to invert the input-output pairs of existing instances with specially designed task descriptions for instruction tuning. For instance, given a question-answer pair, we can create a new instance by predicting the answer-conditioned question (\eg \emph{``Please generate a question based on the answer:''}). 

\paratitle{Formatting Daily Chat Data.}
Despite that a large number of training instances have been formatted with instructions, they mainly come from public NLP datasets, either lacking instruction diversity or mismatching with real human needs~\cite{Ouyang-arxiv-2022-Training}.   
To overcome this issue, InstructGPT~\cite{Ouyang-arxiv-2022-Training} proposes to take the queries that real users have submitted to the OpenAI API as the task descriptions.  %
Additionally, to enrich the task diversity, human labelers are also asked to compose the instructions for real-life tasks, including open-ended generation, open question answering, brainstorming, and chatting. 
Then, they let another group of labelers directly answer these instructions as the output.
Finally, they pair one instruction (\ie the collected user query) and the expected output  (\ie the human-written answer) as a training instance.
Note that InstructGPT also employs these real-world tasks formatted in natural language for alignment tuning (discussed in Section~\ref{sec-alignment}). 
Further, GPT-4~\cite{OpenAI-OpenAI-2023-GPT-4} has designed potentially high-risk instructions and guided the model to reject these instructions through supervised fine-tuning for safety concerns. 
{Considering the absence of high-quality public chat data, several studies have also collected  users' chat requests as input data,  and then utilized ChatGPT or GPT-4 to generate responses as output data. A notable example of such a dataset is the conversational data from ShareGPT~\cite{ShareGPT}. Additionally, Dolly~\cite{Conover-2023-arxiv-Dolly} and OpenAssistant~\cite{kopf-arxiv-2023-openassistant} have further released their conversation data, which has been carefully labeled by human annotators to attain a high level of quality.}



\paratitle{Formatting Synthetic Data.}
To reduce the burden of human annotation or manual collection, several semi-automated approaches~\cite{Wang-arXiv-2022-Self} have been proposed for constructing instances by feeding existing instances into LLMs to synthesize diverse task descriptions and {instances}. As illustrated in Figure~\ref{fig:instruction-tuning}(c), the Self-Instruct method only needs 175 instances as the initial task pool. Then, they randomly select a few instances from the pool as demonstrations and prompt a LLM to generate new instructions and corresponding input-output pairs. After the quality and diversity filtering, newly generated instances would  be added into the task pool. Hence, the synthetic method is an effective and economical way to generate large-scale instruction data for LLMs. However, the instances generated by the Self-Instruct method might be  simplistic or lack the diversity. To improve the quality of  synthetic instructions,  WizardLM~\cite{Xu-arxiv-2023-WizardLM} introduces Evol-Instruct by proposing in-depth and in-breadth evolving to enrich the complexity and diversity of the instances. Furthermore, Self-Align~\cite{Sun-arxiv-2023-Principle} establishes multiple human-aligned principles to filter the synthesized instances. It then employs these instances to train a LLM in order to yield more aligned instances. To enhance the quality of the instance output, researchers directly adopt human-written texts as the output and synthesize corresponding instructions using ICL examples~\cite{Li-arxiv-2023-Self}.




\paratitle{Key Factors for Instruction Dataset Construction.}
The quality of instruction instances has an important impact on the performance of the model. 
Here, we discuss some essential factors for instance construction. 

$\bullet$ \emph{Scaling the instructions.} 
It has been widely shown that scaling the number of tasks can largely enhance the generalization ability of LLMs~\cite{Wei-ICLR-2022-Finetuned,Sanh-ICLR-2022-Multitask,Wang-EMNLP-2022-Super}. 
With the increasing of the task number, the model performance initially shows a continuous growth pattern, while the gain becomes negligible when it reaches a certain level~\cite{Wang-EMNLP-2022-Super,Chung-arxiv-2022-Scaling}. 
A plausible speculation is that a certain number of representative  tasks can provide relatively sufficient  knowledge and adding more tasks  may not bring additional gains~\cite{Chung-arxiv-2022-Scaling}.
Also,  it is  beneficial to enhance the diversity of the task descriptions in several aspects, such as length, structure, and creativity~\cite{Sanh-ICLR-2022-Multitask}.
As for the number of instances per task, it has been found that a small number of instances can usually saturate the generalization performance of the model to perform a specific task~\cite{Wei-ICLR-2022-Finetuned,Chung-arxiv-2022-Scaling}.
{Specially, several recent work~\cite{zhou-arxiv-2023-lima,Chen-arxiv-2023-AlpaGasus} has explored the effect of fine-tuning with a small amount of high-quality instruction data (\eg one or a few thousand instances),  showing very promising results on the evaluation tasks.  
In contrast, another line of studies continue to explore the scaling effect of instruction data~\cite{Mukherjee-arxiv-2023-Orca,YuLan-Chat}. For example, Orca~\cite{Mukherjee-arxiv-2023-Orca} scales up the synthesized instances to 5 million with step-by-step explanations, and it achieves superior performance across a wide range of tasks.}



$\bullet$ \emph{Formatting design.}
As an important factor, the design of natural language format also highly impacts the generalization performance of LLMs~\cite{Wang-EMNLP-2022-Super}.
Typically, we can add task descriptions and optional demonstrations to the input-output pairs of existing datasets, where the task description is the most key part for LLMs to understand the task~\cite{Wang-EMNLP-2022-Super}. 
Further, it can lead to substantial improvements by using an appropriate number of exemplars as demonstrations~\cite{Chung-arxiv-2022-Scaling}, which also alleviates the model sensitivity to instruction engineering~\cite{Wei-ICLR-2022-Finetuned,Chung-arxiv-2022-Scaling}. 
However, incorporating other components (\eg things to avoid, reasons, and suggestions) into instructions may have a negligible or even adverse effect on the performance of LLMs~\cite{Mishra-ACL-2022-Cross, Wang-EMNLP-2022-Super}. 
Recently, to elicit the step-by-step reasoning ability of LLMs, some work~\cite{Chung-arxiv-2022-Scaling}  proposes to include chain-of-thought (CoT) examples for some reasoning datasets, such as arithmetic reasoning.
It has been shown that fine-tuning LLMs with both CoT and non-CoT examples can lead to a good performance across various reasoning tasks,  including those that require multi-hop reasoning ability (\eg commonsense question answering and arithmetic reasoning) as well as those without the need for such a reasoning  way (\eg sentiment analysis and  extractive question answering)~\cite{Chung-arxiv-2022-Scaling,Iyer-arxiv-2022-OPT}.

$\bullet$ {\emph{Instruction quality improvement.}}
Data quality is very important for the performance of instruction tuning, and a surge of work has been proposed to further improve the quality of existing instruction datasets.
Typically, these methods mostly rely on carefully designed prompts, to guide LLMs to refine or rewrite the given instruction.
WizardLM~\cite{Xu-arxiv-2023-WizardLM} aims to complexify and diversify the Alpaca dataset~\cite{alpaca} by devising prompts to widen and deepen the required knowledge of given instructions.
It also crafts the filter strategy to remove the low-quality instructions.
To further provide fine-grained knowledge guidance, recent work also involves the knowledge taxonomy into the input prompt, \eg knowledge key points~\cite{Huang-arxiv-2024-Key} and the human-AI conversation topic taxonomy~\cite{Ding-2023-EMNLP-Enhancing}. 
To guarantee the instruction quality, early methods mainly employ close-source API or powerful open-source LLMs, which would take a huge cost for large-scale instructions synthesis. 
Considering this issue, recent studies widely explore the potential of relatively small models for data synthesis. For instance, 
JiuZhang3.0~\cite{Zhou-arxiv-2024-Jiuzhang3.0} fine-tunes a 7B language model to synthesize questions by distilling the knowledge from GPT-4, and then utilizes it to synthesize massive high-quality instructions based on pre-training corpus.
Such a way can achieve better performance on mathematical reasoning tasks than baseline methods, with only 20\% data synthesis cost.

$\bullet$ {\emph{Instruction selection.}}
{As a surge of instruction datasets are proposed, it is non-trivial to select the high-quality ones from them to construct the training dataset.
Generally, existing work either leverages quality estimation metrics or employs LLMs as the judge model to rank all the instruction instances, and then selects those with relatively higher scores.
Concretely, for metrics, perplexity and other heuristic measurements (\eg length)~\cite{gao-arxiv-2023-instructionmining} have been widely used in practice, \eg  we can  consider removing high-perplexity or very short instructions, which might correspond to low-quality ones.
To better estimate the effect of an instruction for the LLM capability, more complex metrics (\eg IFD~\cite{li-arxiv-2023-boosting}) have also been proposed, which are computed by combining multiple simple metrics.
{
Additionally, diversity-aware sampling methods have been introduced to ensure the overall coverage of representative instruction data~\cite{iclr-2018-sener-active}.
Besides, when downstream task data is available, cross-instance gradient similarity can be employed to measure the value of training instances for the target task. 
LESS~\cite{arxiv-2024-xia-less} computes gradients for both downstream validation and training instruction data, to evaluate the contribution of instruction data based on extensions of influence function~\cite{koh-2017-understanding}.  
}
}

To summarize, diversity and quality of instructions are important factors to consider when scaling the number of instances~\cite{zhou-arxiv-2023-lima}. 
As the capacities of LLMs improve, data synthesis methods have become the mainstream approach for generating large amount of instruction data. 
Following this trend, there are increasingly more automatically generated instruction datasets available, and  selection and refining methods are key to effectively use these  datasets. 
To help readers understand how different factors affect instruction tuning,  we conduct an empirical study by experimenting with multiple specially constructed  instruction datasets in Section~\ref{instruction-results}.



\begin{table*}[t]
    \centering
    \caption{Basic statistics of the required number of GPUs, tuning time, batch size (denoted as BS) per device (full tuning and LoRA tuning), and inference rate (the number of generated tokes per second). Our experiments are conducted based on two Linux servers having 8 A800-80G SXM4 GPUs with 6 NVSwitch and 8 3090-24G GPUs, respectively. {The major difference between A800 and A100 lies in the NVLink interconnect speed. Thus, our estimations about training and inference efficiency would be slightly improved for A100, while the rest memory consumption would remain the same.}  
    {
    For full tuning experiments, we use data parallel training, ZeRO Stage 3, BF16, and gradient checkpointing. Additionally, the LoRA tuning can be executed on one 80G GPU utilizing INT8 quantization with the rank setting set to 16. {All the experiments are conducted with Alpaca-52K dataset by  training LLaMA models three epochs.} The max sequence length for both training settings is set to 512. The inference experiments are performed with the batch size set to 1.}
    }
    \renewcommand\tabcolsep{2.5pt}
    \begin{tabular}{c|ccc|ccc|cc|cc|cc}
    \toprule
    \multirow{2}{*}{\textbf{Models}} & \multicolumn{3}{c|}{\textbf{A800 Full Tuning}} & \multicolumn{3}{c|}{\textbf{A800 LoRA Tuning}} & \multicolumn{2}{c|}{\textbf{A800 Inference (16-bit)}} & \multicolumn{2}{c|}{\textbf{3090 Inference (16-bit)}} & \multicolumn{2}{c}{\textbf{3090 Inference (8-bit)}} \\
    & \#GPU & BS & Time & \#GPU & BS & Time & \#GPU & \#Token/s & \#GPU & \#Token/s & \#GPU & \#Token/s \\
    \midrule
    LLaMA (7B)  &  2 & 8 & 3.0h  &  1 & 80 & 3.5h  & 1 & 36.6 & 1 & 24.3 & 1 & 7.5 \\
    LLaMA (13B) &  4 & 8 & 3.1h  &  1 & 48 & 5.1h  & 1 & 26.8 & 2 & 9.9  & 1 & 4.5 \\
    LLaMA (30B) &  8 & 4 & 6.1h  &  1 & 24 & 14.3h & 1 & 17.7 & 4 & 3.8  & 2 & 2.6 \\
    LLaMA (65B) & 16 & 2 & 11.2h &  1 & 4  & 60.6h & 2 & 8.8  & 8 & 2.0  & 4 & 1.5  \\
    \bottomrule
    \end{tabular}
    \label{tab:instruction-time}
\end{table*}

\subsubsection{Instruction Tuning Strategies}
\label{sec-ituning-strategy}
Unlike pre-training, instruction tuning is often more efficient since only a moderate number of instances are used for training. 
Since instruction tuning can be considered as a supervised training process, its optimization is different from  pre-training in several aspects~\cite{Chung-arxiv-2022-Scaling},   
{such as the training objective (\ie sequence-to-sequence loss) and optimization configuration (\eg smaller batch size and learning rate)}, which require special attention in practice. 
In addition to these  optimization configurations, there are also four important aspects to consider  for instruction tuning:

\paratitle{Balancing the Data Distribution.} 
Since instruction tuning involves a mixture of different tasks, it is important to balance the proportion of different tasks during fine-tuning. A widely used method is the \emph{examples-proportional mixing} strategy~\cite{Raffel-JMLR-2020-Exploring}, \ie combining all the datasets and sampling each instance equally from the mixed datasets. Furthermore, increasing the sampling ratio of high-quality collections (\eg FLAN~\cite{Wei-ICLR-2022-Finetuned} and P3~\cite{Bach-ACL-2022-PromptSource}) can generally lead to performance improvement according to recent findings~\cite{Chung-arxiv-2022-Scaling,Iyer-arxiv-2022-OPT}. Further, it is common to set a \emph{maximum cap} to control the maximum number of examples that  a dataset can contain during instruction tuning~\cite{Raffel-JMLR-2020-Exploring}, which is set to prevent larger datasets from overwhelming the entire distribution~\cite{Raffel-JMLR-2020-Exploring,Iyer-arxiv-2022-OPT}. In practice, the maximum cap is typically set to several thousands or tens of thousands according to different datasets~\cite{Wei-ICLR-2022-Finetuned,Chung-arxiv-2022-Scaling}. 
{Recently, it has been 
empirically found  that existing instruction datasets (Table~\ref{tab:instruction-collection}) mainly focus on  enhancing LLMs' capabilities in certain aspects,  
and a single dataset alone cannot lead to a comprehensive enhancement in model capacity~\cite{Wang-arxiv-2023-How}. Therefore, it is often  suggested to use a mixture of existing instruction datasets to achieve a balanced improvement in different capacities, including NLP task data (\eg FLAN v2~\cite{Liu-arxiv-2023_scaling}), chat data (\eg ShareGPT~\cite{ShareGPT}), and synthetic data (\eg GPT4-Alpaca~\cite{Peng-23-arxiv-Instruction}).}




\paratitle{Combining Instruction Tuning and Pre-Training.} 
To make the tuning process more effective and stable, OPT-IML~\cite{Iyer-arxiv-2022-OPT} incorporates pre-training data during instruction tuning, which can be regarded as regularization for model tuning. 
{Further, instead of using a separate two-stage process (\emph{pre-training} then \emph{instruction tuning}), some studies  attempt to train a model from scratch with a mixture of pre-training data (\ie plain texts) and instruction tuning data (\ie formatted  datasets)} using multi-task learning~\cite{Raffel-JMLR-2020-Exploring}. Specifically, GLM-130B~\cite{Zeng-arxiv-2022-GLM} and Galactica~\cite{Taylor-arxiv-2022-Galactica} integrate instruction-formatted datasets as a small proportion of the pre-training corpora to pre-train LLMs, which potentially achieves the advantages of pre-training and instruction tuning at the same time.

\paratitle{Multi-stage Instruction Tuning.}
For instruction tuning, there are two kinds of important instruction data, namely task-formatted instructions and daily chat instructions. 
Generally, the former has a significantly larger volume than the latter. 
It is important to balance the training with the two kinds of  instruction data. In addition to carefully mixing different instruction data, we can also adopt a multi-stage instruction tuning strategy~\cite{YuLan-Chat}, where LLMs are first fine-tuned with large-scale task-formatted instructions and subsequently fine-tuned on daily chat ones.
To avoid the capacity forgetting issue, it is also useful to add an amount of  task-formatted instructions at the second stage. 
Actually, such a multi-stage tuning strategy can be also applied to other settings for instruction tuning. For example, we can schedule different fine-tuning stages with progressively increased  levels on difficulty and complexity, and gradually improve the capacities of LLMs to follow complex instructions.

\ignore{To improve the capacities of LLMs for generalizing to unseen tasks and handling open-ended generation, task-formatted and daily chat instructions are commonly used for tuning. However, high-quality daily chat instructions are much fewer than task-formatted ones (shown in Table~\ref{tab-instructions}), hence a simple mixture of these instructions would drown the daily chat ones, affecting the chatting performance.
Thus, it is more reasonable to perform the multi-stage instruction tuning strategy~\cite{YuLan-Chat}, where LLMs are first tuned by large-scale task-formatted instructions, and then tuned on daily chat ones.
To avoid forgetting the task generalization capacity, it is also feasible to add a small subset of the task-formatted instructions in the second stage.
Further, it is also applicable to progressively improve the difficulty and complexity of the instructions in the multi-stage tuning process, which can gradually improve the capacities of LLMs to follow complex instructions.}

\paratitle{Other Practical Tricks.} 
{In practice, there are also several useful strategies and tricks that are helpful to improve the fine-tuning performance of LLMs. We  list several  representative ones as follows:}

$\bullet$ \emph{Efficient training for multi-turn chat data.}
Given a multi-turn chat example (the conversation between a user and chatbot), a straightforward fine-tuning way is to split it into multiple context-response pairs for training: a LLM is fine-tuned to generate the response based on the corresponding context for all splits (\ie at each utterance from the user). In such a fine-tuning way, it is apparent that there exist  overlapping utterances in the split examples from a conversation. 
To save the training cost, Vicuna~\cite{vicuna2023} has adopted an efficient way that feeds the whole conversation into the LLM, but relies on a loss mask that only computes the loss on the responses of the chatbot for training.
It can significantly reduce the compute costs derived from the overlapped utterances.


$\bullet$ \emph{Establishing self-identification for LLM.}
To deploy  LLMs for real-world applications, it is necessary to establish its identity and make LLMs aware of these identity information, such as name, developer and affiliation. 
A practical  way is to create identity-related instructions for fine-tuning the LLM. It is also feasible to prefix the input with the self-identification prompt, \eg ``\emph{The following is a conversation between a human and an AI assistant called \textsc{ChatbotName}, developed by \textsc{Developer}.}'', where \textsc{ChatbotName} and \textsc{Developer} refer to the name and developer of the chatbot, respectively.

In addition to the above practical strategies and tricks, existing work has also used other tricks, \eg concatenating multiple examples into a single sequence to approach the max length~\cite{Krell-2021-arxiv-efficient}.


\subsubsection{The Effect of Instruction Tuning}\label{subsec:effectIT}

In this part, we discuss the effect of instruction tuning on LLMs in three major aspects.

\paratitle{Performance Improvement.}
Despite being tuned on a moderate  number of instances, instruction tuning has become an important way to improve or unlock the abilities of LLMs~\cite{Chung-arxiv-2022-Scaling}.  %
Recent studies have experimented with language models in  multiple scales (ranging from  77M to 540B), showing that the models of different scales can all benefit from instruction tuning~\cite{Chung-arxiv-2022-Scaling,Longpre-arxiv-2023-The}, yielding improved performance as the parameter scale  increases~\cite{Muennighoff-2022-arxiv-Crosslingual}.  
Further, smaller models with instruction tuning can even perform better than larger models without fine-tuning~\cite{Sanh-ICLR-2022-Multitask,Chung-arxiv-2022-Scaling}. 
Besides the model scale, instruction tuning demonstrates consistent improvements in various model architectures, pre-training objectives, and model adaptation methods~\cite{Chung-arxiv-2022-Scaling}.
In practice, instruction tuning offers  %
{a general approach to enhancing the abilities of existing language models~\cite{Chung-arxiv-2022-Scaling} (including small-sized PLMs). Also, it is much less costly than pre-training, since the amount of  instruction data required by LLMs is significantly smaller than pre-training data.}

\paratitle{Task Generalization.}
Instruction tuning encourages the model to understand natural language instructions for task completion. 
It endows LLMs with the ability (often considered as an emergent ability) to follow human instructions~\cite{Wei-arxiv-2022-Emergent} to perform specific tasks without demonstrations, even on unseen tasks~\cite{Chung-arxiv-2022-Scaling}.  
A large number of studies have confirmed the effectiveness of instruction tuning to achieve superior performance on both seen and unseen tasks~\cite{Iyer-arxiv-2022-OPT,Longpre-arxiv-2023-The}. 
Also, instruction tuning has been shown to be useful in alleviating several weaknesses of LLMs (\eg repetitive generation or complementing the input without accomplishing a certain task)~\cite{Ouyang-arxiv-2022-Training,Chung-arxiv-2022-Scaling}, leading to a superior capacity to solve real-world tasks for LLMs. Furthermore, LLMs trained with instruction tuning can generalize to related tasks across languages. For example, BLOOMZ-P3~\cite{Muennighoff-2022-arxiv-Crosslingual} is fine-tuned based on BLOOM~\cite{Scao-arxiv-2022-BLOOM} using English-only task collection P3~\cite{Bach-ACL-2022-PromptSource}. Interestingly, BLOOMZ-P3 can achieve a more than 50\% improvement in multilingual sentence completion tasks compared to BLOOM, which shows that instruction tuning can help LLMs acquire general task skills from English-only datasets and transfer such skills into other languages~\cite{Muennighoff-2022-arxiv-Crosslingual}.
In addition, it has been found  that using English-only instructions can produce satisfactory results on multilingual tasks~\cite{Muennighoff-2022-arxiv-Crosslingual}, which helps reduce the effort of instruction engineering for a specific language. 

\paratitle{Domain Specialization.}
Existing LLMs have showcased superior capabilities in traditional NLP tasks (\eg generation and reasoning) and daily questions.  However, they may still lack domain knowledge to accomplish specific tasks, such as medicine, law, and finance (See Section~\ref{sec-application} for a detailed discussion of LLMs in different applications). Instruction tuning is an effective approach to adapting existing general LLMs to be domain-specific experts.
For instance, researchers propose to fine-tune Flan-PaLM~\cite{Chung-arxiv-2022-Scaling} using medical datasets to create Med-PaLM~\cite{singhal-arxiv-2022-large}, a medical knowledge assistant that achieves performance levels comparable to those of expert clinicians.
Furthermore, a recent study~\cite{Zhang-2023-arxiv-recommendation} fine-tunes FLAN-T5 to support e-commerce recommender systems with natural language instructions, showing strong performance in a variety of recommendation tasks.  
There are also several open-sourced medical models instruction-tuned based on LLaMA~\cite{Touvron-arxiv-2023-LLaMA}, such as  BenTsao~\cite{wang-arxiv-2023-huatuo}.
Also, researchers  explore instruction tuning on law~\cite{huang-arxiv-2023-lawyer}, finance~\cite{wu-arxiv-2023-bloomberggpt}, and arithmetic computation~\cite{liu-arxiv-2023-goat}. 

\begin{table*}[htb]
    \centering
    \caption{Results of instruction-tuning experiments (all in a single-turn conversation) based on the LLaMA (7B) and LLaMA (13B) model under the chat  and QA setting. We employ four instruction improvement strategies on the Self-Instruct-52K dataset, \ie enhancing the complexity (\emph{w/ complexity}), increasing the diversity (\emph{w/ diversity}), balancing the difficulty (\emph{w/ difficulty}), and scaling the instruction number (\emph{w/ scaling}). $^*$Since we select the LLaMA (7B)/(13B) model fine-tuned on Self-Instruct-52K as the baseline, we omit the win rate of the fine-tuned model with Self-Instruct-52K against itself.}
    \label{tab-instruction-tuning-res}
\resizebox{1.6\columnwidth}{!}{
\begin{tabular}{llrcH|Hccc}
\toprule
\multirow{2.5}{*}{\textbf{Models}}   & \multirow{2.5}{*}{\begin{tabular}[c]{@{}c@{}}\textbf{Dataset}\\ \textbf{Mixtures}\end{tabular}} &  \multirow{2.5}{*}{\begin{tabular}[c]{@{}c@{}}\textbf{Instruction}\\ \textbf{Numbers}\end{tabular}}& \multirow{2.5}{*}{\begin{tabular}[c]{@{}c@{}}\textbf{Lexical}\\ \textbf{Diversity}\end{tabular}} & \multirow{2.5}{*}{\begin{tabular}[c]{@{}c@{}}\textbf{Topic}~($\uparrow$)\\ \textbf{Diversity}\end{tabular}} & \multicolumn{2}{c}{\textbf{Chat}} & \multicolumn{2}{c}{\textbf{QA}} \\ 
\cmidrule(r){6-7}\cmidrule(r){8-9}
& & & & & Human$^*$ & AlpacaFarm & MMLU & BBH3k \\
\midrule
LLaMA~(7B) & \ding{172}~FLAN-T5 & 80,000 & 48.48 & 26.79 & - & 23.77 & \cellcolor[HTML]{C4DDEC}{38.58} & \cellcolor[HTML]{A7CBE2}{32.79}\\
 & \ding{173}~ShareGPT & 63,184 & 77.31 & 28.86 & - & \cellcolor[HTML]{A7CBE2}{81.30} & \cellcolor[HTML]{E5F0F7}{38.11} & 27.71\\
 & \ding{174}~Self-Instruct-52K & 82,439 & 25.92 & 23.41 & / & /$^*$ & 37.52 & \cellcolor[HTML]{E5F0F7}{29.81} \\
 & \ding{173} + \ding{174} & 145,623 & 48.22 & 26.89 & - & \cellcolor[HTML]{E5F0F7}{71.36} & \cellcolor[HTML]{A7CBE2}{41.26} & 28.36\\
 & \ding{172} + \ding{173} + \ding{174} & 225,623 & 48.28 & 27.32 & - & 70.00 & \cellcolor[HTML]{92BFDB}{43.69} & 29.69\\
\cmidrule{2-9}
 & \ding{174}~Self-Instruct-52K & 82,439 & 25.92 & 23.41 & / & /$^*$ & 37.52 & 29.81\\
 & \makecell[l]{w/ complexity} & 70,000 & 70.43 & 27.97 & - & \cellcolor[HTML]{C4DDEC}{76.96}  & \cellcolor[HTML]{C6DEED}{39.73} & \cellcolor[HTML]{92BFDB}{33.25}\\
 & \makecell[l]{w/ diversity} & 70,000 & 75.59 & 26.10 & - & \cellcolor[HTML]{92BFDB}{81.55}  & 38.01 & \cellcolor[HTML]{C4DDEC}{30.03}\\
 & \makecell[l]{w/ difficulty} & 70,000 & 73.48 & 20.77 & - & \cellcolor[HTML]{C6DEED}{79.15}  & 32.55 & \cellcolor[HTML]{C6DEED}{31.25}\\
 & \makecell[l]{w/ scaling} & 220,000 & 57.78 & 23.78 & - & 51.13  & 33.81 & 26.63\\
 \midrule
LLaMA~(13B) & \ding{172}~FLAN-T5 & 80,000 & 48.48 & 26.79 & - & 22.12 & 34.12 & \cellcolor[HTML]{C4DDEC}{34.05}\\
 & \ding{173}~ShareGPT & 63,184 & 77.31 & 28.86 & - & \cellcolor[HTML]{C4DDEC}{77.13} & \cellcolor[HTML]{92BFDB}{47.49} & \cellcolor[HTML]{E5F0F7}{33.82}\\
 & \ding{174}~Self-Instruct-52K & 82,439 & 25.92 & 23.41 & / & /$^*$ & 36.73 & 25.43 \\
 & \ding{173} + \ding{174} & 145,623 & 48.22 & 26.89 & - & \cellcolor[HTML]{E5F0F7}{72.85} & 41.16 & 29.49\\
 & \ding{172} + \ding{173} + \ding{174} & 225,623 & 48.28 & 27.32 & - & 69.49 & \cellcolor[HTML]{C4DDEC}{43.50} & 31.16\\
\cmidrule{2-9}
 & \ding{174}~Self-Instruct-52K & 82,439 & 25.92 & 23.41 & / & /$^*$ & 36.73 & 25.43\\
 & \makecell[l]{w/ complexity} & 70,000 & 70.43 & 27.97 & - & \cellcolor[HTML]{C6DEED}{77.94} & \cellcolor[HTML]{A7CBE2}{46.89} & \cellcolor[HTML]{A7CBE2}{35.75}\\
 & \makecell[l]{w/ diversity} & 70,000 & 75.59 & 26.10 & - & \cellcolor[HTML]{A7CBE2}{78.92} & \cellcolor[HTML]{C6DEED}{44.97} & \cellcolor[HTML]{92BFDB}{36.40}\\
 & \makecell[l]{w/ difficulty} & 70,000 & 73.48 & 20.77 & - & \cellcolor[HTML]{92BFDB}{80.45} & \cellcolor[HTML]{E5F0F7}{43.15} & \cellcolor[HTML]{C6DEED}{34.59}\\
 & \makecell[l]{w/ scaling} & 220,000 & 57.78 & 23.78 & - & 58.12 & 38.07 & 27.28\\
\bottomrule
\end{tabular}
}
\end{table*}

\subsubsection{Empirical Analysis for Instruction Tuning} \label{instruction-results}

Fine-tuning LLMs with different instruction sets tend to  lead to model  variants with varied performance on downstream tasks.  In this section, we will explore the effect of different types of instructions in  fine-tuning LLMs (\ie LLaMA (7B) and LLaMA (13B)\footnote{Due to the limit of computational resources, we cannot conduct large-scale experiments on larger LLaMA variants right now, which would be scheduled in a future version.}), as well as examine the usefulness of several instruction improvement strategies.

\paratitle{Instruction Datasets.} According to the discussion in  Section~\ref{sec-instruction-formatted}, we mainly consider three common kinds of instructions as follows: 

\textbullet~\emph{Task-specific instructions.} For the first type of   instructions, we adopt the most commonly-used multi-task  instruction dataset, \emph{FLAN-T5}~\cite{Chung-arxiv-2022-Scaling}, which contains 1,836 tasks and over 15M instructions by combining four data mixtures from prior work.

\textbullet~\emph{Daily chat instructions.} This type of instructions are conversations posed by  users about daily life, which are more closely related to real-life scenarios. We adopt the  ShareGPT instruciton set, consisting of 63K real-user instructions. It has been used as the core instructions for  Vicuna.

\textbullet~\emph{Synthetic instructions.} In addition to reusing existing instructions, we can also automatically synthesize massive instructions using LLMs. We adopt the popular synthetic instruction dataset Self-Instruct-52K~\cite{Wang-arXiv-2022-Self}, consisting of  52K instructions paired with about 82K instance inputs and outputs. These generated instructions have a similar data distribution as the human-written seed tasks (\eg grammar checking, brainstorming).

{As the original FLAN-T5 dataset is very large (\ie over 15M), we randomly sample 80,000 instructions from it for conducting a fair comparison with other instruction datasets (\ie ShareGPT and Self-Instruct-52K) at a similar scale.
In our experiments, we test on each individual instruction set to explore their own effects and also examine their combinatorial effects on model performance. 
}


\paratitle{Improvement Strategies.} Although real-world instructions from human users are more suitable for fine-tuning LLMs, it is difficult to collect them at a large scale. 
As alternatives to human-generated instructions, most existing research mainly adopts synthetic instructions generated by LLMs.
However, there are some potential  problems with synthetic instructions,  such as poor topic diversity and uneven instruction difficulty (either too simple or too difficult). 
Thus, it is necessary to improve the quality of the synthetic instructions.
Next, we summarize four major improvement strategies widely used in existing work as follows:

\textbullet~\emph{Enhancing the instruction complexity.}
As discussed in existing work~\cite{Xu-arxiv-2023-WizardLM}, enhancing the complexity of instructions can improve the model capacity of LLMs in following complex instructions, 
{\eg including more task demands or requiring more reasoning steps.}
To validate this strategy, we follow WizardLM~\cite{Xu-arxiv-2023-WizardLM} by gradually increasing the complexity levels, \eg adding constraints, increasing reasoning steps, and complicating the input. 
{We leverage the publicly released WizardLM-70K instructions~\cite{Xu-arxiv-2023-WizardLM} as the complexity-enhanced instruction dataset, which has been generated via the above enhancement approach based on the Self-Instruct-52K dataset~\cite{Xu-arxiv-2023-WizardLM}.}

\textbullet~\emph{Increasing the topic diversity.}
In addition to the complexity, 
improving the topic diversity of the instruction dataset  
{can help elicit different abilities of LLMs on  diverse tasks in real world~\cite{Sun-arxiv-2023-Principle}.}
However,  it is difficult to directly control the self-instruct process for generating diverse instructions. Following YuLan-Chat~\cite{YuLan-Chat}, we employ ChatGPT to rewrite the instructions from Self-Instruct-52K dataset for adapting them into 293 topics via specific prompts.
Finally, we obtain 70K instructions as the diversity-increased dataset.

\textbullet~\emph{Scaling the instruction number.}
In addition to the above aspects, the number of instructions is also an important factor that may affect the model performance.
Specially, using more instructions can extend the task knowledge and improve the ability of instruction following for LLMs~\cite{Chung-arxiv-2022-Scaling}. 
To examine this strategy, we sample new instructions from the synthesized instruction set released from the MOSS project~\cite{sun2023moss}, {as they are also synthesized using the same self-instruct method~\cite{Wang-arXiv-2022-Self}.}
We mix them with the Self-Instruct-52K dataset to compose a larger one containing 220K instructions. 

\textbullet~\emph{Balancing the instruction difficulty.}
As the synthetic instructions tend to contain too easy or too hard ones, it is likely to result in training instability or even overfitting for LLMs. To explore the  potential effects, we leverage the perplexity score of LLMs to estimate the difficulty of instructions and remove  too easy or too hard instructions.  
To generate the same scale  of instructions for fair comparison, we adopt a LLaMA (7B) model to compute the perplexity for the 220K instructions from the large instruction dataset, and then keep 70K instructions of moderate perplexity scores as the difficulty-balanced dataset.

\paratitle{Experimental Setup.} 
To conduct the experiments on the effect of instruction data, we leverage these new instruction datasets for tuning LLaMA, a popular LLM backbone that has been widely used for instruction-tuning. 
We use the code from YuLan-Chat~\cite{YuLan-Chat} for our experiments,  and train LLaMA 7B and 13B on a server of 8 A800-80G GPUs. 
All the hyper-parameters settings remain the same as Stanford Alpaca.
To better evaluate the  {instruction following ability of fine-tuned models, we consider two settings, namely 
\emph{Chat setting} and \emph{QA setting}. 
 The chat setting mainly utilizes user instructions and queries from daily chat, whereas the QA setting mainly employs question answering examples from existing NLP datasets. }
The evaluation on the chat setting is conducted based on the AlpacaFarm evaluation set~\cite{Dubois-arxiv-2023-AlpacaFarm}. 
Instead of using a full pairwise comparison,  
{we select the LLaMA 7B and 13B models fine-tuned on Self-Instruct-52K as the reference baselines, and then compare them with other fine-tuned LLaMA 7B and 13B models using different instructions, respectively.} Since our focus is to examine the usefulness of different strategies to generate the instructions, the model fine-tuned on Self-Instruct-52K can serve as a good reference.  
Following AlpacaFarm~\cite{Dubois-arxiv-2023-AlpacaFarm}, for each comparison,  we employ ChatGPT to automatically annotate which response from two compared models each time is the best for the user query, and report the win rate (\%) as the evaluation metric.
For the QA setting, we select two benchmarks, MMLU~\cite{Hendrycks-ICLR-2021-Measuring} and BBH~\cite{Suzgun-arxiv-2022-Challenging}, and evaluate the accuracy based on their default settings by using heuristic rules to parse the answers from these LLMs.

For both instruction tuning and evaluation, we adopt the following prompt: ``\emph{The following is a conversation between a human and an AI assistant. The AI assistant gives helpful, detailed, and polite answers to the user's questions.$\backslash$n [\textbar Human\textbar]:\{input\}$\backslash$n[\textbar AI\textbar]:}''.
To reproduce our results, we release the  code and data at the link:  \url{https://github.com/RUCAIBox/LLMSurvey/tree/main/Experiments}.

\paratitle{Results and Analysis.} 
The results using different instruction datasets based on 7B and 13B LLaMA are in Table~\ref{tab-instruction-tuning-res}. Next, we summarize and analyze our findings in detail.

\textbullet~\emph{{Task-formatted instructions are more proper for the QA setting, but may not be useful for the chat setting.}}
By comparing the performance of instruction tuning using FLAN-T5 with that of ShareGPT and Self-Instruct-52K, we can observe that FLAN-T5 mostly achieves a better performance on QA benchmarks while underperforms ShareGPT on the chat setting.   
The reason is that FLAN-T5 is composed of a mixture of instructions and examples from existing NLP tasks, \eg translation and reading comprehension. 
As a result, LLaMA  fine-tuned with FLAN-T5 performs better on QA tasks, but poorly on user queries.
In contrast, ShareGPT consists of real-world human-ChatGPT conversations, which is able to better elicit LLaMA to {follow user instructions} in daily life, while may not be suitable for accomplishing the QA tasks.

\textbullet~\emph{A mixture of different kinds of instructions are helpful to improve the comprehensive abilities of LLMs.}
After mixing the three kinds of instructions for fine-tuning, we can see that the derived LLaMA variant (with FLAN-T5, ShareGPT and Self-Instruct-52K) performs well in both task settings.   
In MMLU, the performance of LLaMA (7B) can surpass the ones using individual instruction set by a large margin, \ie 43.69 vs. 38.58 (FLAN-T5).
It shows that mixing multiple sources of instruction datasets is  helpful to improve the performance of instruction-tuned LLMs, which  scales the instruction number as well as increases the  diversity. 

\textbullet~\emph{Enhancing the complexity and diversity of instructions leads to an improved model  performance.}
By increasing the  complexity and diversity of  the Self-Instruct-52K dataset respectively, the chat and QA performance of LLaMA can be consistently improved, {\eg from 37.52 to 39.73 in MMLU for LLaMA (7B)}.
It demonstrates that both strategies are useful to improve the instruction following ability of LLMs.
Further, we can see that improving the complexity yields a larger performance improvement  on QA tasks.
The reason is that the QA tasks mostly consist of difficult questions for evaluating LLMs, which can be better solved by LLMs that have learned complex instructions at the fine-tuning stage.

\textbullet~{\emph{Simply increasing the number of instructions may not be that useful, and balancing the difficulty is not always helpful.}}
As the results shown in Table~\ref{tab-instruction-tuning-res}, balancing the difficulty and increasing the number of fine-tuning  instructions are not very helpful in our experiments. 
{Especially for scaling the instruction number, it even hurts the performance, \eg a decrease from 29.81 to 26.63 in BBH for LLaMA (7B).
It shows that simply scaling the number of synthesized instructions without quality control may not be effective  to improve the performance.}
Furthermore, fine-tuning with  the instructions of moderate difficulty also performs well in the chat setting, while slightly decreasing the performance in the QA setting.
A possible reason is that we filter complex and hard instructions with large perplexity scores, hurting the model performance in answering  complex questions.

\textbullet~{\emph{A larger model scale  leads to a better  instruction following performance.}} 
{By comparing the performance of LLaMA (7B) and LLaMA (13B) models fine-tuned with the same set of instruction data, we can see that LLaMA (13B) mostly achieves a better performance. 
It indicates that scaling the model size is helpful for improving the instruction following capability.
Besides, we can see that the QA performance has been improved a lot, \eg from 38.11 to 47.49 in MMLU.
It is likely  because  that the larger models generally have better knowledge utilization and reasoning capability~\cite{Wei-arxiv-2022-chain,Brown-NeurIPS-2020-Language}, which can  accurately answer more complex questions.
}

\begin{center}
\begin{tcolorbox}[colback=blue!5!white,colframe=blue!55!black,width=0.48\textwidth,title={Instruction Tuning Suggestions}]
{
To conduct instruction tuning on LLMs, one can prepare the computational resources according to the basic statistics about the required number of GPUs and tuning time in Table~\ref{tab:instruction-time}. 
After setting up the development environment,  we recommend beginners to follow the code of Alpaca repository~\cite{alpaca} for instruction tuning. 
Subsequently, one should select the base model and construct the instruction datasets as we discuss in this section.  
When computational resources for training are constrained, users can utilize LoRA for parameter-efficient tuning (see Section~\ref{sec-PEFT}). As for inference, users can further use quantization methods to deploy LLMs on fewer or smaller GPUs (see Section~\ref{sec-PEFT}).} 
\end{tcolorbox}
\end{center}

\subsection{Alignment Tuning}
\label{sec-alignment}

This part first presents the background of alignment with its definition and criteria, then focuses on the collection of human feedback data for aligning LLMs, and finally discusses the key technique of reinforcement learning from human feedback (RLHF) for alignment tuning. 

\subsubsection{Background and Criteria for Alignment}
\label{sec-alignment-background}
\paratitle{Background.} LLMs have shown remarkable capabilities in a wide range of NLP tasks~\cite{Brown-NeurIPS-2020-Language,Chowdhery-arxiv-2022-PaLM,Wei-ICLR-2022-Finetuned,Zhang-arxiv-2022-OPT}. However, these models may sometimes exhibit  unintended behaviors, \eg  fabricating false information, pursuing inaccurate objectives, and producing harmful, misleading, and biased expressions~\cite{Ouyang-arxiv-2022-Training,Kenton-arxiv-2021-Alignment}. For LLMs, the language modeling objective pre-trains the model parameters by word prediction while lacking the consideration of human values or preferences.    
{To avert these unexpected behaviors, human alignment has been proposed to make LLMs act in line with human expectations}~\cite{Ouyang-arxiv-2022-Training,Ziegler-arxiv-2019-Fine-Tuning}.  
However, unlike the original pre-training  and adaptation tuning (\eg instruction tuning), such an alignment requires considering very different criteria  (\eg helpfulness, honesty, and harmlessness).  It has been shown that alignment might harm the general abilities of LLMs to some extent, which is called \emph{alignment tax} in related literature~\cite{Askell-arxiv-2021-A}.  
 
\paratitle{Alignment Criteria.} 
{Recently, there is increasing attention on developing multifarious criteria to regulate the behaviors of LLMs. 
Here, we take three representative alignment criteria (\ie helpful, honest, and harmless) as examples for discussion, which have been widely adopted  in existing literature~\cite{Askell-arxiv-2021-A,Ouyang-arxiv-2022-Training}. 
}
In addition, there are  other alignment criteria for LLMs from different  perspectives including behavior, intent, incentive, and inner aspects~\cite{Kenton-arxiv-2021-Alignment}, which are essentially similar (or at least with similar  alignment techniques) to the above three criteria. 
It is also feasible to modify the three criteria according to specific needs, \eg substituting honesty with correctness~\cite{Glaese-arxiv-2022-Improving}.   
Next, we give brief explanations about the three representative alignment criteria:

$\bullet$ \emph{Helpfulness.} To be helpful, the LLM should demonstrate a clear attempt to assist users in solving their tasks or answering questions in a concise and efficient manner as possible. At a higher level, when further clarification is needed, the LLM should demonstrate the capability of eliciting  additional relevant information through pertinent inquiries and exhibit suitable levels of sensitivity, perceptiveness, and prudence~\cite{Askell-arxiv-2021-A}. Realizing the alignment of helpful behavior is challenging for LLMs since it is difficult to precisely define and measure the intention of users~\cite{Kenton-arxiv-2021-Alignment}.

$\bullet$ \emph{Honesty.} At a basic level, a LLM aligned to be honest should present accurate content to users instead of fabricating information. Additionally, it is crucial for the LLM to convey appropriate degrees of uncertainty in its output, in order to avoid any form of deception or misrepresentation of information. This requires the model to know about its capabilities and levels of knowledge (\eg ``know unknowns''). According to the discussion in \cite{Askell-arxiv-2021-A}, honesty is a more objective criterion compared to helpfulness and harmlessness, hence honesty alignment could potentially be developed with less reliance on human efforts. 

$\bullet$ \emph{Harmlessness.} To be harmless, it requires that the language produced by the model should not be offensive or discriminatory. To the best of its abilities, the model should be capable of detecting covert endeavors aimed at soliciting requests for malicious purposes. Ideally, when the model was  induced to conduct a dangerous action (\eg committing a crime), the LLM should politely refuse. Nonetheless, 
\emph{what behaviors} are deemed harmful and \emph{to what extent} vary amongst individuals or  societies~\cite{Askell-arxiv-2021-A} highly  depend on who is using the LLM, the type of the posed question, and the context (\eg time) at which the LLM is being used.

As we can see, these criteria are quite subjective, and are developed based on human cognition. 
Thus, it is difficult to directly formulate them as optimization objectives for LLMs. 
{In existing work, there are many ways to fulfill these criteria when aligning LLMs. 
A promising technique is \emph{red teaming}~\cite{Perez-EMNLP-2022-Red}, which involves using manual or automated means to probe LLMs in an adversarial way to generate harmful outputs and then updates LLMs to prevent such outputs.

\subsubsection{Collecting Human Feedback}\label{sec-human-feedback}
During the pre-training stage, LLMs are trained using the language modeling objective on a large-scale corpus. However, it cannot take into account the subjective and qualitative evaluations of LLM outputs by humans (called \emph{human feedback} in this survey).
High-quality human feedback is extremely important for aligning LLMs with human preferences and values.
In this part, we discuss how to select a team of human labelers for  feedback data collection.  

\paratitle{Human Labeler Selection.}  
In existing work, the dominant method for generating human feedback data is human annotation~\cite{Ziegler-arxiv-2019-Fine-Tuning,Ouyang-arxiv-2022-Training,Glaese-arxiv-2022-Improving}. This highlights the critical role of selecting appropriate human labelers.
To provide high-quality feedback, human labelers are supposed to have a qualified level of education and excellent proficiency in English. For example, Sparrow~\cite{Glaese-arxiv-2022-Improving} requires human labelers to be UK-based native English speakers who have obtained at least an undergraduate-level educational qualification.
Even then, several studies~\cite{Ziegler-arxiv-2019-Fine-Tuning} have found that there still exists a mismatch between the intentions of researchers and human labelers, which may lead to low-quality human feedback and cause LLMs to produce unexpected output. 
To address this issue, %
InstructGPT~\cite{Ouyang-arxiv-2022-Training} further conducts a screening process to filter labelers by assessing the agreement between human labelers and researchers. Specifically, researchers first label a small amount of data and then measure the agreement between themselves and human labelers. The labelers with the highest agreement will be selected to proceed with the subsequent annotation work. 
In some other work~\cite{Menick-arxiv-2022-teaching},  ``super raters'' are used to ensure the high quality of human feedback. Researchers evaluate the performance of human labelers and select a group of well-performing human labelers (\eg high agreement) as super raters. The super raters will be given priority to collaborate with the researchers in the subsequent study. 
When human labelers annotate the output of LLMs, it is helpful to specify detailed instructions and provide instant guidance for human labelers, which can further regulate the annotation of labelers.

\paratitle{Human Feedback Collection.}  In existing work, there are mainly three kinds of approaches to collecting  feedback and preference data from human labelers.

$\bullet$ \emph{Ranking-based approach.} In early work~\cite{Ziegler-arxiv-2019-Fine-Tuning}, human labelers often evaluate model-generated outputs in a coarse-grained manner (\ie only selecting the best) without taking into account more fine-grained alignment criteria. Nonetheless, different labelers may hold diverse opinions on the selection of the best candidate output, and this method disregards the unselected samples, which may lead to inaccurate or incomplete human feedback.
To address this issue, subsequent studies~\cite{Glaese-arxiv-2022-Improving} introduce {the Elo rating system} to derive the preference ranking by comparing candidate outputs. 
{The ranking of outputs serves as the training signal that guides the model to prefer certain outputs over others, thus inducing outputs that are more reliable and safer.} 

$\bullet$ \emph{Question-based approach.}
Further, human labelers can provide more detailed feedback by answering certain questions designed by researchers~\cite{Nakano-arxiv-2021-WebGPT}, covering the alignment criteria as well as additional constraints for LLMs. Specially, in WebGPT~\cite{Nakano-arxiv-2021-WebGPT}, to assist the model in  filtering and utilizing relevant information from retrieved documents, human labelers are required to answer questions with multiple options about whether the retrieved documents are useful for answering the given input. 

$\bullet$ \emph{Rule-based approach.}
Many studies also develop rule-based methods to provide more detailed human feedback.
As a typical case, Sparrow~\cite{Glaese-arxiv-2022-Improving} not only selects the response that labelers consider the best but also uses a series of rules to test whether model-generated responses meet the alignment criteria of being helpful, correct, and harmless.  
In this way, two kinds of human feedback data can be obtained: (1) the response preference feedback is obtained by comparing the quality of model-generated output in pairs, and (2) the rule violation feedback is obtained by collecting the assessment from human labelers (\ie a score indicating to what extent the generated output has violated the rules). 
{Furthermore,  GPT-4~\cite{OpenAI-OpenAI-2023-GPT-4} utilizes a set of zero-shot classifiers (based on GPT-4 itself) as rule-based reward models, which can automatically determine whether the model-generated outputs violate a set of human-written rules.
}

In the following, we focus on a well-known technique, reinforcement learning from human feedback (RLHF), which has been widely  used in the recent powerful LLMs such as ChatGPT.} As discussed below, the alignment criteria introduced in %
{Section~\ref{sec-alignment-background}} can be fulfilled by learning from human feedback on the responses of LLMs to users' queries. 

\subsubsection{Reinforcement Learning from Human Feedback}\label{sub:RLHF}

\begin{figure}
    \centering
    \includegraphics[width=0.5\textwidth]{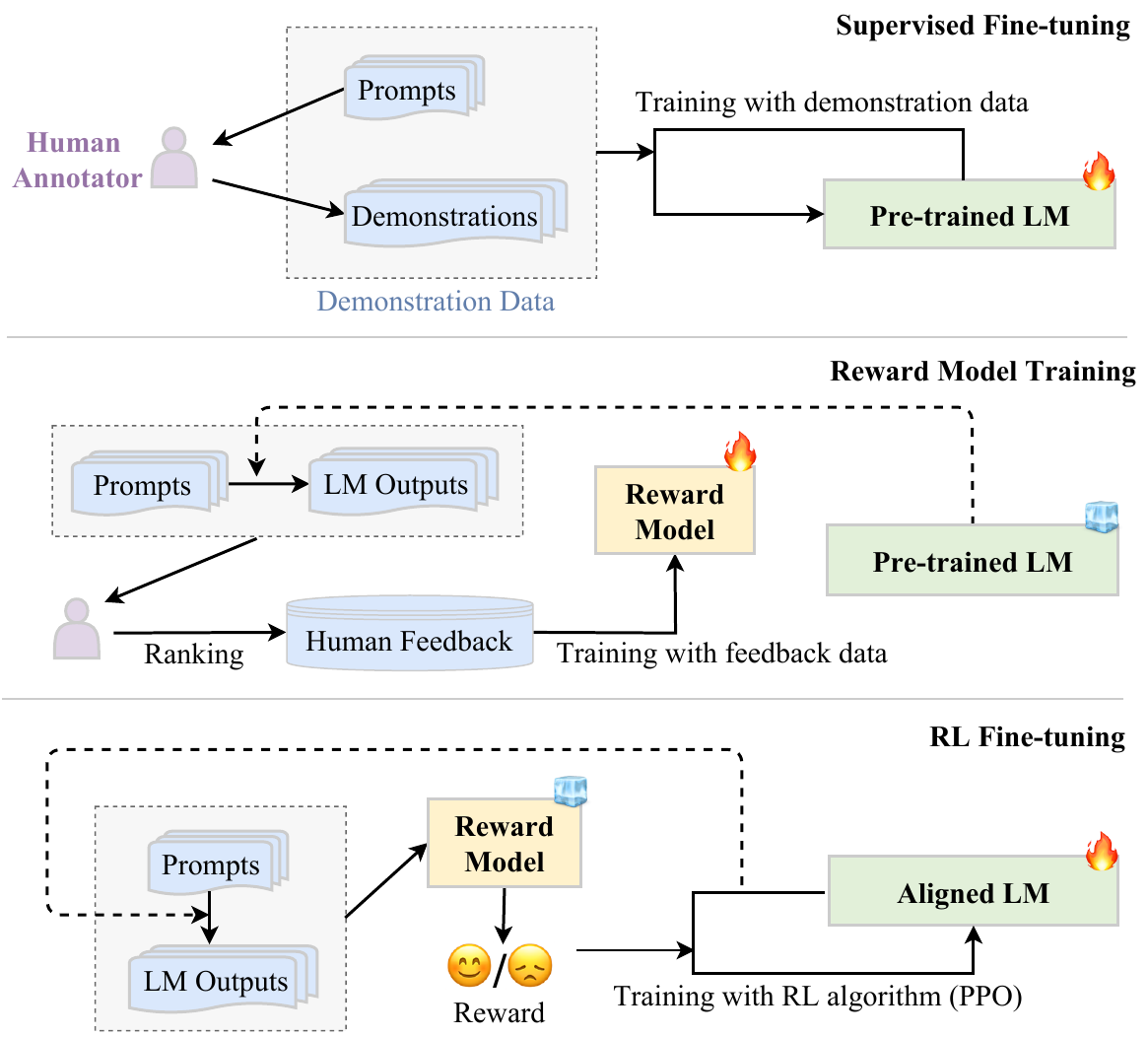}
    \caption{The workflow of the RLHF algorithm.}
    \label{fig:RLHF}
\end{figure}

To align LLMs with human values, reinforcement learning from human feedback (RLHF)~\cite{Christiano-NeurIPS-2017-Deep,Ziegler-arxiv-2019-Fine-Tuning} has been proposed to fine-tune LLMs with the collected human feedback data, which is useful to improve the alignment criteria (\eg helpfulness, honesty, and harmlessness).
RLHF employs reinforcement learning~(RL) algorithms~(\eg Proximal Policy Optimization~(PPO)~\cite{schulman-arxiv-2017-proximal}) to  adapt LLMs to human feedback by learning a reward model. Such an approach incorporates humans in the training loop for developing well-aligned LLMs, as exemplified by InstructGPT~\cite{Ouyang-arxiv-2022-Training}. 

\begin{figure*}[!t]
    \centering
    \includegraphics[width=1\textwidth]{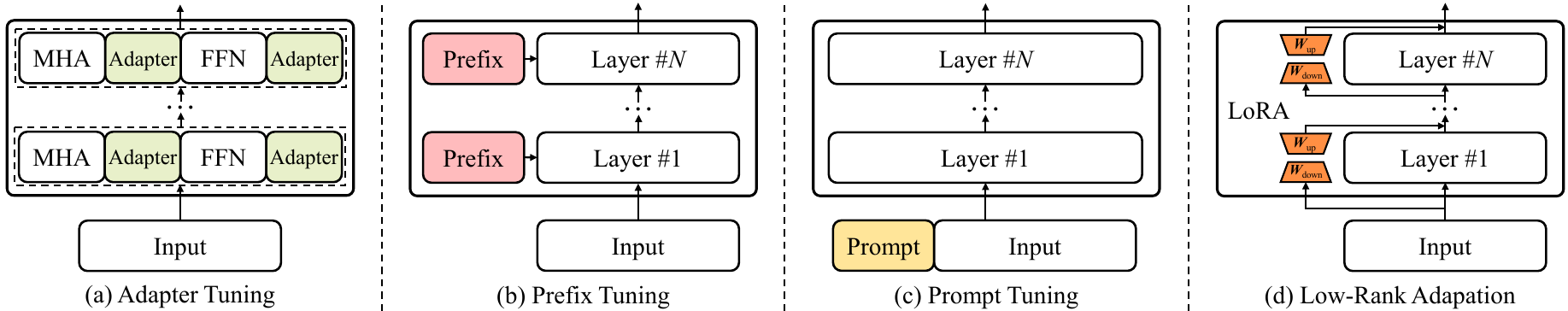}
    \caption{An illustration of four different parameter-efficient fine-tuning methods. MHA and FFN denote the multi-head attention and feed-forward networks in the Transformer layer, respectively.}
    \label{fig:efficient-tuning}
\end{figure*}

\paratitle{RLHF System.}
The RLHF system mainly comprises three key components: a pre-trained LM to be aligned, a reward model learning from human feedback, and a RL algorithm training the LM.  
Specifically, the \textit{pre-trained LM} is typically a generative model that is  initialized with existing pre-trained LM parameters. For example, OpenAI uses 175B GPT-3 for its first popular RLHF model, InstructGPT~\cite{Ouyang-arxiv-2022-Training}, and DeepMind uses the 280 billion parameter model Gopher~\cite{Rae-arxiv-2021-Scaling} for {its GopherCite model~\cite{Menick-arxiv-2022-teaching}. } %
Further, the \textit{reward model~(RM)} provides (learned) guidance signals that reflect human preferences for the text generated by the LM, usually in the form of a scalar value. The reward model can take on two forms: a fine-tuned LM or a LM trained de novo using human preference data.
{Existing work typically employs reward models having  a  parameter scale different from that of the aligned LM~\cite{Menick-arxiv-2022-teaching,Ouyang-arxiv-2022-Training}.} For example, OpenAI uses 6B GPT-3 and DeepMind uses 7B Gopher as the reward model, respectively.
Finally, to optimize the pre-trained LM using the signal from the reward model, a specific \textit{RL algorithm} is designed for large-scale model tuning. Specifically, Proximal Policy Optimization (PPO)~\cite{schulman-arxiv-2017-proximal} is a widely used RL algorithm for alignment in existing work~\cite{Ouyang-arxiv-2022-Training,Menick-arxiv-2022-teaching,Glaese-arxiv-2022-Improving}. 

\paratitle{Key Steps for RLHF.} Figure~\ref{fig:RLHF} illustrates the overall three-step process of RLHF~\cite{Ouyang-arxiv-2022-Training} as introduced below. 

$\bullet$ \textit{Supervised fine-tuning.} To make the LM initially perform desired behaviors, it  usually needs to collect a supervised dataset containing input prompts (instruction) and desired outputs for fine-tuning the LM. These prompts and outputs can be written by human labelers for some specific  tasks while ensuring the diversity of tasks.  
{For example, InstructGPT~\cite{Ouyang-arxiv-2022-Training} asks human labelers to compose prompts (\eg ``\emph{List five ideas for how to regain enthusiasm for my career}'') and desired outputs for several generative tasks such as open QA, brainstorming, chatting, and rewriting. }
Note that the first step is optional in specific settings or scenarios. 

$\bullet$ \textit{Reward model training.} The second step is to train the RM using human feedback data. Specifically, we employ the LM to generate a certain number of output texts using sampled prompts (from either the supervised dataset or the human-generated prompt) as input. We then invite human labelers to annotate the preference for these pairs. The annotation process can be conducted in multiple forms, and a common approach is to annotate by ranking the generated candidate texts, which can reduce  the inconsistency among annotators. Then,  the RM is trained to predict the human-preferred output. In InstructGPT, labelers rank model-generated outputs from best to worst, and the RM (\ie 6B GPT-3) is trained to predict the ranking. Note that, in recent work~\cite{Bai-arXiv-2022-Constitutional}, the annotation of preference on response pairs has been conducted by an AI agent (usually an aligned LLM) instead of humans, which is called ``\emph{reinforcement learning from AI feedback (RLAIF)}''. 
{LLMs trained with typical RLHF algorithms tend to generate harmless responses with less helpfulness, which is  called  \emph{evasion problem}~\cite{Bai-arXiv-2022-Constitutional}. To  guarantee both the harmlessness and helpfulness, RLAIF generates the AI feedback based on pre-set alignment principles in instructions~\cite{Bai-arXiv-2022-Constitutional,Lee-CoRR-2023-RLAIF}, which can also reduce  the efforts of human annotation.} 

$\bullet$ \textit{RL fine-tuning.} At this step, aligning (\ie fine-tuning) the LM is formalized as an RL problem. In this setting, the pre-trained LM acts as the policy that takes as input a prompt and returns an output text, the action space of it is the vocabulary, the state is the currently generated token sequence, and the reward is provided by the RM. To avoid eviating significantly from the initial (before tuning) LM, a penalty term is commonly incorporated into the reward function. 
For example, InstructGPT optimizes the LM against the RM using the PPO algorithm. 
For each input prompt, InstructGPT calculates the KL divergence between the generated results from the current LM and the initial LM as the penalty.
It is noted that the second and final steps can be iterated in multiple turns for better aligning LLMs. 
Due to the instability of the RL algorithm, recent work~\cite{Dong-RAFT-2023-arxiv} replaces the RL tuning with another supervised fine-tuning by reusing the best ranked samples with higher rewards.

\paratitle{Practical Strategies for RLHF.}
{Although RLHF is promising to effectively improve the alignment of LLMs with humans, it is practically challenging for researchers to successfully implement  it.
In this part, we focus on discussing several useful  strategies and tricks for improving the effectiveness and efficiency of RLHF. 
Concretely, we focus on the effective training of reward models, efficient and effective RL training, respectively.}

{$\bullet$ \textit{Effective reward model training.} 
Despite that InstructGPT used a small reward model (6B GPT model), increasing work~\cite{Touvron-2023-llama2-arxiv} has shown it is often more effective to use a large reward model (\eg  equal or greater than the original model size), since large reward models generally perform better in judging the quality of the LLM generated outputs. 
In LLaMa 2~\cite{Touvron-2023-llama2-arxiv},  pretrained chat model checkpoints are used 
to initialize the reward model, they argue that such an approach can effectively reduce the information mismatch between the model to be aligned and the reward model by sharing the same pre-training knowledge.   
Whereas, it is common to encounter the overfitting problem when training large-scale reward models. As a simple yet effective solution, existing work~\cite{askell2021general,zheng2023secrets} has introduced the LM loss on the preferred response of the input prompt from the human-annotated alignment dataset as a regularizer, which alleviates the overfitting of the reward model on the binary classification task. 
In addition, as there are multiple criteria for alignment (\eg helpfulness and honesty), it is often difficult to train a single reward model that can satisfy all the alignment  criteria.
Therefore, it is useful to train multiple reward models that focus on different alignment criteria~\cite{Touvron-2023-llama2-arxiv}, and compute the final reward based on the produced ones from them via special combination strategies (\eg mean pooling and weighted sum).
Such a way enables more  flexible rules or standards on multiple  criteria, \eg  
relaxing the requirement on helpfulness while posing more strict limits on harmfulness.} 

{$\bullet$ \textit{Effective RL training.}
As the RL training process tends to  be unstable and hyper-parameter sensitive,
it is suggested that the language model should be well supervised fine-tuned
 before RL training, so as to reaching a good model capacity.  
A commonly-used way is to fine-tune the LLM on its best outputs of the prompts (referred to as \emph{rejection sampling} or \emph{best-of-$N$}) from the alignment dataset until convergence before RL.
Given a prompt, the LLM would first produce $N$ outputs via the sampling algorithm, and then the best candidate from the model will be selected by the reward model for learning.
After fine-tuning the LLM on the best samples until convergence, the RL  process will be performed to further improve the performance.
LLaMA 2~\cite{Touvron-2023-llama2-arxiv} has successively trained five versions of RLHF models, where the LLM has been progressively improved with the improvement of the reward models.
In this way, the collected prompts and annotations of human preference data can better reflect the issues of the current model checkpoint, thus making special tuning to address these issues. 
In addition, LLaMA 2 also adds samples from prior iterations into the subsequent ones, to alleviate the possible capacity regression issue during iterative optimization.}

{$\bullet$ \textit{Efficient RL training.}
As the RL training requires to iterate the inference process of both the LLM and reward models, it would greatly increase the total memory and computation cost, especially for larger reward models and LLMs.
As a practical trick, we can deploy the reward model on a separate server, and invoke the corresponding API to work with the LLM on its own server. 
In addition, as RLHF requires the LLM to generate multiple candidate  outputs, instead of calling the sample decoding procedure for multiple times, it is more efficient to utilize the  beam search decoding algorithm\footnote{\url{https://huggingface.co/docs/transformers/v4.31.0/en/main\_classes/text\_generation\#transformers.GenerationMixin.group\_beam\_search}}.
It only needs to perform one-pass decoding  for response generation, meanwhile such a strategy can also enhance the diversity of the generated candidate responses.  }


\paratitle{Process-Supervised RLHF.} 
In existing literature of RLHF~\cite{Uesato-arxiv-2022-Solving}, the supervision approach for RL training generally takes two major forms, either using outcome-supervision signals or process-supervision signals.
The outcome-supervised RLHF employs a quantitative score to assess the quality of the whole text generated by LLMs. In contrast, process-supervised RLHF offers an evaluation of each individual component (\eg sentence, word, or reasoning step) within the generated content, which leverage fine-grained supervision signals to guide the training, helping LLMs refine the undesired generation contents~\cite{Uesato-arxiv-2022-Solving, Lightman-arxiv-2023-let}. In what follows, we discuss two key aspects of process-supervised RLHF. 

$\bullet$ \textit{Obtaining Fine-grained Supervision Signals.}
Compared with outcome rewards, it is more difficult to obtain fine-grained supervision signals. 
OpenAI has released a fine-grained annotation dataset named PRM800k~\cite{Lightman-arxiv-2023-let} consisting of 12K process-annotated mathematical problems~(\ie MATH dataset~\cite{Hendrycks-nips-2021-Measuring}) and 75K solutions generated by LLMs of these problems, where each reasoning step of mathematical problems is labeled as \emph{positive}, \emph{negative} or \emph{neutral} in PRM800k.
Considering the cost and efficiency of the human annotation process, several methods aim to automatically annotate the correctness of intermediate reasoning steps, {\eg using powerful LLMs to directly replace human annotators~\cite{Wang-23-arxiv-Shepherd} or Monte Carlo tree search~\cite{Chen-arxiv-2024-AlphaMath}}.
After obtaining fine-grained supervision signals, existing work typically leverages them to train process-supervised reward models~(PRM)~\cite{Ma-arxiv-2023-Let, Lightman-arxiv-2023-let}, which can produce step-level rewards (\eg sentence based or token based rewards) during the RLHF procedure. 
{Furthermore, rather than leveraging the discriminative model to produce the rewards, RLMEC~\cite{Chen-arxiv-2024-Improving} utilizes a generative reward model trained on rewriting tasks with the minimum editing constraint, to provide token-level rewards.
}
{In addition, for the downstream tasks where fine-grained supervision signals are difficult to collected, outcome-supervision signals can also be utilized to perform process-supervised RLHF~\cite{Xi-arxiv-2024-Training}.\ignore{看不懂，还是不清楚。。。是说硬要用吗？}}

$\bullet$ \textit{Utilizing the PRMs.}
{
To effectively leverage process-supervision signals from PRMs, existing work mainly utilizes these fine-grained signals to evaluate individual parts within the LLM responses and then guides LLMs to adjust their generation behaviors to maximize the received reward of the response.
Concretely, expert iteration~\cite{Silver-nat-2017-Mastering,Anthony-nips-2017-Thinking}, an effective RL algorithm, has been utilized to improve the base policy via learning from expert policy~\cite{Uesato-arxiv-2022-Solving}.
Typically, expert iteration contains two main stages: policy improvement and distillation~\cite{Uesato-arxiv-2022-Solving}.
In the policy improvement stage, expert policy processes the systematic search procedure to produce the samples under the guidance of PRMs.
Subsequently, during the distillation stage, the samples generated by expert policy in the first stage are utilized to improve the base policy through supervised fine-tuning.
In addition to expert iteration, PRMs can also be utilized to re-rank the candidates of the final answers generated by LLMs~\cite{Lightman-arxiv-2023-let} or to select better intermediate reasoning steps during step by step reasoning~\cite{Ma-arxiv-2023-Let, Luo-arxiv-2023-WizardMath}.
}

\subsubsection{Alignment without RLHF}
\label{sec-alignment-withoutRL}

Although RLHF has achieved great success in aligning the behaviors of LLMs with human values and preferences, it also suffers from notable limitations. First, RLHF needs to train multiple LMs including the model being aligned, the reward model, and the reference model at the same time,  which is  tedious in algorithmic procedure and memory-consuming in practice. Besides, the commonly-used PPO algorithm in RLHF is rather complex and often sensitive to hyper-parameters. As an alternative, increasing studies explore to directly optimize LLMs to adhere to human preferences, using supervised fine-tuning without reinforcement learning~\cite{zhou-arxiv-2023-lima}.

\paratitle{Overview.} 
The basic idea of non-RL alignment approaches is to directly fine-tune LLMs with \emph{supervised learning} on high-quality \emph{alignment dataset}.
It basically assumes that response feedback or golden rules to avert unsafe behaviors have been injected or included in the specially curated alignment dataset, so that LLMs can directly learn aligned behaviors from these demonstration data via suitable fine-tuning strategies. 
Thus, to implement this approach, two key issues are the construction of alignment dataset and the design of fine-tuning loss.  
{For the first issue, the alignment dataset can be automatically constructed by an aligned LLMs according to human-written safety principles~\cite{Sun-arxiv-2023-Principle} or refining existing examples using edits operations~\cite{Liu-NeurIPS-2022-Second}. In addition, we can also reuse existing reward models to select high-rated responses from existing human feedback data~\cite{Dong-RAFT-2023-arxiv}. For the second issue,  non-RL alignment approaches mainly fine-tune  LLMs  in a  supervised learning way (the same as the original instruction tuning loss) on a high-quality alignment dataset, meanwhile auxiliary learning objectives can be used to enhance the  alignment performance, \eg  ranking responses or contrasting instruction-response pairs. } 



\paratitle{Alignment Data Collection.} {The construction of alignment data is important to effectively align the behaviors of LLMs with human preferences. To collect high-quality alignment data, 
some work tries to reuse existing reward models to {select} high-rated  responses,  and others explore to leverage powerful LLMs (\eg ChatGPT) or build a simulated environment to generate synthetic alignment examples. Next, we will discuss these three lines of research.
}

$\bullet$  {\textit{Reward model based approaches.} The reward model in RLHF has been trained to measure the alignment degree on {the responses of LLMs}. It is straightforward to leverage existing reward models to select high-quality responses as alignment data for subsequent fine-tuning. Based on this idea, RAFT~\cite{Dong-RAFT-2023-arxiv} adopts reward models trained on human preference data to rank the responses of LLMs and collect those with higher rewards for supervised fine-tuning. 
In addition, the reward model can be also used to score model responses and assign them to different quality groups. Quark~\cite{Lu-nips-2022-quark} sorts the responses of LLMs into different quantiles based on the reward scores. Each quantile is attached with a special reward token to represent the reward level of the quantile. Conditioned on the highest-reward tokens, LLMs are subsequently prompted to generate high-quality responses. {Given an initial answer and the corresponding human feedback, ILF~\cite{Scheurer-arxiv-2023-ILF} first adopts LLMs to generate refined answers, then utilizes the reward model to select the answer that best matches the feedback for further training.}
As valuable resources for aligning LLMs, several reward models have been released, including DeBERTa-base/large/xxlarge from OpenAssistant\footnote{https://huggingface.co/OpenAssistant}, Moss-7B from Fudan\footnote{https://github.com/OpenLMLab/MOSS-RLHF}, and Flan-T5-xl from Stanford\footnote{https://huggingface.co/stanfordnlp/SteamSHP-flan-t5-xl}.


$\bullet$ {\textit{LLM based generative approaches.} Reward models help to select aligned data from model responses. However, training reward models itself necessitates substantial high-quality human-labeled data, which is typically expensive and in short supply. 
In addition, although existing reward models can be reused, they might not be able to accurately capture the nonalignment behaviors in another separately trained LLM.   
Therefore, some work explores leveraging powerful LLMs to automatically generate human-aligned data. As a representative work, constitutional AI~\cite{Bai-arXiv-2022-Constitutional} proposes  that human supervision comes from a set of principles (\ie natural language instructions) governing AI behaviors. Based on these principles, LLMs will critique their own harmful responses and revise them repeatedly into finally aligned responses. Similarly, Self-Align~\cite{Sun-arxiv-2023-Principle} first adopts self-instruct~\cite{Wang-arXiv-2022-Self} to generate instructions focusing on covering diverse topics. Then, the model is also prompted with multiple human-written principles that describe the rules of expected model behaviors (also with several in-context exemplars), to generate helpful, ethical, and reliable responses as alignment data. 
{To mitigate the limit that the original SFT method can only learn from positive responses, 
FIGA~\cite{Guo-arxiv-2023-Beyond} develops an improved supervised alignment approach, where both  negative (the original output of low quality) and positive (the refined output by LLMs) responses are leveraged in a contrastive way, to enable LLMs to deeply understand what fine-grained revisions actually  lead to good response. 
}

$\bullet$ {\textit{LLM based interactive approaches.} Most existing approaches train LLMs in isolation, where LLMs are not present in actual  environments to improve themselves through external feedback signals. As a comparison, humans learn social norms and values from interactions with others in social environments~\cite{Krishna-PNAS-2022-Socially}. To mimic such a learning approach, Stable Alignment~\cite{Liu-arxiv-2023-training} builds a simulated interaction environment  consisting of a number of LLM agents, where AI agents keep interacting with and  each other, receiving feedback on improvement.  
Once a central agent receives an instruction, it produces a response and shares it with nearby agents. These critic  agents generate feedback comprising ratings about the response and revision  suggestions. Then the central agent would revise the original response following these suggestions. 
}
Such an alignment approach can be also extended to real-world environment with humans. 
 
\paratitle{Supervised Alignment Tuning.} {After obtaining alignment data,} it is also key to design suitable fine-tuning strategies for direct alignment. A straightforward approach is to optimize LLMs  using the conventional sequence-to-sequence objective based on the alignment data. 
In addition to the conventional optimization objective, several studies further explore auxiliary losses that enhance the learning from the alignment data.}

$\bullet$ \textit{Primary training objective.} Since the alignment data typically consists of an input instruction and an output response, the primary training loss is still the traditional cross-entropy loss for sequence-to-sequence learning. Based on this loss, many studies propose a number of improvement variants for enhancing the supervised alignment tuning. For example, CoH~\cite{Liu-arxiv-2023-Chain} constructs the training data by prepending ``\emph{A helpful answer:}'' and ``\emph{An unhelpful answer:}'' to the annotated good and bad responses, respectively, and only compute losses for those response tokens with special masking.  Quark~\cite{Lu-nips-2022-quark} sorts model responses into different quantiles with varying alignment quality, it prepends a special reward token to each model response to represent the reward level of the response. 
These studies basically adopt the maximum likelihood objective, and employ instruction prefixes to guide the learning of human preference. 

$\bullet$ \emph{Direct preference optimization}. 
To better mimic the learning approach of RLHF in a supervised learning way, DPO~\cite{Rafailov-arxiv-2023-Direct} proposes to reparameterize the response rewards using the policy model (\ie the language model being optimized), and then the original reward modeling objective can be reformulated only based on the policy model. In this way, DPO removes the explicit reward modeling step, and optimizing the new learning objective that only involves the policy model is equivalent to optimizing the rewards.
{
Based on DPO, existing work has proposed several improvement strategies for enhancing the effectiveness or efficiency, \eg decomposing the optimization of positive responses and negative responses into two independent components~\cite{Ethayarajh-arxiv-2024-KTO} or removing the probability of the reference model in the objective function~\cite{Meng-arxiv-2024-SimPO}.
Furthermore, FIGA~\cite{Guo-arxiv-2023-Beyond} designs a token-level contrastive loss that aims to encourage desirable tokens, penalize undesirable ones, and disregard trivial tokens.
}
Despite the effectiveness, recent work has also revealed that DPO may have inherent limitations in several aspects. First, based on the analysis about the magnitude and gradient directions, recent work reveals that DPO might have difficulty in well balancing the learning of positive instances and negative instances~\cite{Feng-arxiv-2024-Towards}.
In addition, as the reference model provides the reward scores for itself in DPO algorithm, a weak reference model would also influence the alignment performance~\cite{Gorbatovski-arxiv-2024-Learn}, which can be enhanced by improved learning strategies~\cite{Kim-arxiv-2024-sDPO} or well-trained policy model~\cite{Gorbatovski-arxiv-2024-Learn}. 


$\bullet$ \textit{Auxiliary optimization objectives.} Besides the primary cross-entropy loss, several studies propose auxiliary training loss to enhance the learning from the alignment data. First, since the responses of each instruction can be scored by the reward model, the ranking loss can be used to train the model to preserve the ranking order of these responses. For example, RRHF~\cite{Yuan-RRHF-2023-arxiv} samples responses from multiple sources, including model-generated responses, such as those derived from the model itself, ChatGPT, and GPT-4, as well as human-written responses, spanning both high-quality and low-quality instances.
To align with the scores from reward models, it further optimizes the ranking loss by encouraging the model to have a higher conditional log probability for the response with a higher ranking. Moreover, SLiC-HF~\cite{Zhao-arxiv-2023-slichf} proposes to assess the similarity between model outputs and human preference via the distance in the latent space,  and introduces specific calibration and regularization loss to calibrate the candidate sequences based on human-preference data. Similarly, the difference between positive and negative responses from the reward model can be employed to construct the regularization loss~\cite{Fisch-arxiv-2024-Robust}, to enhance the discrimination between positive and negative responses by LLMs. 
Second, to enhance the relatedness between the response and the instruction, some work adopts contrastive learning to push up the probability of correct instruction-response pairs while pushing down incorrect instruction-response pairs. Specifically, for an output response, the proposed approach in \cite{Zhang-arxiv-2023-The} contrasts the target instruction to the other irrelevant instructions. By doing so, it can enable the model to learn the right correlation between instructions and responses.

\subsubsection{Remarks on SFT and RLHF}
\label{sec-remarks-SFTRL}
As discussed in Section~\ref{sec-instruction}, instruction  tuning is the process of training pre-trained language models with
formatted demonstration data (instructions paired with desired outputs). At early exploration,   instruction data was mainly collected from NLP  tasks~\cite{Wei-ICLR-2022-Finetuned}, while it has been now extended to more diverse supervision data  that pairs input and output texts (\eg the utterances of open-ended dialogues). Training with such paired texts is also  called \emph{supervised fine-tuning~(SFT)} in the context of LLMs~\cite{Ouyang-arxiv-2022-Training}. 
In this part, we mainly use the abbreviation  \emph{SFT} for discussion but not instruction tuning, due to the simplicity and popularity. 

Since SFT and RLHF are two major adaptation tuning methods for LLMs, it is important to understand the connections and difference between them.  
Next, we make some discussions on this issue\footnote{This part would be somehow subjective, mainly based on the authors' opinions and experiences. Comments or corrections are welcome to enhance this part. }.

\paratitle{Overall Comparison with RL Formulation}. Following the discussion in Section~\ref{sub:RLHF} (the part related to RL training),  the text generation problem can be formulated as a decision-making process based on RL. 
Taking a prompt as input, the task of a LLM is to generate a text completion that appropriately responds to the prompt. This task would be completed step by step. At each step, an  agent (\ie LLM) will perform an action (\ie generating a token) according to the policy (\ie the generative probability distribution of LLM) conditioned on the current state (currently generated token sequence and other available  context information). 
It is expected that a high-quality output text would be produced by the LLM, which can earn a large reward score based on the entire response.  
Overall, RLHF and SFT can be considered as two different training approaches to optimizing  the  above decision making process for LLMs.     
Specially, RLHF 
firstly learns the reward model, and then employs it to improve the LLM with RL training (\eg PPO). As a comparison, SFT adopts  a teacher-forcing approach, which directly optimizes the likelihood of  a demonstration output. 
Such a token-level training way essentially does  \emph{behavior cloning} (a special algorithm of imitation learning~\cite{Ahmed-ACM-2017-Imitation}): it utilizes the expert's action (\ie the target token at each step) as the supervision label and directly learns to imitate the demonstrations from experts without specifying a reward model as in typical RL algorithms. 
To learn the desired policies, 
SFT adopts a ``local'' optimization way (\ie token-level loss) based on demonstration data, while RLHF takes a  ``global'' optimization way (\ie text-level loss) by involving human preference. More theoretical analysis about  imitation learning and reinforcement learning can be referred to the related RL literature \cite{Ahmed-ACM-2017-Imitation,Levine-youtube-2022-Imitate}.  


\ignore{$\bullet$ \emph{General comparison}. Actually, the connections between imitation learning and reinforcement learning can be discussed in a more general way~\cite{}, not limited to the text generation. Firstly, since SFT actually does behavior cloning, a direct problem would be whether we imitate or reinforce for text completion tasks? 
In essence, RL needs to consider two major goals, namely fitting the available data and maximizing the reward; while imitation learning mainly focuses on the first goal. In the context of LLMs, SFT somehow advocates that the LLMs  ``memorize'' the desired output but not necessarily  learning the underlying ground-truth policy, which does not satisfy our expectation. 
}

\paratitle{Pros and Cons of SFT}.  
SFT has been shown to be an effective approach to  boosting the performance of LLMs on various  benchmarks~\cite{Wei-ICLR-2022-Finetuned,Chung-arxiv-2022-Scaling,alpaca,vicuna2023}, which can largely enhance the task generalization ability  and  flexibly endow specific functions (\eg establishing the chatbot's identity). 
More discussions about the usefulness of SFT can be found in Section~\ref{subsec:effectIT}.  
It has been  widely recognized  that SFT mainly \emph{unlocks} the abilities but not \emph{inject} new abilities into LLMs.   
Thus, it might become problematic when one tries to stimulate the non-endogenous abilities of LLMs via SFT.  
As a concrete scenario, it would potentially advocate the  hallucination behaviors when demonstration data is beyond the knowledge or ability scope of LLMs, \eg training a LLM to answer questions about its unknown facts.   
An interesting viewpoint from John Schulman's talk on RLHF~\cite{John-youtube-2023-RLHF} is that distilling superior models to train less capable models (\eg prompting GPT-4 to generate the response as fine-tuning data) might increase the possibilities of generating the hallucinated texts, thus likely affecting the factual accuracy of LLMs. 
Furthermore, as a behavior cloning method, SFT aims to imitate the behaviors (without explorations) of the experts who construct the demonstration data. 
However, there often exist variations among different annotators on the writing styles, quality, and preferences of demonstration data, which tends to affect the learning performance of SFT. 
Thus, high-quality instruction data (but not the quantity) is the primary factor for effective training of LLMs during the SFT stage~\cite{Touvron-2023-llama2-arxiv}.  

\paratitle{Pros and Cons of RLHF}. 
RLHF was early explored in the literature of deep RL~\cite{Christiano-NeurIPS-2017-Deep},  then  borrowed to improve the capacity of language models (\eg summarization~\cite{Stiennon-arxiv-2020-learning}), and subsequently adopted as the fundamental  technique  to develop InstructGPT~\cite{Ouyang-arxiv-2022-Training}. Recently, increasing evidence~\cite{Touvron-2023-llama2-arxiv,Bai-arXiv-2022-Constitutional} has  demonstrated the effectiveness of RLHF in mitigating the  harmful responses and enhancing the model capacity. 
Specially, LLaMA 2 has demonstrated that RLHF can improve both the helpfulness and harmlessness scores~\cite{Touvron-2023-llama2-arxiv}, and  attributed this to a better human-LLM synergy for data annotation. 
They explain this reason in two major aspects as follows. 
First, since human annotators mainly provide preference annotations for RLHF, 
it can largely alleviate the discrepancies of annotators as that in SFT. Secondly, 
preference annotation is much easier than writing the demonstration data, and annotators can even judge the quality of more superior generations than those they create, making it possible to explore a broader state space beyond what can be demonstrated by human annotators. 
Another key point is that RLHF essentially encourages LLMs to learn correct policies by contrasting the  self-generated responses (discriminating between good and bad responses). It no longer forces the model to imitate external demonstration data, and thus can mitigate the hallucination issues with SFT as discussed above\footnote{In RLHF,  it seems to be also important  that reward models should be aware of the knowledge or ability of a LLM to be aligned. For example, LLaMA 2 adopts pre-trained chat model checkpoints  to initialize reward models~\cite{Touvron-2023-llama2-arxiv}. }. Actually, RLHF has been demonstrated to be an important approach to reduce the  hallucination behaviors in GPT-4~\cite{OpenAI-OpenAI-2023-GPT-4}.  
However, RLHF inherits the drawbacks of classic RL algorithms, \eg sample  inefficiency and training instability. 
When adapted to LLMs, RLHF further  relies on a strong SFT model as initial model checkpoint for efficiently achieving good performance.
In addition, human annotators are involved in a complex iterative optimization process, in which a number of important details (\eg the prompt selection, the schedule of reward model training and PPO training, and the settings of hyper-parameters) have important impact on the whole model performance.   

Overall, SFT is particularly useful to increase the model capacity of pre-trained model checkpoints right after pre-training, while RLHF is promising to further improve the model capacity of SFT models. However, RLHF has been difficult to implement, and far from well explored (according to public literature), and more improvements (\eg efficient and reliable annotation~\cite{Bai-arXiv-2022-Constitutional} and simplified optimization~\cite{Rafailov-arxiv-2023-Direct}) are still needed for further   research.

\subsection{Parameter-Efficient Model Adaptation}\label{sec-PEFT}
In the above, we have discussed the approaches of instruction tuning and alignment tuning to adapt LLMs according to specific goals. Since  LLMs consist of a huge amount of model parameters, it would be costly to perform the full-parameter tuning. In this section, we will discuss how to conduct efficient tuning  on LLMs.  We first review several representative parameter-efficient fine-tuning methods for Transformer language models, and then summarize existing work on parameter-efficient fine-tuned LLMs. 

\subsubsection{Parameter-Efficient Fine-Tuning Methods}\label{sec-PEFT-methods}
In existing literature, parameter-efficient fine-tuning~\cite{Li-ACL-2021-prefix,Lester-ACL-2021-The,Hu-ICLR-2022-LoRA} has been an important topic that aims to reduce the number of trainable parameters while retaining a good performance as possible. In what follows, we briefly review four  parameter-efficient fine-tuning methods for Transformer language models, including adapter tuning, prefix tuning, prompt tuning and 
LoRA. The illustration of these four methods are shown in Figure~\ref{fig:efficient-tuning}.

\paratitle{Adapter  Tuning}. Adapter tuning  incorporates small neural network modules (called \emph{adapter}) into the Transformer models~\cite{Houlsby-ICML-2019-Parameter}. To implement  the adapter module, a bottleneck architecture has been proposed in \cite{Houlsby-ICML-2019-Parameter,Hu-arXiv-2023}, which first compresses the original feature vector into a smaller dimension (followed by a nonlinear transformation) and then recovers it to the original dimension.  
The adapter modules would be integrated into each Transformer layer, typically using a serial insertion after each of the two core parts (\ie  attention layer and feed-forward layer) of a Transformer layer. 
Alternatively,  parallel adapters~\cite{He-ICLR-2022-towards} can be also used in Transformer layers, where it places two adapter modules in parallel with the attention layer and feed-forward layer accordingly.   
During fine-tuning, the adapter modules would be optimized according to the  specific task goals, while the parameters of the original language model are  frozen in this process. 
In this way, we can effectively reduce the number of trainable parameters during fine-tuning.

\paratitle{Prefix Tuning}. 
Prefix tuning~\cite{Li-ACL-2021-prefix} prepends a sequence of prefixes, which are a set of trainable continuous vectors,  to each Transformer layer in language models.  These prefix vectors are task-specific,  which can be considered as virtual token embeddings. To optimize the prefix vectors, a reparameterization trick~\cite{Li-ACL-2021-prefix} has been proposed by learning a MLP function that maps a smaller matrix to the parameter matrix of prefixes, instead of directly optimizing the prefixes.   
It has been shown that this trick is useful for stable training.  
After optimization, the mapping function would be discarded, and only the derived prefix vectors are kept to enhance task-specific performance. 
Since only the prefix parameters would be trained, it can lead to a  parameter-efficient model optimization.
Similar to prefix tuning, p-tuning v2~\cite{Liu-arXiv-2021-P-tuning} incorporates layer-wise prompt vectors into the Transformer architecture specially for natural language understanding, which also utilizes multi-task learning for jointly optimizing shared prompts.  
It has been shown to be useful in improving the model performance of different  parameter scales  on  natural language understanding tasks.

\paratitle{Prompt Tuning}. Different from prefix tuning, prompt tuning~\cite{Liu-arXiv-2021-GPT,Lester-ACL-2021-The}  mainly focuses on incorporating trainable prompt vectors at the input layer\footnote{Here, prompt tuning denotes a category of related efficient tuning methods exemplified by the work~\cite{Liu-arXiv-2021-GPT,Lester-ACL-2021-The,gu-ACL-2022-ppt}, instead of a specific method as used in \cite{Lester-ACL-2021-The}. Indeed, the prefix based tuning methods~\cite{Li-ACL-2021-prefix,Liu-arXiv-2021-P-tuning} can be also considered as prompting methods, which are called \emph{deep prompting tuning}   in \cite{Liu-arXiv-2021-P-tuning}. In this survey, prompt tuning specially refer to the methods that only include the prompt tokens at the input layer, in the context of LLMs.
We assign p-tuning v2~\cite{Liu-arXiv-2021-P-tuning} to the category of prefix tuning, because it incorporates layerwise prompts in langauge models. }. 
Based on the discrete prompting methods~\cite{jiang-TACL-2020-how,shin-EMNLP-2020-autoprompt}, it  augments the input text  by including a group of soft prompt tokens (either in a free form~\cite{Liu-arXiv-2021-GPT} or a prefix form~\cite{Lester-ACL-2021-The}), and then takes the prompt-augmented input to solve specific downstream tasks. 
In implementation, task-specific prompt embeddings are combined with the input text embeddings, which are subsequently fed into language models. P-tuning~\cite{Liu-arXiv-2021-GPT} has proposed a free form to combine the context, prompt and target tokens, which can be applied to the  architectures for both natural language understanding and generation. They further learn the representations of soft prompt tokens by a bidirectional LSTM. 
Another representative approach~\cite{Lester-ACL-2021-The}  named \emph{prompt tuning} directly prepends  prefix prompts to the input.
During training, only the prompt embeddings would be learned according to  task-specific supervisions.
Since this method only includes a small number of trainable parameters at the input layer, it has been found that the performance highly relies on the model capacity of the underlying language models~\cite{Lester-ACL-2021-The}.

\paratitle{Low-Rank Adaptation~(LoRA)}. LoRA~\cite{Hu-ICLR-2022-LoRA} imposes  the low-rank constraint for approximating the update matrix at each dense layer, so as to reduce the trainable parameters  for adapting to downstream tasks. 
Consider the case of optimizing a parameter matrix $\mathbf{W}$. The update process can be written in a general form as:   $\mathbf{W} \leftarrow \mathbf{W} + \Delta \mathbf{W}$. 
The basic idea of LoRA is to freeze the original matrix $\mathbf{W} \in \mathbb{R}^{m \times n}$ while approximating the parameter update $\Delta \mathbf{W}$ by low-rank decomposition matrices, \ie $\Delta \mathbf{W}=\mathbf{A}\cdot  \mathbf{B}^\top$, where $\mathbf{A}\in \mathbb{R}^{m \times k}$ and $\mathbf{B}\in \mathbb{R}^{n \times k}$ are the trainable  parameters for task adaptation and $k \ll \min(m,n)$ is the reduced rank. The major merit of LoRA is that it can largely save the memory and storage usage (\eg VRAM). Further, one can only keep a single large model copy, while maintaining a number of task-specific low-rank decomposition matrices for adapting to different downstream tasks.
Further, several studies have also discussed how to set the rank in a more principled approach, \eg  importance score based allocation~\cite{Zhang-arXiv-2023-Adaptive} and search-free optimal rank selection~\cite{Valipour-arXiv-2022-DyLoRA}.

Besides the above  methods, there is  extensive research on efficient tuning of Transformer language models.  
 However, a more comprehensive discussion  of efficient tuning is beyond the scope of this article, which can be found in the related papers on this topic~\cite{He-ICLR-2022-towards,Ding-NMI-2023-Parameter}.

\subsubsection{Parameter-Efficient Fine-Tuning on LLMs}
With the rising of LLMs, efficient tuning  has attracted increasing research attention for developing a more lightweight adaptation approach in downstream tasks.

In particular, LoRA~\cite{Hu-ICLR-2022-LoRA} has been widely applied to open-source LLMs (\eg LLaMA and BLOOM) for parameter-efficient fine-tuning. 
Among these research attempts, LLaMA and its  variants have  gained much attention for parameter-efficient tuning.  
For example, Alpaca-LoRA~\cite{Alpaca-LoRA} has been trained using LoRA as a lightweight tuned  version of  Alpaca~\cite{Taori-github-2023-Stanford} (a fine-tuned 7B LLaMA model with 52K human demonstrations of instruction following).  
There are extensive explorations of Alpaca-LoRA ranging in different languages or model sizes, which can be found in the collection page\footnote{https://github.com/tloen/alpaca-lora}. 
 A recent study LLaMA-Adapter~\cite{Zhang-arXiv-2023-LLaMA-Adapter} inserts  learnable prompt vectors into each Transformer layer, in which zero-initialized attention has been proposed to improve the training by mitigating the influence of under-fitted prompt vectors. They  also extend this approach to a multi-modal setting, \eg visual question answering.   

Further, an empirical study~\cite{Hu-arXiv-2023} has been conducted to examine the effect of  different tuning methods on  language models.
They compare four efficient tuning methods including  serial adapter tuning~\cite{Houlsby-ICML-2019-Parameter},  parallel adapter tuning~\cite{He-ICLR-2022-towards,Pfeiffer-EMNLP-2022-MAD-X}, and LoRA~\cite{Hu-ICLR-2022-LoRA}, on three open-source LLMs,   namely GPT-J (6B), BLOOM (7.1B) and LLaMA (7B), for evaluation.  Based on the experimental results on six math reasoning datasets, they show that these efficient-tuning methods  under-perform the reference baseline GPT-3.5 on difficult tasks, while achieving a comparable performance on simple tasks. 
Overall, LoRA performs relatively  well among these  comparison methods, using significantly fewer trainable parameters.     

As an important resource, the library \emph{PEFT}~\cite{peft-github-2022} (standing for parameter-efficient fine-tuning) has been released on GitHub\footnote{https://github.com/huggingface/peft}. It has included several widely used efficient tuning methods, including LoRA~\cite{Hu-ICLR-2022-LoRA}/AdaLoRA~\cite{Zhang-arXiv-2023-Adaptive}, prefix-tuning~\cite{Li-ACL-2021-prefix,Liu-arXiv-2021-P-tuning}, P-Tuning~\cite{Liu-arXiv-2021-GPT}, and prompt-tuning~\cite{Lester-ACL-2021-The}. Further, it supports a number of language models such as GPT-2 and LLaMA, and also covers several representative vision Transformer models (\eg ViT and Swin Transformer).

As  discussed in Section~\ref{sec-PEFT-methods}, there have been a large number of efficient tuning methods proposed in the existing literature. However,  most of these approaches are tested on  small-sized pre-trained  language models, instead of the LLMs. 
So far, there still lacks a thorough investigation on the effect of  different efficient tuning methods  on large-sized language models at different settings or tasks.





\section{Utilization}
\label{sec-utilization}

\begin{table*}[t]
\centering
\caption{Typical LLM utilization methods and their key points for ICL, CoT, and planning. Note that the key points only highlight the most important technical contribution.} 
\label{tab:utilization}
\resizebox{\textwidth}{!}{%
\begin{tabular}{l|l|l}
\toprule
\textbf{Approach}                              & \textbf{Representative Work}     & \textbf{Key Point}                                                              \\
\midrule
\multirow{6}{*}{\begin{tabular}[c]{@{}l@{}}In-context\\Learning~(ICL)\end{tabular}} & KATE~\cite{Liu-ACL-2022-What}         & Demonstration selection (similar; k-NN)                      \\
                                                                                    & EPR~\cite{Rubin-NAACL-2022-Learning}  & Demonstration selection (dense retrieval; constrative learning)    \\
                                                                                    & SG-ICL~\cite{Kim-2022-arxiv-Self}     & Demonstration selection (LLM as the demonstration generator) \\
                                                                                    & APE~\cite{Zhou-2023-ICLR-Large}       & Demonstration format (automatic generation \& selection)                         \\
                                                                                    & Structured Prompting~\cite{Hao-2022-arxiv-Structured} & Demonstration format (grouped context encoding; rescaled attention) \\
                                                                                    & GlobalE \& LocalE~\cite{Lu-ACL-2022-Fantasically}     & Demonstration order (entropy-based metric; probing set generation with LLM) \\
\midrule
\multirow{6}{*}{\begin{tabular}[c]{@{}l@{}}Chain-of-thought\\Prompting~(CoT)\end{tabular}}  & Complex CoT~\cite{Fu-arxiv-2022-Complexity}                       & Demonstration (complexity-based selection)                          \\
                                                                                            & Auto-CoT~\cite{Zhang-arxiv-2022-Automatic}                        & Demonstration (automatic generation)                             \\
                                                                                            & Selection-Inference~\cite{Creswell-2022-arXiv-selection}          & Generation (alternate between selection and inference)                      \\
                                                                                            & Self-consistency~\cite{Wang-arxiv-2022-Self-Consistency}          & Generation (diverse paths; self-ensemble)             \\
                                                                                            & DIVERSE~\cite{Li-arxiv-2022-On}                                   & Generation (diverse paths); Verification (step-wise voting)           \\
                                                                                            & Rationale-augmented ensembles~\cite{Wang-arxiv-2022-Rationale}    & Generation (rationale sampling) \\
\midrule
\multirow{13}{*}{Planning}            & Least-to-most prompting~\cite{Zhou-arxiv-2022-Least} & Plan generation (text-based; problem decomposition)                                                  \\
                                     & DECOMP~\cite{Khot-2022-arXiv-Decomposed}             & Plan generation (text-based; problem decomposition) \\
                                     & PS~\cite{Wang-arXiv-2023-Plan}                       & Plan generation (text-based) \\
                                     & Faithful CoT~\cite{Lyu-arxiv-2023-Faithful}          & Plan generation (code-based) \\
                                     & PAL~\cite{Gao-arxiv-2022-PAL}                        & Plan generation (code-based; Python) \\
                                     & HuggingGPT~\cite{Shen-2023-arXiv-Hugginggpt}         & Plan generation (code-based; models from HuggingFace) \\
                                     & AdaPlanner~\cite{Sun-2023-arXiv-adaplanner}          & Plan refinement (skill memory) \\
                                     & TIP~\cite{Lu-2023-arXiv-multimodal}                  & Feedback acquisition (visual perception) \\
                                     & RAP~\cite{Hao-2023-arXiv-reasoning}                  & Feedback acquisition (LLM as the world model); Plan refinement (Monte Carlo Tree Search) \\
                                     & ChatCoT~\cite{Chen-2023-arXiv-chatcot}               & Feedback acquisition (tool); Plan refinement (conversation between LLM and tools) \\
                                     & ReAct~\cite{Yao-2022-arXiv-react}                    & Feedback acquisition (tool); Plan refinement (synergizing reasoning and acting)                                       \\
                                     & Reflexion~\cite{Shinn-2023-arXiv-Reflexion}          & Feedback acquisition (text-based self-reflection); Plan refinement (dynamic memory)                   \\
                                     & Tree of Thoughts~\cite{Yao-arxiv-2023-Tree}          & Feedback acquisition (vote comparison); Plan refinement (tree-based search)                                                       \\
\bottomrule
\end{tabular}%
}
\end{table*}

After pre-training or adaptation tuning, 
a major approach to using LLMs is to design suitable \textit{prompting} strategies for solving various tasks. {In existing literature, task-specific prompts can be effectively learned through manual creation and automatic optimization.} 
{A representative prompting method is \textit{in-context learning}~\cite{Brown-NeurIPS-2020-Language, Dong-arxiv-2023-A}, which formulates the task description and/or demonstrations in the form of natural language text.}
In addition, \textit{chain-of-thought prompting}~\cite{Wei-arxiv-2022-chain} can be employed to enhance in-context learning by involving a series of intermediate reasoning steps in prompts.
Furthermore, \textit{planning}~\cite{Zhou-arxiv-2022-Least} is proposed for solving complex tasks, which first breaks them down into smaller sub-tasks and then generates a plan of action to solve these sub-tasks one by one.
We summarize representative work for these prompting approaches in Table~\ref{tab:utilization}.
Next, we will elaborate on the details of the four techniques.

\subsection{Prompting}
As discussed in previous work~\cite{Liu-survey-2023-Pre-train}, prompting is the major approach to utilizing LLMs for solving various tasks. 
Since the quality of prompts will largely influence the  performance of LLMs in specific tasks, there have been a series of studies proposed to generate suitable task prompts through manual creation or automatic optimization, which will be introduced in this section.


\subsubsection{Prompt Creation}\label{subsec:promptdesign}
The process of manually creating a suitable prompt is also called \emph{prompt engineering}~\cite{Liu-arxiv-2022-Design,White-arxiv-2023-Prompt}. 
A well-designed prompt is very helpful to elicit the abilities of LLMs for accomplishing specific tasks.
In this part, we will first introduce the key components of prompts and discuss several principles for prompt design. Then, we evaluate ChatGPT with different prompts to show the results on several representative tasks. 
We are aware that there have been several existing papers~\cite{Santu-arxiv-2023-TELeR,White-arxiv-2023-Prompt} and websites~\cite{OpenAI-OpenAI-2023-PromptGuide,Contributors-AIShort-2023-AIShort,Contributors-Github-2023-Awesome} that present the suggestions and guidelines to design good prompts. 
As a comparison, we mainly aim to discuss the key factors (ingredients and principles) that are useful for prompt creation, and provide experimental results and analysis on popular tasks as the reference to the beginners. 


\paratitle{Key Ingredients.}
Typically, there are four key ingredients that {depict the functionality of a  prompt for eliciting the abilities of LLMs to complete the tasks}, including task description, input data, contextual information, and prompt  style. To have an intuitive understanding of our discussion, we also present three prompt examples for question answering, meta-review generation, and text-to-SQL    in Table~\ref{tab:prompt-examples}.

\textbullet~\emph{Task description.}
A task description is typically a specific instruction that LLMs are expected to follow. 
In general, one  should clearly describe the task goal  in natural language. 
For the tasks with special input or output format, detailed clarifications are often needed, and one can further utilize keywords to highlight the special settings for better guiding LLMs in task completion.

\textbullet~\emph{Input data.} 
In common cases, it is straightforward to describe input data (\eg an instance to be responded by LLMs) in natural language. 
For special input data, such as knowledge graph and table, it is necessary to {apply an appropriate and convenient way} 
to make them readable for LLMs.  
For structured data, linearization is commonly used to transform the original records (\eg knowledge triples) into sequences~\cite{Jiang-2023-arxiv-StructGPT} due to the simplicity.  
Further, the programming language (\eg executable code) has also been utilized to formulate the structured data,  
{which can also support using external tools (\eg program executor) to produce the precise results~\cite{Beurer-arxiv-2023-Prompting,Lu-arxiv-2023-Chameleon}.
}


\textbullet~\emph{Contextual information.}
In addition to the task description and input data, contextual or background information is also essential for specific tasks. For example, retrieved documents are highly useful for open-domain question answering as supporting evidence. Both the quality of the retrieved documents and their relevance to the question have an impact on the generated answers~\cite{Ren-arxiv-2023-Investigating}. 
Thus, it needs to include such information in a proper prompt pattern or expression format.
Furthermore, in-context task exemplars are also helpful for eliciting LLMs to accomplish a complex task, which can better depict the task goal, {the special output formats, and the mapping relation between input and output.}

 
\textbullet~\emph{{Prompt} style.}
For different LLMs, it is important to design a suitable prompt style for eliciting their abilities to solve specific tasks. 
Overall, one should express the prompt as a clear question or detailed instruction that can be well understood and  answered. 
In some cases, it is also useful to add the {prefix or suffix} to better guide LLMs.
For example, using the prefix ``\emph{Let us think step by step}'' can help elicit LLMs perform step-by-step reasoning, and using the prefix ``\emph{You are an expert on this task (or in this domain)}'' can boost the performance of LLMs in some specific tasks. 
Further, 
for chat-based LLMs (\eg ChatGPT), instead of directly feeding a long or complex task prompt, it is suggested to decompose it into multiple prompts for the sub-tasks and then feed them into LLMs via a multi-turn conversation~\cite{Chen-2023-arXiv-chatcot}.

\paratitle{Design Principles.} Based on the key ingredients of prompts, we summarize several critical design principles that can help create more  effective prompts for solving various  tasks.

\textbullet~\emph{Expressing the task goal clearly.}  {Task descriptions should not be ambiguous or unclear, which likely lead to inaccurate or inappropriate responses.} This highlights the need for clear and unambiguous directives when utilizing these models~\cite{Ouyang-arxiv-2022-Training}. A clear and detailed description should contain various elements to explain a task, including task objective, input/output data (\eg ``\emph{Given a long document, I want you to generate a concise summary.}''),  and the response constraints (\eg ``\emph{the length of the summary cannot exceed 50.}''). By providing a well-clarified task description, LLMs can more effectively understand the target task and generate the desired output.

{
\textbullet~\emph{Decomposing into easy, detailed sub-tasks.} To solve complex tasks, it is important to decompose the difficult task into several more easier, detailed sub-tasks for helping LLMs accomplish the goal step by step, which is closely related to the planning technique in Section~\ref{subsec-planning}.
}
For example, following the suggestion~\cite{Santu-arxiv-2023-TELeR}, we can explicitly list the {sub-tasks in the form of multiple numbered items} (\eg ``\emph{Braid a coherent narrative by performing the following tasks: 1. ...; 2. ...; 3. ...}''). By decomposing a target task into sub-tasks, LLMs can focus on solving easier   sub-tasks and finally achieve more accurate results for complex tasks. 

\textbullet~\emph{Providing few-shot demonstrations.} As discussed in Section~\ref{subsec-icl},  LLMs can benefit from in-context learning for solving complex tasks, where the prompts contain a small number of task examples of the desired input-output pairs, \ie few-shot demonstrations. Few-shot demonstrations can help LLMs learn the semantic mapping between input and output without parameter tuning.
In practice, it is suggested that one should generate a few high-quality demonstrations for the target task, which would highly benefit the final task performance. 


\textbullet~\emph{Utilizing model-friendly format.} 
Since LLMs are pretrained on specially constructed datasets, there are some prompt formats that can make LLMs better understand the instruction. For example, as the OpenAI documentation suggests, we can use \texttt{\#\#\#} or \texttt{"""} as a stop symbol to separate the instruction and context, which can be better understood by LLMs. As a general guideline, most existing LLMs perform a task better in English, thus it is useful to employ English instructions to solve difficult tasks based on machine translation.

{
\textbullet~\emph{Adopting role-playing strategies.}
Since LLMs are pretrained on extensive corpora containing diverse characters and dialogues, they possess an inherent ability for role-playing.
This feature can be harnessed through specific prompts to enhance the corresponding capacity for some specific domains~\cite{Amatriain-CoRR-2024-Prompt}.
For instance, when solving a math problem, we can use a prompt prefix like ``\emph{You are an expert in mathematics}''.
This enables LLMs to solve the problem from an expert's perspective, thereby leveraging their pretrained knowledge more effectively. 
By guiding LLMs with role-playing prompts, they can often generate more reasonable and accurate solutions.
}

\paratitle{Useful Tips.}
In addition to the design principles, we also present a collection of  useful prompt tips based on existing work or our empirical experiences in Table~\ref{tab-tips}. 
Note that these tips are suggested in a general manner, it does not indicate that they are  the best prompts for the corresponding tasks.  
This part will be continuously updated with more guidelines or tips. We  welcome readers to contribute to this collection of  prompt tips.
We present the detailed procedure to contribute to the prompt tips, at the link: \url{https://github.com/RUCAIBox/LLMSurvey/tree/main/Prompts}. 

\begin{table*}[htb]
    \centering
    \caption{A collection of useful tips for designing prompts {that are collected from online notes~\cite{Santu-arxiv-2023-TELeR,White-arxiv-2023-Prompt,OpenAI-OpenAI-2023-PromptGuide,Contributors-AIShort-2023-AIShort} and experiences from our authors}, where we also show the related ingredients and principles (introduced in Section~\ref{subsec:promptdesign}). We abbreviate principles as Prin. and list the IDs of the related principles for each prompt.  {\textcircled{1}: expressing the task goal clearly; \textcircled{2}: decomposing into easy, detailed sub-tasks; \textcircled{3}: providing few-shot demonstrations; \textcircled{4}: utilizing model-friendly format.}}
    \label{tab-tips}
\scriptsize 
\begin{tabular}{cp{0.75\textwidth}c}
\toprule
\textbf{Ingredient} & \textbf{Collected Prompts} & \textbf{
Prin.}\\
\midrule
\multirow{4}{*}{\textbf{Task Description}}  & T1. Make your prompt \underline{\textbf{as detailed as possible}}, \eg ``\emph{Summarize the article into a short paragraph within 50 words. The major storyline and conclusion should be included, and the unimportant details can be omitted.}'' & \textcircled{1} \\
& T2. It is helpful to let the LLM know that it is \textbf{\underline{an expert with a prefixed prompt}}, \eg ``\emph{You are a sophisticated expert in the domain of compute science.}'' & \textcircled{1} \\ 
& T3. Tell the model \textbf{\underline{more what it should do}}, but not what it should not do. & \textcircled{1} \\
& T4. To avoid the LLM to generate too long output, you can just use the prompt: ``\emph{Question: {} Short Answer: {}}''. Besides, you can also use the following suffixes, ``\emph{in a or a few words}'', ``\emph{in one of two sentences}''. & \textcircled{1} \\
\midrule
\multirow{2}{*}{\textbf{Input Data}} & I1. For the question required factual knowledge, it is useful to first \underline{\textbf{retrieve relevant documents}} via the search engine, and then \underline{\textbf{concatenate them into the prompt}} as reference. & \textcircled{4}\\
& I2. To highlight some important parts in your prompt, please \underline{\textbf{use special marks}}, \eg \emph{quotation} ($""$) and \emph{line break} ($\backslash$n). You can also use both of them for emphasizing. & \textcircled{4} \\ 
\midrule
\multirow{4}{*}{\textbf{Contextual Information}}  & C1. For complex tasks, you can \textbf{\underline{clearly describe the required intermediate steps}} to accomplish it, \eg ``\emph{Please answer the question step by step as: Step 1 - Decompose the question into several sub-questions, $\cdots$}'' & \textcircled{2} \\
& C2. If you want LLMs to provide the score for a  text, it is necessary to provide a \textbf{\underline{detailed description about the}} \textbf{\underline{scoring standard}} with examples as reference. & \textcircled{1} \\
& C3. When LLMs generate text according to some context (\eg making recommendations according to purchase history), instructing them with \textbf{\underline{the explanation about the generated result}} conditioned on context is helpful to improve the quality of the generated text. & \textcircled{2} \\
& C4. An approach similar to \textbf{\underline{tree-of-thoughts}} but can be \textbf{\underline{done in one prompt}}: \eg \emph{Imagine three different experts are answering this question. All experts will write down one step of their thinking, then share it with the group of experts. Then all experts will go on to the next step, etc. If any expert realizes they're wrong at any point then they leave. The question is} & \textcircled{2} \\
\midrule
\multirow{9}{*}{\textbf{Demonstration}} & D1. \underline{\textbf{Well-formatted in-context exemplars}} are very useful, especially for producing the outputs with complex formats. & \textcircled{3} \\
& D2. For few-shot chain-of-thought prompting, you can also use the prompt ``\emph{Let's think step-by-step}'', and the few-shot examples should be \textbf{\underline{separated by ``$\backslash$n''}} instead of full stop. & \textcircled{1}\textcircled{3} \\
& D3. You can also \textbf{\underline{retrieve similar examples}} in context to supply the useful task-specific knowledge for LLMs. To retrieve more relevant examples, it is useful to \textbf{\underline{first obtain the answer}} of the question, and then concatenate it with the question for retrieval. & \textcircled{3}\textcircled{4} \\
& D4. The \textbf{\underline{diversity of the in-context exemplars}} within the prompt is also useful. If it is not easy to obtain diverse questions, you can also seek to keep the \textbf{\underline{diversity of the solutions}} for the questions. & \textcircled{3} \\
& D5. When using chat-based LLMs, you can \textbf{\underline{decompose in-context exemplars into multi-turn messages}}, to better match the human-chatbot conversation format. Similarly, you can also decompose the reasoning process of an exemplars into multi-turn conversation. & \textcircled{3} \\
& D6. \textbf{\underline{Complex and informative}} in-context exemplars can help LLMs answer complex questions. & \textcircled{3} \\
& D7. As a symbol sequence can typically be divided into multiple segments (\eg $i_1, i_2, i_3$ $\longrightarrow$ $i_1, i_2$ and $i_2, i_3$), the preceding ones can be used \textbf{\underline{as in-context exemplars}} to guide LLMs to predict the subsequent ones,  meanwhile providing  historical information. & \textcircled{2}\textcircled{3} \\ 
& D8. \textbf{\underline{Order matters}} for in-context exemplars and prompts components. For very long input data, the position of the question (first or last) may also affect the performance. & \textcircled{3} \\
& D9. If you can not obtain the in-context exemplars from existing datasets, an alternative way is to use the \textbf{\underline{zero-shot}} \textbf{\underline{generated ones}} from the LLM itself. & \textcircled{3} \\
\midrule
\multirow{8}{*}{\textbf{Other Designs}}  & O1. Let the \underline{\textbf{LLM check its outputs}} before draw the conclusion, \eg ``\emph{Check whether the above solution is correct or not.}'' & \textcircled{2} \\
& O2. If the LLM can not well solve the task, you can \textbf{\underline{seek help from external tools}} by prompting the LLM to manipulate them. In this way, the tools should be encapsulated into callable APIs with detailed description about their functions, to better guide the LLM to utilize the tools. & \textcircled{4} \\
& O3. The prompt should be \textbf{\underline{self-contained}}, and better not include pronouns (\eg it and they) in the context. & \textcircled{1} \\
& O4. When using LLMs for \textbf{\underline{comparing}} two or more examples, the order affects the performance a lot. & \textcircled{1} \\
& O5. Before the prompt, \textbf{\underline{assigning a role for the LLM}} is useful to help it better fulfill the following task instruction, \eg \emph{``I want you to act as a lawyer''}. & \textcircled{1}\\
& O6. OpenAI models can perform a task better in English than other languages. Thus, it is useful to first \textbf{\underline{translate the input into English}} and then feed it to LLMs. & \textcircled{4} \\
& O7. For multi-choice questions, it is useful to \textbf{\underline{constrain the output space}} of the LLM. You can use a more detailed explanation or just imposing constraints on the logits. & \textcircled{1} \\
& O8. For sorting based  tasks (\eg recommendation), instead of directly outputting the complete text of each item after sorting, one can  \textbf{\underline{assign indicators}} (\eg \emph{ABCD}) to the unsorted items and instruct the LLMs to directly output the sorted indicators. & \textcircled{1} \\
\bottomrule
\end{tabular}
\end{table*}

\paratitle{Empirical Analysis.}
We further conduct empirical studies to present the impact of prompts on task performance.
To conduct the experiments, we select a variety of tasks that span language generation, knowledge utilization, complex reasoning, structure data generation, and information retrieval. 
For each task, we manually write a prompt that follows general guidelines introduced above. Note that the tested prompts may not be the optimal for these tasks, since they mainly aim to help readers understand how to write an effective prompt for solving different tasks. 
Also, we add a simplified  prompt as the comparison for most tasks.  
Following the experimental settings in Section~\ref{sec-empirical}, we examine the  3-shot performance of ChatGPT on complex reasoning tasks (Colored Objects and GSM8k), and zero-shot performance on other tasks.
We report the experimental results in Table~\ref{tab-instructions}, where we also include the supervised  performance in existing papers as reference. 

$\bullet$ \emph{Carefully designed prompts can boost the zero-shot or few-shot performance of ChatGPT.} 
By comparing the results of using different prompts on the same task, we can see that using the carefully designed prompts  can achieve better performance than the simpler ones. 
{In the carefully designed prompts, we provide a more clearly expressed task description (\eg WMT and WikiFact), or use a model-friendly format (\eg GSM8k and OBQA). 
For example, for WikiFact task, the prompt with a more detailed task description leads to a performance increase  from 29.25 to 31.21.} 

$\bullet$ {\emph{More complex tasks can benefit more from careful prompt engineering on ChatGPT.}
In the WikiFact and Colored Objects tasks, the designed prompts have greatly improved the performance of ChatGPT, \ie from 23.61 to 28.47 on WikiFact and from 53.20 to 66.75 on Colored Objects.
It indicates the necessity of prompt engineering for LLMs to perform well on complex tasks, since these tasks typically have specific output formats or require background knowledge. 
Our example prompts  provide more detailed task description (\eg output format and task goal),  which can help ChatGPT better understand the complex task requirement  for fulfilling it.}

$\bullet$  {\emph{For mathematical reasoning tasks, it is more effective to design specific prompts based on the format of programming language.}
For GSM8k, the designed prompt employs code-formatted few-shot demonstrations to convert this mathematical reasoning task into code generation task, which can leverage the strong code synthesis ability of ChatGPT for solving mathematical problems. 
Further, with the help of an external program executor, we are able to obtain more precise results instead of using LLMs for arithmetic operation.
As we can see, the performance is boosted from 78.47 to 79.30 on GSM8k, indicating the usefulness of programming language in mathematical reasoning tasks.} 


$\bullet$ {\emph{In knowledge utilization and complex reasoning tasks, ChatGPT with proper prompts achieves comparable performance or even outperforms the supervised baselines methods.}
In knowledge utilization and  complex reasoning tasks, ChatGPT with proper zero-shot or few-shot prompts can achieve comparable performance or even outperform the supervised  methods, {\eg 31.21 (ChatGPT) \emph{v.s.} 34.20 (supervised baseline) on WikiFact.} 
Despite that, ChatGPT still performs worse than supervised baseline models on some specific tasks (\eg ARC and WikiFact), since these supervised models have been specially optimized with task-specific data.  
}


$\bullet$ \emph{Through suitable prompt engineering, LLMs can handle some non-traditional NLP tasks.}  
{With the help of specific prompts, ChatGPT can also accomplish non-traditional NLP tasks, \ie the general recommendation and conversational recommendation. 
A key point is that these tasks can be well expressed or described in natural language. }
However, the performance of ChatGPT is still far from the referenced performance in these tasks, as LLMs cannot directly fit these tasks, which require specific domain knowledge and task adaptation~\cite{Zhang-2023-arxiv-recommendation,Hou-2023-arxiv-large}.



\begin{table*}[ht]
	\footnotesize
	\centering
 \caption{Example instructions collected from \cite{Santu-arxiv-2023-TELeR,Chang-arxiv-2023-How}. The \colorbox{LightSkyBlue1}{blue} text denotes the task description, the \colorbox{tPink}{red} text denotes the contextual information, the \colorbox{tGreen}{green} text denotes the demonstrations, and the \colorbox{Khaki1}{gold} text denotes the prompt style.} 
	\label{tab:prompt-examples}
	\begin{tabular}{p\textwidth}
		\toprule
            \rowcolor{LightSkyBlue1}{\fontfamily{ppl}\selectfont Use the provided articles delimited by triple quotes to answer questions. If the answer cannot be found in the articles, write ``I could not find an answer.''} \\
            \specialrule{0em}{1pt}{1pt}
            \rowcolor{tGreen}
            \rowcolor{tPink}{\fontfamily{ppl}\selectfont \textbf{Articles:} ``````Joao Moutinho is a Portuguese footballer who last played as a central midfielder for Premier League club Wolverhampton Wanderers and the Portugal national team.'''''' } \\
            \specialrule{0em}{1pt}{1pt}
            \rowcolor{tGreen}{\fontfamily{ppl}\selectfont \textbf{Question:} Is the following sentence plausible?
'Joao Moutinho was out at third.' } \\
            \rowcolor{tGreen}{\fontfamily{ppl}\selectfont \textbf{Answer:} \colorbox{Khaki2}{Let's think step by step. Joao Moutinho is a soccer player. Being out at third is part of baseball, not soccer.} So the answer is No.} \\
            \rowcolor{tGreen}{\fontfamily{ppl}\selectfont ...} \\
            \rowcolor{tGreen}{\fontfamily{ppl}\selectfont $<$Demonstrations$>$} \\
            \\
            {\fontfamily{ppl}\selectfont \textbf{Articles:} $<$insert articles, each delimited by triple quotes$>$} \\
            {\fontfamily{ppl}\selectfont \textbf{Question:} $<$insert question$>$} \\
            {\fontfamily{ppl}\selectfont \textbf{Answer:}} \\
            
            \midrule[0.9pt]
            \midrule[0.9pt]
            
            \rowcolor{LightSkyBlue1}{\fontfamily{ppl}\selectfont Prepare a meta-review by answering the following questions from the reviewer comments (provided after the questions).} \\
            \specialrule{0em}{1pt}{1pt}
            \rowcolor{Khaki1}{\fontfamily{ppl}\selectfont 1. Based on the reviewer’s comments, what are the core contributions made by this manuscript?} \\
            \rowcolor{Khaki1}{\fontfamily{ppl}\selectfont 2. What are the common strengths of this work, as mentioned by multiple reviewers?} \\
            \rowcolor{Khaki1}{\fontfamily{ppl}\selectfont 3. What are the common weaknesses of this work, as highlighted by multiple reviewers?} \\
            \rowcolor{Khaki1}{\fontfamily{ppl}\selectfont 4. What suggestions would you provide for improving this paper?} \\
            \rowcolor{Khaki1}{\fontfamily{ppl}\selectfont 5. What are the missing references mentioned by the individual reviews?} \\
            \specialrule{0em}{1pt}{1pt}
            \rowcolor{tGreen}{\fontfamily{ppl}\selectfont \textbf{The review texts are below:}  $<$insert three comments $R_1$, $R_2$, $R_3$ from the reviewers$>$} \\
            \rowcolor{tGreen}{\fontfamily{ppl}\selectfont \textbf{Meta-review:} $<$insert meta-review$>$} \\
            \rowcolor{tGreen}{\fontfamily{ppl}\selectfont ...} \\
            \rowcolor{tGreen}{\fontfamily{ppl}\selectfont $<$Demonstrations$>$} \\
            \\
            {\fontfamily{ppl}\selectfont Provide justification for your response in detail by explaining why you made the choices you actually made. A good output should be coherent, highlight major strengths/issues mentioned by multiple reviewers, be less than 400 words in length, and finally, the response should be in English only.} \\ 
            \\
            {\fontfamily{ppl}\selectfont \textbf{The review texts are below:} $<$insert three comments $R_1$, $R_2$, $R_3$ from the reviewers$>$} \\
            {\fontfamily{ppl}\selectfont \textbf{Meta-review:} } \\

            \midrule[0.9pt]
            \midrule[0.9pt]

            \rowcolor{tPink}{\fontfamily{ppl}\selectfont CREATE TABLE Highschooler (} \\
            \rowcolor{tPink}{\fontfamily{ppl}\selectfont ID int primary key,} \\
            \rowcolor{tPink}{\fontfamily{ppl}\selectfont name text,} \\
            \rowcolor{tPink}{\fontfamily{ppl}\selectfont grade int} \\
            \rowcolor{tPink}{\fontfamily{ppl}\selectfont );} \\
            \rowcolor{tPink}{\fontfamily{ppl}\selectfont /*} \\
            \rowcolor{tPink}{\fontfamily{ppl}\selectfont 3 example rows:} \\
            \rowcolor{tPink}{\fontfamily{ppl}\selectfont SELECT * FROM Highschooler LIMIT 3;} \\
            \rowcolor{tPink}{\fontfamily{ppl}\selectfont ID \quad name \quad grade} \\
            \rowcolor{tPink}{\fontfamily{ppl}\selectfont 1234 \quad Janie \quad 8} \\
            \rowcolor{tPink}{\fontfamily{ppl}\selectfont 5678 \quad Mary \quad 8} \\
            \rowcolor{tPink}{\fontfamily{ppl}\selectfont 9012 \quad Mike \quad 9} \\
            \rowcolor{tPink}{\fontfamily{ppl}\selectfont */} \\
            \specialrule{0em}{1pt}{1pt}
            \rowcolor{LightSkyBlue1}{\fontfamily{ppl}\selectfont Using valid SQLite, answer the following questions for the tables provided above.} \\
            \specialrule{0em}{1pt}{1pt}
            \rowcolor{tGreen}{\fontfamily{ppl}\selectfont \textbf{Question:}  What is Kyle's id?} \\
            \rowcolor{tGreen}{\fontfamily{ppl}\selectfont \textbf{SQL:} SELECT ID FROM Highschooler WHERE name=``Kyle'';} \\
            \rowcolor{tGreen}{\fontfamily{ppl}\selectfont ...} \\
            \rowcolor{tGreen}{\fontfamily{ppl}\selectfont $<$Demonstrations$>$} \\
            \\
            {\fontfamily{ppl}\selectfont \textbf{Question:} $<$insert question$>$} \\
            {\fontfamily{ppl}\selectfont \textbf{SQL:} } \\
            
            \bottomrule
	\end{tabular}
\end{table*}

\subsubsection{Prompt Optimization} \label{sec:prompt_opt} 
{Although manually creating task prompts is more intuitive, it is time consuming and, more importantly, models are highly sensitive to the crafted prompts---improper prompts will lead to low task performance (as shown in Table~\ref{tab-instructions}). Therefore, a large body of studies propose automatic optimization approaches for discrete prompts and continuous prompts to achieve the optimal performance~\cite{shin-EMNLP-2020-autoprompt,Li-ACL-2021-prefix}. In this part, we will detail these studies from two perspectives, \ie discrete prompts and continuous prompts.}

\paratitle{Discrete Prompt Optimization.}  {Discrete prompt is typically composed of a sequence of natural language tokens. Despite that the form is simple and flexible, optimizing prompts in discrete space is a challenging problem due to the combinatorial huge search space. To automatically search effective prompts for downstream tasks, existing studies propose a wide spectrum of discrete prompt optimization approaches, which are detailed as follows.}

$\bullet$  {\textit{Gradient-based approaches.} This kind of approaches aims to optimize the prompt search process by maximizing the output likelihood via gradient update~\cite{shin-EMNLP-2020-autoprompt,Wen-CoRR-2023-Hard,Gao-ACL-2021-Making,Zhou-CoRR-2023-InstructZero, Lin-arxiv-2023-Use}.
As a representative work, Auto-Prompt~\cite{shin-EMNLP-2020-autoprompt} proposes a gradient-guided method to greedily search the optimal token for each position of the prompt,  leveraging the gradient approximated by the change in the log-likelihood when replacing a prompt token with another candidate token from vocabulary.
However, such a search process can be extremely expensive since it needs to evaluate each candidate token for each position of the prompt, leading to a number of additional forward passes.
Therefore, an improved gradient method~\cite{Wen-CoRR-2023-Hard} has been proposed by transforming discrete tokens into continuous embeddings and computing the gradient on continuous space during optimization. 

$\bullet$ {\textit{RL-based approaches.} 
Since discrete prompts are difficult to be learned through gradient back-propagation, a number of studies propose to formulate the discrete prompt optimization as a reinforcement learning (RL) problem and leverage RL algorithms for optimization~\cite{Deng-EMNLP-2022-RLPrompt, Zhang-ICLR-2023-TEMPERA,Jafari-arxiv-2024-MORL,Kong-arxiv-2024-PRewrite}. For example, RLPrompt~\cite{Deng-EMNLP-2022-RLPrompt} {trains a policy network to generate desired prompts with multiple reward functions}. In this approach, several effective reward stabilization strategies are also proposed to enhance the RL training efficiency.  Compared to previous work that requires sufficient data for training, TEMPERA~\cite{Zhang-ICLR-2023-TEMPERA} proposes to edit prompts at test time {by utilizing a pre-trained RL agent to sequentially edit different parts of a manually-written initial prompt. 
{
Although these methods are simple and effective, they explore a manually defined edit space (\eg add, swap and delete) and focus on modifying the original prompt, which limits the flexibility of prompt search.
In contrast, PRewrite~\cite{Kong-arxiv-2024-PRewrite} employs RL to train a prompt rewriter for generating new prompts instead of modification, which does not impose any restrictions in the prompt rewriting and offers improved  flexibility in the action space.
}}

$\bullet$  {\textit{Edit-based approaches.} 
For the above methods, gradient-based and RL-based tuning can be extremely computationally demanding for ever larger models, and may not be feasible for API-based model calls (\eg ChatGPT). Therefore, another line of work aims to directly edit existing prompts based on the task performance. Specifically, GPS~\cite{Xu-EMNLP-2022-GPS} borrows an idea from the genetic algorithm and proposes a genetic prompt search method that utilizes a language model (\ie T5) to edit prompts by taking the cloze task form. In addition to model based edit methods, human-defined operations can be also employed for prompt editing~\cite{Prasad-EACL-2023-GrIPS}, including delete, swap, paraphrase, and addition. Based on these operations, they iteratively edit the prompts and greedily search for the best prompt guided by the model performance on a small pool of examples.

$\bullet$  {\textit{LLM-based approaches.} Due to the exceptional capacities of LLMs, an increasing number of studies directly leverage LLMs as prompt generator~\cite{Zhou-ICLR-2023-Large,Pryzant-CoRR-2023-Automatic,Yang-CoRR-2023-Large,Ye-arxiv-2023-prompt,Tang-arxiv-2024-unleashing,Yang-EMNLP-2023-InstOptima,Guo-arxiv-2023-connecting,Do-arxiv-2023-prompt}. 
Specifically, APE~\cite{Zhou-ICLR-2023-Large} utilizes an LLM to generate initial prompts, then selects the best prompt with the highest accuracy, and finally improves the best candidate through an iterative Monte Carlo search method.
{
\ignore{莫名其妙，没有任何转折和过渡}
However, this method does not effectively constrain the prompt search space, which might likely lead to unstable results.
To achieve good performance and fast convergence, one line of work utilizes heuristic methods (\eg evolutionary algorithms~\cite{Yang-EMNLP-2023-InstOptima,Guo-arxiv-2023-connecting} and adversarial learning~\cite{Do-arxiv-2023-prompt}) for prompt optimization.
Another line of work draws an analogy to gradient-based model optimizers for LLM-based prompt optimization.
}
{
For example, APO~\cite{Pryzant-CoRR-2023-Automatic} instructs the LLM to generate text feedback on how to refine an old prompt into new improved prompts and then execute textual gradient descent. 
}
However, their search in the prompt space might be inefficient {without fully considering the whole refinement trace of previous prompts}, thus potentially leading to sub-optimal results. 
{
Therefore, some recent studies~\cite{Yang-CoRR-2023-Large,Ye-arxiv-2023-prompt}  incorporate the previous prompts with their scores to instruct LLMs for progressively generating better new prompts.
}
To further design formalized guidelines about the design of prompt optimizers, GPO~\cite{Tang-arxiv-2024-unleashing} conducts a systematic analogy for LLM-based prompt optimizers with gradient-based model optimizers.
It further develops a {more formal} LLM-based prompt optimization framework, which extensively borrows the idea of machine learning optimization. 
Specifally, it 
retrieves relevant prompts from the previous prompts and utilizes the generation-based refinement strategy to perform the update. In order to avoid large variation at each iteration, GPO further adopts a cosine-based decay strategy to control the edit distance. 
However, these approaches still struggle in exploring the vast space of effective prompts. Inspired by human-like trial-and-error, prompt optimization is further formulated as a strategic planning problem~\cite{Wang-CoRR-2023-PromptAgent} and uses Monte Carlo tree search to navigate the vast prompt space.}

\paratitle{Continuous Prompt Optimization.} 
{Different from discrete prompts, continuous prompts consist of a set of continuous embeddings, which can be directly optimized through the gradient update based on the loss of downstream tasks. Note that continuous prompt optimization has been  mainly studied in PLMs, but draws limited attention in era of LLMs due to their massive magnitudes of parameters. We include the discussion of this part for content completeness. In prior work, most studies typically rely on supervised learning to train continuous prompts based on task data. Furthermore, in data-scarce scenarios, transfer learning methods can be employed to alleviate the lack of labeled data on target tasks. These two approaches are detailed below. 
}

$\bullet$ {\textit{Prompt learning  with sufficient data.} In this approach, most existing methods regard continuous prompts as trainable model parameters and then leverage supervised learning to optimize the continuous prompts by minimizing the cross-entropy loss based on sufficient downstream task data~\cite{Li-ACL-2021-prefix,Lester-ACL-2021-The,Tang-COLING-2022-Context,Liu-arXiv-2021-P-tuning}. As discussed in Section~\ref{sec-PEFT-methods}, prefix tuning~\cite{Li-ACL-2021-prefix} prepends a sequence of prefixes (\ie a set of trainable continuous vectors) to each Transformer layer in language models, while prompt tuning~\cite{Lester-ACL-2021-The} only incorporates trainable prompt vectors at the input layer. By fixing the large-scale parameters of LLMs and only tuning continuous prompt vector, this kind of approaches can be extremely parameter-efficient (Section~\ref{sec-PEFT}). However, these approaches are typically independent of the inputs, lacking sufficient consideration of input semantics. Therefore, the authors in \cite{Tang-COLING-2022-Context} propose context tuning, where the continuous prompts are derived based on the input text and learned through the downstream task losses.}

$\bullet$  {\textit{Prompt transferring  with scarce data.} Supervised learning approaches demand in sufficient training data to learn optimal continuous prompts, which may not work well in data-scarce domains and tasks. To address this problem, SPoT~\cite{Vu-ACL-2022-SPoT} proposes a prompt-based transfer learning approach, which first learns  
{a single continuous prompt} for several representative source tasks and then uses this prompt to initialize the prompt for a target task.  {However, this approach leverages the same prompt for solving all instances of the target task. For a single task, even a well-learned prompt may not be suitable for all the data instances from a large population.} To address this issue, an improved method~\cite{Li-NAACL-2022-Learning} designs an adaptive attention mechanism during the prompt transfer process to derive the target prompts,  considering both task- and instance-level information.  
{The prompt transfer paradigm can leverage the knowledge of data-sufficient source tasks encoded in source prompts for  solving  data-scarce target tasks.}
}
\subsection{In-Context Learning}
\label{subsec-icl}
As a special prompting form, in-context learning~(ICL) is first proposed along with GPT-3~\cite{Brown-NeurIPS-2020-Language}, which has become a typical approach to utilizing LLMs. 

\subsubsection{ICL Formulation}
\label{subsubsec-icl-formulation}

As stated in~\cite{Brown-NeurIPS-2020-Language}, ICL uses a formatted natural language prompt, consisting of the task description and/or a few task examples as demonstrations.
Figure~\ref{fig:utilization} presents an illustration of ICL.
First, starting with a task description, a few examples are selected from the task dataset as demonstrations.
Then, they are combined in a specific order to form natural language prompts with specially designed templates.
Finally, the test instance is appended to the demonstration as the input for LLMs to generate the output.
Based on task demonstrations, LLMs can recognize and perform a new task without explicit gradient update.

Formally, let $D_k = \{ f(x_1, y_1), \dots, f(x_k, y_k) \}$ represent a set of demonstrations with $k$ examples, where $f(x_k, y_k)$ is the prompt function that transforms the $k$-th task example into natural language prompts.
Given the task description $I$, demonstration $D_k$, and a new input query $x_{k+1}$, the prediction of the output $\hat{y}_{k+1}$ generated from LLMs can be formulated as follows\footnote{
When ICL was introduced in the GPT-3's paper~\cite{Brown-NeurIPS-2020-Language}, it was originally defined to be a combination of the task description and demonstration examples, wherein either component is dispensable. Following this definition, when a LLM is required to solve an unseen task by using only task descriptions, it can be also considered to perform ICL for task solving, whereas the ICL ability can be enhanced by instruction tuning.   
}: 
\begin{equation}\label{eq-ICL-prompting}
     \text{LLM} \big(I, \underbrace{ f(x_1, y_1), \dots, f(x_k, y_k)}_{\text{demonstrations}}, f(\underbrace{x_{k+1}}_{\text{input}}, \underbrace{\vphantom{\hat{y}_{k+1}} \_\_\_}_{\text{answer}}) \big) \rightarrow \hat{y}_{k+1}.
\end{equation}
where the actual answer $y_{k+1}$ is left as a blank to be predicted by the  LLM. %
Since the performance of ICL heavily relies on demonstrations, it is important to properly design them in the prompts. 
According to the construction process in Equation~\eqref{eq-ICL-prompting}, we focus on three major aspects of formatting demonstrations in the prompts, including how to select examples that make up demonstrations, format each example into the prompt with the function $f(\cdot)$, and arrange demonstrations in a reasonable order.  

{
A comprehensive review of ICL has been presented in the survey paper~\cite{Dong-arxiv-2023-A}, and we suggest the readers referring to it for a more general, detailed discussion on this topic. Compared with this survey, we specially focus on the discussion of applying ICL to LLMs in two major aspects, \ie demonstration design and the underlying mechanism of ICL. 
}
Also, ICL has a close connection with instruction tuning (discussed in Section~\ref{sec-instruction}) in that  
{both utilize natural language to format the task or instances}. 
However, instruction tuning needs to fine-tune LLMs for adaptation, while ICL only prompts LLMs for utilization.  
{Furthermore, instruction tuning can enhance the ICL ability of LLMs to perform target tasks, especially in the zero-shot setting (only using task descriptions)~\cite{Chung-arxiv-2022-Scaling}.  
}

\begin{figure*}[t]
    \centering
    \includegraphics[width=\textwidth]{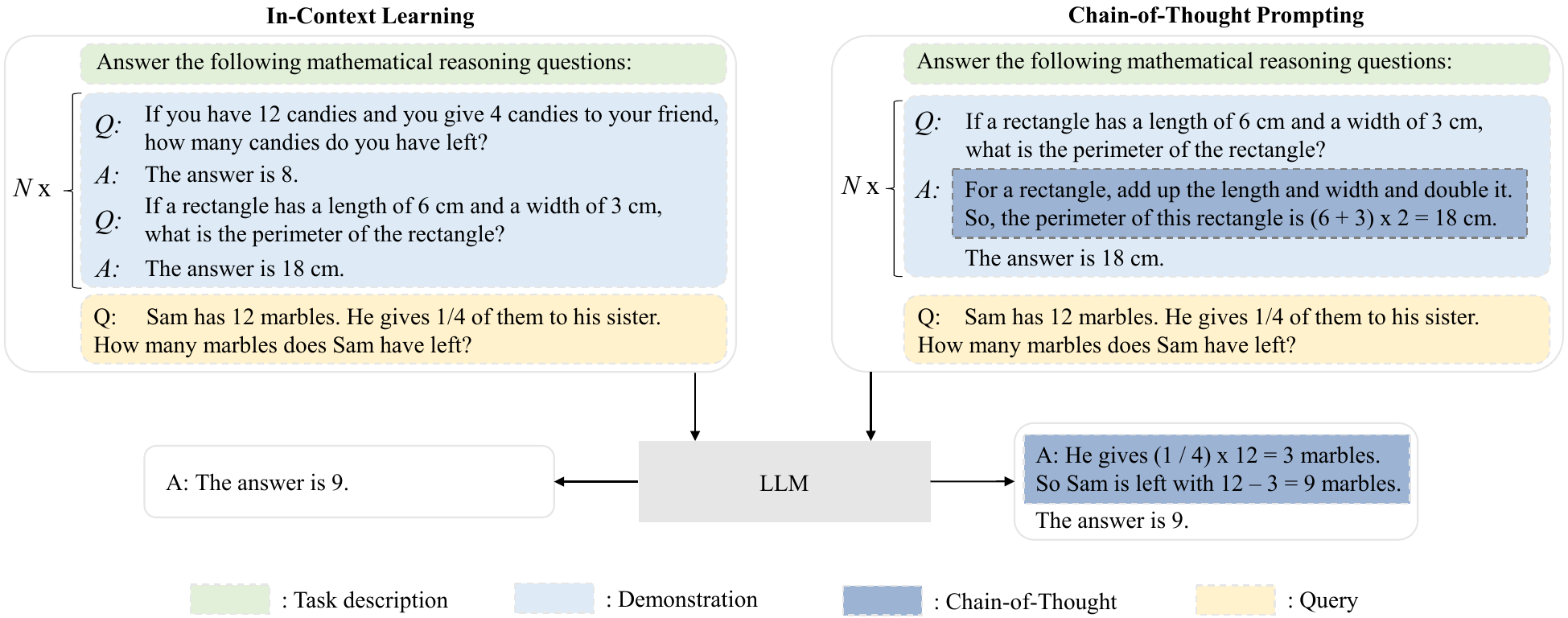}
    \caption{
        A comparative illustration of in-context learning~(ICL) and chain-of-thought~(CoT) prompting. 
        ICL prompts LLMs with a natural language description, several demonstrations, and a test query, while  
        CoT prompting involves a series of intermediate reasoning steps in prompts.
    }
    \label{fig:utilization}
\end{figure*}

\subsubsection{Demonstration Design}
{Several studies have shown that the effectiveness of ICL is highly affected by the design of demonstrations~\cite{Min-EMNLP-2022-Rethinking, Lu-ACL-2022-Fantasically,Zhao-ICML-2021-Calibrate}}
Following the discussion in {Section~\ref{subsubsec-icl-formulation}}, we will introduce the demonstration design of ICL from three major aspects, \ie demonstration selection, format, and order.

\paratitle{Demonstration Selection.}
{
The performance of ICL tends to have a large variance with different demonstration examples~\cite{Liu-ACL-2022-What}, so it is important to select a subset of examples that can effectively leverage the ICL capability of LLMs.}
There are two main demonstration selection approaches, namely heuristic and LLM-based approaches:

$\bullet$~\emph{Heuristic approaches.}  
{Due to their simplicity and low costs,} existing work widely adopts heuristic methods to select demonstrations.
Several studies employ a $k$-NN based retriever to select examples that are semantically relevant to the query~\cite{Liu-ACL-2022-What, Lee-COLING-2022-Does}.
{However, they perform the selection individually for each example, rather than evaluating the example set as a whole.}
To resolve this issue, diversity-based selection strategies are proposed to choose the most representative set of examples for specific tasks~\cite{Levy-arxiv-2022-Diverse, Su-arxiv-2022-selective}.
Furthermore, in~\cite{Ye-arxiv-2022-Complementary}, both relevance and diversity are taken into consideration when selecting demonstrations.

$\bullet$~\emph{LLM-based approaches.}  
Another line of work selects demonstrations by making use of LLMs. 
For example, LLMs can be utilized to directly measure the informativeness of each example according to the performance gain after adding the example~\cite{Li-arxiv-2023-Finding}. 
In addition, EPR~\cite{Rubin-NAACL-2022-Learning} proposes a two-stage retrieval approach that first recalls similar examples with an unsupervised method (\eg BM25) and then ranks them using a dense retriever (trained with positive and negative examples labeled by LLMs).
As an alternative approach, the task of demonstration selection can be formulated into a RL problem, where LLMs serve as the reward function to provide feedback for training the policy model~\cite{Zhang-EMNLP-2022-Active}. Since LLMs perform well for text annotation~\cite{Gilardi-arXiv-2023-Crowd}, some recent studies employ LLM itself as the demonstration generator without human intervention~\cite{Kim-arxiv-2022-Self-Generated}. 

{To summarize, as discussed in~\cite{Michael-ICLR-2022-An}, the selected demonstration examples in ICL should contain sufficient information about the task to solve as well as be relevant to the test query, for the above two selection approaches.} 

\paratitle{Demonstration Format.}
After selecting task examples, the next step is to integrate and format them into a natural language prompt for LLMs. 
A straightforward method is to instantiate a pre-defined template with the corresponding input-output pairs~\cite{Liu-survey-2023-Pre-train}.
To construct more informative templates, recent studies consider adding task descriptions~\cite{Chung-arxiv-2022-Scaling} or enhancing the reasoning capability of LLMs with chain-of-thought prompts~\cite{Wei-arxiv-2022-chain}.
For instance, in~\cite{Mishra-ACL-2022-Cross}, the authors collect a large-scale dataset with task descriptions written by humans.
After tuning with this dataset, the performance on seen tasks can be boosted, and LLMs can also generalize to unseen tasks to some extent.
To reduce the annotation costs, a semi-automated approach has been proposed in~\cite{Wang-arXiv-2022-Self} by employing a seed set consisting of human-written task descriptions to guide LLMs to generate task descriptions for new tasks. 
Since it is costly to manually annotate demonstration formats for different tasks, some work also studies how to automatically generate high-quality ones. 
As two representative methods, Auto-CoT~\cite{Zhang-arxiv-2022-Automatic} leverages LLMs with the zero-shot prompt ``\emph{Let’s think step by step}'' for generating intermediate reasoning steps, while least-to-most prompting~\cite{Zhou-arxiv-2022-Least} first queries LLMs to perform problem decomposition and then utilizes LLMs to sequentially solve sub-problems based on the intermediate answers to previously solved ones.  

\paratitle{Demonstration Order.}
LLMs are shown to sometimes suffer from the {recency} bias, \ie they are prone to repeat answers that are near the end of demonstrations~\cite{Zhao-ICML-2021-Calibrate}. 
Thus, it is important to arrange demonstrations (\ie task examples) in a reasonable order.
Early work proposes several heuristic methods to quickly find a good order.  
For example, demonstrations can be directly organized according to their similarity to the query in the embedding space~\cite{Liu-ACL-2022-What}: the more similar, the closer to the end.
In addition, global and local entropy metrics can be used to score different demonstration orders~\cite{Lu-ACL-2022-Fantasically}. 
To integrate more task information, some recent studies propose to minimize the  
{code length} required to compress and transmit task labels, which is inspired by information theory~\cite{Wu-arxiv-2022-Self}.
However, these methods need additional labeled data as the {validation set to evaluate the performance of specific demonstration orders}. 
To eliminate this need, the authors in~\cite{Lu-ACL-2022-Fantasically} propose to sample the validation data from the LLM itself. 

\subsubsection{Underlying Mechanism}
\label{sec-ICL-mechanism}
After pre-training, LLMs can exhibit intriguing ICL capability without being updated.  
In what follows, we discuss two key questions about the ICL ability of LLMs, \ie ``\emph{how does pre-training affect the ICL ability}'' and ``\emph{how do LLMs perform ICL during inference}''.

\paratitle{How Pre-Training Affects ICL?} 
ICL is first proposed in GPT-3~\cite{Brown-NeurIPS-2020-Language}, and it has been shown that the ICL ability becomes more significant with a larger model size.
Further, some studies reveal that small-scale PLMs can also demonstrate a strong ICL ability by continual pre-training~\cite{Gu-arXiv-2023-Pre} or fine-tuning~\cite{Min-NAACL-2022-MetaICL} on specially designed training tasks, which typically involve additional task examples in the input during the training process.
It suggests that the design of training tasks is an important influence factor on the ICL capability of LLMs. 
Besides training tasks, recent studies have also investigated the relationship between ICL and pre-training corpora~\cite{Michael-ICLR-2022-An, Hahn-2023-arXiv-a}.
For example, ICL can be theoretically explained as the product of pre-training on documents that exhibit long-range coherence~\cite{Michael-ICLR-2022-An}. 
{
Further, another study~\cite{Hahn-2023-arXiv-a} theoretically analyzes  that when scaling parameters and data, LLMs based on next-word prediction can emerge the ability of ICL by learning from the compositional structure (\eg how words and phrases are combined to form larger linguistic units like sentences) present in language data.  
}

\paratitle{How LLMs Perform ICL?}
At the inference stage, researchers focus on analyzing how the ICL capability operates based on given demonstrations since no explicit learning or updating is involved.
According to the discussion in~\cite{Pan-2023-arXiv-what}, there are two main ways for LLMs to utilize demonstrations: task recognition and task learning.  

$\bullet$~\emph{Task recognition.}
{In the first way, LLMs recognize the task from demonstrations and utilize the prior knowledge obtained from pre-training to solve new test tasks. 
A Probably Approximately Correct~(PAC) framework~\cite{Wies-2023-arXiv-the} has been proposed to assess the learnability of ICL.
It assumes that there exists a latent variable representing the task in the pre-training data, and LLMs have been shown to be capable of capturing this variable from demonstrations, enabling them to recognize the task in ICL.
Also, the interpretation of ICL as task recognition is  supported by several empirical studies~\cite{Min-EMNLP-2022-Rethinking, Webson-2022-NAACL-do}.
For example, it has been observed that replacing the inputs or labels of demonstrations with random ones sampled from the input or label space does not seriously hurt the performance of LLMs, indicating that LLMs mainly recognize the target task from demonstrations instead of learning from them~\cite{Min-EMNLP-2022-Rethinking, Pan-2023-arXiv-what}.
Similarly, LLMs can exhibit decent performance even if the prompt template is irrelevant or misleading~\cite{Webson-2022-NAACL-do}.}

$\bullet$~\emph{Task learning.}
{In the second way, LLMs learn new tasks unseen in the pre-training stage only through demonstrations.
Specially, task learning is   analyzed mainly from the perspective of gradient descent and considered as implicit fine-tuning~\cite{Oswald-arxiv-2022-Transformers, Dai-arxiv-2022-Why}.}
Then, ICL can be explained as follows: by means of forward computation, LLMs generate meta-gradients with respect to demonstrations and implicitly perform gradient descent via the attention mechanism.
Experiments also show that certain attention heads in LLMs are capable of performing task-agnostic atomic operations~(\eg copying and prefix matching), which are closely related to the ICL ability~\cite{Olsson-arxiv-2022-In}.
Furthermore, some studies abstract ICL as an algorithm learning process~\cite{rek-arxiv-2022-what}. 
For example, the authors in~\cite{rek-arxiv-2022-what} find that LLMs essentially encode implicit models through their parameters during pre-training.
With the examples provided in ICL, LLMs can implement learning algorithms such as gradient descent or directly compute the closed-form solution to update these models during forward computation.
Under this explanation framework, it has been shown that LLMs can effectively learn simple linear functions and even some complex functions like decision trees with ICL~\cite{rek-arxiv-2022-what}.

As discussed in a recent study~\cite{Pan-2023-arXiv-what}, LLMs exhibit the abilities of both task recognition and task learning in ICL, but the two abilities seem to be possessed with different model scales.
As shown in the experiments~\cite{Pan-2023-arXiv-what}, the ability of task recognition is easier to obtain, and even a small LM with only 350M parameters can exhibit this ability, while task learning can only emerge for LLMs with at least 66B parameters.
Another study~\cite{Wei-arxiv-2023-Larger} also supports this finding with specially designed experiments.
They set up the tasks with flipped and semantically unrelated labels in the experiment, which require task learning when performing ICL.
The results suggest that small LMs tend to disregard the labels and mainly depend on their prior knowledge to accomplish the task, while LLMs have the ability to surpass their prior knowledge and acquire new knowledge from demonstrations, resulting in better outcomes. 
Furthermore, to improve the task learning ability, Meta-In-Context Learning~\cite{Forno-2023-arXiv-meta} proposes to include multiple related tasks instead of just a single one in the prompt.
In addition, Symbol Tuning~\cite{Wei-2023-arXiv-symbol} fine-tunes LLMs on demonstrations with semantically unrelated labels (\eg foo/bar instead of positive/negative for sentiment analysis), forcing LLMs to learn the task from demonstrations instead of relying on prior knowledge.
\begin{figure*}[t]
    \centering
    \includegraphics[width=\textwidth]{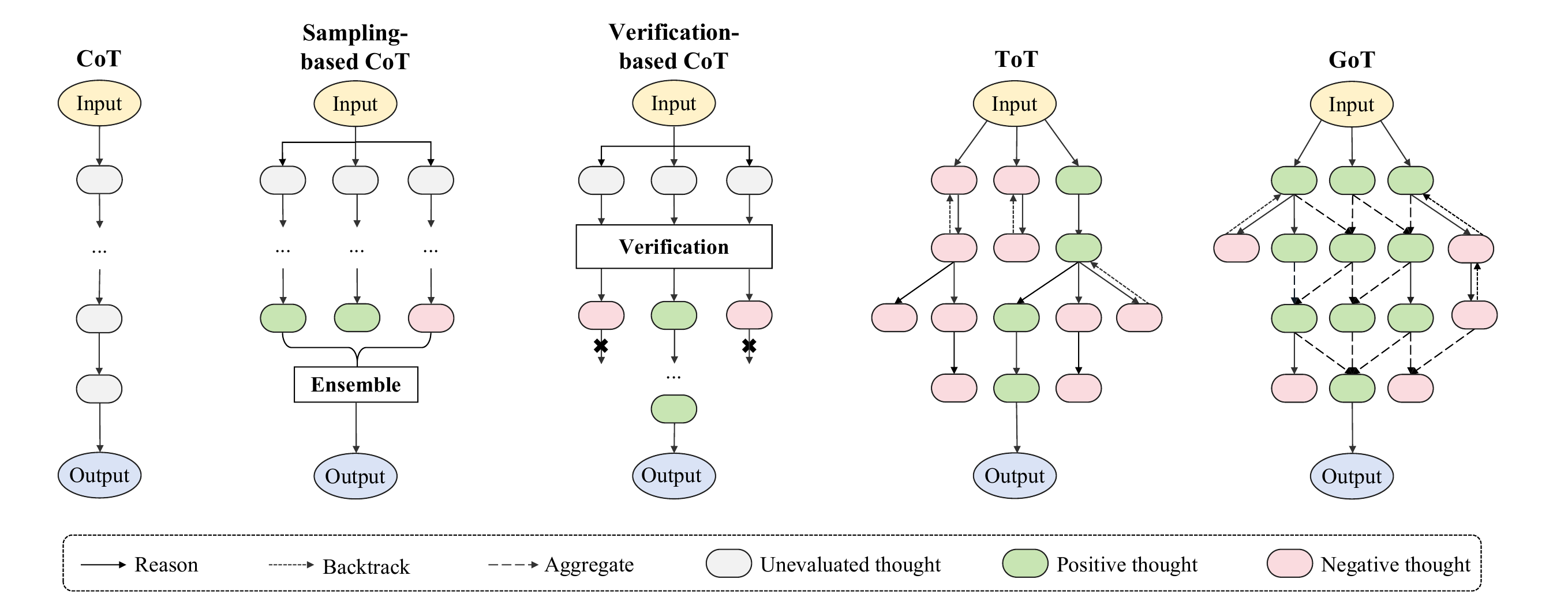}
    \caption{
        An illustration of the evolution of CoT prompting strategies. It begins with the basic CoT approach and progresses to enhanced CoT generation techniques,  including sampling-based and verification-based methods. Finally, it extends to variations of the chain structure, such as trees and graphs. Here, ``thought'' refers to an intermediate reasoning step as stated in~\cite{Wei-arxiv-2022-chain, Yao-arxiv-2023-Tree}.
    }
\label{fig:extension_of_CoT}
\end{figure*}

\subsection{Chain-of-Thought Prompting}
\label{subsec-cot}

Chain-of-Thought~(CoT) prompting~\cite{Wei-arxiv-2022-chain, Chu-arxiv-2023-A} is an improved prompting strategy to boost the performance of LLMs on complex reasoning tasks, such as arithmetic reasoning~\cite{Miao-ACL-2020-A}, commonsense reasoning~\cite{Talmor-naacl-2019-CommonsenseQA}, and symbolic reasoning~\cite{Wei-arxiv-2022-chain}.
Instead of simply constructing the prompts with input-output pairs like ICL, CoT prompting further incorporates intermediate reasoning steps, which serve as the bridge between inputs and outputs.
{
Figure~\ref{fig:utilization} presents an illustration of CoT.
In the following part, we will first elaborate on the basic CoT prompting approach and its improved strategies, then discuss when and why CoT prompting works.
}

\subsubsection{Basic CoT Prompting Approach}

CoT prompting is first proposed as an extension of ICL~\cite{Wei-arxiv-2022-chain}, which augments each demonstration $\langle$\emph{input, output}$\rangle$ as $\langle$\emph{input, CoT, output}$\rangle$.
A \textit{CoT} is a series of intermediate reasoning steps for connecting the \textit{input} and \textit{output}.
With these augmented demonstrations, LLMs can follow them to  
{generate CoTs and the answer for a new input.} 
However, unlike $\langle$\emph{input, output}$\rangle$ pairs in ICL, CoTs are difficult to obtain and usually require human annotation.
Fortunately, it has been found that LLMs can be triggered to generate CoTs through simple instructions like ``\emph{Let's think step by step.}''~\cite{Kojima-arxiv-2022-Large}, making CoT prompting easy to use.
There are also alternative magic prompts that {can elicit the ability of CoT reasoning and further improve the performance of LLMs}, such as ``\emph{Take a deep breath and work on this problem step-by-step.}''~\cite{Yang-CoRR-2023-Large}.

{
As illustrated in Figure~\ref{fig:extension_of_CoT}, the generation process of CoT follows a chain structure in the basic CoT prompting approach, where LLMs generate CoTs step by step.
Typically, CoT takes the format of natural language text.
However, textual CoTs may not work well on complex tasks that require rigorous logic for reasoning.
Considering this, some work uses code~\cite{Chen-arxiv-2022-Program, Gao-ICML-2023-PAL} due to its structured and precise nature.
Furthermore, the authors in~\cite{Zhao-arxiv-2023-Automatic} propose to dynamically select text or code as the format of CoTs to combine their advantages.
}

\subsubsection{Improved CoT Prompting Strategies}

{
Despite the performance improvement in complex reasoning tasks, CoT prompting still suffers from problems like incorrect reasoning and instability.
In this part, we first introduce how to design better CoT prompts and enhanced CoT generation strategies, and then introduce the extension of the basic chain structure of CoT.
Figure~\ref{fig:extension_of_CoT} illustrates the evolution of representative CoT prompting strategies.
}

\paratitle{Better Prompt Design.}
Since CoT prompting relies on prompts to elicit the reasoning capabilities of LLMs, the design of prompts is critical to its performance.
As a direct approach, it is shown that using diverse CoTs (\ie multiple reasoning paths for each problem) can effectively enhance the performance~\cite{Li-arxiv-2022-On}.
Another intuitive idea is that prompts with more complex reasoning paths are more likely to elicit the reasoning ability of LLMs~\cite{Fu-arxiv-2022-Complexity}, which can result in higher accuracy in generating correct answers.
However, all these approaches rely on annotated CoT datasets, which limits their use in practice. 
To overcome this limitation, magic instructions such as  ``\emph{Let's think step by step}'' can be used to automatically construct CoTs by prompting LLMs~\cite{Zhang-arxiv-2022-Automatic}. 


\paratitle{Enhanced CoT Generation.}
{
Since LLMs are prone to producing incorrect reasoning steps and exhibiting instability in the generation process, there are a number of studies~\cite{Li-arxiv-2023-Making, Wang-arxiv-2022-Self-Consistency} to improve the generation of CoT.
In this part, we will introduce two typical approaches to enhancing the generation of CoT: sampling- and verification-based methods.
}

$\bullet$ \emph{Sampling-based methods.}
{
LLMs are known to suffer from instability during inference, which can lead to unfaithfulness in the generated reasoning steps.
To address this issue, some work proposes to sample multiple reasoning paths instead of using greedy decoding.
As a representative solution, self-consistency~\cite{Wang-arxiv-2022-Self-Consistency} 
first generates several reasoning paths and then takes an ensemble over the corresponding answers,  selecting the most consistent one through majority voting.
However, such a method can still lead to wrong answers when most of the reasoning paths are misled.
Considering this, the authors in~\cite{Fu-arxiv-2022-Complexity} only vote on the $k$ most complex reasoning paths based on their observation that reasoning paths with higher complexity (\eg more reasoning steps) usually have better performance. 
{Furthermore, MCR~\cite{Yoran-arxiv-2023-Answering} proposes referring to the steps from other reasoning paths when generating the next step, and performs reasoning across multiple reasoning paths to generate the final answer.}
}

$\bullet$ \emph{Verification-based methods.} {
The sequential nature of reasoning steps in CoTs can lead to the accumulation of errors in the generated CoTs when certain steps are incorrect. 
To mitigate this problem, recent studies propose to verify the correctness of generated reasoning steps with either trained verifiers or LLMs themselves. 
For example, DIVERSE~\cite{Li-arxiv-2023-Making} trains solution-level and step-level verifiers respectively to examine the reasoning steps at different granularities. 
Another approach~\cite{Ling-arxiv-2023-Deductive} utilizes LLMs to verify the correctness of reasoning steps through step-by-step self-verification with a specially designed reasoning format.
In addition, several studies propose backward reasoning for verification: 
it first deduces the necessary question conditions~\cite{Xue-arxiv-2023-RCOT, Weng-arxiv-2023-Large} or variables~\cite{Jiang-arxiv-2023-Forward} {from the model's predictions}, and then compares them with the original ones.
}

\paratitle{Reasoning Structure Extension.}   
{
Despite the generality, the chain reasoning structure of basic CoT prompting limits its effectiveness in solving complex tasks, which require exploration like foresight and backtracking during inference.
Therefore, many studies have been devoted to extending the reasoning structure by designing more intricate thought processes, \eg tree- and graph-structured reasoning. 
}

$\bullet$ \emph{Tree-structured reasoning.} 
This approach (exemplified by Tree of Thoughts~(ToT)~\cite{Yao-arxiv-2023-Tree, Long-arxiv-2023-Large}) formulates the reasoning process in a hierarchical tree structure, where intermediate thoughts are nodes.
{In this way, it enables  LLMs to explore multiple reasoning paths in parallel and further supports the operation of lookahead and backtracking to facilitate more comprehensive decisions.} 
In addition, TouT~\cite{Mo-arxiv-2023-Tree} takes the uncertainty of intermediate thoughts into account for thought evaluation based on Monte Carlo Dropout.

$\bullet$ \emph{Graph-structured reasoning.} {
Although the tree structure facilitates parallel reasoning, it also imposes restrictions on the reasoning process.
With more complex topological structures, graphs offer greater flexibility in reasoning, enabling the characterization of more intricate relationships and interactions.
For instance, Graph of Thoughts~(GoT)~\cite{Besta-arxiv-2023-Graph, Lei-arxiv-2023-Boosting} conceptualizes the reasoning process as an arbitrary graph, where vertices denote intermediate thoughts and edges denote the interdependence between these thoughts. 
{Compared with ToT, it can further utilize thoughts from other reasoning paths when generating new thoughts.}
However, such an approach requires a large number of interactions with LLMs, making the thought exploration process highly inefficient. 
} 
{
To reduce potentially meaningless thought exploration, XoT~\cite{ding-arxiv-2023-everything} further proposes to guide the search of thoughts with pre-trained policy and value networks.
}

\subsubsection{Further Discussion on CoT Prompting}
In this part, we present discussions regarding two fundamental questions related to CoT prompting, \ie ``\textit{when does CoT prompting work for LLMs}'' and ``\textit{why can LLMs perform CoT reasoning}''.

\paratitle{{When CoT Prompting Works For LLMs?}} 
Since CoT reasoning is an emergent ability~\cite{Wei-arxiv-2022-Emergent}, it only has a positive effect on sufficiently large models (typically containing 10B or more parameters~\cite{Wei-arxiv-2022-chain}) but not on small models. 
Moreover, since CoT prompting augments the standard prompting with intermediate reasoning steps, it is mainly effective for the tasks that require step-by-step reasoning~\cite{Wei-arxiv-2022-chain}, \eg arithmetic reasoning, commonsense reasoning, and symbolic reasoning.
Whereas, for other tasks that do not rely on complex reasoning, CoT prompting might lead to worse performance than standard prompting~\cite{Wang-arxiv-2022-Rationale}, \eg MNLI-m/mm, SST-2, and QQP from GLUE~\cite{Wang-EMNLP-2018-GLUE}.   
Interestingly, it seems that the performance gain brought by CoT prompting could be significant only when standard prompting yields poor results~\cite{Wei-arxiv-2022-chain}.

\paratitle{{Why LLMs Can Perform CoT Reasoning?}} 
As the second question, we discuss the underlying mechanism of CoT prompting in the following two aspects. 

$\bullet$ \emph{The source of CoT reasoning ability}. 
Regarding the source of CoT reasoning capability, it is widely hypothesized that it can be attributed to training on code since models trained on it show a strong reasoning ability~\cite{Liang-arxiv-2022-Holistic, FU-blog-2022-how, Bi-arxiv-2023-When}.   
Intuitively, code data is well organized with algorithmic logic and programming flow, which may be useful to improve the reasoning performance of LLMs. 
However, this hypothesis still lacks publicly reported evidence of ablation experiments (\emph{with} and \emph{without} training on code). 
In addition, instruction tuning seems not to be the key reason for obtaining the CoT reasoning ability, since 
it has been empirically shown that instruction tuning on non-CoT data does not improve the performance on held-out CoT reasoning benchmarks~\cite{Chung-arxiv-2022-Scaling}.

$\bullet$ \emph{The effect of CoT prompting components}. 
The major distinction between CoT prompting and standard prompting is the incorporation of reasoning paths prior to the final answer. 
Thus, some researchers investigate the effects of different components in the reasoning paths. 
Specifically, a recent study identifies three key components in CoT prompting, namely  \emph{symbols}~(\eg numerical quantities in arithmetic reasoning), \emph{patterns}~(\eg equations in arithmetic reasoning), and \emph{text}~(\ie the rest of tokens that are not symbols or patterns)~\cite{Madaan-arxiv-2022-Text}. 
It is shown that the latter two parts (\ie patterns and text) are essential to the model performance, and removing either one would lead to a significant performance drop. 
However, the correctness of symbols and patterns does not seem critical. 
Further, there exists a symbiotic relationship between text and patterns:   the text helps LLMs to generate useful patterns, and patterns aid LLMs to understand tasks and generate texts that help solve them~\cite{Madaan-arxiv-2022-Text}.


In summary, CoT prompting provides a general and flexible approach to eliciting the reasoning ability of LLMs. 
There are also some preliminary attempts to extend this technique to solve multimodal~\cite{Zhang-arxiv-2022-Multimodal} and multilingual tasks~\cite{Shi-arxiv-2022-Language}.
\subsection{Planning}
\label{subsec-planning}
Prompting with ICL and CoT is a conceptually simple yet general approach to solving various tasks. 
However, this approach struggles with complex tasks like mathematical reasoning~\cite{Qian-2022-arXiv-limitations} and multi-hop question answering~\cite{Ning-arxiv-2023-ChatGPT}.
As an enhanced approach, prompt-based planning has been proposed to break down complex tasks into smaller sub-tasks and generate a plan of actions to accomplish the task.

\subsubsection{The Overall Framework}

\begin{figure}[t]
    \centering
    \includegraphics[width=\linewidth]{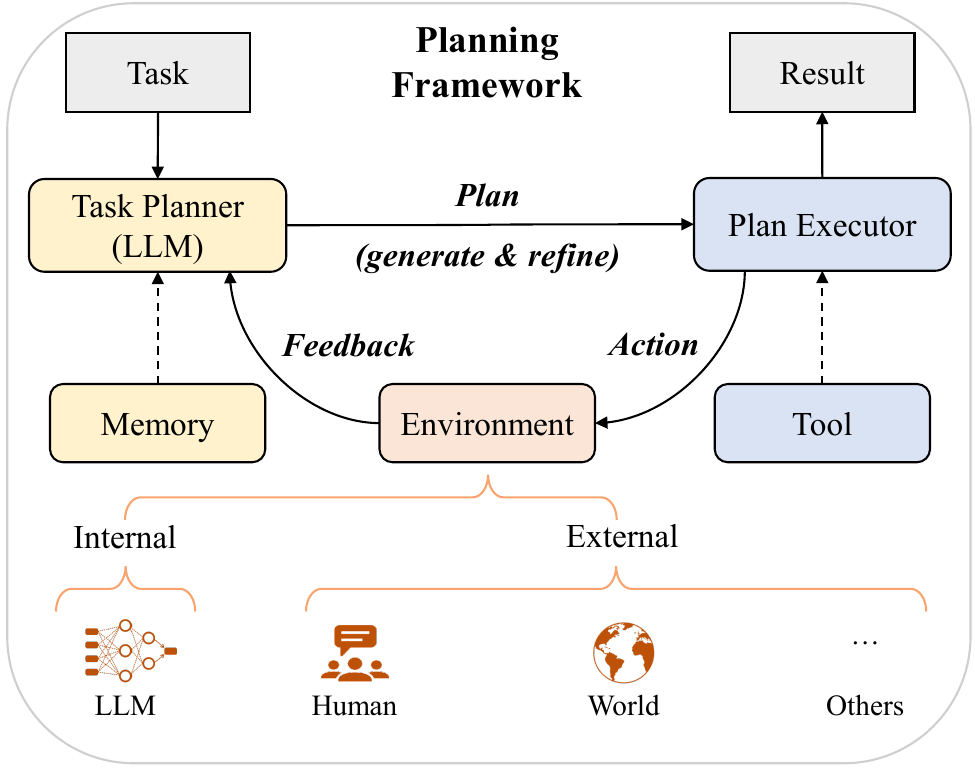}
    \caption{An illustration of the formulation for prompt based planning by LLMs for solving complex tasks.}
    \label{fig:planning}
\end{figure}

In this part, we first formulate the general planning paradigm of LLMs for solving complex tasks, which is illustrated in Figure~\ref{fig:planning}.

In this paradigm, there are typically three components: \emph{task planner}, \emph{plan executor}, and \emph{environment}\footnote{
Despite the similarity with RL, our formulation decouples the planning and execution phases, whereas in RL, they are typically interleaved in the agent.
This paradigm is defined in a general yet slightly loose way, and it mainly aims to help readers understand the key idea underlying the planning approaches of LLMs.
}.
Specifically, task planner, which is played by LLMs, aims to generate the whole plan to solve a target task.
The plan can be presented in various forms, \eg an action sequence in the form of natural language~\cite{Zhou-arxiv-2022-Least} or an executable program written in programming language~\cite{Gao-arxiv-2022-PAL}.
The LLM-based task planner can be enhanced with the memory mechanism for plan storage and retrieval, which is helpful for long-horizon tasks.
Then, plan executor is responsible for executing the actions in the plan.
It can be implemented by models like LLMs for textual tasks~\cite{Wang-arXiv-2023-Plan} or by tools like code interpreters for coding tasks~\cite{Shinn-2023-arXiv-Reflexion}.
Furthermore, environment refers to where the plan executor carries out the actions, which can be set differently according to specific tasks, \eg the LLM itself~\cite{Yao-2023-arXiv-tree} or an external virtual world like Minecraft~\cite{Wang-2023-arXiv-voyager}.
It provides \textit{feedback} about the execution result of the action to the task planner, either in the form of natural language~\cite{Shinn-2023-arXiv-Reflexion} or from other multimodal signals~\cite{Lu-2023-arXiv-multimodal}.

For solving a complex task, the task planner first needs to clearly understand the task goal and generate a reasonable plan based on the reasoning of LLMs (See Section~\ref{sec:plan-gen}).
Then, the plan executor acts according to the plan in the environment, and the environment will produce feedback for the task planner (See Section~\ref{sec:feedback}).
The task planner can further incorporate the feedback obtained from the environment to refine its initial plan and iteratively perform the above process to get better results as the task solution (See Section~\ref{sec:plan-refine}).


\subsubsection{Plan Generation}
\label{sec:plan-gen}
Plan generation focuses on directly generating action sequences by prompting LLMs.
Based on the format of the generated plans, existing work can be divided into two groups: text-based and code-based approaches. 

\paratitle{Text-based Approaches.}
It is straightforward for LLMs to generate plans in the form of natural language.
In this approach, LLMs are prompted to generate a sequence of actions for the plan executor to perform and solve the complex task.
For example, Plan-and-Solve~\cite{Wang-arXiv-2023-Plan} adds explicit instructions like ``\texttt{devise a plan}'' to directly prompt the LLM for planning in a zero-shot manner, while Self-planning~\cite{Jiang-arXiv-2023-Self} and DECOMP~\cite{Khot-2022-arXiv-Decomposed} add demonstrations in the prompt to guide the LLM to devise a plan through ICL.
Following this way, some work further considers incorporating extra tools or models when planning. 
For example, ToolFormer~\cite{Schick-arxiv-2023-Toolformer} first annotates a pre-training corpus with potential API calls using LLMs, and then fine-tunes LLMs on it,  so that LLMs can learn when and how to call APIs and incorporate the results returned by APIs during generation.
HuggingGPT~\cite{Shen-2023-arXiv-Hugginggpt} introduces the models available in HuggingFace and regards LLMs as the controller to select suitable models based on their descriptions and aggregate their results as the final solution.

\paratitle{Code-based Approaches.}
Although text-based approaches sound intuitive, they cannot guarantee faithful execution of the plan, which may lead to failure even when the plan is sound. 
To address this issue, code-based approaches have been proposed to generate more verifiable plans in the form of executable code in  programming languages, \eg Python or PDDL.
In this way, LLMs are first prompted to generate the program and then utilize a deterministic solver to execute it.
For example, Faithful CoT~\cite{Lyu-arxiv-2023-Faithful} and PAL~\cite{Gao-arxiv-2022-PAL} decompose a reasoning task into two stages: at the first stage, the LLM generates a plan conditioned on the query; at the second stage, a deterministic solver executes the plan to derive the final answer. 
Furthermore, code-based approaches can be applied to embodied agents in a similar way. 
For example, PROGPROMPT~\cite{Singh-arxiv-2022-ProgPrompt} and LLM+P~\cite{Liu-2023-arXiv-LLM+P} first utilize LLMs to generate plans in the form of python functions or PDDL files, and then leverage a virtual agent or classical planner to solve the problem according to the code-based plans.

\subsubsection{Feedback Acquisition}
\label{sec:feedback}
After executing the generated plan, the environment would produce the feedback signal to the LLM-based task planner, which can be used to refine its initial plan for better results.
In existing work, there are typically two sources of feedback from the environment, depending on their relationship with the LLM-based task planner: internal (\ie the LLM itself) and external (\eg tools or virtual worlds) feedback.

\paratitle{Internal Feedback.}
The LLM itself can be utilized as a feedback provider.
One straightforward way is to directly evaluate the quality of the generated plans through prompting.
For example, RAP~\cite{Hao-2023-arXiv-reasoning} evaluate the likelihood that each candidate plan can lead to task success, while Tree of Thoughts~\cite{Yao-2023-arXiv-tree} proposes to vote across plans by making comparisons between them.
Further, LLMs can provide feedback based on the intermediate results from the plan executor.
For example, Reflexion~\cite{Shinn-2023-arXiv-Reflexion} utilizes LLMs to transform sparse result signals (\eg success or failure) into concrete {text-based feedback (\eg ``\emph{You should recommend comedies that the user mentions in the query instead of horror movies}'') and stores this feedback in long-term memory for future planning.}

\paratitle{External Feedback.}
In addition to LLMs, external objects can also provide feedback signals.
For example, tools like code interpreters are widely used in programming tasks to provide real-time error messages~\cite{Shinn-2023-arXiv-Reflexion}, models like stable diffusion~\cite{Rombach-2022-CVPR-high} can be used in multimodal tasks to provide visual perception~\cite{Lu-2023-arXiv-multimodal}, and {virtual worlds} like Minecraft can provide immersive experiences~\cite{Wang-2023-arXiv-voyager}.
Besides, some work (\eg Generative Agents~\cite{Park-arxiv-2023-Generative}) explores multi-agent collaboration in simulated environments, where each agent receives feedback not only from interaction with the environment but also from communication with other agents.

\subsubsection{Plan Refinement}
\label{sec:plan-refine}
With access to feedback from the environment, the task planner can accordingly refine its current plan and iteratively go through the ``\emph{planning -- execution -- refinement}'' loop for better results.
In this part, we summarizes three major refinement approaches in existing work. 

\paratitle{Reasoning.}
The feedback data from the environment may not be directly suitable to be utilized by LLMs for plan refinement, \eg containing irrelevant information or taking a non-language form.
To solve this, some work adds the explicit reasoning process to extract critical information from feedback~\cite{Yao-2022-arXiv-react, Chen-2023-arXiv-chatcot}.
For example, React~\cite{Yao-2022-arXiv-react} prompts LLMs with demonstrations to generate reasoning traces over feedback.
It has been widely used in autonomous agent projects, such as AutoGPT~\cite{AutoGPT}, which can automatically reason over the observed feedback to revise the initial plan for solving various user requests.
However, these approaches typically fix the order of reasoning and planning.
To support flexible switching between the two processes for better performance, ChatCoT~\cite{Chen-2023-arXiv-chatcot} further unifies the tool-augmented reasoning process into a multi-turn conversation between the LLM-based task planner and the tool-based environment.

\paratitle{Backtracking.}
Early methods mainly consider planning forward actions while maintaining the existing plan, thus likely leading to local optimal plans based on a short-term evaluation.
To solve this, Tree of Thoughts~\cite{Yao-2023-arXiv-tree} allows backtracking with search algorithms like breadth-first and depth-first search to make global planning.
It refines the plan step by step by backtracking to the last state in the initial plan and choosing the next unexplored action.
Furthermore, some studies~\cite{Wang-2023-arXiv-describe, Lu-2023-arXiv-multimodal} utilize feedback signals to revise the entire plan.
For example, DEPS~\cite{Wang-2023-arXiv-describe} selects a better plan according to feedback signals, while TIP~\cite{Lu-2023-arXiv-multimodal} adds feedback signals to prompts for the LLM-based planner to revise each step in the initial plan.

\paratitle{Memorization.}
In order to handle long-horizon tasks, it has become a key approach to aid plan refinement with \emph{long-term memory} in addition to utilizing the \emph{short-term memory} of LLMs through ICL.
For example, Reflexion~\cite{Shinn-2023-arXiv-Reflexion} stores the feedback from self-reflection into the memory, so previous feedback can be retrieved for plan refinement.
Generative Agents~\cite{Park-arxiv-2023-Generative} designs the memory stream mechanism as the core component of agents for action planning and reflection.
Further, the skill library mechanism~\cite{Wang-2023-arXiv-voyager, Sun-2023-arXiv-adaplanner} is proposed to store successful plans in the library, which can be reused and synthesized as complex plans for novel tasks.
To implement the long-term memory mechanism, tools like vector databases (\eg milvus~\cite{Wang-2021-ICDM-Milvus}) can be used to encode plans or feedbacks into high-dimensional vectors for efficient storage and retrieval at a large scale.
MemoryBank~\cite{Zhong-2023-arxiv-MemoryBank} further proposes the memory updating mechanism to allow memory forgetting and strengthening following the Ebbinghaus Forgetting Curve theory.

\section{Capacity and Evaluation}
\label{sec-evaluation}
To examine the effectiveness and superiority of LLMs, a surge of tasks and benchmarks have been proposed for conducting empirical ability evaluation and analysis.
In this section, we first introduce three types of basic ability evaluation of LLMs for language generation and understanding, then present several advanced ability evaluations with more complicated settings or goals, and finally discuss existing benchmarks, evaluation approaches, and empirical analysis.

\begin{table*}[htbp]
    \centering  
    \caption{Representative basic and advanced abilities and corresponding representative datasets for evaluating.}
    \footnotesize
    \renewcommand\tabcolsep{2.5pt}
    \begin{tabular}{cccc}
        \toprule
        \textbf{Level} & \textbf{Ability} & \textbf{Task} & \textbf{Dataset} \\
        \midrule
        \multirow{27}{*}{Basic} & \multirow{6}{*}{Language Generation}    &Language Modeling             &Penn Treebank~\cite{Marcus-CL-1993-Building}, WikiText-103~\cite{Merity-ICLR-2017-Pointer}, the Pile~\cite{Gao-arxiv-2021-Pile}, LAMBADA~\cite{Paperno-ACL-2016-LAMBADA} \\ 
        \addlinespace
                                            &    &\multirow{3}{*}{Conditional Text Generation}   & WMT'14,16,19,20,21,22~\cite{Bojar-WMT-2014-Findings,Bojar-WMT-2016-Findings,Barrault-WMT-2019-Findings,Barrault-WMT-2020-Findings,Akhbardeh-WMT-2021-Findings,Kocmi-WMT-2022-Findings}, Flores-101~\cite{Goyal-TACL-2022-The}, DiaBLa~\cite{Bawden-journal-2021-DiaBLa}, \\ 
                                              &  & &  
                                                CNN/DailyMail~\cite{Nallapati-acl-2016-Abstractive}, XSum~\cite{Naryan-EMNLP-2018-XSUM}, WikiLingua~\cite{Ladhak-EMNLP-2020-WikiLingua}\\
                                            & & & OpenDialKG~\cite{Moon-ACL-2019-OpenDialKG} \\
                                                
                                                 \addlinespace
                                             &   &\multirow{2}{*}{Code Synthesis}                & APPS~\cite{Hendrycks-nips-2021-Measuring}, HumanEval~\cite{Chen-arxiv-2021-evaluating}, MBPP~\cite{Austin-arxiv-2021-Program}, CodeContest~\cite{Li-Science-2022-AlphaCode},
                                                MTPB~\cite{nijkamp-arxiv-2022-Codegen},
                                                \\ 
                                             &   &   &
                                                DS-1000~\cite{Lai-arxiv-2022-DS}, ODEX~\cite{Wang-arxiv-2022-Execution} \\
                                                \cmidrule(r){2-4}
        &\multirow{9}{*}{Knowledge Utilization}  
            &\multirow{3}{*}{Closed-Book QA}                &  
                Natural Questions~\cite{Kwiatkowski-ACL-2019-Natural}, 
                ARC~\cite{Clark-arxiv-2018-Think},  
                TruthfulQA~\cite{Lin-ACL-2022-TruthfulQA}, 
                Web Questions~\cite{Berant-EMNLP-2013-Semantic},\\
          &  &           & 
                TriviaQA~\cite{Joshi-ACL-2017-TriviaQA}, 
                PIQA~\cite{Bisk-AAAI-2020-PIQA}, 
                LC-quad2.0~\cite{Dubey-ISWC-2019-LC}, 
                GrailQA~\cite{Gu-WWW-2021-Beyond}, 
                KQApro~\cite{Cao-ACL-2022-KQA},\\ 
          &  &           &
                CWQ~\cite{Hu-COLING-2022-Logical}, 
                MKQA~\cite{Longpre-TACL-2021-MKQA}, 
                ScienceQA~\cite{Saikh-IJDL-2022-ScienceQA} \\
            \addlinespace
         &   &\multirow{3}{*}{Open-Book QA}                  &  
                Natural Questions~\cite{Kwiatkowski-ACL-2019-Natural}, 
                OpenBookQA~\cite{Mihaylov-EMNLP-2018-Can}, 
                ARC~\cite{Clark-arxiv-2018-Think}, 
                TriviaQA~\cite{Joshi-ACL-2017-TriviaQA}, \\
          &  &           &  
                 Web Questions~\cite{Berant-EMNLP-2013-Semantic},
                MS MARCO~\cite{Nguyen-NIPS-2016-MS}, 
                QASC~\cite{Khot-AAAI-2020-QASC}, 
                SQuAD~\cite{Rajpurkar-EMNLP-2016-SQuAD}, 
                \\
        &  &           &  
                WikiMovies~\cite{Miller-EMNLP-2016-Key} \\
            \addlinespace
          &  &\multirow{2}{*}{Knowledge Completion}          & 
                WikiFact~\cite{Goodrich-KDD-2019-Assessing}, 
                FB15k-237~\cite{Toutanova-CVSC-2015-Observed}, 
                Freebase~\cite{Bollacker-SIGMOD-2008-Freebase}, 
                WN18RR~\cite{Dettmers-AAAI-2018-Convolutional}, \\
          &  &           &
                WordNet~\cite{Miller-Commun-1995-WordNet},
                LAMA~\cite{Petroni-EMNLP-2019-Language}, 
                YAGO3-10~\cite{Mahdisoltani-CIDR-2015-YAGO3}, 
                YAGO~\cite{Suchanek-WWW-2007-Yago}\\ 
        \cmidrule(r){2-4}
        & \multirow{12}{*}{Complex Reasoning}      &\multirow{4}{*}{Knowledge Reasoning}           &  CSQA~\cite{Talmor-naacl-2019-CommonsenseQA}, StrategyQA~\cite{Geva-tacl-2021-Did}, HotpotQA~\cite{yang-2018-acl-HotpotQA}, ARC~\cite{Clark-arxiv-2018-Think}, BoolQ~\cite{Clark-naacl-2019-BoolQ},  \\ 
                                              &  &   &
                                                PIQA~\cite{Bisk-AAAI-2020-PIQA},
                                                SIQA~\cite{Sap-arxiv-2019-SocialIQA}, HellaSwag~\cite{Zellers-acl-2019-HellaSwag}, WinoGrande~\cite{Sakaguchi-aaai-2020-WinoGrande}, 
                                                COPA~\cite{Roemmele-aaai-2011-Choice}, \\
                                             &  &   &
                                                OpenBookQA~\cite{Mihaylov-EMNLP-2018-Can},
                                                 ScienceQA~\cite{Saikh-IJDL-2022-ScienceQA}, proScript~\cite{Sakaguchi-acl-2021-proScript}, ProPara~\cite{Dalvi-acl-2018-Tracking},  \\ 
                                              &  &   &
                                                ExplaGraphs~\cite{Saha-acl-2021-ExplaGraphs},
                                                ProofWriter~\cite{Tafjord-acl-2021-ProofWriter},  EntailmentBank~\cite{Dalvi-acl-2021-Explaining},  \\ 
                                            &  &   &
                                                ProOntoQA~\cite{Saparov-arxiv-2022-Language}  \\ 
                                                \addlinespace
                                             &   &\multirow{3}{*}{Symbolic Reasoning}            & 
                                                CoinFlip~\cite{Wei-arxiv-2022-chain}, ReverseList~\cite{Wei-arxiv-2022-chain},  LastLetter~\cite{Wei-arxiv-2022-chain}, Boolean Assignment~\cite{Anil-arxiv-2022-Exploring},
                                                \\
                                             &   &    &
                                                Parity~\cite{Anil-arxiv-2022-Exploring}, Colored Object~\cite{Srivastava-arxiv-2022-Beyond}, Penguins in a Table~\cite{Srivastava-arxiv-2022-Beyond},   \\
                                             &   &   &
                                                Repeat Copy~\cite{Gao-arxiv-2022-PAL}, Object Counting~\cite{Gao-arxiv-2022-PAL}  \\
                                                \addlinespace
                                             &   &\multirow{3}{*}{Mathematical Reasoning}        &  MATH~\cite{Hendrycks-ICLR-2021-Measuring}, GSM8k~\cite{Cobbe-arxiv-2021-Training}, SVAMP~\cite{Patel-NAACL-2021-Are}, MultiArith~\cite{Roy-acl-2015-Solving}, ASDiv~\cite{Miao-ACL-2020-A}, \\
                                             &   &  &
                                                MathQA~\cite{Amini-acl-2019-MathQA}, 
                                                AQUA-RAT~\cite{Ling-acl-2017-Program}, MAWPS~\cite{Koncel-NAACL-2016-MAWPS}, DROP~\cite{Dua-NAACL-2019-DROP},  \\
                                             &   &  & 
                                                NaturalProofs~\cite{Welleck-NIPS-2021-NaturalProofs},
                                                PISA~\cite{Jiang-AITP-2021-LISA},
                                                miniF2F~\cite{Zheng-ICLR-2022-miniF2F}, ProofNet~\cite{Azerbayev-arxiv-2023-ProofNet}  \\
                                                \midrule
    \multirow{18}{*}{Advanced} & \multirow{5}{*}{Human Alignment}    & Honestness             &TruthfulQA~\cite{Lin-ACL-2022-TruthfulQA}, HaluEval~\cite{Li-arxiv-2023-HaluEval} \\
        \addlinespace
        & & Helpfulness             & HH-RLHF~\cite{Bai-arxiv-2022-Training} \\
        \addlinespace
        & & \multirow{2}{*}{Harmlessness}             & HH-RLHF~\cite{Bai-arxiv-2022-Training}, Crows-Pairs~\cite{Nangia-EMNLP-2020-CrowS} \\
        & & & WinoGender~\cite{Rudinger-NAACL-2018-Gender}, RealToxicityPrompts~\cite{Gehman-2023-arxiv-RealToxicityPrompts} \\
        \cmidrule(r){2-4}
        & \multirow{4}{*}{\makecell[c]{Interaction with \\ External Environment}}    &       \multirow{1}{*}  Household       & VirtualHome~\cite{Puig-CVPR-2018-VirtualHome}, BEHAVIOR~\cite{Srivastava-CoRL-2021-BEHAVIOR}, ALFRED~\cite{Shridhar-CVPR-2020-ALFRED},ALFWorld~\cite{Shridhar-2021-iclr-ALFWorld} \\
        \addlinespace
        & &       \multirow{1}{*}  Website Environment   & WebShop~\cite{Yao-2022-nips-WebShop}, Mind2Web~\cite{Deng-2023-arxiv-Mind2Web}  \\
        \addlinespace
        & &       \multirow{1}{*}  Open World       &MineRL~\cite{Guss-2019-ijcai-MineRL}, MineDojo~\cite{Fan-2022-nips-minedojo} \\
        \cmidrule(r){2-4}
        &  \multirow{8}{*}{Tool Manipulation}    & \multirow{1}{*} Search Engine             & HotpotQA~\cite{yang-2018-acl-HotpotQA}, TriviaQA~\cite{Joshi-ACL-2017-TriviaQA}, Natural Questions~\cite{Kwiatkowski-ACL-2019-Natural} \\
        \addlinespace
        & & \multirow{1}{*} Code Executor             & GSM8k~\cite{Cobbe-arxiv-2021-Training}, TabMWP~\cite{Lu-2022-arxiv-Dynamic}, Date Understanding~\cite{Srivastava-arxiv-2022-Beyond} \\
        \addlinespace
        & & \multirow{1}{*} Calculator             & GSM8k~\cite{Cobbe-arxiv-2021-Training}, MATH~\cite{Hendrycks-ICLR-2021-Measuring}, CARP~\cite{Zhang-2023-arxiv-Evaluating} \\
        \addlinespace
        & & \multirow{1}{*} Model Interface             & GPT4Tools~\cite{yang-2023-arxiv-GPT4Tools}, Gorilla~\cite{Patil-2023-arxiv-Gorilla} \\
        \addlinespace
        & & \multirow{2}{*}{Data Interface}             & WebQSP~\cite{Yih-2016-acl-The}, MetaQA~\cite{Puerto-2023-eacl-MetaQA}, WTQ~\cite{Pasupat-2015-acl-Compositional} \\
        & & & WikiSQL~\cite{Zhang-2017-arxiv-Seq2SQL}, TabFact~\cite{Chen-2020-iclr-TabFact}, Spider~\cite{Yu-2018-emnlp-Spider} \\
        \addlinespace
        \bottomrule
    \end{tabular}
    \label{tab:dataset}
\end{table*}

\subsection{Basic Ability}\label{sec:basicability}
In this part, we mainly focus on three basic types of ability evaluation for LLMs, \ie language generation, knowledge utilization, and complex reasoning. 
It is noted that we do not intend to have complete coverage of all the related tasks, but instead only focus on the most widely discussed or studied tasks for LLMs.  Next, we introduce these tasks in detail.

\subsubsection{Language Generation}
\label{sec-langauge-generation}
According to the task definition, existing tasks about language generation can be roughly categorized into language modeling, conditional text generation, and code synthesis tasks. Note that code synthesis is not a typical NLP task, we include it for discussion because it can be directly solved by a number of LLMs (trained on code data) in a similar generation approach as natural language text.   

\paratitle{Language Modeling.}
As the most fundamental ability of LLMs, \emph{language modeling} aims  to predict the next token based on the previous tokens~\cite{Bengio-JMLR-2003-A}, which mainly focuses on the capacity of basic language understanding and generation.  For evaluating such an ability, typical language modeling datasets that existing work uses include Penn Treebank~\cite{Marcus-CL-1993-Building}, WikiText-103~\cite{Merity-ICLR-2017-Pointer}, and the Pile~\cite{Gao-arxiv-2021-Pile}, where the metric of \emph{perplexity} is commonly used for evaluating the model performance under the zero-shot setting. 
Empirical studies~\cite{Brown-NeurIPS-2020-Language,Zeng-arxiv-2022-GLM} show that LLMs bring substantial performance gains over the previous state-of-the-art methods on these evaluation datasets. 
To better test the modeling capacity of long-range dependencies in text, the LAMBADA dataset~\cite{Paperno-ACL-2016-LAMBADA} has been introduced, where LLMs are required to predict the last word of sentences based on a paragraph of context.
Then, the accuracy and perplexity of the predicted last words are employed  to evaluate LLMs. 
As shown in existing work, the performance on the language modeling tasks typically follows the scaling law~\cite{Kaplan-arxiv-2020-Scaling}, which means that scaling language models would improve the accuracy and reduce the perplexity.

\paratitle{Conditional Text Generation.}
As an important topic in language generation, conditional text generation~\cite{Li-IJCAI-2021-Pretrained} focuses on generating texts satisfying specific task demands based on the given conditions, typically including machine translation~\cite{Bahdanau-ICLR-2015-Neural}, text summarization~\cite{Nallapati-acl-2016-Abstractive}, and question answering~\cite{Berant-EMNLP-2013-Semantic}.
{To measure the quality of the generated text, automatic metrics (\eg Accuracy, BLEU~\cite{Papineni-acl-2002-bleu} and ROUGE~\cite{lin-acl-2004-rouge}) and human ratings have been typically used for evaluating the performance. } 
Due to the powerful language generation capabilities, LLMs have achieved remarkable performance on existing datasets and benchmarks.
{For instance, GPT-4 exhibits comparable performance as commercial translation products, {even for the translation task of languages that are with significant linguistic distance~\cite{Jiao-arxiv-2023-mt}.}
On news summarization tasks~(\ie CNN/DM and XSUM), LLMs also demonstrate comparable performance with human freelance writers~\cite{Zhang-2023-arxiv-Benchmarking}.
Despite the rapid progress on model capacity, there are increasing concerns on the feasibility of existing automatic metrics to faithfully assess the performance of LLMs in conditional text generation tasks~\cite{Zhang-2023-arxiv-Benchmarking,Goyal-2023-arxiv-News,Gehrmann-2022-arxiv-Repairing}. 
As the alternatives to automatic metrics, recent studies also propose to incorporate LLMs as generation evaluators to examine the quality of the generated content~\cite{Wang-2023-arxiv-Is,Liu-2023-arxiv-G-Eval,vicuna2023}. 
Moreover, researchers also explore more challenging language generation tasks for LLMs, such as  structured data generation~\cite{Jiang-2023-arxiv-StructGPT} and long text generation~\cite{Yang-EMNLP-2022-Re3,OpenAI-OpenAI-2023-GPT-4,Zhou-2023-arxiv-RecurrentGPT}.}

\paratitle{Code Synthesis.}
In addition to generating high-quality natural language text, existing LLMs also show strong abilities to generate {formal language, especially  computer programs (\ie code)} that satisfy specific conditions, called \emph{code synthesis}~\cite{Gulwani-Found-2017-Program}. 
Unlike natural language generation, as the generated code can be directly checked by execution with corresponding compilers or interpreters, existing work mostly evaluates the quality of the generated code from LLMs by calculating the pass rate against the test cases, \ie pass@$k$\footnote{Given $k$ programs generated by the LLM, pass@$k$ is computed as 1 when at least one program passes all test cases, or else 0}. %
Recently, several code benchmarks focusing on functional correctness are proposed to assess the code synthesis abilities of LLMs, such as APPS~\cite{Hendrycks-nips-2021-Measuring}, HumanEval~\cite{Chen-arxiv-2021-evaluating}, and MBPP~\cite{Austin-arxiv-2021-Program}. 
Typically, they consist of diverse programming problems, with text specification and test cases for correctness checking. 
To improve such an ability, it is key to fine-tuning (or pre-training) LLMs on code data, which can effectively adapt LLMs to code synthesis tasks~\cite{nijkamp-arxiv-2022-Codegen}.
{In addition, existing work  has proposed new strategies to generate code, \eg sampling multiple candidate solutions~\cite{Austin-arxiv-2021-Program} and planning-guided decoding~\cite{Zhang-ICLR-2023-Planning}, which can be considered as the imitation of bug-fixing and code-planning processes by  programmers.}
Impressively,  LLMs have recently shown competitive performance with humans by achieving a ranking of the top 28\% among users on the programming contest platform Codeforces~\cite{Li-Science-2022-AlphaCode}. 
Further, GitHub Copilot has been released to assist programming in coding IDEs (\eg Visual Studio and JetBrains IDEs), which can support  a variety of  languages including Python, JavaScript, and Java. 
A viewpoint article entitled ``\emph{The End of Programming}''~\cite{Welsh-ACM-2023-The} in Communications of the ACM  has discussed the impact of AI programming in the field of computer science, emphasizing an important shift towards the highly adaptive LLM as a new atomic  unit of computation. 

\paratitle{Major Issues.} Although LLMs have achieved splendid performance in generating human-like text, they are susceptible to suffering from two major issues in language generation as discussed below.

$\bullet$ \emph{Unreliable generation evaluation.}
\label{sec:generation-eval}
With the advancement of  language generation ability of LLMs, existing studies find that 
the generated texts from LLMs have reached a comparable quality to the reference texts on a variety of text  generation tasks. However, due to the intrinsic weakness of existing evaluation benchmarks, there exists pronounced inconsistency between human evaluation and automatic reference-based metrics~\cite{Zhang-2023-arxiv-Benchmarking,Goyal-2023-arxiv-News,Gehrmann-2022-arxiv-Repairing,Bang-arxiv-2023-A}. For example, 
in OpenDialKG~\cite{Moon-ACL-2019-OpenDialKG}, ChatGPT underperforms a fine-tuned GPT-2 on BLEU and ROUGE-L metrics, while earning more favor from human judgment~\cite{Bang-arxiv-2023-A}. Furthermore, existing work argues that even human evaluation may  not be robust enough~\cite{Goyal-2023-arxiv-News,Zhang-2023-arxiv-Benchmarking,Liu-arxiv-2022-Revisiting,Fabri-2021-tacl-SummEval}.
In some cases, it is difficult to achieve a high level of consensus among human annotators~\cite{Goyal-2023-arxiv-News}, and there is also a large gap between the annotation quality of crowdworkers and experts~\cite{Fabri-2021-tacl-SummEval,Liu-arxiv-2022-Revisiting}.
Thus, how to conduct reliable evaluation for language generation tasks in the era of LLMs has become a fundamental yet challenging research topic. 
Recently, increasing research work proposes to leverage LLMs to improve the evaluation quality of the generated texts.  
Specially, LLMs can be   used to improve the evaluation quality of existing metrics.
For example, Para-Ref~\cite{Tang-2023-arxiv-Not} augments various automatic metrics by leveraging LLMs to paraphrase existing references into semantically equivalent references with diverse expressions. 
Further, LLMs are widely employed as the evaluators of text generation in a reference-free manner, including evaluating a single prediction~\cite{Wang-2023-arxiv-Is,Liu-2023-arxiv-G-Eval,Wang-2023-arxiv-rethinking} or comparing several candidates~\cite{Gao-arxiv-2023-Human,Ji-2023-arxiv-Exploring,vicuna2023,Bai-2023-arxiv-Benchmarking}. 
Nevertheless, LLMs may expose bias (\eg order bias or preference for LLM-generated texts over human-written texts) as language generation evaluators, demonstrating disparities when compared to human evaluation~\cite{Liu-2023-arxiv-Evaluate,Liu-2023-arxiv-G-Eval,Wang-2023-arxiv-Large}. 

\begin{center}
\begin{tcolorbox}[colback=blue!5!white,colframe=blue!55!black,width=0.46\textwidth,title={Unreliable Generation Evaluation}]
LLMs have been capable of generating texts with a comparable quality to human-written texts, which however might be underestimated by automatic reference-based metrics. %
As an alternative evaluation approach, 
LLMs can serve as language generation evaluators to evaluate a single text, compare multiple candidates, and improve existing metrics. However, this evaluation approach still needs more inspections and examinations in real-world tasks.  
\end{tcolorbox}
\end{center}

$\bullet$ {\emph{Underperforming specialized generation}}. Although LLMs have learned general language patterns to generate coherent text, 
their proficiency in generation might be constrained when dealing  with a specialized domain or task. %
{For instance, a language model that has been trained on general web articles may face challenges when generating a medical report  which involves many medical jargon and methods.} 
Intuitively, domain knowledge should be critical for model specialization. However, it is not easy to inject such specialized knowledge into LLMs.   
As discussed in recent analyses~\cite{FU-blog-2022-how,Ye-arxiv-2023-A}, when LLMs are trained to exhibit some specific ability that allows them to excel in some areas, they might struggle in others. Such an issue is related to \emph{catastrophic forgetting}~\cite{Michael-Psychology-1989-Catastrophic, Kemker-AAAI-2018-Measuring} in training neural networks, which refers to the conflict phenomenon of integrating new and old knowledge.  
Similar cases also occur in human alignment of LLMs, where ``\emph{alignment tax}''~\cite{Ouyang-arxiv-2022-Training} (\eg a potential loss in the in-context learning  ability) has to be paid for aligning to human values and needs.  
{
Moreover, due to the limitations of sequence modeling architecture, LLMs still face challenges in the understanding and generation of structured data. Consequently, they often fall behind task-specific models on complex structured data tasks, such as knowledge-base question answering and semantic parsing~\cite{Xie-EMNLP-2022-UnifiedSKG,Jiang-2023-arxiv-StructGPT}.
}
Therefore, it is important to develop effective model specialization methods  that can flexibly adapt LLMs to various task scenarios,  meanwhile retaining the original abilities as possible. 

\begin{center}
\begin{tcolorbox}[colback=blue!5!white,colframe=blue!55!black,width=0.46\textwidth,title={Underperforming Specialized Generation}]
LLMs may fall short in mastering generation tasks that require domain-specific knowledge or generating structured data. It is non-trivial to inject specialized knowledge into LLMs, meanwhile maintaining the original abilities of LLMs.
\end{tcolorbox}
\end{center}

\begin{figure*}[h]
    \centering
        \includegraphics[width=1\textwidth]{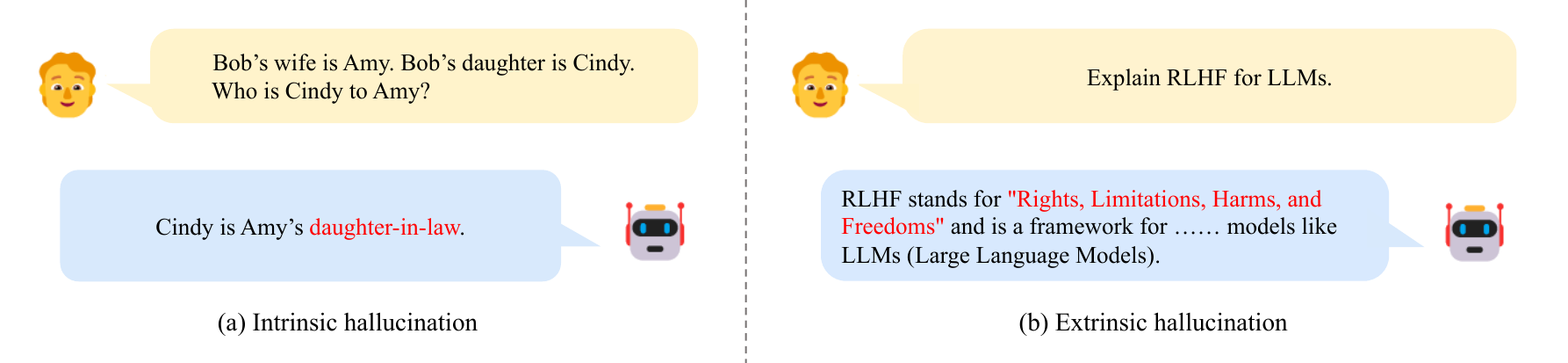}
    \caption{Examples of intrinsic and extrinsic hallucination for a public LLM (access date: March 19, 2023). As an example of intrinsic hallucination, the LLM gives a conflicting judgment about the relationship between Cindy and Amy, which  contradicts the input. 
{For extrinsic hallucination, in this example, the LLM seems to have an incorrect understanding of the meaning of RLHF (reinforcement learning from human feedback), though it can correctly understand the meaning of LLMs (in this context). } }
    \label{fig:hallucination}
\end{figure*}

\subsubsection{Knowledge Utilization}

Knowledge utilization is an important ability of intelligent systems to accomplish knowledge-intensive tasks (\eg commonsense question answering and fact completion) based on supporting factual evidence. 
Concretely, it requires LLMs to {properly utilize the rich factual knowledge from the pre-training corpus or retrieve external data when necessary.} 
In particular, question answering~(QA) and knowledge completion have been two commonly used tasks for evaluating this ability. 
According to the test tasks (question answering or knowledge completion) and evaluation settings (\emph{with} or \emph{without} external resources), we categorize existing knowledge utilization tasks into three types, namely closed-book QA, open-book QA\footnote{In this part, open-book QA refers to the QA tasks that require to extract and utilize useful information from external knowledge resources, as the antithesis of closed-book QA (only using the encoded information from  pre-training corpus). Note that there is a dataset  also named OpenBookQA~\cite{Mihaylov-EMNLP-2018-Can}, which  follows the settings of open-book QA tasks by extracting and utilizing  external science facts.}, and knowledge completion.

\paratitle{Closed-Book QA.}
Closed-book QA tasks~\cite{Roberts-EMNLP-2020-How} test the acquired factual knowledge of LLMs from the pre-training corpus, where LLMs should answer the question only based on the given context without using external resources. 
For evaluating this ability, there are  several datasets that can be leveraged, including  Natural Questions~\cite{Kwiatkowski-ACL-2019-Natural}, Web Questions~\cite{Berant-EMNLP-2013-Semantic}, and TriviaQA~\cite{Joshi-ACL-2017-TriviaQA}, %
{where the accuracy metric is widely adopted.}
Empirical results have revealed that  %
LLMs can perform well in this setting and even match the performance of state-of-the-art {open-domain} QA systems~\cite{Chowdhery-arxiv-2022-PaLM}.
Also, the performance of LLMs on closed-book QA tasks shows a scaling law pattern in terms of both model size and data size: 
scaling the parameters and training tokens can increase the capacity of LLMs and help them learn (or memorize) more knowledge from the pre-training data~\cite{Chowdhery-arxiv-2022-PaLM}. 
Further, under a similar parameter scale, LLMs with more pre-training data relevant to the evaluated tasks would achieve better performance~\cite{Nakano-arxiv-2021-WebGPT}. 
Also, the closed-book QA setting  provides a testbed for probing the accuracy of the factual knowledge encoded by LLMs. 
{However, as shown in existing work~\cite{Brown-NeurIPS-2020-Language}, LLMs might perform less well on QA tasks relying on fine-grained knowledge, even when it exists in the pre-training data.}

\paratitle{Open-Book QA.} %
Unlike closed-book QA, in open-book QA tasks, LLMs can extract useful evidence from the external knowledge base or document collections,  and then answer the question based on the extracted evidence~\cite{Izacard-arxiv-2022-Few, Guu-ICML-2020-Retrieval, Lewis-NeurIPS-2020-Retrieval,Lan-2021-arxiv-Complex}. 
{Typical open-book QA datasets (\eg Natural Questions~\cite{Kwiatkowski-ACL-2019-Natural}, OpenBookQA~\cite{Mihaylov-EMNLP-2018-Can}, and SQuAD~\cite{Rajpurkar-EMNLP-2016-SQuAD}) have overlap with closed-book QA datasets, but they incorporate external data sources, \eg Wikipedia.
The metrics of accuracy and F1 score  are widely used in open-book QA tasks for evaluation.}
To select relevant knowledge from external resources, LLMs are often paired with a text retriever (or even a  search engine), which is trained independently or jointly with LLMs~\cite{Izacard-arxiv-2022-Few,Borgeaud-icml-2022-Improving,Nakano-arxiv-2021-WebGPT}. 
{
Also, previous work~\cite{Xu-arxiv-2023-Search,Peng-arxiv-2023-Check,Jiang-2023-arxiv-Active} has indicated that retrievers can  assist LLMs in verifying and rectifying the reasoning path.
}
In evaluation, existing studies mainly focus on testing how LLMs utilize the extracted knowledge to answer the question and show that the retrieved evidence can largely improve the accuracy of the generated answers, even enabling a smaller LLM to outperform $10\times$ larger ones~\cite{Izacard-arxiv-2022-Few,Borgeaud-icml-2022-Improving}.
Further, open-book QA tasks can be also employed to evaluate the recency of  knowledge information. 
Pre-training or retrieving from outdated knowledge resources may cause LLMs to generate incorrect answers for time-sensitive  questions~\cite{Izacard-arxiv-2022-Few}.

\paratitle{Knowledge Completion.}
In knowledge completion tasks, LLMs might be (to some extent) considered as a knowledge base~\cite{Petroni-EMNLP-2019-Language}, which can be leveraged to complete or predict the missing parts of knowledge units (\eg knowledge triples).
Such tasks can probe and evaluate \emph{how much} and \emph{what kind of} knowledge LLMs have learned from the pre-training data.
Existing knowledge completion tasks can be roughly divided into knowledge graph completion tasks (\eg FB15k-237~\cite{Toutanova-CVSC-2015-Observed} and  WN18RR~\cite{Dettmers-AAAI-2018-Convolutional}) and fact completion tasks (\eg WikiFact~\cite{Goodrich-KDD-2019-Assessing}), which aim to complete the triples from a knowledge graph and incomplete sentences about specific facts, respectively. 
Empirical studies have revealed that it is difficult for existing LLMs to accomplish knowledge completion tasks {related to specific relation types~\cite{Liang-arxiv-2022-Holistic}.}
As shown in the evaluation results on WikiFact, LLMs perform well on several frequent relations that occur in the pre-training data (\eg \texttt{currency} and \texttt{author}), while not well on rare ones (\eg \texttt{discoverer\_or\_inventor} and \texttt{place\_of\_birth}). 
Interestingly, %
{under the same evaluation settings (\eg in-context learning), InstructGPT (\ie \texttt{text-davinci-002)}} outperforms GPT-3 in all subsets of WikiFact.

\paratitle{Major Issues}. Although LLMs have achieved key progress in capturing and utilizing knowledge information, they suffer from two major issues as discussed below.

$\bullet$ \emph{Hallucination}.  In generating factual texts, a challenging issue is \emph{hallucination generations}~\cite{Bang-arxiv-2023-A, Huang-arxiv-2023-A}, where the generated information is either in conflict with the existing source (\emph{intrinsic hallucination}) or cannot be verified by the available source (\emph{extrinsic hallucination}), which are illustrated by two examples in Figure~\ref{fig:hallucination}.
Hallucination widely occurs in existing LLMs, even the most superior LLMs such as GPT-4~\cite{OpenAI-OpenAI-2023-GPT-4}. 
{
Furthermore, existing work shows that LLMs encounter difficulties in recognizing the hallucinated content in text~\cite{Li-arxiv-2023-HaluEval}, even the  powerful ChatGPT. 
Additionally, beyond language tasks, a recent study has shown that large vision-language models (LVLM) also face  challenges with hallucination, \ie  generating objects that are not present in the accompanying images~\cite{Li-arxiv-2023-Evaluating}.
}
In essence, LLMs seem to  ``unconsciously''  utilize the knowledge in task solving, which still lack an ability to accurately control the use of internal or external knowledge.   
Hallucinations would mislead LLMs to generate undesired outputs and mostly degrade the performance, leading to potential risks when deploying LLMs in real-world applications.
To alleviate this problem,  alignment tuning strategies (as discussed in Section~\ref{sec-alignment}) have been widely utilized in existing work~\cite{Ouyang-arxiv-2022-Training}, which rely on tuning LLMs on high-quality data or using human feedback. 
{
Moreover, the integration of external tools for the provision of credible information sources can help alleviate the hallucination issue~\cite{Li-arxiv-2023-HaluEval,Peng-arxiv-2023-Check,Nakano-arxiv-2021-WebGPT}.
Another line of research work leverages uncertainty estimation of LLMs to identify hallucinations~\cite{Kadavath-arxiv-2023-Language,Manakul-arxiv-2023-SelfCheckGPT}.
For instance, considering that hallucinated facts are prone to exhibit inconsistency across different sampled outputs, SelfCheckGPT~\cite{Manakul-arxiv-2023-SelfCheckGPT} detects hallucination by measuring information inconsistency within sampled outputs. 
} 
For the evaluation of the hallucination problem, a set of hallucination detection tasks have been proposed, \eg TruthfulQA~\cite{Lin-ACL-2022-TruthfulQA} for detecting human falsehood mimicked by models. More recently, {HaluEval~\cite{Li-arxiv-2023-HaluEval} creates a large-scale LLM-generated and human-annotated
hallucinated samples to evaluate the ability of language models to recognize hallucination in both task-specific and general scenarios.} 
\begin{center}
\begin{tcolorbox}[colback=blue!5!white,colframe=blue!55!black,width=0.46\textwidth,title={Hallucination}]
LLMs are prone to generate untruthful information that either conflicts with the existing source or cannot be verified by the available source. Even the most powerful LLMs such as ChatGPT face great challenges in migrating the hallucinations of the generated texts.  
This issue can be partially alleviated by special approaches such as alignment tuning and  tool utilization. 
\end{tcolorbox}
\end{center}

$\bullet$ \emph{Knowledge recency}. %
{As another major challenge, LLMs would encounter difficulties when solving tasks that require the latest knowledge beyond the training data.  %
To tackle this issue, a straightforward approach is to regularly update LLMs with new data.  
However, it is very costly to fine-tune LLMs, and also likely to cause the catastrophic forgetting issue when incrementally training LLMs. 
Therefore, it is necessary to develop efficient and effective approaches that can integrate new knowledge into existing LLMs, making them up-to-date.
Existing studies have explored how to utilize the external knowledge source (\eg search engine) to complement LLMs, which can be either jointly optimized with LLMs~\cite{Izacard-arxiv-2022-Few} or used as a plug-and-play module~\cite{Peng-arxiv-2023-Check}. For instance, ChatGPT utilizes a retrieval plugin to access up-to-date information sources~\cite{OpenAI-blog-2023-plugins}.
By incorporating the extracted relevant information into the context~\cite{Lazaridou-arxiv-2022-Internet,Qian-2023-arxiv-WebBrain,Liu-2023-arxiv-RETA-LLM}, LLMs can acquire new factual knowledge and perform better on relevant tasks.
However, such an approach seems to be still at a superficial level. In addition, 
{existing studies also explore editing parameters of language models to update intrinsic knowledge~\cite{Dai-ACL-2022-Knowledge,Meng-NIPS-2022-Locating,Geva-2021-emnlp-Transformer}.
Nevertheless, previous work~\cite{Yao-arxiv-2023-Editing} has shown that several parameter editing methods   perform not well on LLMs, though they can improve the performance of small language models.
Therefore, it is still difficult to directly amend intrinsic knowledge or inject specific knowledge into LLMs, which remains {an open research problem~\cite{Yao-arxiv-2023-Editing}}.
Recently, a useful framework \emph{EasyEdit}~\cite{wang-CoRR-2023-EasyEdit} has been released to facilitate the research of knowledge editing for LLMs. 
}%
}
%

\begin{center}
\begin{tcolorbox}[colback=blue!5!white,colframe=blue!55!black,width=0.46\textwidth,title={Knowledge Recency}]
The parametric knowledge of LLMs is hard to be updated in a timely manner.
Augmenting LLMs with external knowledge sources is a practical approach to tackling the issue.
However, how to effectively update knowledge within LLMs remains an open research problem.
\end{tcolorbox}
\end{center}

\subsubsection{Complex Reasoning} 

Complex reasoning refers to the ability of understanding and utilizing supporting evidence or logic to derive conclusions or make decisions~\cite{Huang-arxiv-2022-Towards,Qiao-arxiv-2022-Reasoning}. 
According to the type of involved  logic and evidence in the reasoning process, 
we consider dividing existing evaluation tasks into three major categories, namely 
knowledge reasoning, symbolic reasoning, and mathematical reasoning.

\paratitle{Knowledge Reasoning.}
The knowledge reasoning tasks rely on logical relations and evidence about factual knowledge to answer the given question.
Existing work mainly uses specific datasets to evaluate the reasoning capacity of the corresponding type of knowledge, \eg CSQA~\cite{Talmor-naacl-2019-CommonsenseQA}/StrategyQA~\cite{Geva-tacl-2021-Did} for commonsense knowledge reasoning and ScienceQA~\cite{Saikh-IJDL-2022-ScienceQA} for science knowledge reasoning. 
{In addition to the accuracy of the predicted results, existing work~\cite{Saikh-IJDL-2022-ScienceQA} has  also evaluated the quality of the generated reasoning process, via automatic metrics (\eg BLEU) or human evaluation.}
Typically, these tasks require LLMs to perform step-by-step reasoning based on factual knowledge, until reaching the answer to the given question. 
To elicit the  %
{step-by-step reasoning} ability, chain-of-thought~(CoT) prompting strategy~\cite{Wei-arxiv-2022-chain} has been proposed for enhancing the complex reasoning capacity of LLMs. 
As discussed in Section~\ref{subsec-cot}, CoT involves the intermediate reasoning steps, which can be manually created~\cite{Wei-arxiv-2022-chain} or automatically generated~\cite{Shao-arxiv-2023-Synthetic}, into the prompts to guide LLMs to perform multi-step reasoning.
Such a way largely improves the reasoning performance of LLMs, leading to new state-of-the-art results on several complex knowledge reasoning tasks~\cite{Wei-arxiv-2022-chain,Chowdhery-arxiv-2022-PaLM,Ning-arxiv-2023-ChatGPT}. 
Further, after reformulating knowledge reasoning tasks into code generation tasks, researchers have found that the performance of LLMs can be further improved~\cite{Madaan-emnlp-2022-Language}, especially with the LLMs pre-trained on code. 
{However, due to the complexity of knowledge reasoning tasks, the  performance  of current LLMs still lags behind human results on tasks such as commonsense reasoning~\cite{Wei-arxiv-2022-chain,Chowdhery-arxiv-2022-PaLM,Sifatkaur-arxiv-2023-Mind}.}
As a common type of   mistakes, LLMs might generate inaccurate  {intermediate steps},  leading to a wrong final result.
To address this issue, existing work has proposed special decoding or ensemble strategies to improve the accuracy of the whole reasoning chain~\cite{Wang-arxiv-2022-Self-Consistency,Li-arxiv-2022-On}. 
\ignore{
More recently, an empirical study~\cite{Ning-arxiv-2023-ChatGPT} reveals that LLMs may have difficulty in explicitly inferring the commonsense knowledge required by a specific task, though they can successfully solve it.  Further, it shows  that leveraging self-generated knowledge may not be  beneficial for improving the  reasoning performance.  
}

\paratitle{Symbolic Reasoning\footnote{{Following~\cite{Wei-arxiv-2022-chain}, we mainly discuss symbolic reasoning tasks specially designed for evaluating LLMs. We do not consider symbolic reasoning methods in traditional NLP tasks, such as deducing logical rules from the knowledge graphs in KBQA.}}.}  
{
The symbolic reasoning tasks mainly focus on manipulating the  symbols in a formal rule setting to fulfill some specific goal~\cite{Huang-arxiv-2022-Towards}, where the operations and rules may have never been seen by LLMs during pre-training. }
{
Existing work~\cite{Wei-arxiv-2022-chain,Kojima-arxiv-2022-Large,Zhou-arxiv-2022-Least} commonly evaluates LLMs on the task of last letter concatenation and coin flip, where the evaluation examples require the same reasoning steps as the in-context examples (called \emph{in-domain test}) or more steps (called \emph{out-of-domain test}).
For an example of the out-of-domain test, LLMs could only see the examples with two  words in context, but it requires LLMs to concatenate the last letters of three or more words.
Typically, the accuracy of the generated symbols is adopted to evaluate the performance of LLMs on these tasks.} 
Thus, LLMs need to understand the semantic relations among the symbolic operations and  %
{their composition in complex scenarios. 
However, under the out-of-domain setting, as LLMs have not seen the complex compositions of symbolic operations and rules (\eg twice the number of operations in context examples), it is hard for LLMs to capture their accurate meanings.} 
To solve this issue, existing studies incorporate scratchpad~\cite{Anil-arxiv-2022-Exploring,Nye-arxiv-2021-Show} and tutor~\cite{Qian-arxiv-2022-Limitations} strategies to help LLMs better manipulate symbolic operations, for generating longer and more complex reasoning processes.
Another line of research work utilizes the formal programming language to represent the symbolic operations and rules, which requires LLMs to generate code and perform the reasoning process by executing it with external interpreters.
Such a way can decompose the complex reasoning process into code synthesis and program execution for LLMs and interpreters, respectively, leading to a simplified  reasoning process with yet more accurate results~\cite{Gao-arxiv-2022-PAL}.

\paratitle{Mathematical Reasoning.}
The mathematical reasoning tasks need to comprehensively utilize mathematical knowledge, logic, and computation for solving problems or generating proof statements. 
Existing mathematical reasoning tasks can be mainly categorized into math problem solving and automated theorem proving.  
{
For math problem solving tasks, SVAMP~\cite{Patel-NAACL-2021-Are}, GSM8k~\cite{Cobbe-arxiv-2021-Training} and MATH~\cite{Hendrycks-ICLR-2021-Measuring} datasets are commonly used for evaluation, where LLMs need to generate accurate concrete numbers or equations to answer the mathematical problem.
As these tasks also require multi-step reasoning, the CoT prompting strategy has been widely adopted for LLMs to improve the reasoning performance~\cite{Wei-arxiv-2022-chain}.
}
As another practical strategy, continually pre-training LLMs on large-scale mathematical corpora can largely boost their performance on  {mathematical reasoning tasks~\cite{Zhao-KDD-2022-JiuZhang,Taylor-arxiv-2022-Galactica,Lewkowycz-arxiv-2022-Solving}.}
Further, since math problems in different languages share the same mathematical logic, researchers also propose a multilingual math word problem benchmark~\cite{Shi-arxiv-2022-Language} to evaluate the multilingual mathematical reasoning capacity of LLMs.
 As another challenging task,  automated theorem proving (ATP)~\cite{Zheng-ICLR-2022-miniF2F,Welleck-NIPS-2021-NaturalProofs,Wang-CICM-2018-First} requires the reasoning model to strictly follow the reasoning logic and mathematical skills. %
{To evaluate the performance on this task, PISA~\cite{Jiang-AITP-2021-LISA} and miniF2F~\cite{Zheng-ICLR-2022-miniF2F} are two typical ATP datasets with the \emph{proof success rate} as the evaluation metric.}
As a typical approach, existing work on ATP utilizes LLMs to aid the search for proofs using an interactive theorem prover (ITP), such as Lean, Metamath, and Isabelle~\cite{Polu-arxiv-2020-Generative,Jiang-arxiv-2022-Thor,Polu-arxiv-2022-Formal}.  
{A major limitation of ATP research is the lack of related corpora in formal language. 
To tackle it, several studies utilize LLMs to convert informal statements into formal proofs for augmenting new data~\cite{Wu-arxiv-2022-Autoformalization} or generate drafts and proof sketches to} %
{reduce the search space of the proofs}~\cite{Jiang-arxiv-2022-Draft}.

\paratitle{Major Issues.}
In spite of the advancements, LLMs still have several limitations in solving complex reasoning tasks.

$\bullet$ \emph{Reasoning inconsistency}.
With improved reasoning strategies (\eg CoT prompting), LLMs can solve some complex reasoning tasks, by performing   
step-by-step reasoning based on the supporting logic and evidence.
Despite the effectiveness, 
the  \emph{reasoning inconsistency} issue often occurs in the decomposed reasoning process. 
Concretely, LLMs may generate the correct answer following an invalid reasoning path, or produce a wrong answer after a correct reasoning process~\cite{Wei-arxiv-2022-chain,Lyu-arxiv-2023-Faithful}, leading to inconsistency between the derived answer and the reasoning process.
{
To alleviate this problem, existing work has proposed to guide the whole generation process of LLMs via external tools or models~\cite{Zhang-ICLR-2023-Planning,Li-arxiv-2022-On,Yao-arxiv-2023-Tree}, to re-check the reasoning process and final answer for correcting the potential errors~\cite{Madaan-arxiv-2023-Refine,Shinn-arxiv-2023-Reflexion,Gou-arxiv-2023-Critic} or fine-tune LLMs with process-based feedback~\cite{Uesate-2023-arxiv-Solving,Lightman-2023-arxiv-Let}. 
For instance, \emph{Tree of Thoughts~(ToT)}~\cite{Yao-arxiv-2023-Tree} empowers LLMs to engage in the decision-making process by concurrently  {exploring and self-evaluating various reasoning paths}.
To refine the {reasoning processes}, Self-Refine~\cite{Madaan-arxiv-2023-Refine} elicits feedback from LLMs on self-generated solutions, {enabling} the iterative refinement of solutions based on the feedback.
Moreover, several studies improve the consistency in the reasoning chain of LLMs through the integration of process-based supervision during training~\cite{Uesate-2023-arxiv-Solving,Lightman-2023-arxiv-Let}.
}
{
As a promising solution, 
recent approaches reformulate the complex reasoning tasks into code generation tasks, where the strict execution of the generated code ensures the consistency between the reasoning process and the outcome.
}
Also, it has been revealed that there might exist   inconsistency between tasks with  similar inputs, where small changes in the task description may cause the model to produce different results~\cite{Patel-NAACL-2021-Are,Lu-arxiv-2022-Survey}.  
{
To mitigate this problem, self-consistency~\cite{Wang-arxiv-2022-Self-Consistency} adopts the ensemble of multiple reasoning paths  to enhance the decoding process of LLMs.
}
\begin{center}
\begin{tcolorbox}[colback=blue!5!white,colframe=blue!55!black,width=0.46\textwidth,title={Reasoning Inconsistency}]
LLMs may generate the correct answer following an invalid reasoning path, or produce a wrong answer after a correct reasoning process, leading to inconsistency between the derived answer and the reasoning process.
{The issue can be alleviated by fine-tuning LLMs with process-level  feedback,  using an ensemble of diverse reasoning paths, and refining the reasoning process with self-reflection or external feedback.}
\end{tcolorbox}
\end{center}

$\bullet$ \emph{Numerical  computation}.
{For complex reasoning tasks, LLMs still face difficulties in the involved numerical computation, especially for the symbols that are seldom  encountered during pre-training, such as arithmetic with large numbers~\cite{Qian-arxiv-2022-Limitations,Lu-arxiv-2022-Survey,Yuan-arxiv-2023-Arithmetic}. 
To tackle this issue, a direct way is to tune LLMs on synthesized arithmetic problems~\cite{Pi-EMNLP-2022-Reasoning,liu-arxiv-2023-goat}. {Also, a surge of studies improve the numerical computation performance by tracing intermediate calculation steps in training and inference stages~\cite{Nye-arxiv-2021-Show,Zhou-2023-arxiv-Teaching,liu-arxiv-2023-goat}, \eg scratchpad tracing.} 
In addition, existing work~\cite{Schick-arxiv-2023-Toolformer} has also  incorporated  external tools (\eg calculator),  especially for handling arithmetic operations. 
More recently, ChatGPT has provided a plugin mechanism to use external  tools~\cite{OpenAI-blog-2023-plugins}.  
In this way, LLMs need to learn how to properly manipulate the tools. For this purpose,   researchers have augmented  the examples using tools (even the LLM itself) for tuning the LLM~\cite{Parisi-arxiv-2022-TALM,Schick-arxiv-2023-Toolformer}, or devised  instructions and exemplars for in-context learning~\cite{Gao-arxiv-2022-PAL}.} 
{
In addition to the aid of external tools, recent studies find that tokenizing digits into individual tokens (\eg LLaMA and Galactica tokenizers) is a useful approach to enhancing  the inherent arithmetic ability of LLMs~\cite{Yuan-arxiv-2023-Arithmetic,liu-arxiv-2023-goat}.  
{One possible explanation is that subword tokenization techniques can result in inconsistent sequences when tokenizing numbers. For instance, with a subword tokenizer the integer 7481 may be tokenized as $7\_481$, while 74815 may be tokenized as $748\_15$ (the same numerical substrings with different splits)~\cite{liu-arxiv-2023-goat}.} 
As a comparison, digit-based tokenization for numbers can avoid such an inconsistency, thus likely improving the numerical computation ability of LLMs. 
}

\begin{center}
\begin{tcolorbox}[colback=blue!5!white,colframe=blue!55!black,width=0.46\textwidth,title={Numerical Computation}]
LLMs face difficulties in numerical computation, especially for the symbols that are seldom  encountered during pre-training.
In addition to using mathematical tools, tokenizing digits into individual tokens is also an effective design choice for improving the arithmetic ability of LLMs.
\end{tcolorbox}
\end{center}


\subsection{Advanced Ability}\label{sec:superior}
{In addition to the above basic evaluation tasks, LLMs also exhibit some superior abilities that require special considerations for evaluation.  
In this part, we discuss several representative advanced  abilities and the corresponding evaluation approaches, including human alignment, interaction with the external environment, and tool manipulation. %
Next, we discuss these advanced abilities in detail. 
}%

\subsubsection{Human Alignment}
It is desired that LLMs could  well conform to human values and needs, \ie human alignment, which is a key ability  for the broad use of LLMs in real-world applications. %

To evaluate this ability, existing studies consider multiple criteria for human alignment, such as helpfulness, honesty, and safety~\cite{Askell-arxiv-2021-A,OpenAI-OpenAI-2023-GPT-4,Bai-arxiv-2022-Training}.
For helpfulness and honesty,  adversarial question answering tasks (\eg TruthfulQA~\cite{Lin-ACL-2022-TruthfulQA})  can be utilized to examine LLM's ability in detecting possible falsehood in the text~\cite{Nakano-arxiv-2021-WebGPT,OpenAI-OpenAI-2023-GPT-4}. 
Furthermore, harmlessness can  be also evaluated by several existing benchmarks, \eg CrowS-Pairs~\cite{Nangia-EMNLP-2020-CrowS} and  Winogender~\cite{Rudinger-NAACL-2018-Gender}.
Despite the automatic evaluation with the above datasets, human evaluation is still a more direct way to effectively test the human alignment ability of LLMs.
{OpenAI invites  many experts in domains related to AI risks to evaluate and improve the behaviors of GPT-4 when encountering risky contents~\cite{OpenAI-OpenAI-2023-GPT-4}.
}
In addition, for other aspects of human alignment (\eg truthfulness),  several studies propose to use  specific instructions and devise annotation rules to guide the annotation process~\cite{Nakano-arxiv-2021-WebGPT}. 
Empirical studies have revealed that these strategies can greatly improve the human alignment ability of LLMs~\cite{Bai-arxiv-2022-Training}.  %
{For instance, after alignment tuning on data collected through interactions with experts, the incorrect behavior rate of GPT-4 can be largely reduced  when it deals with sensitive or disallowed prompts. } 
{In addition, high-quality pre-training data can reduce the effort required for alignment~\cite{OpenAI-OpenAI-2023-GPT-4}.} 
For instance, Galactica is potentially more  harmless due to the less biased contents in the scientific corpus~\cite{Taylor-arxiv-2022-Galactica}.

\subsubsection{Interaction with External Environment}
In addition to standard evaluation tasks, LLMs have the ability to receive feedback from the external environment and perform actions  according to the behavior instruction, \eg generating action plans in natural language to manipulate agents~\cite{Huang-ICML-2022-Language,Carta-arxiv-2023-Grounding}.
Such an ability is also emergent in LLMs  that can generate detailed and highly realistic action plans, while smaller models (\eg GPT-2) tend to generate shorter or meaningless plans~\cite{Huang-ICML-2022-Language}. 

To test this ability,  several embodied AI environments and benchmarks can be  used for evaluation, described as follows. VirtualHome~\cite{Puig-CVPR-2018-VirtualHome} builds a 3D simulator for household tasks such as cleaning and cooking, in which the agent can execute natural language actions generated by LLMs. ALFRED~\cite{Shridhar-CVPR-2020-ALFRED} includes more challenging tasks that require LLMs to accomplish compositional targets. BEHAVIOR~\cite{Srivastava-CoRL-2021-BEHAVIOR} focuses on  {everyday chores} in simulation environments and requires LLMs to generate complex solutions, \eg changing the internal status of objects. 
{
Apart from restricted environments such as household tasks, a line of research work investigates the proficiency of LLM-based agents to explore open-world environments, such as Minecraft and the Internet~\cite{Zhu-arxiv-2023-Ghost, Wang-arxiv-2023-Voyager}.
Voyager~\cite{Wang-arxiv-2023-Voyager} introduces an automatic curriculum module that enables LLMs to continuously acquire new skills based on feedback from the environment.  
GITM~\cite{Zhu-arxiv-2023-Ghost} focuses on solving  various challenges in Minecraft based on LLM,  through task decomposition, planning, and invocation of interfaces. 
}
Based on the generated action plans or task completions, existing work either adopts the regular metrics (\eg executability and correctness of the generated action plans)~\cite{Huang-ICML-2022-Language} in the benchmark or directly conducts real-world experiments and measures the success rate~\cite{Ahn-arxiv-2022-Do}, to evaluate such ability.
It has been shown that  LLMs are  capable in interacting with the external environment and generating accurate action plans~\cite{Liang-arxiv-2022-Code}.
Recently, several improvement methods have  been proposed to enhance the interaction ability of LLMs, \eg designing code-like prompts~\cite{Singh-arxiv-2022-ProgPrompt} and providing real-world grounding~\cite{Ahn-arxiv-2022-Do}.  

{
In addition, recent work also explores multi-agent collaboration based on LLMs in simulated environments~\cite{Park-arxiv-2023-Generative,Fu-arxiv-2023-Improving,Metha-arxiv-2023-Improving}.
These studies simulate human social behaviors by instantiating multiple LLM-based agents with observations, planning, and memories in a sandbox environment.
In controlled evaluation, the abilities of generative agents to search, plan, and think  are evaluated by humans in an interview-like manner.
Further, they also conduct descriptive measurements on multiple agents within a simulated environment to examine emergent social behaviors.
}

\subsubsection{Tool Manipulation}

When solving complex problems, LLMs can turn to external tools if they determine it is necessary. %
By encapsulating available tools with API calls, existing work has involved a variety of  external tools, \eg search engine~\cite{Nakano-arxiv-2021-WebGPT}, calculator~\cite{Schick-arxiv-2023-Toolformer}, and compiler~\cite{Gao-arxiv-2022-PAL}, to enhance the performance of LLMs on several  specific tasks. Recently, OpenAI has supported the use of  plugins in ChatGPT~\cite{OpenAI-blog-2023-plugins}, which can equip LLMs with broader capacities beyond language modeling. For example, the web browser plugin enables ChatGPT to access fresh information. Further, incorporating third-party plugins is particularly key for creating a prosperous ecosystem of applications based on LLMs. 

{To examine the ability of tool manipulation, existing work mostly adopts complex reasoning tasks for evaluation, such as mathematical problem solving (\eg GSM8k~\cite{Cobbe-arxiv-2021-Training} and  SVAMP~\cite{Patel-NAACL-2021-Are}) or knowledge question answering (\eg TruthfulQA~\cite{Lin-ACL-2022-TruthfulQA}), where the successful utilization of tools is very important for enhancing the required skills that LLMs are  incapable in (\eg numerical calculation).
In this way, the evaluated performance on these tasks can reflect the ability of LLMs in tool manipulation.}
To teach LLMs to utilize tools, existing studies add exemplars using tools in context to elicit LLMs~\cite{Gao-arxiv-2022-PAL}, or fine-tune LLMs on simulated data about tool utilization~\cite{Parisi-arxiv-2022-TALM, Schick-arxiv-2023-Toolformer}.
It has been found that with the help of tools, LLMs become  more capable of handling the issues that they are not good at,  \eg equation calculation and answering timely questions~\cite{Schick-arxiv-2023-Toolformer,Chen-2023-arXiv-chatcot}.
{
However, as the number of available tools increases, the limited context length of LLMs may pose challenges in describing and demonstrating extensive tool APIs.
To address this issue, existing work retrieves the usage of relevant tools, or encoding tool information as tokens within the embedding space~\cite{Shishir-2023-arxiv-Gorilla,Hao-2023-arxiv-ToolkenGPT,Liang-2023-arxiv-TaskMatrix}.}

{
In addition to existing tools developed by humans, LLMs possess the capability to make their own tools for specific tasks autonomously~\cite{Cai-arxiv-2023-Tool}. 
This enables the models to independently explore and manipulate these self-created tools, thereby expanding their potential for autonomous exploration  in solving a wide range of real-world tasks.}

{\emph{Summary}. The above three abilities are of great value to the practical performance of LLMs: conforming  to 
human values and preferences (human alignment), acting properly in real-world scenarios (interaction with the external environment), and expanding the ability scope (tool manipulation). 
In addition to the above three advanced abilities, LLMs might  also show other    abilities that are specially related to some tasks (\eg data annotation~\cite{Gilardi-arXiv-2023-Crowd}) or learning mechanisms (\eg self-improvement~\cite{Huang-arxiv-2022-Large}).
It will be an open direction to discover, measure and evaluate these newly emerging abilities, so as to better utilize  and improve LLMs.  
}
\begin{table*}[htbp]
    \centering  
    \caption{{A category of existing evaluation work. ``General'' denotes that the evaluation focuses on an overall performance of multiple abilities. The evaluated abilities are not limited to the representative basic and advanced abilities mentioned in Section~\ref{sec:basicability} and \ref{sec:superior}.}}
    \label{tab-category-evaluation}
    \footnotesize
    \begin{tabular}{ccccc}
    \toprule
    \textbf{Method} & \textbf{Evaluation} & \textbf{Model Types} & \textbf{Abilities/Domain} & \textbf{Data Source}\\
    \midrule
    \multirow{28}{*}{Benchmark} & MMLU~\cite{Hendrycks-ICLR-2021-Measuring} & Base/Fine-tuned/Specialized & General & Human exam/practice \\
    & BIG-bench~\cite{Srivastava-arxiv-2022-Beyond} & Base/Fine-tuned/Specialized & General & Human annotation \\
    & HELM~\cite{Liang-arxiv-2022-Holistic} & Base/Fine-tuned/Specialized & General & Benchmark collection \\
    & Open LLM Leaderboard~\cite{Edward-2023-hf-open} & Base/Fine-tuned/Specialized & General & Benchmark collection \\
    & AGIEval~\cite{Zhong-2023-arxiv-AGIEval} & Base/Fine-tuned/Specialized & General & Human exam/practice \\
    & MMCU~\cite{Zeng-arxiv-2023-MMCU} & Base/Fine-tuned/Specialized & General & Human exam/practice \\
    & M3KE~\cite{Liu-2023-arxiv-M3KE} & Base/Fine-tuned/Specialized & General & Human exam/practice \\
    & C-Eval~\cite{Huang-arxiv-2023-CEval} & Base/Fine-tuned/Specialized & General & Human exam/practice \\
    & Xiezhi~\cite{Gu-2023-arxiv-Xiezhi} & Base/Fine-tuned/Specialized & General & Human exam/practice \\
    & OpenCompass~\cite{2023opencompass} & Base/Fine-tuned/Specialized & General & Benchmark collection \\
    & Chain-of-Thought Hub~\cite{Fu-arxiv-2023-Chain} & Base/Fine-tuned & General & Benchmark collection \\
    & KoLA~\cite{Yu-arxiv-2023-KoLA} & Base/Fine-tuned & Knowledge utilization & Web \\
    & ARB~\cite{Sawada-arxiv-2023-ARB} & Fine-tuned & Complex reasoning & Human exam/practice \\
    & APIBench~\cite{Peng-arxiv-2023-Revisiting} & Base/Fine-tuned & Tool manipulation & Web \\
    & APIBank~\cite{Li-arxiv-2023-API-Bank} & Fine-tuned & Tool manipulation & Synthesis \\
    & ToolAlpaca~\cite{Tang-arxiv-2023-ToolAlpaca} & Base/Fine-tuned & Tool manipulation & Synthesis \\
    & T-Bench~\cite{Xu-arxiv-2023-On} & Fine-tuned & Tool manipulation & Synthesis \\
    & ToolBench~\cite{Qin-arxiv-2023-ToolLLM} & Fine-tuned & Tool manipulation & Synthesis \\
    & BOLAA~\cite{Liu-arxiv-2023-BOLAA} & Base/Fine-tuned &  Environment interaction & Benchmark collection \\
    & AgentBench~\cite{Liu-arxiv-2023-AgentBench} & Base/Fine-tuned & Environment interaction & Human annotation/Synthesis \\
    & HaluEval~\cite{Li-arxiv-2023-HaluEval} & Base/Fine-tuned & Human alignment & Human annotation/Synthesis \\
    & PromptBench~\cite{Zhu-arxiv-2023-PromptBench} & Base/Fine-tuned & Robustness & Benchmark collection \\
    & HumanEval~\cite{Chen-arxiv-2021-evaluating} & Base/Fine-tuned/Specialized & Code synthesis & Human annotation \\
    & MultiMedQA~\cite{singhal-arxiv-2022-large} & Specialized & Healthcare & Benchmark collection \\
    & FLUE~\cite{Shah-arxiv-2023-FLUE} & Specialized & Finance & Benchmark collection \\
    & LegalBench~\cite{Guha-arxiv-2022-LegalBench} & Specialized & Legal & Human annotation \\
    \midrule
    \multirow{2}{*}{Human} & Chatbot Arena~\cite{Zheng-2023-arxiv-Judging} & Base/Fine-tuned/Specialized & Human Alignment & Human annotation \\
    & SciBench~\cite{Wang-arxiv-2023-SciBench} & Fine-tuned & Complex reasoning & Human exam/practice \\
    \midrule
    \multirow{5}{*}{Model} & AlpacaEval~\cite{Li-2023-github-alpaca_eval} & Fine-tuned & Instruction following & Synthesis \\
    & MT-bench~\cite{Zheng-2023-arxiv-Judging} & Fine-tuned & Human alignment & Human annotation \\
    & TrustGPT~\cite{Huang-arxiv-2023-TrustGPT} & Base/Fine-tuned & Human alignment & Benchmark collection \\
    & LMExamQA~\cite{Bai-arxiv-2023-Benchmarking} & Base/Fine-tuned & Knowledge utilization & Synthesis \\
    & ChatEval~\cite{Chan-arixiv-2023-ChatEval} & Base/Fine-tuned & Knowledge utilization & Benchmark collection \\
    \bottomrule
    \end{tabular}
\label{tab:benchmark}
\end{table*}

\subsection{Benchmarks and Evaluation Approaches}
In the above, we have discussed the basic and advanced abilities of LLMs.
Next, we will introduce existing evaluation benchmarks and approaches~\cite{Chang-2023-arxiv-A,Zhuang-2023-arxiv-Through}.

\subsubsection{Comprehensive Evaluation Benchmarks}
Recently, several comprehensive benchmarks~\cite{Hendrycks-ICLR-2021-Measuring,Srivastava-arxiv-2022-Beyond,Liang-arxiv-2022-Holistic} have been released for the evaluation of LLMs.
In this part, we introduce several widely used benchmarks, \ie MMLU, BIG-bench, HELM, and a series of human exam benchmarks.

$\bullet$ \emph{MMLU}~\cite{Hendrycks-ICLR-2021-Measuring} is a versatile benchmark for large-scale evaluation of multi-task knowledge understanding, covering a wide range of knowledge domains {from mathematics and computer science to humanities and social sciences. The difficulties of these tasks vary from basic to advanced.} 
{As shown in existing work, LLMs mostly outperform small models by a substantial margin on this benchmark~\cite{Chowdhery-arxiv-2022-PaLM,Chung-arxiv-2022-Scaling,Taylor-arxiv-2022-Galactica,Touvron-arxiv-2023-LLaMA}, which shows the scaling law in model size. %
{More recently, GPT-4 achieves a remarkable record (86.4\% in 5-shot setting) in MMLU, which is significantly  better than the previous state-of-the-art models~\cite{OpenAI-OpenAI-2023-GPT-4}.}}

$\bullet$ \emph{BIG-bench}~\cite{Srivastava-arxiv-2022-Beyond} is a collaborative benchmark intended to probe existing LLMs from various aspects.
{It comprises 204 tasks that encompass a broad range of topics, including linguistics, childhood development, mathematics, commonsense reasoning, biology, physics, social bias, software development, and so on.}
{By scaling  the model size, LLMs can even outperform the average human performance under the few-shot setting on 65\% of tasks in BIG-bench~\cite{Chowdhery-arxiv-2022-PaLM}.}
{Considering the high evaluation cost of the entire benchmark,  a lightweight benchmark BIG-bench-Lite has been proposed, which contains  24 small yet diverse and challenging tasks from BIG-bench.}
Additionally, the BIG-bench hard (BBH) benchmark~\cite{Suzgun-arxiv-2022-Challenging} has been proposed to concentrate on investigating the currently unsolvable tasks of LLMs by selecting the challenging tasks in which LLMs exhibit inferior performance compared to humans.
Since BBH becomes more difficult, small models mostly achieve performance close to random.  
{As a comparison, CoT prompting can elicit the abilities of LLMs to perform step-by-step reasoning for enhancing the performance, even exceeding the average human performance in BBH.}

$\bullet$ \emph{HELM}~\cite{Liang-arxiv-2022-Holistic} is a comprehensive benchmark that currently implements a core set of 16 scenarios and 7 categories of metrics. It is built on top of many prior studies, conducting a holistic evaluation of language models. 
As shown in the experimental results of HELM, instruction tuning can consistently boost the performance of LLMs in terms of accuracy, robustness, and fairness. Further, for reasoning tasks, the LLMs that have been pre-trained on the code corpus show superior performance.

\ignore{
$\bullet$ \emph{Xiezhi}~\cite{Gu-2023-arxiv-Xiezhi} is an extensive benchmark for domain knowledge evaluation, consisting of a vast collection of multiple-choice questions across 516 diverse disciplines spanning 13 different subjects. 
Considering different domain knowledge utilization scenarios, Xiezhi also provides two specific benchmarks, \ie Xiezhi-Specialty and Xiezhi-Interdiscipline.
Experimental results reveal that the top-performing LLMs outperform average human performance in the fields of science, engineering, agronomy, and medicine, while humans still exhibit significantly higher performance than all LLMs in the domains such as economics, jurisprudence, and literature.
}

$\bullet$ \emph{Human-level test benchmarks} aim to evaluate the comprehensive  ability of LLMs with questions designed for testing humans, such as AGIEval~\cite{Zhong-2023-arxiv-AGIEval}, MMCU~\cite{Zeng-arxiv-2023-MMCU}, M3KE~\cite{Liu-2023-arxiv-M3KE},  C-Eval~\cite{Huang-arxiv-2023-CEval} and {Xiezhi}~\cite{Gu-2023-arxiv-Xiezhi}.
These benchmarks encompass a wide range of domains, difficulty levels, and languages to provide a comprehensive evaluation of LLMs' general capabilities.
Compared to publicly available models, models offering API services (\eg  GPT-4, ChatGPT, Claude) demonstrate superior performance compared to publicly available models on these evaluation benchmarks.
As the best-performing model in evaluations, GPT-4 surpasses average human performance in AGIEval~\cite{Zhong-2023-arxiv-AGIEval}.
{However, it still lags behind the top human performance on these challenging benchmarks.} 
Hence, there remains ample room for further enhancements in the overall  abilities of LLMs, particularly for publicly accessible models. 

The above benchmarks cover a variety of mainstream evaluation tasks and real-world human exam questions for the evaluation of LLMs. 
Also, there are several benchmarks that focus on evaluating specific abilities of LLMs, such as TyDiQA~\cite{Clark-trans-2020-TyDi} for multilingual knowledge utilization and MGSM~\cite{Shi-arxiv-2022-Language} for multilingual mathematical reasoning. To conduct the  evaluation, one can select suitable benchmarks according to specific goals.
{In addition, there are also several open-source evaluation frameworks for researchers to evaluate LLMs on existing benchmarks or extend new tasks for customized evaluations, such as Language Model Evaluation Harness~\cite{Leo-zenodo-2021-A} and OpenAI Evals~\cite{OpenAI-OpenAI-2023-GPT-4}.}
{Further, some researchers
also construct continuously updated leaderboards by aggregating representative benchmarks, to compare the performance of existing LLMs, such as Open LLM Leaderboard~\cite{Edward-2023-hf-open}.
The above benchmarks and leaderboards provide important references to demonstrate the basic and advanced abilities of LLMs. We will give more deep discussions on pros and cons on evaluation approaches in Section~\ref{subsec-evaapp}. 
}

\subsubsection{Evaluation Approaches}\label{subsec-evaapp}
After introducing existing benchmarks, in this part, we will review existing evaluation approaches for assessing the performance of LLMs.
To organize our discussion, we categorize LLMs into three different types: \emph{base LLMs} (pre-trained model checkpoints), \emph{fine-tuned LLMs} (instruction or alignment fine-tuned model checkpoints), and \emph{specialized LLMs} (adapted model checkpoints for some specific task or domain).
Here, we keep both fine-tuned LLMs and specialized LLMs, to distinguish the different purposes of LLMs: general or specific  task solvers.
To evaluate the three types of LLMs, we can test the LLM's performance related to  different abilities (\eg basic or advanced abilities as discussed in Section~\ref{sec:basicability} and \ref{sec:superior}).
In general, there are three main approaches to evaluating LLMs, namely benchmark-based approach~\cite{Hendrycks-ICLR-2021-Measuring}, human-based approach~\cite{Zheng-2023-arxiv-Judging}, and model-based approach~\cite{Li-2023-github-alpaca_eval}.
Table~\ref{tab:benchmark} shows an illustration of the relationship among LLM type, evaluation approach, and tested abilities. 
Next, we will discuss the evaluation approaches for different types of LLMs.

\paratitle{Evaluation of Base LLMs.}
Base LLMs refer to the model checkpoints obtained right after pre-training.
For base LLMs, we mainly focus on examining the basic abilities (Section~\ref{sec:basicability}), such as complex reasoning and knowledge utilization.
Since most of these basic abilities can be assessed with well-defined tasks, benchmark-based approaches have been widely used to evaluate base LLMs.
Next, we will introduce common evaluation benchmarks and evaluation procedures for base LLMs.

$\bullet$~\emph{Common benchmarks.}
To evaluate base LLMs, typical benchmarks are designed in the form of close-ended problems like multiple-choice questions.
These commonly used benchmarks can be mainly divided into two categories: knowledge-oriented and reasoning-oriented benchmarks.
Knowledge-oriented benchmarks (\eg MMLU~\cite{Hendrycks-ICLR-2021-Measuring} and C-Eval~\cite{Huang-arxiv-2023-CEval}) aim to evaluate the capacity of world knowledge, while reasoning-oriented benchmarks (\eg GSM8K~\cite{Gao-arxiv-2023-Human}, BBH~\cite{Suzgun-arxiv-2022-Challenging}, and MATH~\cite{Hendrycks-ICLR-2021-Measuring}) focus on evaluating the capability of solving complex reasoning tasks.
Further, some recently proposed benchmarks (\eg OpenCompass~\cite{2023opencompass}) combine these two types for a comprehensive comparison.

$\bullet$~\emph{Benchmark based evaluation procedure.}
To perform the benchmark evaluation, each problem will first be formatted into a prompt for LLMs to generate the result text.
Then, the generated result text will be parsed with human-written rules to get the predicted answer.
Finally, the performance of LLMs can be automatically calculated using standard metrics like accuracy by comparing the predicted answer with the ground-truth one.
The evaluation approach can be conducted in either the few-shot or zero-shot setting, which might lead to different evaluation results or rankings.
Since base LLMs have not been instruction fine-tuned (with relatively weak task generalization ability), the few-shot setting is often more suitable for evaluation.
For some complex reasoning tasks, CoT prompts also need to be used to fully exhibit the capacity during evaluation.
Another note is that this evaluation approach can also be applied to assess the abilities of fine-tuned LLMs.
Actually, several leaderboards (\eg Open LLM Leaderboard~\cite{Edward-2023-hf-open}) are built upon this approach, evaluating both base and fine-tuned LLMs.

\paratitle{Evaluation of Fine-tuned LLMs.} 
Fine-tuned LLMs in this part refer to the model checkpoints obtained after instruction tuning or alignment tuning based on pre-trained model weights\footnote{In some cases, it is also called \emph{chat models}.}.
Typically, fine-tuned LLMs will be tested on various abilities (\eg knowledge utilization and human alignment), and thus it is common that they are assessed with multiple evaluation approaches.
In addition to benchmark-based evaluation, human-based and model-based approaches have also been widely used to evaluate the advanced abilities of fine-tuned LLMs.
Next, we will introduce the two evaluation methods.

$\bullet$~\emph{Human-based evaluation.}
Unlike automatic evaluation for basic abilities, human evaluation typically considers more factors or abilities  in real-world use, such as human alignment and tool manipulation.
In this evaluation approach, test tasks are usually in the form of open-ended questions, and human evaluators are invited to  make judgments on the quality of answers generated by LLMs.
Typically, there are two main types of scoring methods for human evaluators: pairwise comparison and single-answer grading.
In pairwise comparison, given the same question, humans are assigned two answers from different models to determine which one is better, while in single-answer grading, they only need to score a single answer at a time.
For example, HELM~\cite{Liang-arxiv-2022-Holistic} employs humans to perform single-answer grading on summarization and disinformation tasks, while Chatbot Arena~\cite{Zheng-2023-arxiv-Judging} constructs a crowdsourcing platform that allows users to engage in conversations with two anonymous chat LLMs and report pairwise comparison results.

$\bullet$~\emph{Model-based evaluation.}
Since human-based evaluation is both expensive and time-consuming, some work has proposed leveraging powerful closed-source LLMs such as ChatGPT and GPT-4 as a surrogate for human evaluators~\cite{Zheng-2023-arxiv-Judging, Li-2023-github-alpaca_eval}.
For example, AlpacaEval~\cite{Li-2023-github-alpaca_eval} collects a set of instructions and utilizes a capable LLM (\eg GPT-4) as the judge to perform pair-wise comparisons against the reference outputs.
Furthermore, MT-bench~\cite{Zheng-2023-arxiv-Judging} collects a set of multi-turn questions for evaluation and improves the reliability of LLM-based evaluators through methods like ICL and CoT.
Compared with human evaluators, LLMs such as ChatGPT and GPT-4 can achieve high agreement with humans, in both small-scale handcrafted and large-scale crowdsourced evaluation tasks.
Despite this, these closed-source LLMs are limited in access and have the potential risk of data leakage.
To address this, recent work~\cite{Zheng-2023-arxiv-Judging} has explored fine-tuning open-source LLMs (\eg Vicuna~\cite{vicuna2023}) as model evaluators using scoring data from human evaluators, which has  narrowed the gap with powerful closed-source LLMs (\eg GPT-4).

\ignore{
In addition to these two main evaluation approaches, the authors in~\cite{Jain-2023-arxiv-Bring} further propose self-supervised evaluation to get rid of the involvement of humans or models, which performs interventions on the input text and takes the sensitivity of LLMs as the evaluation metric.
}

\paratitle{Evaluation of Specialized LLMs.}
Specialized LLMs refer to the model checkpoints specially adapted to some domains or applications like healthcare~\cite{singhal-arxiv-2022-large} and finance~\cite{Shah-2022-EMNLP-When}.
As special task solvers, specialized LLMs will be tested not only on general abilities (\eg basic ability like complex reasoning and advanced ability like human alignment), but also on specific abilities related to their designated domains or applications.
For this purpose, one often needs to construct specific benchmarks tailored for the target domains or applications. 
Then, these domain-specific benchmarks can be combined with general benchmarks to conduct both comprehensive and targeted evaluation for specialized LLMs.
For example, MultiMedQA~\cite{singhal-arxiv-2022-large} is a specific benchmark in healthcare, which includes medical examinations and healthcare questions.
In this work~\cite{singhal-arxiv-2022-large}, MultiMedQA has been combined  with MMLU~\cite{Hendrycks-ICLR-2021-Measuring} to assess the performance of specialized LLMs for healthcare, such as Med-PaLM~\cite{singhal-arxiv-2022-large}.
Similarly, FLUE~\cite{Shah-2022-EMNLP-When} constructs a benchmark for finance, spanning from financial sentiment analysis to question answering.
It has been used collaboratively with BBH~\cite{Suzgun-arxiv-2022-Challenging} to evaluate finical LLMs like BloombergGPT~\cite{wu-arxiv-2023-bloomberggpt}.

\paratitle{Pros and Cons of Different Evaluation Approaches}. 
In the above, we have discussed different evaluation approaches to assess the abilities of LLMs.
Next, we simply analyze the pros and cons of each evaluation approach.  

$\bullet$~\emph{Benchmark-based approach}.
This evaluation approach can leverage existing benchmarks for assessing the performance of LLMs.
The tasks involved in these benchmarks often contain sufficient test samples to measure the core abilities (\eg reasoning).
The whole evaluation procedure can be (almost) automatic, and it is convenient to carry out test experiments for various base LLMs, especially useful for monitoring the performance of model checkpoints during pre-training.
However, LLMs are often sensitive to the evaluation settings, including the question prompts, zero-shot or few-shot tests, and the answer parsing methods.
Thus, one should take possible influencing factors into consideration when conducting the evaluation experiments.
The evaluation results should be noted with the adopted evaluation settings.
Another issue is the data contamination~\cite{Chowdhery-arxiv-2022-PaLM,zhou-arxiv-2023-dont}, \ie the test data itself or relevant content has been contained in the pre-training corpora.
This phenomenon has become increasingly severe since more and more open data has been collected for developing LLMs.

$\bullet$~\emph{Human-based approach}. 
Human evaluation offers several advantages when assessing the capabilities of LLMs to solve real-world tasks.
One of the key benefits is its ability to directly reflect the actual abilities of LLMs.
Based on feedback and experiences from real users, human evaluation provides a more direct measure of LLMs' performance in real-world scenarios.
Further, it can conduct more flexible and diverse evaluation tasks based on human evaluators.
For instance, users can submit various queries and test the abilities of LLMs according to their own task cognition.
It allows for a deep understanding of the strengths and weaknesses of LLMs across different types of tasks and contexts.
However, human evaluation also has inherent limitations that could potentially affect its accuracy and consistency.
Factors such as personalized tastes and varying education levels among evaluators can introduce biases or even inconsistencies in the evaluation process.
In some cases, users' judgments are likely to be subjective, which may not reflect the true capabilities of the LLMs.
Moreover, conducting robust and reliable human evaluations often requires a large number of evaluators, which can be very expensive and time-consuming.
In addition, human evaluation is often not reproducible, making it infeasible to extend existing evaluation results or track the progress of LLMs.

$\bullet$~\emph{Model-based approach}.
As a surrogate for human-based approaches, model-based approaches serve to diminish the reliance on human involvement, and enable more efficient and scalable evaluation.
In addition, LLMs can provide meaningful explanations for the assigned rating scores, thereby enhancing the interpretability of evaluations.
Despite their scalability and explanability, model-based approaches have been found to suffer from several issues, including position, verbosity, and self-enhancement bias~\cite{Zheng-2023-arxiv-Judging}.
Specially, position bias (\ie the order to present the responses) refers to the fact that LLMs tend to assign high scores for the answers at specific positions over others, 
 verbosity bias means that LLMs favor verbose answers even if they are short in quality compared with shorter answers, and 
self-enhancement bias indicates that LLMs often overrate in their own generations. 
In addition, since LLMs have limited capacities in  solving complex reasoning problems, they cannot serve as qualified evaluators for some difficult tasks (\eg mathematical reasoning).
These limitations can be mitigated to some extent by specific prompt engineering and fine-tuning strategies~\cite{Zheng-2023-arxiv-Judging}.


To summarize, our categorization (Table~\ref{tab:benchmark}) of existing work on LLM evaluation is mainly based on two major dimensions, namely evaluation methodology and model type, which are further extended with the test abilities.
There are some recent work~\cite{Chang-2023-arxiv-A,Zhuang-2023-arxiv-Through} that also has discussed the categorization or taxonomies of existing work for LLM evaluation.

\begin{table*}[htb]
\renewcommand\arraystretch{1.1}
\setlength\tabcolsep{2.5pt}
    \centering
    \caption{Evaluation on the eight abilities of  LLMs with specially selected tasks. The shade of the \colorbox[HTML]{FC8D59}{Orange} and \colorbox[HTML]{92BFDB}{Blue} fonts denote the performance orders of the results in closed-source and open-source models, respectively. This table will be continuously updated by incorporating the results of more models.
    }
    \label{tab-experimental-res}
\resizebox{1.85\columnwidth}{!}{
\begin{tabular}{lccccccccc}
\toprule
\multirow{2.5}{*}{\textbf{Models}}   & \multicolumn{4}{c}{\textbf{Language Generation}} & \multicolumn{5}{c}{\textbf{Knowledge Utilization}} \\ 
\cmidrule(r){2-5}\cmidrule(r){6-10}
& LBD$\uparrow$ & WMT$\uparrow$ & XSum$\uparrow$ & HumanEval$\uparrow$ & TriviaQA$\uparrow$ & NaturalQ$\uparrow$ & WebQ$\uparrow$ & ARC$\uparrow$ & WikiFact$\uparrow$ \\
\midrule
ChatGPT & \cellcolor[HTML]{FEE8DD}55.81 & \cellcolor[HTML]{FCA77F}{36.44} & \cellcolor[HTML]{FC8D59}{21.71} & \cellcolor[HTML]{FC8D59}{79.88} & \cellcolor[HTML]{FC8D59}{54.54} & \cellcolor[HTML]{FC8D59}{21.52} & \cellcolor[HTML]{FEDCCC}17.77 & \cellcolor[HTML]{FC8D59}{93.69} & \cellcolor[HTML]{FEDCCC}{29.25} \\
Claude & \cellcolor[HTML]{FCA77F}64.47 & \cellcolor[HTML]{FEE8DD}{31.23} & \cellcolor[HTML]{FEE8DD}{18.63} & \cellcolor[HTML]{FEF7F3}{51.22} & \cellcolor[HTML]{FEF7F3}40.92 & \cellcolor[HTML]{FEF7F3}13.77 & \cellcolor[HTML]{FEF7F3}14.57 & \cellcolor[HTML]{FEF7F3}66.62 & \cellcolor[HTML]{FCA77F}{34.34} \\
Claude 2 & \cellcolor[HTML]{FEF7F3}45.20 & \cellcolor[HTML]{FEF7F3}12.93 & \cellcolor[HTML]{FEDCCC}19.13 & \cellcolor[HTML]{FCA77F}78.04 & \cellcolor[HTML]{FCA77F}54.30 & \cellcolor[HTML]{FCA77F}21.30 & \cellcolor[HTML]{FC8D59}21.06 & \cellcolor[HTML]{FEE8DD}79.97 & \cellcolor[HTML]{FC8D59}35.83\\
Davinci003 & \cellcolor[HTML]{FC8D59}69.98 & \cellcolor[HTML]{FC8D59}{37.46} & \cellcolor[HTML]{FEF7F3}{18.19} & \cellcolor[HTML]{FEDCCC}{67.07} & \cellcolor[HTML]{FEE8DD}51.51 & \cellcolor[HTML]{FEE8DD}17.76 & \cellcolor[HTML]{FEE8DD}16.68 & \cellcolor[HTML]{FEDCCC}88.47 & \cellcolor[HTML]{FEF7F3}{28.29}  \\
Davinci002 &  \cellcolor[HTML]{FEDCCC}58.85 & \cellcolor[HTML]{FEDCCC}{35.11} & \cellcolor[HTML]{FCA77F}{19.15} & \cellcolor[HTML]{FEE8DD}{56.70} & \cellcolor[HTML]{FEDCCC}52.11 & \cellcolor[HTML]{FEDCCC}20.47 & \cellcolor[HTML]{FCA77F}{18.45} & \cellcolor[HTML]{FCA77F}89.23 & \cellcolor[HTML]{FEE8DD}{29.15}  \\
\midrule
LLaMA 2-Chat~(7B) & 56.12 & 12.62 & \cellcolor[HTML]{A7CBE2}16.00 & 11.59 & \cellcolor[HTML]{92BFDB}38.93 & \cellcolor[HTML]{92BFDB}12.96 & \cellcolor[HTML]{A7CBE2}11.32 & \cellcolor[HTML]{92BFDB}72.35 & 23.37 \\
Vicuna~(13B) & 62.45 & \cellcolor[HTML]{A7CBE2}20.49 & \cellcolor[HTML]{92BFDB}17.87 & \cellcolor[HTML]{92BFDB}20.73 & \cellcolor[HTML]{C4DDEC}29.04 & \cellcolor[HTML]{C6DEED}10.75 & \cellcolor[HTML]{92BFDB}11.52 & \cellcolor[HTML]{E5F0F7}
20.69 & \cellcolor[HTML]{92BFDB}28.76 \\
Vicuna~(7B) & \cellcolor[HTML]{C4DDEC}63.90 & \cellcolor[HTML]{C6DEED}{19.95} & \cellcolor[HTML]{C6DEED}{13.59} & \cellcolor[HTML]{A7CBE2}{17.07} & 28.58 & \cellcolor[HTML]{C4DDEC}9.17 & 6.64 & 16.96 & \cellcolor[HTML]{C6DEED}{26.95} \\
Alpaca~(7B) & \cellcolor[HTML]{E5F0F7}63.35 & \cellcolor[HTML]{92BFDB}{21.52} & {8.74} & {13.41} & 17.14 & 3.24 & 3.00 & \cellcolor[HTML]{C6DEED}49.75 & \cellcolor[HTML]{C4DDEC}{26.05}\\
ChatGLM~(6B) & 33.34 & \cellcolor[HTML]{C4DDEC}{16.58} & \cellcolor[HTML]{C4DDEC}{13.48} & {13.42} & 13.42 & 
4.40 & \cellcolor[HTML]{C4DDEC}9.20 & \cellcolor[HTML]{A7CBE2}{55.39} & {16.01} \\
\midrule
LLaMA 2~(7B) & \cellcolor[HTML]{C6DEED}66.39 & 11.57 & \cellcolor[HTML]{E5F0F7}11.57 & \cellcolor[HTML]{A7CBE2}17.07 & \cellcolor[HTML]{C6DEED}30.92 & 5.15 & 2.51 & \cellcolor[HTML]{C4DDEC}24.16 & \cellcolor[HTML]{A7CBE2}28.06\\
LLaMA~(7B) & \cellcolor[HTML]{92BFDB}67.68 & \cellcolor[HTML]{E5F0F7}{13.84} & {8.77} & \cellcolor[HTML]{C4DDEC}{15.24} & \cellcolor[HTML]{A7CBE2}{34.62} & \cellcolor[HTML]{E5F0F7}7.92 & \cellcolor[HTML]{C6DEED}{11.12} & 4.88 & {19.78} \\
Falcon~(7B) & \cellcolor[HTML]{A7CBE2}66.89 & {4.05} & {10.00} & {10.37} & \cellcolor[HTML]{E5F0F7}28.74 & \cellcolor[HTML]{A7CBE2}{10.78} & \cellcolor[HTML]{E5F0F7}8.46 & 4.08 & \cellcolor[HTML]{E5F0F7}{23.91} \\
Pythia~(12B) & 61.19 & {5.43} & {8.87} & \cellcolor[HTML]{E5F0F7}{14.63} & 15.73 &
1.99 & 4.72 & 11.66 & {20.57} \\
Pythia~(7B) & 56.96 & {3.68} & {8.23} & {9.15} & 10.16 & 1.77 & 3.74 & 11.03 & {15.75}\\
\midrule[0.8pt]
\multirow{2.5}{*}{\textbf{Models}} & \multicolumn{3}{c}{\textbf{Knowledge Reasoning}} & \multicolumn{2}{c}{\textbf{Symbolic Reasoning}} & \multicolumn{2}{c}{\textbf{Mathematical Reasoning}} & \multicolumn{2}{c}{\textbf{Interaction with Environment}}\\ 
\cmidrule(r){2-4}\cmidrule(r){5-6}\cmidrule(r){7-8}\cmidrule(r){9-10}
& OBQA$\uparrow$ & HellaSwag$\uparrow$ & SocialIQA$\uparrow$ & C-Objects$\uparrow$ & Penguins$\uparrow$ & GSM8k$\uparrow$ & MATH$\uparrow$ & ALFW$\uparrow$ & WebShop$\uparrow$ \\
\midrule
ChatGPT & \cellcolor[HTML]{FCA77F}81.20 & \cellcolor[HTML]{FCA77F}61.43 & \cellcolor[HTML]{FC8D59}73.23 & \cellcolor[HTML]{FEF7F3}53.20 & \cellcolor[HTML]{FEF7F3}40.27 & \cellcolor[HTML]{FCA77F}{78.47} & \cellcolor[HTML]{FC8D59}{33.78} & \cellcolor[HTML]{FEF7F3}58.96 & \cellcolor[HTML]{FEDCCC}45.12/15.60 \\
Claude & \cellcolor[HTML]{FC8D59}81.80 & \cellcolor[HTML]{FEDCCC}54.95 & \cellcolor[HTML]{FC8D59}73.23 & \cellcolor[HTML]{FEE8DD}59.95 & \cellcolor[HTML]{FEE8DD}47.65 & \cellcolor[HTML]{FEDCCC}70.81 & \cellcolor[HTML]{FEDCCC}20.18 & \cellcolor[HTML]{FCA77F}76.87 & \cellcolor[HTML]{FCA77F}47.72/23.00 \\
Claude 2 & \cellcolor[HTML]{FEE8DD}71.60 & \cellcolor[HTML]{FEE8DD}50.75 & \cellcolor[HTML]{FEE8DD}58.34 & \cellcolor[HTML]{FC8D59}{66.76} & \cellcolor[HTML]{FC8D59}{74.50} & \cellcolor[HTML]{FC8D59}{82.87} & \cellcolor[HTML]{FCA77F}{32.24} & \cellcolor[HTML]{FC8D59}77.61 & \cellcolor[HTML]{FEE8DD}34.96/19.20 \\
Davinci003 & \cellcolor[HTML]{FEDCCC}74.40 & \cellcolor[HTML]{FC8D59}62.65 & \cellcolor[HTML]{FEDCCC}69.70 & \cellcolor[HTML]{FCA77F}{64.60} & \cellcolor[HTML]{FEDCCC}61.07 & \cellcolor[HTML]{FEE8DD}57.16 & \cellcolor[HTML]{FEE8DD}17.66 & \cellcolor[HTML]{FEDCCC}65.67 & \cellcolor[HTML]{FC8D59}{64.08/32.40} \\
Davinci002 & \cellcolor[HTML]{FEF7F3}69.80 & \cellcolor[HTML]{FEF7F3}47.81 & \cellcolor[HTML]{FEF7F3}57.01 & \cellcolor[HTML]{FEDCCC}62.55 & \cellcolor[HTML]{FCA77F}{67.11} & \cellcolor[HTML]{FEF7F3}49.96 & \cellcolor[HTML]{FEF7F3}14.28 & \cellcolor[HTML]{FCA77F}{76.87} & \cellcolor[HTML]{FEF7F3}29.66/15.20 \\
\midrule
LLaMA 2-Chat~(7B) & \cellcolor[HTML]{A7CBE2}45.62 & \cellcolor[HTML]{C6DEED}
74.01 & \cellcolor[HTML]{C4DDEC}43.84 & \cellcolor[HTML]{C4DDEC}43.40 & \cellcolor[HTML]{A7CBE2}38.93 & 9.63 & 2.22 & \cellcolor[HTML]{92BFDB}11.19 & \cellcolor[HTML]{92BFDB}{24.51/5.60} \\
Vicuna~(13B) & \cellcolor[HTML]{E5F0F7}43.65 & \cellcolor[HTML]{E5F0F7}70.51 & \cellcolor[HTML]{C6DEED}45.97 & \cellcolor[HTML]{92BFDB}53.55 & \cellcolor[HTML]{C6DEED}36.91 & \cellcolor[HTML]{92BFDB}18.50 & \cellcolor[HTML]{A7CBE2}3.72 & \cellcolor[HTML]{A7CBE2}8.96 & \cellcolor[HTML]{A7CBE2}{22.74/5.00} \\
Vicuna~(7B) & \cellcolor[HTML]{C4DDEC}43.84 & 69.25 & \cellcolor[HTML]{A7CBE2}
46.27 & \cellcolor[HTML]{A7CBE2}44.25 & \cellcolor[HTML]{C4DDEC}36.24 & \cellcolor[HTML]{A7CBE2}14.03 & \cellcolor[HTML]{C6DEED}3.54 & 1.49 & \cellcolor[HTML]{C4DDEC}
6.90/1.40\\
Alpaca~(7B) & \cellcolor[HTML]{92BFDB}47.82 & 69.81 & \cellcolor[HTML]{92BFDB}47.55 & 39.35 & \cellcolor[HTML]{92BFDB}40.27 & \cellcolor[HTML]{E5F0F7}4.93 & \cellcolor[HTML]{92BFDB}4.16 & 4.48 & 0.00/0.00\\
ChatGLM~(6B) & 30.42 & 29.27 & 33.18 & 14.05 & 14.09 & 3.41 & 1.10 & 0.00 & 0.00/0.00 \\
\midrule
LLaMA 2~(7B) & \cellcolor[HTML]{C6DEED}44.81 & \cellcolor[HTML]{A7CBE2}74.25 &41.72 & \cellcolor[HTML]{C6DEED}43.95 & \cellcolor[HTML]{E5F0F7}35.75 & \cellcolor[HTML]{C6DEED}10.99 & \cellcolor[HTML]{E5F0F7}2.64 & \cellcolor[HTML]{A7CBE2}8.96 & 0.00/0.00 \\
LLaMA~(7B) & 42.42 & \cellcolor[HTML]{C4DDEC}73.91 & 41.46 & \cellcolor[HTML]{E5F0F7}39.95 & 34.90 & \cellcolor[HTML]{C6DEED}10.99 & \cellcolor[HTML]{C4DDEC}3.12 & 2.24 & 0.00/0.00 \\
Falcon~(7B) & 39.46 & \cellcolor[HTML]{92BFDB}74.58 & \cellcolor[HTML]{E5F0F7}
42.53 & 29.80 & 24.16 & 1.67 & 0.94 & \cellcolor[HTML]{C4DDEC}
{7.46} & 0.00/0.00 \\
Pythia~(12B) & 37.02 & 65.45 & 41.53 & 32.40 & 26.17 & 2.88 & 1.96 & 5.22 & \cellcolor[HTML]{E5F0F7}3.68/0.60 \\
Pythia~(7B) & 34.88 & 61.82 & 41.01 & 29.05 & 27.52 & 1.82 & 1.46  & \cellcolor[HTML]{C4DDEC}{7.46} & \cellcolor[HTML]{C6DEED}{10.75/1.80}\\
\midrule[0.8pt]
\multirow{2.5}{*}{\textbf{Models}}  &  \multicolumn{5}{c}{\textbf{Human Alignment}} & \multicolumn{4}{c}{\textbf{Tool Manipulation}} \\
\cmidrule(r){2-6}\cmidrule(r){7-10}
& TfQA$\uparrow$ & C-Pairs$\downarrow$ & WinoGender$\uparrow$ & RTP$\downarrow$ & HaluEval$\uparrow$ & HotpotQA$\uparrow$ & Gorilla-TH$\uparrow$ & Gorilla-TF$\uparrow$ & Gorilla-HF$\uparrow$ \\
\midrule
ChatGPT & \cellcolor[HTML]{FCA77F}{69.16} & \cellcolor[HTML]{FEE8DD}{18.60} & \cellcolor[HTML]{FCA77F}{62.50}/\cellcolor[HTML]{FCA77F}{72.50}/\cellcolor[HTML]{FCA77F}{79.17} & \cellcolor[HTML]{FC8D59}{3.07} & \cellcolor[HTML]{FC8D59}{66.64} & \cellcolor[HTML]{FEF7F3}{23.80} & \cellcolor[HTML]{FCA77F}{67.20} & \cellcolor[HTML]{FC8D59}
{44.53} & \cellcolor[HTML]{FCA77F}{19.36}\\
Claude & \cellcolor[HTML]{FEDCCC}{67.93} & \cellcolor[HTML]{FEF7F3}{32.73} & \cellcolor[HTML]{FEE8DD}{71.67}/\cellcolor[HTML]{FEE8DD}{55.00}/\cellcolor[HTML]{FEE8DD}{52.50} & \cellcolor[HTML]{FEDCCC}{3.75} & \cellcolor[HTML]{FCA77F}{63.75}  & \cellcolor[HTML]{FEDCCC}{33.80} & \cellcolor[HTML]{FEE8DD}{22.04} & \cellcolor[HTML]{FEDCCC}{7.74} & \cellcolor[HTML]{FEDCCC}{7.08}\\
Claude 2 & \cellcolor[HTML]{FC8D59}{71.11} & \cellcolor[HTML]{FEDCCC}{10.67} & \cellcolor[HTML]{FEF7F3}{60.00}/60.00/55.83 & \cellcolor[HTML]{FCA77F}{3.20} & \cellcolor[HTML]{FEF7F3}50.63 & \cellcolor[HTML]{FC8D59}{36.4} & \cellcolor[HTML]{FEDCCC}{61.29} & \cellcolor[HTML]{FCA77F}{22.19} & \cellcolor[HTML]{FC8D59}{23.67} \\
Davinci003 & \cellcolor[HTML]{FEE8DD}{60.83} & \cellcolor[HTML]{FC8D59}{0.99} & \cellcolor[HTML]{FC8D59}67.50/68.33/{79.17} & \cellcolor[HTML]{FEE8DD}{8.81} & \cellcolor[HTML]{FEE8DD}{58.94}  & \cellcolor[HTML]{FCA77F}
{34.40} & \cellcolor[HTML]{FC8D59}
{72.58} & \cellcolor[HTML]{FEE8DD}{3.80} & \cellcolor[HTML]{FEE8DD}{6.42}\\
Davinci002 & \cellcolor[HTML]{FEF7F3}{53.73} & \cellcolor[HTML]{FCA77F}{7.56} & \cellcolor[HTML]{FEDCCC}{72.50}/70.00/64.17 & \cellcolor[HTML]{FEF7F3}{10.65} & \cellcolor[HTML]{FEDCCC}{59.67}  & \cellcolor[HTML]{FEE8DD}{26.00} & \cellcolor[HTML]{FEF7F3}{2.69} & \cellcolor[HTML]{FEF7F3}{1.02} & \cellcolor[HTML]{FEF7F3}{1.00}\\
\midrule
LLaMA 2-Chat~(7B) & \cellcolor[HTML]{92BFDB}{69.77} & \cellcolor[HTML]{A7CBE2}{48.54} & 47.50/46.67/46.67 & \cellcolor[HTML]{A7CBE2}{4.61} & \cellcolor[HTML]{C6DEED}43.82 & \cellcolor[HTML]{C4DDEC}{4.40} & 0.00 & 0.00 & \cellcolor[HTML]{C6DEED}{0.22} \\
Vicuna~(13B) & \cellcolor[HTML]{C6DEED}{62.30} & \cellcolor[HTML]{92BFDB}{45.95} & \cellcolor[HTML]{C6DEED}{50.83}/50.83/52.50 & \cellcolor[HTML]{E5F0F7}{5.00} & \cellcolor[HTML]{92BFDB}49.01 & \cellcolor[HTML]{A7CBE2}{11.20} & 0.00 & \cellcolor[HTML]{92BFDB}{0.44} & \cellcolor[HTML]{92BFDB}{0.89} \\
Vicuna~(7B) & \cellcolor[HTML]{C4DDEC}{57.77} & {67.44} & 49.17/49.17/49.17 & \cellcolor[HTML]{C6DEED}{4.70} & \cellcolor[HTML]{C4DDEC}{43.44} & \cellcolor[HTML]{C6DEED}{6.20} & {0.00} & {0.00} & \cellcolor[HTML]{A7CBE2}{0.33}\\
Alpaca~(7B) & {46.14} & {65.45} & \cellcolor[HTML]{A7CBE2}{53.33}/51.67/{53.33} & \cellcolor[HTML]{C4DDEC}{4.78} & \cellcolor[HTML]{A7CBE2}{44.16}  & \cellcolor[HTML]{92BFDB}{11.60} & {0.00} & {0.00} & \cellcolor[HTML]{C4DDEC}{0.11}\\
ChatGLM~(6B) & \cellcolor[HTML]{A7CBE2}{63.53} & \cellcolor[HTML]{C6DEED}{50.53} & 47.50/47.50/46.67 & \cellcolor[HTML]{92BFDB}{2.89} & {41.82} & \cellcolor[HTML]{E5F0F7}{4.00} & {0.00} & {0.00} & {0.00}\\
\midrule
LLaMA 2~(7B) & 50.06 & \cellcolor[HTML]{C4DDEC}{51.39} & 48.83/48.83/50.83 & 6.17 & \cellcolor[HTML]{E5F0F7}42.23 & 3.80 & 0.00 & 0.00 & \cellcolor[HTML]{C4DDEC}{0.11} \\
LLaMA~(7B) & {47.86} & {67.84} & \cellcolor[HTML]{92BFDB}54.17/{52.50}/51.67 & {5.94} & {14.18} & {1.60} & {0.00} & {0.00} & \cellcolor[HTML]{C4DDEC}{0.11}\\
Falcon~(7B) & {53.24} & {68.04} & \cellcolor[HTML]{E5F0F7}50.00/50.83/50.00 & {6.71} & {37.41} & {1.00} & {0.00} & {0.00} & {0.00}\\
Pythia~(12B) & \cellcolor[HTML]{E5F0F7}{54.47} & {65.78} & 49.17/48.33/49.17 & {6.59} & {27.09} & {0.40} & {0.00} & {0.00} & {0.00}\\
Pythia~(7B) & {50.92} & \cellcolor[HTML]{E5F0F7}{64.79} & \cellcolor[HTML]{C4DDEC}51.67/49.17/50.00 & {13.02} & {25.84}  & {0.20} & {0.00} & {0.00} & {0.00}\\
\bottomrule\end{tabular}
}
\end{table*}

\begin{table*}[!h]
    \centering
    \caption{Prompt examples and their performance of ChatGPT on representative tasks. For most  tasks, we compare the performance for \emph{simple} and \emph{complex}  prompts. We also present the  reported performance of supervised methods. ``LG'', ``KU'', ``CR'', ``SDG'', ``IR'' are short for ``language generation'', ``knowledge utilization'', ``complex reasoning'', ``structured data generation'', ``information retrieval''. ``-'' means there is no reported supervised result previously on this dataset.}
    \label{tab-instructions}
\scriptsize 
\begin{tabular}{cp{0.10\textwidth}c p{0.55\textwidth} rr}
\toprule
\multicolumn{2}{c}{\textbf{Tasks}}   & \textbf{Datasets} & \makecell[c]{\textbf{Instructions}} &  \textbf{ChatGPT} & \textbf{Supervised} \\ 
\midrule
        \multirow{13.5}{*}{LG} & \multirow{4.5}{*}{Translation} & \multirow{4.5}{*}{WMT} & \texttt{I want you to act as a translator. Please translate the English sentence into Czech.} & 20.66 & \multirow{4.5}{*}{41.40~\cite{Zan-WMT-2022-Vega-MT}}\\
        \cmidrule{4-5}
        & & & \texttt{I want you to act as a translator. Translate the given English sentence into Czech, and ensure that the translated sentence is semantically consistent with the given sentence. $\backslash$n Sentence: \{source sentence\} $\backslash$n Translation:} & 21.12 \\
\cmidrule{2-6}
& \multirow{4.5}{*}{Summarization} & \multirow{4.5}{*}{XSum} & \texttt{Please generate a one-sentence summary for the given document.} & 21.71 & \multirow{4.5}{*}{42.08~\cite{Zhao-arxiv-2022-Calibrating}} \\
\cmidrule{4-5}
        & & & \texttt{\{document\} Try your best to summarize the main content of the given document. And generate a short summary in 1 sentence for it.$\backslash$n Summary:
} & 23.01 &\\
\midrule
\multirow{13}{*}{KU} & \multirow{2.5}{*}{Closed-Book QA} & \multirow{2.5}{*}{ARC} & \texttt{Choose your answer to the question. \{query\} \{options\}} & 85.19 & \multirow{2.5}{*}{92.00~\cite{Khashabi-EMNLP-2020-UnifiedQA}} \\
\cmidrule{4-5}
& & & \texttt{Choose a correct answer according to the given question, and output the corresponding id, do not answer other content except the answer id.} & 85.86 \\
\cmidrule{2-6}
& \multirow{6.5}{*}{Open-Book QA} & \multirow{6.5}{*}{OBQA} & \texttt{Choose your answer to the question: \{question\} \{choices\}. You must only output A, B, C, or D without any extra explanation. The answer is} & 81.20 & \multirow{6.5}{*}{87.20~\cite{Khashabi-EMNLP-2020-UnifiedQA}} \\
\cmidrule{4-5}
& & & \texttt{Following is a question that requires multi-step reasoning, use of additional common and commonsense knowledge, and rich text comprehension. Choose your answer to the question: $\backslash$n Question: Frilled sharks and angler fish live far beneath the surface of the ocean, which is why they are known as $\backslash$n Choices: $\backslash$n A. Deep sea animals $\backslash$n B. fish $\backslash$n C. Long Sea Fish $\backslash$n D. Far Sea Animals $\backslash$n You must only output A, B, C, or D without any extra explanation. The answer is} & 82.20 \\
\cmidrule{2-6}
& \multirow{2.5}{*}{Fact Extraction} & \multirow{2.5}{*}{WikiF} & \texttt{Complete the sentence with one or a few words.} & 29.25 & \multirow{2.5}{*}{34.20~\cite{Liang-arxiv-2022-Holistic}}  \\
\cmidrule{4-5}
& & & \texttt{Complete the given sentence with one entity name in Wikipedia (MUST be a noun) as short as possible, and ensure that the completed sentence conforms to the facts.} & 31.21\\
\midrule
\multirow{13}{*}{CR} & \multirow{2.5}{*}{Symbolic Reasoning} & \multirow{2.5}{*}{C-Objects} & \texttt{Problem: \{problem\}$\backslash$n
Answer:} & 53.20 & \multirow{2.5}{*}{---} \\
\cmidrule{4-5}
& & & \texttt{You are an expert in reasoning problem. Here are some examples about symbolic reasoning. You can use the knowledge in examples and solve the last problem. You should follow the examples and generate the final answer without external solution or words.} & 66.75\\
\cmidrule{2-6}
\cmidrule{2-6}
& \multirow{4.5}{*}{Math Word Problems} & \multirow{4.5}{*}{GSM8k} & \texttt{Problem: \{problem\}$\backslash$n
Solution: Let's think step by step.
} & 78.47 & \multirow{4.5}{*}{63.20~\cite{Zhu-arxiv-2022-Solving}} \\
\cmidrule{4-5}
& & & \texttt{Let's use python to solve math problems. Here are three examples how to do it,$\backslash$n Q: Olivia has \$23. She bought five bagels for \$3 each. How much money does she have left?$\backslash$n```def solution():$\backslash$n ~~~~"""Olivia has \$23. She bought five bagels for \$3 each. How much money does she have left?"""$\backslash$n ~~~~money\_initial = 23$\backslash$n ~~~~bagels = 5$\backslash$n ~~~~bagel\_cost = 3$\backslash$n ~~~~money\_spent = bagels * bagel\_cost$\backslash$n ~~~~money\_left = money\_initial - money\_spent$\backslash$n ~~~~result = money\_left$\backslash$n ~~~~return result```$\backslash$n ...... $\backslash$n How about this question?$\backslash$n Q:} & 79.30\\
\midrule
\multirow{7}{*}{SDG} & Code Synthesis & HumanEval & \texttt{I want you act as a code completer. Given a code snippet, your objective is to complete the code and ensure that it can achieve the described functionality.} & 79.88 & 48.20~\cite{Nguyen-arxiv-2023-Meet} \\
\cmidrule{2-6}
& \makecell[l]{Text-to-SQL} & Spider & \texttt{\#\#\# Complete sqlite SQL query only and with no explanation.$\backslash$n \#$\backslash$n\#\#\# Sqlite SQL tables, with their properties: $\backslash$n\#$\backslash$n\{table\}$\backslash$n\# \{foreign\_key\}$\backslash$n\#$\backslash$n\#\#\# \{question\}$\backslash$n SELECT} & 70.10 & 84.10~\cite{Li-arxiv-2023-RESDSQL} \\
\midrule
\multirow{15}{*}{IR} & Recommendation & MovieLens & \texttt{I've watched the following movies in the past in order: $\backslash$n \{user\_his\_text\} $\backslash$n$\backslash$n Now there are \{recall\_budget\} candidate movies that I can watch next: $\backslash$n \{candidate\_text\_order\} $\backslash$n Please rank these \{recall\_budget\} movies by measuring the possibilities that I would like to watch next most, according to my watching history. Please think step by step. $\backslash$n Note that my most recently watched movie is \{recent\_item\}. Please show me your ranking results with order numbers. Split your output with line break. You MUST rank the given candidate movies. You can not generate movies that are not in the given candidate list.} & 48.80 & 76.25~\cite{Kang-ICDM-2018-Self}
\\
\cmidrule{2-6}
& Conversational \quad Recommendation & ReDial & \texttt{Recommend 10 items that are consistent with user preference. The recommendation list can contain items that the dialog mentioned before. The format of the recommendation list is: no. title (year). Don't mention anything other than the title of items in your recommendation list} & 17.20 & 25.60~\cite{Yang-NAACL-2022-Improving} \\
\bottomrule\end{tabular}
\end{table*}



\subsection{Empirical Evaluation}\label{sec-empirical}
The above evaluation benchmarks and approaches are mainly employed to evaluate the overall abilities of LLMs. 
In this part, we conduct a fine-grained evaluation of the abilities discussed in Section~\ref{sec:basicability} and Section~\ref{sec:superior}.
For each kind of ability, we select representative tasks and datasets for conducting evaluation experiments to examine the corresponding performance of LLMs. 

\subsubsection{Experimental Settings}
In this part, we introduce the experimental settings for our evaluation.

\paratitle{Evaluation Models.} 
To conduct the evaluation, we consider representative LLMs from open-source models to closed-source API-accessing models as follows:

\textbullet~\emph{Open-source models.} 
Existing open-source models can be categorized into base models and instruction-tuned models. Base models are only pre-trained on a large general-purpose corpus with the language modeling objective, but without further supervised fine-tuning. In our evaluation, we select four representative base models including LLaMA (7B)~\cite{Touvron-arxiv-2023-LLaMA}, LLaMA 2 (7B)~\cite{Touvron-2023-llama2-arxiv}, Pythia (7B and 12B)~\cite{Biderman-arxiv-2023-Pythia}, and Falcon (7B)~\cite{Ebtesam-arxiv-2023-Falcon}\footnote{Experiments with larger models are still in schedule due to the limit of computational resources. }. 
Instruction-tuned models are those fine-tuned using instructions (\ie task datasets, daily chat, or synthetic instructions). In our experiments, we select four representative instruction-tuned models including Vicuna (7B and 13B)~\cite{vicuna2023}, Alpaca (7B)~\cite{alpaca}, and ChatGLM (6B)~\cite{Zeng-arxiv-2022-GLM}.
{In addition, we also include LLaMA 2-Chat (7B)~\cite{Touvron-2023-llama2-arxiv} for comparison, and it is a representative model that has been aligned with human via instruction tuning and RLHF, based on LLaMA 2 (7B).}

\textbullet~\emph{Closed-source models.} 
In addition to the open-source models, there are also closed-source models that can only be accessed via APIs, which have gained much attention from both developers and researchers. 
Here, we select four representative closed-source models including text-davinci-002/003 (short as \emph{Davinci002/003}), ChatGPT, Claude, 
{and Claude 2, where the first three models are developed by OpenAI and the other two are developed by Anthropic.}

\paratitle{Tasks and Datasets.} Next, we set up the evaluation tasks and datasets for the abilities discussed in  Section~\ref{sec:basicability} and Section~\ref{sec:superior}.  
{We mainly evaluate the zero-shot performance of LLMs on these datasets. For more complex tasks that are hard to be solved in the zero-shot manner (\eg mathematical reasoning and tool manipulation), we mainly report the 3-shot performance, considering the context length limit of open-source models. 
}

\textbullet~\emph{Language generation.} As discussed before, for language generation, we consider evaluating three kinds of tasks, \ie language modeling, conditional text generation, and code synthesis. Specially, we select four commonly-used datasets, namely LAMBADA~\cite{Paperno-ACL-2016-LAMBADA} (language modeling), WMT'22~\cite{Kocmi-WMT-2022-Findings} (machine translation), XSum~\cite{Naryan-EMNLP-2018-XSUM} (text summarization), and HumanEval~\cite{Chen-arxiv-2021-evaluating} (code synthesis) for evaluation. 
{
In WMT'22, we construct a new evaluation set by selecting 1000 examples for each language pair from the original large-scale test set to examine the average performance of LLMs in machine translation.}
We evaluate the zero-shot performance of LLMs on these datasets, and compute the \emph{accuracy} of predicting words for LAMBADA, \emph{BLEU-4} for WMT'22, \emph{ROUGE-L} for XSum, and \emph{pass@$10$} for HumanEval.

\textbullet~\emph{Knowledge utilization.} 
To evaluate the ability of knowledge utilization, 
we select four question answering datasets (\ie TriviaQA~\cite{Joshi-ACL-2017-TriviaQA}, Natural Questions~\cite{Kwiatkowski-ACL-2019-Natural}, Web Questions~\cite{Berant-EMNLP-2013-Semantic}, and ARC~\cite{Clark-arxiv-2018-Think}), and a fact extraction dataset, WikiFact~\cite{Goodrich-KDD-2019-Assessing}. 
We also report the zero-shot performance of LLMs on these datasets, {and compute \emph{accuracy} for ARC and \emph{exact match} for other datasets.} 

\textbullet~\emph{Complex reasoning.} 
For complex reasoning, we evaluate the comparison models on OpenbookQA~\cite{Mihaylov-EMNLP-2018-Can}, HellaSwag~\cite{Zellers-acl-2019-HellaSwag}, and SocialIQA~\cite{Sap-arxiv-2019-SocialIQA} for knowledge reasoning; Colored Objects~\cite{Srivastava-arxiv-2022-Beyond} and Penguins in the Table~\cite{Srivastava-arxiv-2022-Beyond} for symbolic reasoning; GSM8k~\cite{Cobbe-arxiv-2021-Training} and MATH~\cite{Hendrycks-ICLR-2021-Measuring} for mathematical reasoning. We compute the \emph{accuracy} for OpenbookQA, HellaSwag, and SocialIQA; \emph{solve rate} for Colored Objects and Penguins in the Table; and \emph{accuracy} for GSM8k and MATH. 
{For knowledge reasoning tasks, we evaluate the zero-shot performance, since they are all QA tasks that can be solved in a zero-shot setting.
For complex symbolic reasoning and mathematical reasoning tasks, we leverage 3-shot in-context exemplars to better elicit LLMs to accomplish them.
Following existing work~\cite{Gao-arxiv-2022-PAL,Wei-arxiv-2022-chain}, we also utilize the chain-of-thought prompting strategy for better solving the mathematical reasoning tasks.}

\textbullet~\emph{Human alignment.} For human alignment, 
we select TruthfulQA~\cite{Lin-ACL-2022-TruthfulQA} to measure whether a LLM is truthful in generating answers to questions, CrowS-Pairs~\cite{Nangia-EMNLP-2020-CrowS} and WinoGender~\cite{Rudinger-NAACL-2018-Gender} to assess the stereotypes in LLMs, RealToxityPrompts~\cite{Gehman-2023-arxiv-RealToxicityPrompts} to evaluate the extent to which LLMs generate toxic language, and HaluEval~\cite{Li-arxiv-2023-HaluEval} to test the ability of LLMs to recognize hallucination.
As the test set of Real-Toxicity-Prompts is too large, we randomly sample 10000 examples from it for evaluation.
We follow LLaMA~\cite{Touvron-arxiv-2023-LLaMA} to report the zero-shot performance, and compute the \emph{accuracy} of identifying a claim as true for TruthfulQA, \emph{accuracy} of recognizing biased sentences (high perplexity) for CrowS-Pairs, \emph{coreference resolution accuracy (he/she/they)} for WinoGender, \emph{toxicity score} for RealToxityPrompts, and \emph{average accuracy} of recognizing hallucinations for HaluEval. 
{For TruthfulQA, we follow existing work~\cite{Touvron-arxiv-2023-LLaMA} that utilizes text-davinci-003 to replace humans for scoring.}
For Crows-Pairs and WinoGender, we follow the experimental settings of LLaMA~\cite{Touvron-arxiv-2023-LLaMA} to compute the perplexity and coreference resolution score.
For RealToxityPrompts, we utilize the  {Perspective-API\footnote{\url{https://perspectiveapi.com/}}} for toxicity evaluation.

\textbullet~\emph{Interaction with environment.} 
To test this ability, we select ALFWorld~\cite{Shridhar-2021-iclr-ALFWorld} and WebShop~\cite{Yao-2022-nips-WebShop} for evaluation, which simulate real-world scenarios such as household and e-commerce environments. 
We follow the setting of ReAct~\cite{Yao-2022-arXiv-react} that evaluate the 1-shot and 2-shot performance of LLMs on WebShop and ALFWorld respectively, and compute \emph{success rate} for ALFWorld and \emph{average score/success rate} for WebShop.
Further, we also follow ReAct~\cite{Yao-2022-arXiv-react} to reduce the length of the input prompt and utilize line break as the EOS token.

\textbullet~\emph{Tool manipulation.} 
For tool manipulation, we consider two kinds of tools including search engine and model interfaces. Therefore, we adopt two tool manipulation benchmarks, \ie HotpotQA~\cite{yang-2018-acl-HotpotQA} and Gorilla~\cite{Patil-2023-arxiv-Gorilla}. HotpotQA requires LLMs to use search engine to retrieve documents from the web, and Gorilla to invoke model APIs from three hubs of TorchHub, TensorHub and HuggingFace. We compute \emph{exact match} for HotpotQA and \emph{accuracy} for Gorilla.
For HotpotQA, we follow ReAct~\cite{Yao-2022-arXiv-react} to report the 3-shot performance.
For Gorilla, we follow the code released by its paper~\cite{Patil-2023-arxiv-Gorilla}, and evaluate the zero-shot performance.

\paratitle{Implementation Details.} For each task and dataset, we evaluate the compared LLMs using the same 
prompts and results parsing method provided by existing work  {(\ie TruthfulQA, HotPotQA, Gorilla, HaluEval)} or designed according to our empirical experience  {(\ie TriviaQA, Natural Questions, Web Questions, ARC, WikiFact, GSM8k, MATH, C-Objects, Penguins, LAMBADA, WMT'22, XSum, HumanEval, CrowS-Pairs, WinoGender, RealToxityPrompt).}
Specifically, all the experiments about closed-source models are based on invoking their official APIs, while for open-source models, we utilize their publicly available code and model parameters, and perform the inference on 8 A800-80G GPUs.
For TriviaQA, OpenbookQA, HellaSwag, and SocialIQA, we experiment on the development set since the test set is not publicly released. While for other datasets, we experiment on the test set. 
To reproduce our experiments, we also publicly release our experimental code and data in \url{https://github.com/RUCAIBox/LLMSurvey/tree/main/Experiments}.

\subsubsection{Results Analysis and Findings}
We report the experimental results in Table~\ref{tab-experimental-res}, and analyze the results in the following.

\paratitle{Analysis of Closed-Source Models.}
We summarize our analysis and findings of  {the four} closed-source models (\ie ChatGPT, Claude, Davinci003 and Davinci002) as follows:

$\bullet$ 
{\emph{These five closed-source models achieve promising results as general-purpose task solvers, in which ChatGPT mostly performs the best.} ChatGPT, Claude, Claude 2, Davinci003 and Davinci002 perform well in most of tasks, including complex tasks (\eg GSM8k), which have shown great potential to be general-purpose task solvers. 
Among them,  ChatGPT exhibits a more superior model capacity on the evaluation tasks, winning the most across all tasks.
In some evaluation tasks, the performance gap between ChatGPT and other closed-source models is very large, especially for  complex tasks \eg 78.47 (ChatGPT) \emph{v.s.} 49.96 (Davinci002) on GSM8k, and 79.88 (ChatGPT) \emph{v.s.}  51.22 (Claude) on HumanEval.
}

$\bullet$ \emph{{Claude 2, ChatGPT and Davinci003} perform better on interaction with environment and tool manipulation tasks.}
On the two evaluation tasks, Claude 2, ChatGPT and Davinci003, perform better than other models by a large margin, \eg 36.40 (Claude 2) \emph{v.s.} 26.00 (Davinci002) on HotpotQA, 44.53 (ChatGPT) \emph{v.s.} 7.74 (Claude) on Gorilla-TF, and 72.58 (Davinci003) \emph{v.s.} 22.04 (Claude) on Gorilla-TH. 
{A possible reason is that these three models have been specially optimized towards these advanced abilities, \eg supporting the use of external plugins.  }

\ignore{
\textcolor{blue}{
$\bullet$ \emph{Davinci002 performs better on symbolic reasoning tasks.} 
{Davinci002 performs well on symbolic reasoning tasks, \eg outperforming ChatGPT on C-Objects and Penguins. 
Compared with Davinci002,  the performance degradation in ChatGPT might be caused by the fact that RLHF (without that in Davinci002) may affect the general abilities (\eg symbolic reasoning) of LLMs to some extent, (\eg alignment tax~\cite{Ouyang-arxiv-2022-Training}), as it mainly focuses on aligning with  human expectations (\eg helpfulness, honesty and harmfulness).} 
}}

$\bullet$ 
\emph{All the comparison models perform not well on very difficult reasoning tasks.} 
On MATH and HotpotQA, all models (including ChatGPT) perform not well. The two tasks are very difficult to solve, requiring accurate understanding of complex mathematical knowledge and performing multi-hop reasoning across documents, respectively.
{Further, these models also have a relatively weak performance on machine translation task (WMT). 
A possible reason is that  WMT also contains many  evaluation examples in minor languages, which might not be well covered in the pre-training data of these LLMs. 
}

\paratitle{Analysis of Open-Source Models.}
Next, we continue to show our analysis and findings about eight open-source models (\ie LLaMA 2-Chat, Vicuna, Alpaca, ChatGLM, LLaMA 2, LLaMA, Pythia and Falcon) as follows: 

$\bullet$ \emph{Instruction-tuned models mostly perform better than the base models.} 
{Among all the compared open-source methods, the instruction-tuned models (\ie LLaMA 2-Chat, Vicuna, Alpaca and ChatGLM) mostly perform better than  non-instruction-tuned models (\ie LLaMA 2, LLaMA, Pythia and Falcon).}
It indicates that instruction tuning is generally capable of improving  the few-shot or zero-shot ability of LLMs in solving various tasks. 
However, after instruction tuning, Vicuna (7B) and Alpaca (7B) suffer from performance degradations on LAMBADA, a language modeling task. 
{The reason may be that the instruction data mainly focuses on enabling LLMs to follow human instructions, which is not always useful  for the general language generation task. }

 
$\bullet$ \emph{These small-sized open-source models perform not well on {mathematical reasoning, interaction with environment, and tool manipulation tasks.}} 
On the tasks of mathematical reasoning, interaction with environment and tool manipulation, all these evaluated open-source models perform not well, including instruction-tuned ones. 
A possible reason is that the instruction data for fine-tuning these models is not specifically designed for these tasks. In addition, these closed-source models may have limited model capacities due to small model sizes.     

$\bullet$ \emph{The top-performing model varies on different human alignment tasks.}
For different human alignment tasks, we can see that these models achieve inconsistent performance rankings.  
For example, LLaMA 2-Chat (7B) performs the best among the compared open-source models on TruthfulQA, while Vicuna (13B) performs the best on CrowS-Pairs. A possible reason is that these tasks are designed with specific purposes for evaluating different aspects of human alignment, and these models exhibit varied performance on different tasks, even for the variants of the same model (\eg Pythia (7B) and Pythia (12B)). More experiments and analysis on human alignment evaluation are needed to reveal more detailed findings.  

$\bullet$ \emph{As a more recently released model, LLaMA 2 (7B) overall achieves a good  performance, especially on complex reasoning tasks.} 
{For complex reasoning tasks, LLaMA 2 (7B) mostly performs better than other base models, \eg 43.95 (LLaMA 2 (7B)) \emph{v.s.} 29.80 (Falcon (7B)) in C-Objects.
For other tasks (\eg language generation and knowledge utilization), LLaMA 2 (7B) can also achieve comparable performance as the best-performing base models. It has used more data for pre-training (\ie about 2 trillion tokens), which mainly contributes to the excellent performance. Furthermore, it also conducts a more robust data cleaning process.}



$\bullet$ \emph{Scaling the open-source modes can improve the performance consistently.} 
{By comparing the performance of Vicuna (7B) and Vicuna (13B), Pythia (7B) and Pythia (13B), we can see that the models with larger scales mostly perform better than smaller ones on these evaluation  tasks, indicating the effectiveness of scaling up the model size. 
Across different tasks, scaling model is more beneficial for more complex tasks (\eg symbolic and mathematical reasoning), where the larger models mostly outperform smaller ones in a large margin.
}


The readers should be note that these findings about open-source language models are  limited to the model sizes. We will continually update this part by including the results of larger versions of these models, and also call for the support of computational resources for more experiments. 

\section{Applications}\label{sec-application}

\begin{figure*}[h]
    \centering
        \includegraphics[width=1\textwidth]{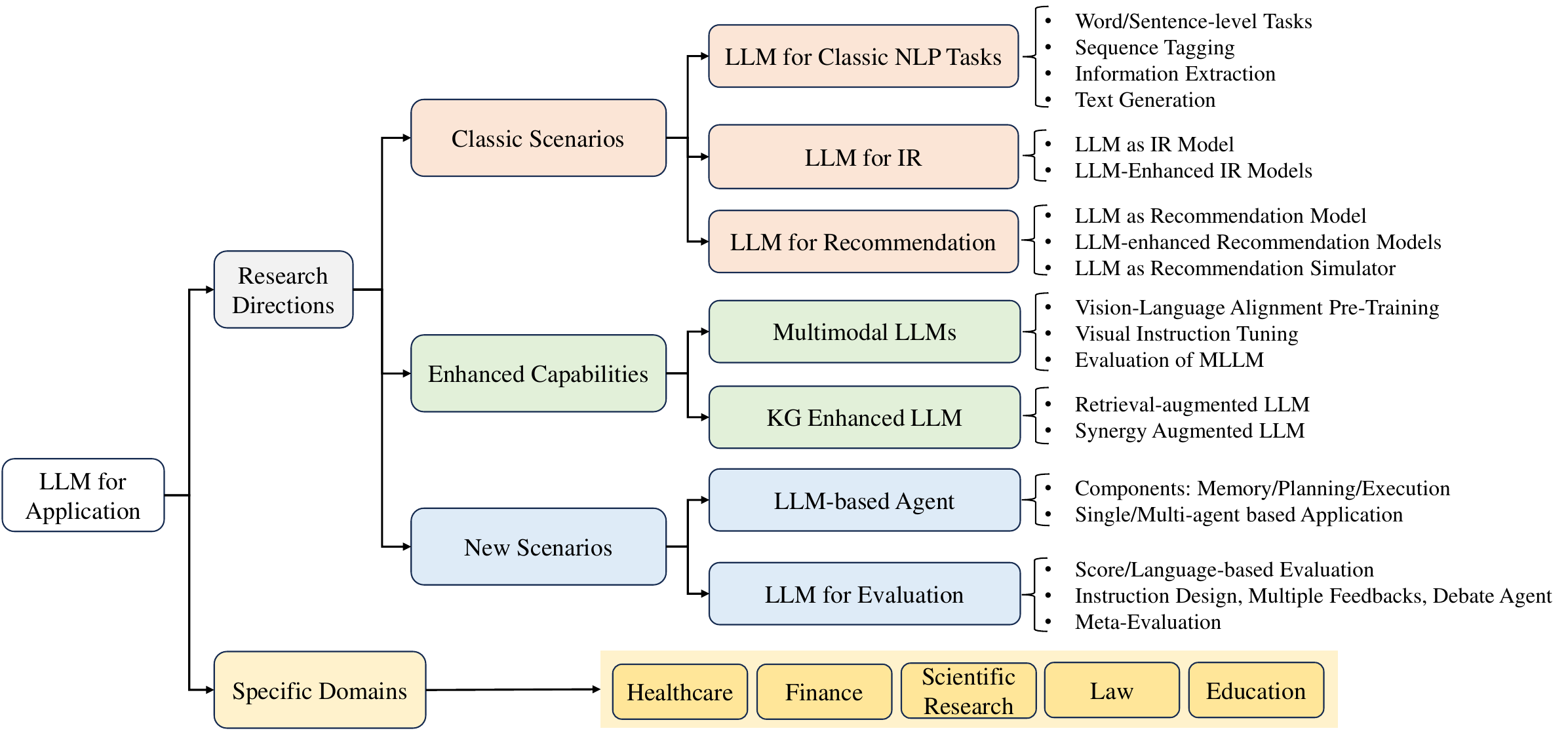}
    \caption{The applications of LLMs in representative research directions and downstream domains. }
    \label{fig:application}
\end{figure*}

{In this section, we briefly review the recent progress on the applications of LLMs in two aspects, namely the impact to research community and representative domains.}
Figure~\ref{fig:application} shows a content organization of this section\footnote{Note that we don't aim to cover all the related research directions or domains, but instead demonstrating the use or impact of LLMs via these selected examples. }. 

\subsection{LLM for Research Community} \label{sec:llm4community}
As LLMs have revolutionized the way how we develop AI algorithms, it poses significant impact on the research community. In this part, we briefly review the advances that led by LLMs for  several representative research directions.  


\subsubsection{LLM for Classic NLP Tasks}
As pre-trained language models (\eg BERT) have originated in the field of NLP, the technical advances of language models has an important impact on the research of NLP. In this part, we discuss the application of LLMs on five kinds of classic NLP tasks, including word-level, sentence-level, sequence tagging, relation extraction, and text generation tasks, which had been the foundation of many existing NLP systems and applications.   
Note that we do not intend to comprehensively cover all NLP tasks, but instead try to analyze the impact of LLMs for fundamental NLP research through the basic tasks.
We also omit the discussion of several tasks (\eg language modeling) that have been discussed early in this survey.

\paratitle{Word/Sentence-level Tasks.}
As long-standing NLP tasks, word-level (\eg word clustering~\cite{Martin-Speech-1998-clustering} and sense disambiguation~\cite{Navigli-ACM-2009-disambiguation}) and sentence-level tasks (sentence matching~\cite{Gomaa-International-2013-similarity} and sentiment classification~\cite{Minaee-ACM-2021-classification}) have been widely studied in the literature and applied in real-world platforms. 
To solve these tasks, the key is to accurately understand the semantic information about the words or sentences.
As rich high-quality labeled data about these tasks has been accumulated so far, existing work~\cite{qiu-CoRR-2020-PTM,Devlin-NAACL-2019-BERT} finds that small language models can achieve very good performance by fine-tuning on it.
Recent studies~\cite{Brown-NeurIPS-2020-Language,Alex-NIPS-2021-RAFT} have also tested the performance of LLMs on these tasks, showing that LLMs can also perform well via in-context learning (with very few examples).   
Whereas,  {as small models can be specially optimized on these tasks to learn the specific task requirement  and domain knowledge}, full-data fine-tuned small models can mostly outperform LLMs using in-context learning on several classic tasks~\cite{Qin-arxiv-2023-Is,Chen-arxiv-2023-Robust}, \eg semantic matching and sentiment analysis. 


\paratitle{Sequence Tagging.}
The sequence tagging tasks, \eg named entity recognition~(NER)~\cite{Nadeau-Lingvisticae-2007-NER} and part-of-speech~(POS) tagging~\cite{Ratnaparkhi-EMNLP-1996-maximum}, are also fundamental tasks. 
Typically, such tasks require assigning each token in the input sequence a proper semantic category label, \eg the classic B-I-O (\emph{Beginning}, \emph{Inside} and \emph{Outside}) tagging scheme for NER tasks.
In the era of deep learning, early efforts~\cite{Yadav-COLING-2018-survey,Souza-arxiv-2019-portuguese} mainly integrate the learned sequence representations (\eg using CNN, LSTM, and BERT) into the classic conditional random field model~(CRF), which performs the tagging task based on structural  prediction.
Recently, researchers have tested the performance of  LLMs in sequence tagging tasks, but observed that LLMs still face challenges in solving them using in-context learning~\cite{Qin-arxiv-2023-Is}, especially for special categories with ambiguous or rare names, \eg the ``MISC'' (\emph{miscellaneous entity}) and ``ORG'' (\emph{organization}) classes.
A possible reason is that LLMs may misunderstand the meanings of these classes in the human-annotated dataset, making it difficult to accurately  understand their semantics according to the instruction and limited examples in the context.

\paratitle{Information Extraction.}
The information extraction task focuses on automatically extracting useful structured information from unstructured text data, such as relation extraction~\cite{Pawar-arxiv-2017-relation} and event extraction~\cite{walker-ldc-2006-ace}, which is also a crucial task relating to many NLP applications.
Typically, previous studies formulate this task as a text classification task or a sequential labeling task.
{As information extraction often needs to accurately understand and process complex semantic relations  (multiple relations within one sentence),  in-context learning with LLMs typically underperform state-of-the-art full-data fine-tuning methods~\cite{gao-arxiv-2023-exploring,ma-arxiv-2023-Large}.}
Whereas, it is shown that enabling collaboration between LLMs and small models can further boost the performance of specific tasks~\cite{tang-arxiv-2023-does,ma-arxiv-2023-Large}. 
In addition, a recent study~\cite{wei-arxiv-2023-zero} also reveals that LLMs can achieve competitive zero-shot performance for information extraction with a two-stage workflow, making this approach attractive in future applications.

\paratitle{Text Generation.}
{Text generation tasks, \eg machine translation~\cite{Bahdanau-ICLR-2015-Neural} and automatic summarization~\cite{Nallapati-acl-2016-Abstractive}, are long-standing NLP tasks that have been widely studied, and there have been a number of deployed products and systems based on fine-tuned small models~\cite{Wu-arxiv-2016-Google,vaswani-CAMT-2018-tensor2tensor}.
Since the pre-training of LLMs is established on text prediction, they exhibit strong language generation abilities  as commercial products~\cite{Jiao-arxiv-2023-mt} and humans~\cite{Zhang-2023-arxiv-Benchmarking}, with the help of proper prompts~\cite{zhang-arxiv-2023-prompting,ghazvininejad-arxiv-2023-dictionary}. 
Additionally, LLMs are flexible to effectively handle special requirement in real-world application scenarios, \eg document-level translation~\cite{wang-arxiv-2023-document}, and also enable natural language interaction with users to further improve the generation quality~\cite{jiao-arxiv-2023-parrot}.
Despite the above success, recent work also reveals that LLMs are hard to well address the generation tasks about low-resource languages and domains, \eg Marathi-to-English translation~\cite{yang-arxiv-2023-bigtrans}, due to their unbalanced training data across different languages.
}

\paratitle{Summary}. 
{Based on the above discussion, we summarize the suggestions, and future direction about the use of LLMs in classic NLP tasks as follows:}

$\bullet$ \emph{Suggestions:}
LLMs and small models have their own merits in different aspects: LLMs are can provide unified solutions to various NLP tasks and achieve competitive performance (especially in the zero/few-shot setting), while small models are economical to develop and can be specially tuned according to target tasks, which can achieve good performance with sufficient high-quality labeled data~\cite{Qin-arxiv-2023-Is,Chen-arxiv-2023-Robust,Kocon-arxiv-2023-ChatGPT,Zhong-arxiv-2023-Can}.  
In applications, one can make suitable choices based on the actual needs, comprehensively considering flexibility, data availability, training compute, and efficiency.

$\bullet$ \emph{Future direction:} 
{
{Despite the excellent general capacities, LLMs still cannot effectively process the NLP tasks in low-resource domains, \eg minor language translation. 
To tackle such tasks, it needs to develop effective approaches to injecting necessary task information or domain-specific knowledge into LLMs, either through fine-tuning or prompting. 
In addition, it is still challenging for LLMs to handle complex semantic relations in classic NLP tasks (\eg nested entity extraction), which is worth more exploration from the underlying working mechanism of LLMs. 
}
It is also promising to combine LLMs and fine-tuned small language models for complementing with each other in solving complex cases of classic NLP tasks~\cite{Cheng-arxiv-2023-UPRISE}. 
Another promising direction is to conduct human-machine collaborative research (\eg conversational translation~\cite{jiao-arxiv-2023-parrot}) on NLP tasks, since LLMs can effectively understand human instructions and make meaningful responses. 
}

\subsubsection{LLM for Information Retrieval}
{
The goal of information retrieval~(IR) systems is to assist users in discovering ideal information resources~(typically documents) and mitigating the information overload issue.
Typically, contemporary IR systems adopt a  retrieve-then-rerank pipeline framework~\cite{Zhao-arxiv-2022-Dense}. Within this framework, the retriever initially retrieves relevant information from a large-scale corpus, and the reranker subsequently performs multi-stage  ranking procedure to acquire the most relevant information~\cite{Ren-EMNLP-2021-rocketqav2}.
Since the advent of LLMs has significant impact on the way of information access, we discuss how it advances the development of IR from two main aspects, namely LLMs as IR models and LLM-enhanced IR models.   
}

\paratitle{LLMs as IR Models.}
{Existing IR models can be overall categorized into \emph{sparse models} (relying on term-based lexical similarity) and \emph{dense models} (relying on embedding based semantic similarity)~\cite{Zhao-arxiv-2022-Dense}.  Specially, dense models are mainly implemented by fine-tuned PLMs (\eg BERT).   
Compared to PLMs, LLMs have more strong model capacities in capturing text semantics, thus having the potential to improve existing dense IR models. 
However, due to the high overhead of LLMs, 
the majority of studies concentrate on employing LLMs as rerankers, aiming to refine the ranking of retrieved candidates. 
To achieve this, recent efforts often formulate special instructions that enable LLMs to perform reranking on a small set of provided candidate documents. Typically, such an approach does not necessitate model training, and achieve promising results compared with well-trained reranking methods~\cite{sun-arxiv-2023-chatgpt, qin-arxiv-2023-large}. 
Specially, the LLM-based reranking approach can be implemented in different ways by zero-shot or few-shot instruction, including  pointwise (\emph{estimating the relevance scores for query-document pairs})~\cite{Cho-ACL-2023-Discrete}, pairwise (\emph{determining the  relevance order of two documents})~\cite{qin-arxiv-2023-large}, or listwise ranking (\emph{sorting a subset of candidate documents})~\cite{tang-arxiv-2023-found}.  
{The essence of these methods lies in the special design of  instructions for text reranking, such as sliding window strategy for document lists~\cite{sun-arxiv-2023-chatgpt, ma-arxiv-2023-zero}, setwise selection prompting~\cite{zhuang-arxiv-2023-setwise}, fine-grained relevance labels incorporation~\cite{zhuang-arxiv-2023-beyond}, and pairwise comparison prompting~\cite{qin-arxiv-2023-large}.}
In addition, recent efforts employ LLMs to generate intermediate texts (\eg URLs) as retrieval results using few-shot demonstrations~\cite{ziems-arxiv-2023-large}.
To further enhance the model performance, LLMs can be specially fine-tuned as  backbones for reranking~\cite{Ma-arxiv-2023-fine, pradeep-arxiv-2023-rankvicuna} or retrieval~(including dense retrieval~\cite{Zhao-arxiv-2022-Dense} and model-based retrieval~\cite{Tay-NIPS-2022-transformer, Ren-ACL-2023-tome}), similar to the fine-tuning process for traditional PLM-based IR models~\cite{Ma-arxiv-2023-fine}. However, fine-tuning LLMs as IR models entails considerable expenses given the huge parameter scale of LLMs.
}


\paratitle{LLM-Enhanced IR Models.}
{As another major research direction, LLMs can be employed to improve existing IR models (\eg small models).  
A common challenge faced by existing IR models is the lack of relevant judgment annotation~\cite{Qu-NAACL-2021-rocketqa, Ren-ACL-2021-PAIR}. 
To tackle this problem, LLMs can be instructed to annotate positive or negative documents for a given query~\cite{peng-arxiv-2023-soft}, or to generate corresponding queries based on a set of documents in the corpus by referring to a few demonstrations~\cite{Dai-ICLR-2023-promptagator, askari-arxiv-2023-generating}.
In addition to training data augmentation, LLM has the potential to improve existing IR models by refining the search-oriented informativeness of both queries and documents.   
In IR systems, the input queries may be constrained by a user's cognitive and cultural competency, making it challenging to accurately  express the real intent, and irrelevant content  present in documents can also impact the relevance evaluation with the query.
As a solution, LLM can be utilized to rewrite the query for enhancing the understanding of the query intent and incorporating additional knowledge into the query through well-designed instructions. The rewritten query can take the form of an improved version of the original query~\cite{mao-arxiv-2023-large}, {a document in the corpus that related to the query~\cite{Gao-ACL-2023-precise}, or an expansion of the query that concatenated with a pseudo generated document~\cite{Wang-arxiv-2023-query2doc}.} 
In addition, documents can also be expanded with queries that are generated based on the original documents using LLMs for context extension~\cite{ma-arxiv-2023-pre}.}

\paratitle{Remaining Issues.} 
{
In this part, we further discuss several important issues to apply LLMs to improve IR  systems. 
First, though LLMs are capable of being as general-purpose task solvers, they are not directly well suited for existing IR systems: they require high overhead for inference~\cite{sun-arxiv-2023-chatgpt, Ma-arxiv-2023-fine}, have limitations in modeling long texts or document lists~\cite{ma-arxiv-2023-zero}, and need special adaptation (\eg instruction tuning) to perform the text ranking task~\cite{sun-arxiv-2023-instruction}.  
Therefore, more systematic approaches to adapt LLMs for modern IR systems should be investigated, to leverage their benefits and meanwhile overcome these limitations. 
Secondly, the advent of LLMs sheds lights on the development of new information seeking ways (\eg New Bing). 
It is meaningful to explore how to reshape the architecture and paradigm of IR by integrating the LLMs' capacities and the merits of existing IR systems~\cite{wang-arxiv-2023-largesearchmodel}. 
Thirdly, existing work mainly focuses on text retrieval tasks, lacking a comprehensive consideration of multimodal information sources. 
As will be discussed in Section~\ref{sec-MLLM}, multimodal large language models~\cite{Li-arXiv-2023-Multimodal} 
are also widely studied, making it feasible to develop more powerful multimedia retrieval systems. 
}


\subsubsection{LLM for Recommender Systems} 
{
Unlike IR systems that analyze user search queries to retrieve relevant documents, recommender systems (RS) aim to capture the underlying user preference and provide appropriate information resources to users~\cite{zhao-cikm-2021-recbole, zhou-cikm-2021-s3rec, Zhao-cikm-2022-recbole-2, Xu-sigir-2023-towards}. 
Typically, existing studies train a recommendation model (either classic or deep learning model) by fitting it over the user's logged data (\eg click data)~\cite{Rendle-arxiv-2012-bpr, Kang-ICDM-2018-Self}.  
However, these models often suffer from a series of technical issues, \eg cold-start recommendation, domain transfer, and poor explainability. 
Recently, LLMs have demonstrated the potential to alleviate these issues of recommendation models~\cite{Zhang-2023-arxiv-recommendation, fan-arxiv-2023-recommender, wu-arixv-2023-a}, due to the strong  capacities of domain generalization and language generation.   
In this part, we briefly
review the recent progress of LLMs in recommender systems, from the following three aspects, namely LLMs as recommendation models, LLM-enhanced recommendation models, and LLMs as recommendation simulators. 
}

\paratitle{LLMs as Recommendation Models.}
{
 With specific methods or mechanisms, LLMs can be adapted to serve as recommendation models.  
Existing work along this line can be generally divided into two main categories. First, some methods prompt LLMs for completing the recommendation task in a zero-shot paradigm (\ie without parameter tuning)~\cite{Gao-arxiv-2023-chat-rec, dai-recsys-2023-uncovering}. A series of prompt engineering methods like recency-focused and in-context learning are introduced to improve recommendation performance as well as alleviate the potential model biases~\cite{hou-arxiv-2023-large, Liu-arxiv-2023-is}.
Second, another category of studies aim to specialize LLMs for personalized recommendation through instruction tuning~\cite{Zhang-2023-arxiv-recommendation,bao-recsys-2023-tallrec}. Specially, high-quality instruction data is key to adapt LLMs to the recommendation tasks, which can be constructed based on user-item interactions with heuristic templates. To further improve the instruction diversity, InstructRec~\cite{Zhang-2023-arxiv-recommendation} employs self-instruct technique to simulate large amounts of potential user instructions in various scenarios like product search and personalized recommendations. In addition to representing each item by its text description, there is also growing attention on extending LLM's vocabulary  with semantic identifiers in recommender systems~\cite{Zhu-arxiv-2023-Collaborative,Zheng-2023-arxiv-Adapting}, to incorporate collaborative semantics into LLMs.}

\paratitle{LLM-enhanced  Recommendation Models.}
{
In addition to instructing LLMs to directly provide recommendations, researchers also propose leveraging the universal knowledge encoded in LLMs to improve traditional recommender systems. 
Existing approaches in this line can be divided into three main categories. The first category employs LLMs to infer users' potential intention from their historical interaction data. Furthermore, traditional recommendation/search models employ the inferred intentions to improve the retrieval of  relevant items~\cite{xi-arxiv-2023-towards, liu-arxiv-2023-a}. Additionally, several studies explore the use of LLMs as feature encoders. 
They employ LLMs to encode the side information of items and users (\eg item's descriptions and user's reviews), 
thus deriving more informative representations of users and items. These representations are then fed into traditional recommender systems as augmented  input~\cite{li-arxiv-2023-exploring, Wei-arixiv-2023-llmrec}. 
{
As another alternative approach, 
several studies~\cite{Li-arxiv-2023-ctrl, Muhamed-nips-2021-ctr-bert} adopt a distillation-like way to transfer LLM's    capacities (\eg semantic encoding) to improve traditional recommenders (\ie small models). 
Specially, they align the hidden states of LLMs and traditional recommendation models via joint training. After training, since only the enhanced small model will be deployed online,  it can avoid the huge overhead of LLMs in online service.    
}

\paratitle{LLM as Recommendation Simulator.} 
{Inspired by the recent success of autonomous AI agents~\cite{wang-arxiv-2023-a}, LLMs have been also utilized to develop recommendation simulators~\cite{Wang-arxiv-2023-RecAgent, Ie-arxiv-2019-recsim} (exemplified by RecAgent~\cite{Wang-arxiv-2023-RecAgent}), showing  great potential to simulate user real behaviors in recommender systems~\cite{Wang-arxiv-2023-RecAgent, Zhang-arxiv-2023-AgentCF, zhang-arxiv-2023-on}.   
Specifically, to make  personalized simulation, an agent will be equipped with a profiling module that encompasses relevant identity information.  Then, a memory module is introduced to store agents' past interaction experiences.  
During the process of simulation, agents are further prompted to conduct self-reflection based on their past experiences, to capture their underlying user preference.  
Most of existing recommendation simulators are conducted in a user-oriented way, without explicitly modeling the items in the interaction process.  
To address this, AgentCF~\cite{Zhang-arxiv-2023-AgentCF} models both users and items as agents,  and further facilitates collaborative reflections to simulate user-item interactions, so as to capturing the two-sided relations between users and items.}


\paratitle{Remaining Issues.} 
{
Despite these efforts, there are still several challenges to address when applying LLMs in recommender systems. First, existing studies have shown that LLM-based  recommendation models in zero/few-shot settings tend to perform worse than traditional ID-based recommenders~\cite{hou-arxiv-2023-large, dai-recsys-2023-uncovering}. This indicates that LLMs might lack an understanding of personalized user behaviors and domain-specific collaborative semantics. Although instruction tuning  alleviates this issue to some extent~\cite{bao-recsys-2023-tallrec, Zhang-2023-arxiv-recommendation}, it can't fully reduce the semantic gap between LLMs and recommender systems, and also suffers from high tuning costs.  
Furthermore, recommender systems prioritize minimizing inference latency to enhance users' experience in low-resourced environments (\eg phones), which poses a challenge to LLMs'  inference speed as well as memory overhead. Therefore, it is important to explore  improvement  techniques, such as efficient tuning and quantization methods, to deploy LLMs efficiently and effectively in real-world recommender systems. In addition, existing LLMs
have limited capacities in long context modeling, make it difficult to process the huge amount of user-item interaction data. Improved context length  extension and context information utilization approaches should be developed to improve the modeling capacities of LLMs in long interaction sequences. 
}


\subsubsection{Multimodal Large Language Model}\label{sec-MLLM}
{In existing literature~\cite{Du-arxiv-2023-survey,Gan-Foundation-2022-vision}, multimodal models mainly refer to the models that can process and integrate information of various  modalities (\eg text, image, and audio) from input, and further produce corresponding output in certain modalities. In this part, we mainly focus on the multimodal extension of LLMs by enabling the information modeling of non-textual modalities, especially the vision modality, called \emph{multimodal large language models~(MLLMs)}~\cite{Li-arXiv-2023-Multimodal}\footnote{In existing work, large vision language models~(LVLMs)~\cite{Li-arxiv-2023-Evaluating} are also used to term such bimodal models that are developed based on LLMs. We use the naming of MLLMs in this part due to its wide use in existing literature. }. 
To start our discussion, we specify the input to be text-image pairs and the output to be text responses. Similar discussions can be made for other modalities, \eg language-audio  models~\cite{Rubenstein-2023-arxiv-audiopalm}, which is beyond our scope here.  
In essence, MLLMs are developed by adapting the information from other modalities to the text modality, so as to leverage the excellent model capacities of LLMs that are learned based on world text. Typically, a MLLM comprises an image encoder for image encoding and a LLM for text generation, associated by a connection module that aligns vision and language representations. During generation, the image is first split into patches, and then transformed into patch embeddings by the image encoder and the connection module, to derive a visual representation that can be understood by the LLM. Subsequently, the patch embeddings and text embeddings are concatenated, and fed into the MLLM, allowing the language model to generate the response autoregressively. In the following, we will discuss the training, evaluation, and key points to develop capable MLLMs.}

\paratitle{Training Process.}
{The training process of the MLLM includes two major stages: vision-language alignment pre-training and visual instruction tuning.} 

\textbullet~\emph{Vision-language alignment pre-training.}
 {To develop MLLMs, existing work mostly initializes the vision encoder and the LLM with pre-trained models~\cite{Alayrac-nips-2022-flamingo, Liu-arxiv-2023-Visual, Zhu-arxiv-2023-MiniGPT-4}. These models retain excellent vision and language capacities, but span different semantic spaces. Thus, the goal of vision-language alignment pre-training (\ie the first-stage training) is to align the vision encoder and the LLM through end-to-end training on large-scale image-text pairs~\cite{Schuhmann-nips-2022-laion5b, Changpinyo-cvpr-2023-conceptual}. However, directly tuning these two models on image-text pairs may cause the degradation of the original representation capacities. 
 To improve the alignment performance, it is crucial to design effective training strategies and select appropriate pre-training data~\cite{Ye-arxiv-2023-mplug, Bai-arxiv-2023-qwen}. Existing work mainly employs the following strategies for cross-modality alignment: (1) if the number of image-text pairs is not sufficiently large~(\eg less than 1M), it is often suggested to only update the connection module~\cite{Liu-arxiv-2023-improved}; (2) if the training data includes high-quality text corpora~\cite{Zhang-arxiv-2023-internlm} or image-text pairs with fine-grained annotations~\cite{Chen-arxiv-2023-shikra},  fine-tuning the LLM can be conducted to boost the performance;  (3) if the number of image-text pairs is very large~(\eg about 1B), fine-tuning the vision encoder is also plausible~\cite{Ye-arxiv-2023-mplug,Bai-arxiv-2023-qwen}, but the benefit remains further verification.} 

\textbullet~\emph{Visual instruction tuning.}  
{After vision-language pre-training, the second-stage training, \ie visual instruction tuning, aims to improve the instruction-following and task-solving abilities of MLLMs. Generally, the input of visual instruction tuning consists of an image and a task description, and the task is to generate a corresponding text output. To boost the performance, high-quality visual instruction data is key to eliciting and enhancing the abilities of MLLMs. Therefore, most studies are dedicated to constructing various visual instruction datasets. As the basic approaches, early studies construct visual instructions by distilling from GPT-4~\cite{Liu-arxiv-2023-Visual} or reformulating vision-language task datasets~\cite{Dai-2023-arxiv-InstructBLIP}. To enhance the quality of instruction data, recent work further proposes improved strategies by increasing the instruction diversity~\cite{liu-arxiv-2023-aligning}, incorporating fine-grained information~(\eg coordinate of objects) into the instruction~\cite{Chen-arxiv-2023-shikra}, or synthesizing complex visual reasoning instructions~\cite{Du-arxiv-2023-What}.  
\ignore{Regarding trainable parameters, the authors in~\cite{Wang-arxiv-2023-what} find that updating the parameters of vision encoder on small-scale instruction tuning dataset results in the performance degradation on semantic understanding tasks. Freezing the vision encoder and fine-tuning the connection module and the LLM~(either full parameter or use LoRA) gives promising results~\cite{Liu-arxiv-2023-improved, Zhang-arxiv-2023-internlm} on the public benchmarks~(\eg SEED-Bench~\cite{Li-arxiv-2023-seed}, MME~\cite{Fu-arxiv-2023-mme}, and MMBench~\cite{Liu-arxiv-2023-mmbench}).}}

\paratitle{Evaluation of MLLM.} {After introducing the approaches to developing MLLMs, we further discuss how to effectively assess the multimodal capabilities of MLLMs from the following three aspects.  
}

\textbullet~\emph{Evaluation perspectives.} 
{The evaluation tasks for MLLMs can be categorized into two main types: \emph{perception} and \emph{cognition} tasks. Specifically, \emph{perception} tasks aim to assess the model's abilities in understanding the basic semantics of the image content, while \emph{cognition} tasks evaluate models with more complex tasks that require reasoning based on perception results. 
The perception ability is typically evaluated through classification tasks about attributes of image (\eg topic and style) and object (\eg existence and color) or OCR-related tasks, based on existing datasets or new datasets derived from existing images with annotations by humans or LLMs~\cite{Gurari-cvpr-2018-vizwiz, Mishra-cvpr-2012-top, Liu-arxiv-2023-mmbench, Fu-arxiv-2023-mme}. 
A notable perception issue is hallucination~\cite{Zhang-arxiv-2023-siren}, where the model's responses contain inconsistent content with the image. 
Among existing studies about hallucination in MLLMs~\cite{liu-arxiv-2023-aligning, Gunjal-arxiv-2023-detecting, Lu-arxiv-2023-evaluation}, object hallucination~\cite{Rohrbach-emnlp-2018-object} has received much research attention.  
To conduct a stable, robust evaluation of object hallucination, POPE~\cite{Li-emnlp-2023-evaluating} proposes a polling-based object probing approach for converting object recognition into a series of binary questions, and the results indicate that current MLLMs often struggle with object hallucination.
\emph{Cognition} tasks, on the other hand, require MLLMs to perform reasoning based on image perception. A common reasoning task is visual question answering (VQA), where models answer questions about images that demand reasoning about spatial relationships~\cite{Hudson-cvpr-2019-gqa}, general knowledge~\cite{Lu-nips-2022-learn}, or scene text~\cite{Amanpreet-cvpr-2019-textvqa}. To fully explore the capabilities of MLLMs, HallusionBench~\cite{Liu-arxiv-2023-hallusionbench} collects 200 sophisticated visual dependent or supplement questions, on which even the most advanced MLLMs like LLaVA-1.5~\cite{Liu-arxiv-2023-improved} and GPT-4V~\cite{OpenAI-OpenAI-2023-GPT-4v} fail to achieve good performance.}

\textbullet~\emph{Evaluation paradigms.}{
The responses of MLLMs can be evaluated either in a closed-ended or an open-ended manner. 
Traditional multimodal tasks often rely on a closed-ended evaluation framework, where the assessment is based on the exact match between the model's response and the ground-truth answer. Examples include the VQA score~\cite{Antol-iccv-2015-vqa} for visual question answering tasks and the CIDEr~\cite{Vedantam-cvpr-2015-cider} score for captioning tasks. However, MLLMs generate responses in an open-ended way, which may contain the correct answer but not exactly match the ground-truth perfectly.
This discrepancy can lead to the underestimation of the model's performance in previous evaluation paradigms. To address this issue, recent approaches have incorporated humans or LLMs as evaluators~\cite {Ye-arxiv-2023-mplug}. For instance, MMBench~\cite{Liu-arxiv-2023-mmbench} employs ChatGPT to align the model responses with the most relevant option in a set of multiple-choice questions. Similarly, LLaVA~\cite{Liu-2023-arxiv-Visual} utilizes GPT-4 for evaluating MLLMs' output, where GPT-4 takes the generated image captions and object bounding boxes as visual inputs for assessment. Such open-ended evaluation methods can improve assessment accuracy while incurring higher costs due to the involvement of humans or LLMs.}

\textbullet~\emph{Evaluation benchmarks.}  {To facilitate a more thorough evaluation of MLLMs, various benchmarks have been developed. Part of them collect existing vision-language tasks for comprehensive evaluation. For instance, LVLM-eHub~\cite{Xu-arxiv-2023-lvlm} aggregates 47 existing text-related visual tasks to assess six distinct capabilities of MLLMs, and Reform-Eval~\cite{Li-arxiv-2023-reform} takes this a step further by standardizing questions from existing benchmarks into a uniform format and discusses how the backbone models influence MLLMs' performance. In addition to incorporating existing tasks, several work also derives new questions annotated by humans or with the help of LLMs. MME~\cite{Fu-arxiv-2023-mme} creates a dataset by pairing images from public sources with manually-collected text instructions for perception and cognition evaluations. MMBench~\cite{Liu-arxiv-2023-mmbench} transforms these instructions into multiple-choice questions and introduces CircularEval to ensure evaluation consistency. SEED-Bench~\cite{Li-arxiv-2023-seed}  further considers temporal understanding tasks and enlarges the evaluation scale to 19K multiple-choice questions with the assistance of LLMs. MM-Vet~\cite{Yu-arxiv-2023-mmvet}  presents more complex tasks to assess the integrated multimodal capabilities of MLLMs. It starts by defining six essential multimodal abilities and then creates intricate questions by combining multiple abilities. In summary, the above benchmarks collectively contribute to the comprehensive evaluation and improved  development of MLLMs.}

\paratitle{Key Points for Improving MLLMs.}
To develop capable MLLMs, we continue to discuss three key points to improve the model capacities, from the perspectives of instruction data, training strategy, and safety and alignment.  

\textbullet~\emph{Visual instruction data}. 
{
Extensive work~\cite{Liu-arxiv-2023-improved,Wang-arxiv-2023-To} has empirically found that both quantity and quality of visual instructions have an important impact on model performance of MLLMs. One basic way to construct visual instructions is to leverage the exceptional capability of LLMs to synthesize instructions  based on text descriptions of images~\cite{Liu-2023-arxiv-Visual}. 
To further enhance the quality of instructions, one can 
construct fine-grained visual instructions with the help of human annotation~\cite{Chen-arxiv-2023-shikra,Zhang-arxiv-2023-LLaVAR} or synthesize more complex data through carefully-designed prompts~\cite{Du-arxiv-2023-What}. 
Despite the effectiveness of the above LLM-based approaches, 
one primary question emerges as to whether a LLM (\ie text generation model without training on any images) possesses the ability to generate sufficiently good visual instructions solely based on verbalized visual information~(\eg captions and coordinates). Specially, existing work has also revealed that visual instructions generated by LLMs sometimes contain misinterpretations about the visual information, \eg object  hallucination~\cite{Li-emnlp-2023-evaluating}.  Therefore, it is crucial to design effective verification methods to control the quality of instruction data generated by LLMs~\cite{Du-arxiv-2023-What}. Furthermore, it still needs   more investigation about  what makes good visual instructions and how visual instructions elicit specific  multimodal abilities in  MLLMs. 
}

\textbullet~\emph{Model training.} Different from LLMs, MLLMs are not trained from scratch, but instead developed based on pre-trained language and vision models. Existing work employs a typical two-stage  approach for training MLLMs, \ie vision-language alignment pre-training and visual instruction tuning. 
In essence,  existing MLLMs aim to (1) preserve the inherent capabilities and parametric knowledge of LLMs as possible, and meanwhile (2) effectively adapt to  multimodal tasks by leveraging the pre-trained LLMs and visual encoders.   
 To achieve the above two goals, two typical  training strategies are often employed for visual instruction tuning, either only optimizing the connection module~\cite{Dai-2023-arxiv-InstructBLIP} or fine-tuning  both the connector module and LLM component~\cite{Liu-2023-arxiv-Visual}. 
 As we can see, the former can reserve  the original capacities of LLMs but likely have a weak an adaptation performance, while the latter can fully adapt to multimodal tasks but suffer from the loss of original capacities of LLMs.
 More efforts should be made to investigate how to effectively balance the two aspects, so as to achieving improved multimodal capacities.
 In addition, existing MLLMs are still overly dependent on the capacities of LLMs, which pose the limits on many multimodal tasks (\eg space positioning). 
 It will be meaningful to explore  improved training approaches of language models, so that multimodal information can be also utilized in  this process.     

\textbullet~\emph{Safety and alignment.}
{
Safety and alignment has been widely discussed in LLMs, which aim to regulate the behaviors of models by technical approaches~\cite{Ouyang-arxiv-2022-Training}. This topic is also important to MLLMs. 
Even a highly advanced MLLM~(\eg GPT-4V~\cite{OpenAI-OpenAI-2023-GPT-4v}) can be susceptible to safety issues. For example, GPT-4V might occasionally exhibit   factual inaccuracies and baseless inferences about images. In some cases, it may even generate harmful content targeting specific individuals or groups~\cite{OpenAI-OpenAI-2023-GPT-4v}. Furthermore,  open-sourced MLLMs are also prone to generate hallucinated response~\cite{Li-emnlp-2023-evaluating} and can be easily manipulated to produce harmful content~\cite{Qi-2023-NAML-Visual}. To address the aforementioned issues, some studies collect  specialized visual instructions to mitigate the problem of hallucination~\cite{liu-arxiv-2023-aligning}. Another alternative approach is to train a revision model to rectify hallucinated response generated by MLLMs in a post-hoc way~\cite{Zhou-arxiv-2023-analyzing}. Additionally, aligning MLLMs with RLHF can also assist MLLMs in generating responses with improved factuality~\cite{Sun-arxiv-2023-Aligning}. Despite these efforts, existing alignment techniques for  MLLMs mainly concentrate on  several specific aspects~(\eg hallucination), lacking a  comprehensive consideration of alignment criteria. 
More efforts should be made to promote the research of safety and alignment for MLLMs. 
}
As a promising solution, knowledge graphs (KGs),  which store enormous knowledge in the triple format, \ie \textit{$\langle$ head\_entity, relation, tail\_entity $\rangle$}, can be utilized to enhance the task performance of LLMs by providing precise and necessary knowledge.
Generally, knowledge enhanced approaches can be expanded into other forms of structured data (\eg tables and databases)~\cite{Ruiz-arxiv-2023-SemTab}, while we limit our discussion to the integration of KG for improving LLMs, which are detailed in two aspects, namely retrieval-augmented LLM and synergy-augmented LLM. 
}
\subsubsection{KG-Enhanced LLM}
\label{sec-KGLLM}
Despite the excellent capacities, LLMs often suffer from challenges on knowledge-intensive tasks, such as 
the potential to generate hallucinated content~\cite{Li-arxiv-2023-HaluEval} and the lack of domain-specific knowledge~\cite{Pan-arxiv-2023-Unifying}. 
As a promising solution, knowledge graphs (KGs),  which store enormous knowledge in the triple format, \ie \textit{$\langle$ head\_entity, relation, tail\_entity $\rangle$}, can be utilized to enhance the task performance of LLMs by providing precise and necessary knowledge.
Generally, knowledge enhanced approaches can be expanded into other forms of structured data (\eg tables and databases)~\cite{Ruiz-arxiv-2023-SemTab}, while we limit our discussion to the integration of KG for improving LLMs, which are detailed in two aspects, namely retrieval-augmented LLM and synergy-augmented LLM. 

\paratitle{Retrieval-Augmented LLM.}
{
Due to the huge amount of fact records in a KG, existing work typically adopts a retrieval model to first obtain a relatively small subgraph from KG, and then leverages it to enhance LLMs by enriching the relevant knowledge.
Before the advent of LLMs, the retrieved subgraphs are often supplemented into training data,  injecting knowledge information into PLMs via parameter learning~\cite{ERNIE3, Zhang-ACL-19-ERNIE, Wang-TACL-21-KEPLER}. 
In contrast, to leverage the retrieved knowledge, LLMs mainly incorporate it as part of the prompt, without  parameter update.  
To implement this approach, there are two main technical problems,  \ie how to retrieve relevant knowledge from KGs and how to make better use of  the structured data by LLMs.  
For the first issue (\ie retrieving relevant knowledge), 
a typical approach is to train a small language model (\eg RoBERTa) to identify question-related fact triples~\cite{Zhang-ACL-2022-Subgraph}. 
To further improve the retrieval performance, several studies also propose an iterative reading-then-reasoning framework, enabling the LLM to interact with the KG multiple times and acquire the required knowledge in a more accurate way~\cite{Jiang-2023-arxiv-StructGPT}.
For the second issue (\ie utilizing retrieved knowledge), a straightforward approach is to serialize the retrieved subgraph and craft specific prompts to include it as the input of LLMs~\cite{Xie-EMNLP-2022-UnifiedSKG, Zhou-ICLR-2023-Large}.
However, due to the loss of structured information in knowledge serialization, LLMs cannot fully capture the structural semantics conveyed by original KGs.  
To address this issue, several model-based approaches train a specialized language model (\eg T5) to transform the subgraph into the natural language text~\cite{Ke-ACL-21-JointGT}.
To guarantee the transformation accuracy, it  relies on sufficient training pairs (often unsupervised constructed)~\cite{Agarwal-arxiv-2020-Large} and excellent model  capability~\cite{Chen-EMNLP-2020-KGPT}.  
}


\paratitle{Synergy-Augmented LLM.}
{
To solve complex tasks (\eg multi-hop question answering~\cite{Lan-2021-arxiv-Complex}), it often requires LLMs to query a KG multiple times, following a systematic solution plan. 
We call such a multi-turn interaction approach to enhancing LLM \emph{synergy-augmented LLM}.   
To better synergize the LLM and KG in a complementary manner, recent studies propose to decompose the complex task into multiple sub-goals and iteratively solve each one by leveraging the necessary knowledge from KG~\cite{Jiang-2023-arxiv-StructGPT, Gu-ACL-23-Pangu, Luo-arxiv-23-Reasoning}.
In this process, the LLM can be regarded as an autonomous agent~(detailed in Section~\ref{sec:llm_based_agent}), which automatically generates the plan and executes it through interaction with the KG environment~\cite{Gu-ACL-23-Pangu}. 
{Specially, the mainstream approaches typically start by enumerating the candidates   using the available knowledge information at the current step, and then retrieve the most appropriate candidates for the next step according to the question~\cite{Gu-ACL-23-Pangu, Luo-arxiv-23-Reasoning}.} 
By iterating the above two steps, LLMs can gradually collect relevant evidence~\cite{Gu-ACL-23-Pangu,Luo-arxiv-23-Reasoning}, and finally approach the correct solution.  
{
Despite the effectiveness, enumeration of the candidates over the KG would lead to a vast search space~\cite{Lan-2020-ACL-Query}.} 
To address it, StructGPT~\cite{Jiang-2023-arxiv-StructGPT} proposes a more efficient way to access knowledge information using the specialized interfaces for KGs.
Specifically, it carefully designs the specialized interfaces according to the common data operations on KG (\eg relation extraction and triple extraction), to ensure efficient and accurate data extraction. 
In this way, LLMs can be instructed to better manipulate and process the structural information of KGs, thus achieving improved task performance. 
}

\paratitle{Future Directions.}
{
Besides the above approaches, there are several promising directions for KG-enhanced LLM remaining underexplored. 
{
First, due to the variety of structured data, it is still difficult for LLMs to directly leverage various kinds of knowledge sources, \eg domain-specific KGs. 
Therefore, it is essential to explore the unified way to manipulate and utilize different knowledge sources by LLMs. 
As a potential solution, it is promising to develop effective approaches to help LLMs comprehend and make use of the access interfaces provided by specific knowledge sources to acquire precise knowledge~\cite{Jiang-2023-arxiv-StructGPT}, while more efforts should be made to investigate how to adapt to the data variety in a cost-effective way. 
}
Second, with the evolution of real-world information, the knowledge stored in LLMs may become outdated or incorrect.
It is necessary to explore how to synchronize the updated knowledge into LLMs through a cost-effective manner~\cite{Wang-arxiv-23-easyedit, Yao-arxiv-23-editing}.
Third, it is promising to investigate the use of factual information from KG to align LLMs in generating more faithful content~\cite{Choi-arxiv-23-KCTs, Zhang-arxiv-23-Mitigating}, which can help reduce the hallucination of LLMs.
}

In addition to exploring KG-enhanced LLMs, it is also meaningful to leverage LLMs to improve the tasks on the KG side (\ie LLM4KG)~\cite{Pan-arxiv-2023-Unifying,Zhu-arxiv-23-LLMs}. 
A typical example is that LLMs can help supplement or construct the KG. 
We omit the discussion of this part, since it is beyond our scope. 

\subsubsection{LLM for Evaluation} 
{
While human evaluation can generally offer reliable quality assessment, it is also often hindered by high annotation costs, significant time requirements, and annotation inconsistencies~\cite{Karpinska-arxiv-23-The}.
In contrast, automatic evaluation can be employed  as a scalable alternative to human evaluation. 
Traditional automatic evaluations have relied on reference-based metrics (\eg BLEU and ROUGE). 
Recently, with the emergence of LLMs as general task solvers highlights their potential as automatic evaluators~\cite{Zheng-2023-arxiv-Judging,Wang-2023-arxiv-Large}, making it promising to conduct LLM based evaluation.
In the following part, we will introduce the recent progress on LLM for evaluation, including evaluation formats, methods, meta-evaluation, and the remaining issues. 
}

\paratitle{Evaluation Formats.}
{
Depending on the type of evaluation outcome, the evaluation format can be categorized into \emph{score-based evaluation} and \emph{language-based evaluation}.
Score-based evaluation employs measurable metrics to assign quality scores (\eg ratings or rankings)  for evaluated texts.
A prevalent way is to conduct pairwise comparison, where LLMs are used to determine the partial order relation of candidate texts following specific guidelines~\cite{Peng-23-arxiv-Instruction,Zheng-2023-arxiv-Judging,Wang-2023-arxiv-Large}, which greatly simplifies the evaluation task.  
However, it may face the inefficiency issue when scaling up the number of candidates~\cite{Zheng-2023-arxiv-Judging}.
{When high-quality reference texts are available during evaluation, LLMs can be instructed to score texts under the guidance provided by references~\cite{Zheng-2023-arxiv-Judging,Wang-arxiv-2023-SciBench,Sawada-arxiv-2023-ARB}.} 
On the other hand, language-based evaluation focuses on generating critiques and suggestions, offering qualitative explanation beyond simple quantitative scoring~\cite{Bai-arXiv-2022-Constitutional,Lee-23-arxiv-RLAIF,Wang-23-arxiv-Shepherd,Cui-23-arxiv-UltraFeedback}. 
It is particularly useful for gathering language feedback signals for human alignment tuning~\cite{Bai-arXiv-2022-Constitutional,Lee-23-arxiv-RLAIF}. 
{Furthermore, it can evolve into a multi-turn interaction framework, where LLM-based evaluators provide natural language feedback to existing solutions from task solvers~\cite{Wang-23-arxiv-MINT}.
This framework evaluates the ability of LLMs to leverage language feedback for refining self-generated solutions. 
}

\paratitle{Evaluation Methods.} 
{
{A common method for LLM-based evaluation involves prompting LLMs with specific instructions.
To further improve the quality of LLM-based evaluation, recent work proposes to prompt LLMs with varied contexts to generate diverse evaluation feedback. 
These contexts vary in aspects such as the candidate order~\cite{Zheng-2023-arxiv-Judging,Wang-2023-arxiv-Large}, evaluation perspectives~\cite{Saha-23-arxiv-Branch,Zhang-23-arxiv-Wider} (\eg relevance, clarity, originality), and evaluation explanation~\cite{Wang-2023-arxiv-Large}.
The generated multiple evaluation feedbacks are then aggregated to produce a final evaluation result, which makes the evaluation process less prone to biases from individual feedback and allows for a more thorough evaluation by covering a wider range of evaluation aspects.}
To further improve the quality of the single-model evaluation, recent studies also develop multi-agent collaboration frameworks~\cite{Chan-23-arxiv-ChatEval,Li-23-arxiv-PRD,Zhang-23-arxiv-Wider} or fine-tune LLMs as specified evaluators~\cite{Bai-arXiv-2022-Constitutional,Lee-23-arxiv-RLAIF,Wang-23-arxiv-Shepherd,Cui-23-arxiv-UltraFeedback,Zhu-23-arxiv-Judge}.
In a multi-model collaboration mode, different LLMs evaluate the  candidates by engaging in discussions to align preferences and reach a consensus~\cite{Chan-23-arxiv-ChatEval,Li-23-arxiv-PRD}.
This method helps reduce the potential biases in individual models through the consensus reached by multiple agents.
Another approach to improving single-model evaluation is to specialize LLMs as scores or critics through fine-tuning~\cite{Bai-arXiv-2022-Constitutional,Lee-23-arxiv-RLAIF,Wang-23-arxiv-Shepherd,Cui-23-arxiv-UltraFeedback,Zhu-23-arxiv-Judge}.
 This process involves creating datasets annotated with preferences and feedback from humans or proficient LLMs. These datasets are then used to train evaluation-oriented  models, enabling them to generate pairwise preference or language  feedback.
The specialized LLM evaluators demonstrate competitive performance with fewer parameters~\cite{Wang-23-arxiv-Shepherd,Cui-23-arxiv-UltraFeedback,Zhu-23-arxiv-Judge}.
}

\paratitle{Meta-Evaluation.} 
{
To effectively assess the quality of LLM-based evaluators, meta-evaluation benchmarks have been introduced, for gauging the agreement with human preferences and the fairness of the    evaluations made by LLMs~\cite{Zheng-2023-arxiv-Judging,Wang-2023-arxiv-Large,Zeng-23-arxiv-Evaluating,Zhang-23-arxiv-Wider,Koo-23-arxiv-Benchmarking}.
As a representative benchmark, MT-Bench~\cite{Zheng-2023-arxiv-Judging} evaluates the agreement between LLMs and human judgments, demonstrating that GPT-4 aligns closely with human preferences in no-tie comparisons on 80 multi-turn questions.
In addition, to address potential biases arising from subjective human evaluations, LLMBar~\cite{Zeng-23-arxiv-Evaluating} manually designs outputs that are objectively worse but superficially appealing, which could mislead evaluators.
The evaluation results reveal that even the most advanced LLMs still fall short of human-level evaluation in the challenging setting.
}

\paratitle{Remaining Issues.}
{
As discussed in Section~\ref{sec-langauge-generation}, recent studies demonstrate that LLM-based evaluators expose multiple types of bias,  such as order bias, self-preference bias, and length bias~\cite{Zheng-2023-arxiv-Judging,Wang-2023-arxiv-Large}.
Although some biases can be mitigated through methods like multi-path ensemble or multi-agent collaboration, they remain inherent to LLM-based evaluators. Consequently, addressing these biases intrinsically within the models continues to be an a challenging issue. 
{In addition, recent work has revealed that LLMs may be incapable of  understanding the self-generated content, exhibiting a weaker understanding capacity  compared to their generation capabilities~\cite{West-23-arxiv-The}. 
Even the most advanced LLMs still struggle identifying their reasoning or factual errors without external feedback~\cite{Huang-23-arixv-Large,Stechly-23-arxiv-GPT4}.} 
Consequently, current LLM-based evaluators might not be adequate for evaluating top-tier LLMs or complex tasks. 
This underscores the importance of improvement approaches for LLM-based evaluators, especially for evaluating capable LLMs and complex tasks demanding sophisticated reasoning, planning, and domain-specific knowledge.
}

\subsection{LLM for Specific Domains}
{In this part, we discuss the applications of LLMs on several representative domains, including healthcare, education, law, finance, and scientific research assistance.}

\paratitle{Healthcare} is a vital application field closely related to human life. Ever since the advent of ChatGPT, a number of studies have applied ChatGPT or other LLMs to the medical domain. 
It has been shown that LLMs are capable of handling  %
{a variety of healthcare tasks, \eg biology information extraction~\cite{tang-arxiv-2023-does}, medical advice consultation~\cite{Nov-arxiv-2023-Medical}, mental health analysis~\cite{Yang-arxiv-2023-mental}, and report simplification~\cite{Jeblick-arxiv-2023-Medicine}}. 
{As the major technical approach, researchers typically design specific prompts or instructions to guide LLMs to perform a wide range of medical tasks. }
To further harness the power of LLMs in the healthcare domain, researchers propose to develop healthcare-related  LLMs~\cite{singhal-arxiv-2022-large,Singhal-2023-arxiv-Towards,Yang-arxiv-2023-zhongjing}.
Specifically, the Med-PaLM models~\cite{singhal-arxiv-2022-large,Singhal-2023-arxiv-Towards} achieves  expert-level performance  on the {United States Medical Licensing Examination (USMLE)}, and earns greater approval from physicians in answering consumer's medical questions.
However, LLMs may fabricate medical misinformation~\cite{Jeblick-arxiv-2023-Medicine,Chen-medrxiv-2023-cancer}, \eg misinterpreting  medical terms and suggesting advice inconsistent  with medical guidelines.  In addition, it would also raise privacy concerns to upload the health information of patients~\cite{tang-arxiv-2023-does} into a commercial server that support the LLM.

{
\paratitle{Education}
is also an important application domain where LLMs potentially exert significant influence.  %
Existing work has found that LLMs can achieve student-level performance on standardized tests~\cite{OpenAI-OpenAI-2023-GPT-4} in a variety of  subjects of mathematics (\eg physics, computer science) on both multiple-choice and free-response problems.
In addition, empirical studies have shown that LLMs can  serve as writing or reading assistant  for education~\cite{Malinka-arxiv-2023-Education,Susnjak-arxiv-2022-Education}.
A recent study~\cite{Susnjak-arxiv-2022-Education} reveals that 
ChatGPT is capable of generating logically consistent answers across disciplines, balancing both depth and breadth.
Another quantitative analysis~\cite{Malinka-arxiv-2023-Education} shows that {students utilizing ChatGPT (either keeping or refining the results from LLMs as their own answers) perform better than average students in some courses from the computer security field. 
{Recently, several perspective papers~\cite{Tan-arxiv-2023-towards,Kamalov-2023-arxiv-A} also explore various application scenarios of LLMs in classroom teaching, such as teacher-student collaboration, personalized learning, and assessment automation.}
{However, the application of LLMs in education may lead to a series of practical issues, \eg  plagiarism, potential bias in AI-generated content, overreliance  on LLMs, and inequitable access for non-English speaking individuals~\cite{Kasneci-learning-2023-chatgpt}.}
}

\paratitle{Law}
is a specialized domain that is built on professional domain knowledge. 
{Recently, a number of studies have applied LLMs} to solve various legal tasks, \eg legal document analysis~\cite{Stanek-arxiv-2023-Can}, legal judgment prediction~\cite{Trautmann-arxiv-2022-Legal}, and legal document writing~\cite{Choi-SSRN-2023-Chatgpt}. A recent study~\cite{Nay-arxiv-2022-Law} has found that 
{LLMs exhibit powerful abilities of legal interpretation and reasoning.} 
Moreover, the latest GPT-4 model achieves a top 10\% score in a simulated bar exam compared with human test-takers~\cite{OpenAI-OpenAI-2023-GPT-4}. 
{
To further improve the performance of LLMs in the law domain,  specially designed legal prompt engineering are employed to  yield advanced performance in long legal document comprehension and complex legal reasoning~\cite{Yu-2022-arxiv-Legal,Trautmann-2022-arxiv-Legal}.
To summarize the progress, LLMs can act as helpful assistants to legal profession.  
Despite the progress, the use of LLMs in law raises concerns about legal challenges, including copyright issues~\cite{Tamkin-arxiv-2021-Understanding}, personal information leakage~\cite{Sun-arxiv-2023-A}, or bias and discrimination~\cite{Abid-AIES-2021-Persistent}.
}

\paratitle{Finance}
is an important field where LLMs have promising application prospects. 
LLMs have been employed on various finance related tasks, such as numerical claim detection~\cite{Shah-arxiv-2023-Zero}, financial sentiment analysis~\cite{Araci-arxiv-2023-FinBERT}, financial named entity recognition~\cite{Alvarado-ALTA-2015-Domain}, and financial reasoning~\cite{Son-arxiv-2023-Beyond}.
Despite the competitive zero-shot performance exhibited by general-purpose LLMs in the finance tasks, they still underperform domain-specific PLMs containing million-scale  parameters~\cite{Shah-arxiv-2023-Zero}.
To leverage the scaling effect of LLMs, researchers collect large-scale finance corpora for continually pre-training LLMs (\eg BloombergGPT~\cite{wu-arxiv-2023-bloomberggpt}, XuanYuan 2.0~\cite{zhang-arxiv-2023-xuanyuan}, and FinGPT~\cite{Yang-2023-arxiv-FinGPT}).
BloombergGPT has demonstrated remarkable performance across a diverse range of financial tasks while maintaining competitive performance in general-purpose tasks~\cite{wu-arxiv-2023-bloomberggpt}.
Nevertheless, it is imperative to consider the potential risks in the application of LLMs in finance, as the generation of inaccurate or harmful content by LLMs could have significant adverse implications for financial markets~\cite{wu-arxiv-2023-bloomberggpt}.
Therefore, it needs more strict reviewing and monitoring on the use of LLMs in the financial field. 


\paratitle{Scientific research} is another 
promising field that LLMs can empower the development progress. 
Prior research demonstrates the effectiveness of LLMs in handling knowledge-intensive scientific tasks (\eg PubMedQA~\cite{Jin-emnlp-2019-PubMedQA}, BioASQ~\cite{Anastasia-blog-2022-BioASQ}), especially for LLMs that are trained on scientific-related corpora~\cite{Taylor-arxiv-2022-Galactica,Lewkowycz-arxiv-2022-Solving,Bi-arxiv-2023-OceanGPT}.
Given the excellent general abilities and broad scientific knowledge, LLMs hold significant potential as helpful assistants across various stages of the scientific research pipeline~\cite{Zhang-arxiv-2023-One}. 
First, during the literature survey stage, LLMs can help conduct a comprehensive overview of the  progress in a specific research  field~\cite{Haman-2023-air-Using,Aydn-2022-ssrn-OpenAI}.
Second, during the research idea generation stage, LLMs demonstrate the ability to generate intriguing scientific hypotheses~\cite{Part-2023-arxiv-Can}.
Third, during the data analysis stage, LLMs can be employed to conduct automatic approaches to analyzing the data characteristics,  including data exploration, visualization, and deriving analytical conclusions~\cite{Hasaan-2023-arxiv-ChatGPT,Cheng-2023-arxiv-Is}.
Fourth, during the paper writing stage,  researchers can also benefit from the assistance of LLMs in scientific writing~\cite{Alkaissi-pubmed-2023-Artificial,Azaria-2023-arxiv-ChatGPT}, in which  {LLMs can offer valuable support for scientific writing through diverse means, such as summarizing the existing  content and polishing the writing~\cite{Buruk-2023-arxiv-Academic}. } 
In addition, LLMs can aid in the automated paper review process, encompassing tasks such as error detection, checklist verification, and candidate ranking~\cite{Liu-2023-arxiv-ReviewerGPT}.
Despite these advances, 
there is much room for improving the capacities of LLMs to serve as helpful, trustworthy scientific assistants, to both increase the quality of the generated scientific content  and reduce the harmful hallucinations. 
\ignore{
\textcolor{blue}{However, when it comes to the depth and overall quality of scientific content, scientific content generated by LLMs still lag behind those produced by human researchers~\cite{Ma-arxiv-2023-AI}. (seem to be strange to mention this point, since it just follows paper review)
}
Furthermore, since LLMs suffer from the issue of hallucinations, they may generate mixed or even completely fabricated information, which significantly reduces the trustworthiness of LLMs as research assistants~\cite{Alkaissi-pubmed-2023-Artificial}. 
}

\emph{Summary}.
In addition to the aforementioned work, 
the applications of LLMs have been also discussed in several other domains.  
For instance, in the psychologic domain, some recent work has studied the human-like characteristics of LLMs, such as self-awareness, theory of mind~(ToM), and affective computing~\cite{Kosinski-arxiv-2023-tom,Amin-arxiv-2023-affective}.
In particular, %
{an  empirical evaluation of ToM conducted on two classic false-belief tasks} speculates that LLMs may have ToM-like abilities since the model in the GPT-3.5 series achieves comparable performance with nine-year-old children in ToM task~\cite{Kosinski-arxiv-2023-tom}.
In addition, another line of work has investigated applying LLMs into the software development domain, \eg code suggestion~\cite{Sridhara-2023-arxiv-ChatGPT}, code summarization~\cite{Sun-2023-arxiv-Automatic}, and automated program repair~\cite{Xia-2023-arxiv-Conversational}.
To summarize, to assist humans by LLMs in real-world tasks has become a significant area of research.
However, it also presents challenges. Ensuring the accuracy of LLM-generated content, addressing biases, and maintaining user privacy and data security are crucial considerations when applying LLMs to real-world scenarios.

\section{Advanced Topics}\label{sec-advanced-topics}

In this section, we focus on discussing several advanced topics that have attracted extensive attention in the research community, and these topics are related to challenging technical issues that largely limit LLM's capacity.  Next, we will introduce these issues and discuss how to address them with feasible approaches. 


\subsection{Long Context Modeling}
\label{sec:long_context}

In real-world application scenarios, there are increasing demands for long context modeling capacities of LLMs,  especially for text file processing (\eg information parsing, extraction, and summarization). Many mainstream LLMs have provided support for long context window. To enhance the long context modeling abilities, there are generally two widely used approaches, namely {scaling position embeddings and adapting context window}. Next, we introduce the two approaches in detail.  

\subsubsection{Scaling Position Embeddings}
{
Transformer-based LLMs can learn effective position embeddings within the maximum training length.  When adapting LLMs to language tasks beyond the maximum training length, it is necessary to scale to larger position indices. 
Specially, some position embedding methods have been shown to possess a certain degree of ability to generalize to text beyond the training length,  which is termed as \textit{extrapolation capability}, including  T5 bias~\cite{Raffel-JMLR-2020-Exploring}, ALiBi~\cite{Press-ICLR-2022-Train}, xPos~\cite{Sun-2022-arxiv-Length} and even NoPE~\cite{kazemnejad-arxiv-2023-impact}. 
{However, as one of the mainstream position embedding methods, RoPE exhibits limited extrapolation ability in empirical studies~\cite{Chen-arxiv-2023-Extending}.}  
In the following, we discuss several methods that adapt RoPE to longer texts. 
}

{
$\bullet$~\emph{Direct model fine-tuning.} 
To adapt LLMs to a long context window, a straightforward approach is to directly fine-tune the models on long texts with the target  length. 
The context extension can be scheduled with gradually increased lengths in a multi-stage manner (\eg 2K $\rightarrow$ 8K $\rightarrow$ 32K). 
To conduct effective extension, it often requires specially prepared long text data for training (Section~\ref{subsec-longtextdata}), and data quality plays a critical role in improving LLM's long context capacities~\cite{xiong-arxiv-2023-effective}. 
However, such a direct 
 fine-tuning approach tends to be inherently slow when adapting LLMs for long texts~\cite{Chen-arxiv-2023-Extending}.
}


$\bullet$~\emph{Position interpolation.} This method downscales the position indices within the original context window, to avoid out-of-distribution rotation angles during  pre-training~\cite{Chen-arxiv-2023-Extending, kaiokendev-github-2023-Things}. Specifically, this approach multiplies all position indices by a scaling coefficient $L/L'$ ($L < L'$), where $L$ and $L'$ denote the original and target context window length, respectively. 
Experimental results~\cite{Chen-arxiv-2023-Extending} have shown that this method can extend the context window effectively and efficiently,  compared to the above approach of direct model fine-tuning. However, it is worth noting that this technique may have an adverse impact on the model's performance when handling normal texts within the original context window~\cite{Chen-arxiv-2023-Extending,Dong-arxiv-2023-BAMBOO}.



$\bullet$~\emph{Position truncation.} To mitigate the challenges posed by out-of-distribution rotation angles, another practical approach is to truncate longer relative positions to satisfy the requirement of the maximum training length. ReRoPE and LeakyReRoPE~\cite{su-online-2023-Rerope} introduce a pre-defined window length for truncation, which is smaller than the maximum training length. 
Specifically, position indices within this pre-defined window would be retained, while those indices beyond the window are either truncated to the pre-defined window length or interpolated to align with the maximum training length. {This strategy can preserve the attention mechanism with the neighbor tokens (within the window length), 
} and further enhance the extrapolation capacity. 
{However, this approach needs to compute the attention matrices twice, accommodating additional computational costs.}


{
$\bullet$ \emph{Base modification.} 
Since LLMs are usually trained with a pre-set maximum training length, wavelengths in certain dimensions of RoPE may exceed the training length for longer text~\cite{Peng-arxiv-2023-Yarn}, on which language models may not be sufficiently trained, \ie training data can't cover a complete rotation cycle. Thus, when processing long text, 
 some rotation angles for certain dimensions would never be seen in the training phase~\cite{Liu-arxiv-2023_scaling}. 
Formally, given a fixed rotation angle $t\cdot \theta_i$,  a smaller basis $\theta_i$ allows for a greater distance $t$, \ie enabling the modeling of longer texts~\cite{Peng-arxiv-2023-Yarn, Roziere-arxiv-2023-codellama,xiong-arxiv-2023-effective}.
According to the formula $\theta_i=b^{-2(i-1)/d}$ in Equation~\ref{eq:basis}, decreasing the basis can be achieved by increasing the value of the base. In addition,
decreasing the base can also help re-scale the wavelengths of all dimensions below the training length, while it often needs continual pre-training to adapt the LLMs to long context windows~\cite{Liu-arxiv-2023_scaling}.
A recent study~\cite{Liu-arxiv-2023_scaling} has empirically compared these two base modification methods, and shown that decreasing the base demonstrates  better extrapolation performance, while increasing the base performs better within the training length.
}

$\bullet$ \emph{Basis truncation.} Similar to the base modification, the truncation of the basis also concentrates on dealing with the singular dimensions with wavelengths exceeding the training length~\cite{Pal-arxiv-2023-giraffe}. According to the definition $\lambda_i=2\pi/\theta_i$ in Equation~\ref{eq:wavelength}, the dimension with a large wavelength $\lambda_i$ has a small basis $\theta_i$ accordingly. Based on this observation, this approach first defines a basis range $[a, c]$. 
{
Given the basis range, the value of basis is modified according to the following ways: (1) when $\theta_i\geq  c$, the value is retained, (2) when $\theta_i \leq a$, the value is set to zero, and (3) when $a < \theta_i < c$, the value is truncated to a fixed small value.} 
{Via basis truncation, the out-of-distribution rotation angles can be avoided at larger position indices. However, this approach does not perform very well at long context tasks~\cite{Pal-arxiv-2023-giraffe}.}

\subsubsection{Adapting Context Window}
{Since Transformer-based LLMs have limited context windows, they can not directly integrate or utilize the entire information of the long sequences exceeding the context window. To alleviate the limitation, several methods have been proposed to adapt LLMs to long context, as discussed below.
}


{
$\bullet$ \emph{Parallel context window.} Inspired by fusion-in-decoder~\cite{Izacard-EACL-2021-Leveraging}, parallel context window methods~\cite{Partner-ACL-2023-Parallel,Hao-2022-arxiv-Structured} adopt a divide-and-conquer strategy to process input text. Specially, it divides the input text into multiple segments, each independently encoded with shared position embeddings. At the generation stage, the attention masks are modified to make that subsequent tokens can access to previous tokens in each segment.
Nevertheless, this method cannot distinguish the order of different segments, resulting in a limited model capacity on certain tasks.}

{
$\bullet$ \emph{$\Lambda$-shaped context window.} 
Some prior work has revealed that LLMs tend to allocate greater attention weights to the starting and nearest tokens among all previous tokens~\cite{Beltagy-arxiv-2020-longformer,Xiao-arxiv-2023-Efficient}, and it potentially results in the ``\emph{lost in the middle}'' phenomenon~\cite{Liu-arxiv-2023-Lost}.}
Based on this observation, 
LM-Infinite~\cite{Han-arxiv-2023-LMinfinite} and StreamingLLM~\cite{Xiao-arxiv-2023-Efficient} propose to employ a ``$\Lambda$-shaped'' attention mask, which selectively preserves the initial tokens and the nearest tokens that each query can attend to and then discards any tokens beyond this scope.  
{Experiments demonstrate that this method can facilitate extra-long text generation with a fixed memory~\cite{Xiao-arxiv-2023-Efficient}.}  However, it may struggle to model the long-range dependency in the context window, {since it cannot effectively utilize the information from the discarded tokens~\cite{Xiao-arxiv-2023-Efficient}.}

$\bullet$ \emph{Token selection.}
It has been shown that a relatively small subset of tokens can effectively capture the majority of attention patterns in a Transformer~\cite{Bertsch-arxiv-2023-Unlimiformer}, \eg  {the top-$k$ attention scores} can well approximate the original full attention. Therefore, a number of studies propose different methods to select the most relevant tokens from token-level or block-level memory units for generation. Token-level selection methods store the past keys in external memory and utilize a $k$-NN search method to retrieve the $k$ most relevant tokens for generation~\cite{Wu-ICLR-2022-Memorizing, Bertsch-arxiv-2023-Unlimiformer, Tworkowski-arxiv-2023-Focused}. For a decoder model, 
it typically employs one certain layer to access these top-$k$ external tokens, while still adopting the normal context window in the rest layers~\cite{Wu-ICLR-2022-Memorizing,Tworkowski-arxiv-2023-Focused}. Block-level selection methods~\cite{Lu-arxiv-2024-LongHeads,Xiao-arxiv-2024-InfLLM} first segment the long sequence into blocks with the same length and {represent each block into several key vectors for retrieval}. Then, the most relevant blocks to the query as well as the neighbor and initial blocks will be selected for attention computations. Unlike token-level selection methods, {block-level selection methods typically retrieve different tokens with specific heads.} 

\subsubsection{Long Text Data}\label{subsec-longtextdata}
To further enhance the long context modeling capacity, it typically requires continual pre-training with specially curated long text data. Next, we discuss how to prepare the long text data from the two aspects of quantity and quality. 


{
$\bullet$ \emph{Quantity effect.} Different from the pre-training phase that requires vast amounts of data,  a small amount of long-text data for continual pre-training is sufficient for context window extension~\cite{Chen-arxiv-2023-Extending}. Several studies show that LLMs have obtained the capability of utilizing {distant information via large-scale pre-training data, and thus it only needs to adapt for extended context windows during continual pre-training}~\cite{fu-icml-2024-data}. Typically, it has shown that LLaMA-2-7B or LLaMA-2-13B can achieve a context window length of over 100K tokens and effective context utilization~\cite{fu-icml-2024-data} with the training on several billion tokens. 
However, the ability to handle short text of LLMs may be affected to some extent~\cite{Chen-arxiv-2023-Extending}.

$\bullet$ \emph{Quality effect.} In addition to the quantity, the quality of long text data is essential to  long context modeling for LLMs. For instance, LongWanjuan~\cite{Lv-arxiv-2024-Longwanjuan} categorize long texts into holistic, aggregated, and chaotic long texts based on three metrics, \ie coherence, cohesion, and complexity, and they show that removing chaotic data and keeping coherent and cohesive data are useful to enhance the long text modeling capacities of LLMs. Further, up-sampling cohesive data can lead to further improvement.  
In addition, when preparing long text data,  data mixture should be carefully adjusted for avoiding large distribution drift with the original  pre-training data. 


{In addition to the studies based on vanilla Transformer, there are a surge of Transformer variants with efficient attentions and other efficient architectures, aiming to alleviate the high computational costs for modeling long texts. These studies are discussed in Section~\ref{sec:archs} and Section~\ref{sec:configuration}. 
{Furthermore,  context compression and prompting techniques (\eg iterative reasoning~\cite{Chen-arxiv-2023-Walking}) have also been proven to be a viable strategy for handling long text tasks}~\cite{zhou-arxiv-2023-recurrentgpt,Packer-arxiv-2023-MemGPT,Chen-arxiv-2023-Walking,Xu-arxiv-2023-retrieval}, without the need of model adaption.}

\subsection{LLM-empowered Agent}
\label{sec:llm_based_agent}
{The research on agents in AI aims to develop entities that can perceive the environment, make decisions, and take actions to achieve specific goals~\cite{Russell-Pearson-2020-Artificial}.
However, traditional agents are often limited to heuristic rules or specific environments, which constrain their generalization to open-domain scenarios~\cite{Lake-arxiv-2016-Building}.
Given that LLMs possess excellent capacities in solving complex tasks, they have rapidly emerged as promising solutions for serving as the core computation unit of agents~\cite{wang-arxiv-2023-a}.
In this part, we will first introduce the framework for LLM-based agents, then explore their applications, and finally discuss the future directions.}

\subsubsection{Overall Framework.}
Next, we first detail the key components of an LLM-based agent and then present the typical workflow.

\paratitle{Components.}
{
Typically, there are three main components in an LLM-based agent: \textit{memory}, \textit{planning}\footnote{
Section~\ref{subsec-planning} introduces planning as a utilization approach for LLMs, while in this section, we describe its utilization as a functional component in LLM-based agents.
}, and \textit{execution}.
Specifically, the \textit{memory} component aims to store the information perceived from the environment and can be utilized to support decision-making.
In particular, LLM-based agents usually maintain information in both short-term memory and long-term memory with the operations of reading and writing.
Short-term memory usually refers to the internal context window of LLMs (\ie input), where LLMs can read and write through actions like reasoning~\cite{Yao-arxiv-2022-ReAct}. 
While long-term memory can be mapped to the external storage like vector databases~\cite{Zhong-2023-arxiv-MemoryBank}, where LLMs can read through retrieval and write with reflection~\cite{Shinn-arxiv-2023-Reflexion}.
Specially, profiles are usually implemented with long-term memory, which is an important feature for an agent that specifies its role and function~\cite{wang-arxiv-2023-a}. 
The \textit{planning} component  
{is responsible for generating the action plan} based on the information from the memory component.
In data format, the plan usually takes the form of text-based instructions~\cite{Wang-arXiv-2023-Plan} or code-based programs~\cite{Gao-arxiv-2022-PAL}.
To generate it, LLM-based agents will first propose several candidates and then select a more suitable one among  them~\cite{Wang-arxiv-2022-Self-Consistency}.
The initial plan can be further refined with execution feedback from the environment~\cite{Wang-2023-arXiv-voyager}.
The \textit{execution} component is in charge of carrying out the plan from the planning component, 
{which can be fulfilled by  the internal LLM~\cite{Wang-arXiv-2023-Plan} or external tools~\cite{Yao-arxiv-2022-ReAct}.} 
} 

\paratitle{Workflow.}
{
With the three components mentioned above, a  typical workflow of an LLM-based agent is as follows. 
First, it receives information from the environment and writes it into short-term memory. 
{Then, the agent processes the newly received information in the short-term memory.
Such a process can be enhanced with information retrieved from long-term memory.
Subsequently, the planning component utilizes the processed information from short-term memory to generate the next plan.}
Finally, the execution component carries out the plan generated from the planning component, which can be further assisted with external tools.
By repeating the aforementioned process, the LLM-based agent can autonomously adjust its behavior in response to feedback from the environment and ultimately achieve its goal.
Once LLM-based agents receive user requests or are assigned goals, they follow the above workflow to accomplish tasks through multi-turn interactions with the environment.
}

{
To summarize, in an LLM-based agent, the LLM serves as the core computation unit and is equipped with components including \textit{memory}, \textit{planning}, and \textit{execution}.
These components are integrated in a systematic way under the control of the LLM during interactions with the environment.
For more details, the readers might refer to the comprehensive survey for LLM-based AI agents~\cite{wang-arxiv-2023-a}. 
}

\subsubsection{Applications}
Recently, LLM-based agents have shown great potential in autonomously solving complex tasks, making it feasible to rapidly develop capable applications for specific domains or tasks. In this section, we will discuss the applications in single-agent and multi-agent scenarios. 
\ignore{
With the amazing capabilities of LLM-based agents, they have revolutionized numerous tasks and domains.
In this section, we will discuss the applications in single-agent and multi-agent scenarios.
}

\paratitle{Single-agent based Applications.}
{
Applications based on a single-agent mode mainly aim to develop capable task solvers that can autonomously complete user requests.
A large number of single-agent projects have been developed, which focus on general-purpose task solving. 
As a representative project, AutoGPT~\cite{AutoGPT} empowers LLMs with long/short-term memory management and external tools like search engines.
In order to autonomously address a user request, AutoGPT understands the request with knowledge from its memory and actions like reasoning, decomposes it into a detailed plan, executes the plan step-by-step with the assistance of tools, and refines the rest plan based on feedback from the environment. 
Such an iterative process continues until the user request is successfully resolved.
Other similar projects include GPT-Engineer~\cite{GPT-Engineer} and XAgent~\cite{xagent2023}.
In addition, there is also some work that aims to develop autonomous agents for specific domains, such as WebGPT~\cite{Nakano-arxiv-2021-WebGPT} for the web-browsing environment, ProgPrompt~\cite{Singh-arxiv-2022-ProgPrompt} for the real-life environment, and Voyager~\cite{Wang-arxiv-2023-Voyager} for the Minecraft environment.
}

\paratitle{Multi-agent based Applications.}
{
Different from single-agent systems where agents work independently, multi-agent systems work in collaboration to unleash collective intelligence.
Typically, multiple agents can be instantiated from the same or different LLMs, each with their respective roles and functions.
According to the coordinating strategies among these agents, multi-agent systems can be divided into two categories: cooperation-based and competition-based. 
In the cooperation-based mode, to share information and seek collaborative actions among agents, various communication protocols have been proposed, including {free-form dialogue~\cite{Li-arxiv-2023-CAMEL}, structured document~\cite{Hone-arxiv-2023-MetaGPT}, and data embedding~\cite{Pham-arxiv-2023-Let}.}
Based on the communication protocol, agents can be effectively organized for downstream applications, such as software engineering~\cite{Hone-arxiv-2023-MetaGPT}, user behavior analysis~\cite{Wang-arxiv-2023-RecAgent, Zhang-arxiv-2023-AgentCF}, and society simulation~\cite{Park-arxiv-2023-Generative}.
As a representative project, LangChain\footnote{https://www.langchain.com/} is a framework for developing multi-agent based applications powered by LLMs. It enables users to deploy different roles of LLM-based agents and utilize them to solve tasks via working in collaboration.
In addition, other similar frameworks, such as AgentVerse~\cite{Chen-arxiv-2023-agentverse} and AutoGen~\cite{Wu-arxiv-2023-autogen}, can also be utilized for developing multi-agent collaborative systems. 
In the competition-based mode, debate serves as one of the popular communication protocols to foster divergent thinking and elicit valuable external feedback among agents.
Such a way is beneficial for domains that demand precise decision-making and accurate responses, such as mathematical reasoning~\cite{Du-arxiv-2023-Improving} and evaluation~\cite{Chan-arixiv-2023-ChatEval}.
}

\subsubsection{Discussion}
Despite the huge success, there still remain several technical  challenges that limit the development and application of LLM-based agents. In this part, we discuss the remaining challenges from the perspective of computational burden, human alignment, complex capability extension, and robustness.}

\paratitle{Computational Costs.}
With the ever-increasing capabilities of LLMs~\cite{wang-arxiv-2023-a}, their performance on agent applications demonstrate promising performance. However, it also introduces significant issues in terms of efficiency due to the high computational demands and intricate interaction mechanisms involved. Furthermore, in multi-agent systems with numerous LLM instances, as the number of agents increases, this issue would be  more severe, since the communication network within multi-agent systems also becomes increasingly complex. Therefore, more effective and efficient communication protocols and architectures are essential to support the heightened coordination demands among agents.


\paratitle{{Alignment with Human Sociality.}}
LLM-based agents can be conceptualized as individual entities, with the emergence of sociability resulting from the interaction among these agents. Autonomous agents often assume specific roles such as coders or researchers, making role-playing a vital capability for agents to solve downstream tasks~\cite{Shao-Character-LLM-2023-arxiv}. However, LLMs, typically trained on web corpora, face difficulties in accurately mimicking roles that are infrequently discussed online or are emergent. They also lack self-awareness in conversational scenarios due to inadequate modeling of human cognitive psychology. Thus, it is imperative to develop improved agent technique, including both training methods and architectures, to better align LLMs with human preferences and enhance their role-playing abilities.

\paratitle{Capability Extension.}
LLM-based agents, similar to humans, require advanced capabilities (\eg tool learning) to fulfill complex functions or tasks, which might be beyond their capacity scope. To address this issue, tool use has become a widely-used approach to enhancing LLMs' capacities in various complex tasks. 
For example, when answering informative user questions, they use search engines to retrieve information from the internet. {However, the quality and quantity of existing available tools impose limitations on their accessibility and comprehensiveness. And it would become more difficult for LLM-based agents to use such limited tools when interacting with dynamic and changing environments}.  In addition, as the scale of tools expands, the compatibility and extensibility between the agents and tools must be further improved to facilitate complex task resolution.

\paratitle{Robustness and Trustworthiness.}
The deployment of LLM-based agent systems necessitates robustness and trustworthiness~\cite{Hua-2024-TrustAgent-arxiv}. The system should be resilient against adversarial inputs from various modalities such as text, image, or audio. Incorporating existing techniques like adversarial training, data augmentation, and sample detection to increase sensitivity to aggressive information in the input can fortify the system's security. Concurrently, it is challenging to ensure the credibility of LLM-based agents given the severe hallucination issues inherently rooted in LLMs. While existing methods such as constrained decoding during inference and external knowledge integration can mitigate these issues to some extent~\cite{Huang-2023-A-arxiv}, further exploration of efficient and effective alignment methods is necessary to develop reliable agent systems.

\subsection{Analysis and Optimization for Model Training} 
\label{sec:training-effi}
{In Section~\ref{sec:training_settings}, we have introduced  basic  techniques for training LLMs.  
As the scale of model parameters and data continues to expand, efficiently training larger models with limited computational resources has become a critical technical challenge in the development of LLMs. 
This challenge primarily encompasses two technical issues: firstly, how to optimize memory usage when loading and processing models across GPU clusters, and secondly, how to maintain or improve training efficiency as models scale.
Next, we will conduct quantitative analyses and introduce advanced training techniques addressing the two aforementioned issues.} 

\subsubsection{Estimation of Training Memory Consumption}
In this part, we will first estimate the GPU memory consumption for training LLMs.

\paratitle{Model States Cost.}
Model states often occupy the majority of memory during training, typically consisting of model parameters, gradients, and optimizer states. As introduced in Section~\ref{subsub:scalable}, mixed precision training has been widely utilized in LLM training. 
{For a model containing $P$ parameters, both the model parameters and their gradients are typically stored as 16-bit floating-point numbers, requiring a total storage of $4P$ bytes ($2P$ for the parameters and $2P$ for the gradients). When using optimizers such as Adam~\cite{Kingma-arXiv-2015-Adam} or AdamW~\cite{Loshchilov-ICLR-2019-Decoupled}, an additional set of 32-bit floating-point numbers are needed to store the optimizer states, including the copy of model parameters, gradient momenta, and gradient variances, which leads to a total storage of $12P$ bytes ($4P$ each for each of these components).
Consequently, the total memory required for storing the model states during training is $16P$ bytes. For instance, training LLaMA-7B ($P \approx 6.7 \times 10^9$) requires around 100GB memory to store the model states alone.}

\paratitle{Activations Cost.}
Activations are the intermediate states that require to be stored in the forward pass for gradient computation during backpropagation. For example, for a binary operation $\bm{Y}=\bm{W}\bm{X}$, calculating the gradient $\frac{\partial Y}{\partial W}$ necessitates the input $\bm{X}$, which should be preserved during the forward pass.
In Table~\ref{tab:activation}, we list the estimation of the activation memory consumption for different components within the Transformer model. 
Take LLaMA-7B ($V=32,000,L=32,H=4,096,H'=11,008,N=32$)  as an example, it would take 16GB memory to store activations per device under the setting $B=1,T=2,048$. 
}

\begin{table}[ht]
\centering
\small
\caption{The activation memory consumption of each computation within the LLaMA model based on research work~\cite{Korthikanti-mlsys-2023-reducing}. We denote batch size by $B$, sequence length by $T$, the vocabulary size by $V$, the number of head in the attention module by $N$, the dimension of each head by $D$, the hidden size by $H$ ($H=ND$), and the intermediate size inside FFN by $H'$. Equations \ding{192}-\ding{200} are layer-wise and need to be multiplied by the number of the layers $L$ when computing the total consumption.}
\label{tab:activation}
\begin{tabular}{lp{0.51\columnwidth}}
\toprule
\textbf{Equations}                                                 & \textbf{Activation consumption}                                                                               \\ \midrule
\ding{192} $\bm{Q}, \bm{K}, \bm{V} = \bm{X} \bm{W}^{Q,K,V}$   & store $\bm{X}$ with size $2BTH$                \\
\ding{193} $\bm{Q}, \bm{K} = \mathrm{RoPE}(\bm{Q}, \bm{K})$ & store $\bm{Q}$ and $\bm{K}$ with size $4BTH$\\
\ding{194} $\bm{O} = \operatorname{Attn} \left(\bm{Q}, \bm{K}, \bm{V} \right)$ & store $\bm{Q}$, $\bm{K}$, and $\bm{V}$ with size $6TH$ and results of softmax with size $2BT^2N$ \\
\ding{195} $\bm{X} = \bm{O} \bm{W}^O$                       & store $\bm{O}$ with size $2BTH$                \\
\ding{196} $\bm{X} = \mathrm{Add}\&\mathrm{Norm}(\bm{X})$   & store $\bm{X}$ with size $2BTH$ \\
\ding{197} $\bm{G}, \bm{U} = \bm X [\bm{W}^G, \bm{W}^U]$        & store $\bm{X}$ with size $2BTH$ \\
\ding{198} $\bm{D} = \mathrm{Swish}(\bm{G}) \cdot \bm{U}$    & store $\bm{G}$ and $\bm{U}$ with size $4BTH'$\\
\ding{199} $\bm{X} = \bm{D} \bm{W}^D$                       & store $\bm{D}$ with size $2BTH'$\\
\ding{200} $\bm{X} = \mathrm{Add}\&\mathrm{Norm}(\bm{X})$   & store $\bm{X}$ with size $2BTH$ \\
\ding{201} $\mathrm{CE}(\mathrm{softmax}(\bm{X}\bm{W}^L))$   & store $\bm{X}$ with size $2BTH$ and results of softmax with size $4BTV$ \\

\bottomrule
\end{tabular}
\end{table}

\paratitle{Other Memory Cost.}
In addition to the main factors affecting GPU memory consumption discussed above, the memory usage also includes the following aspects:

$\bullet$ \emph{Deep learning frameworks.}
The PyTorch framework requires approximately 1GB of GPU memory when loading its core functions. This is the essential overhead for the framework to operate.

$\bullet$ \emph{Distributed frameworks.}
When utilizing distributed training frameworks (\eg DeepSpeed), its GPU memory usage can fluctuate between 1GB and 4GB. The exact amount depends on the level of optimization and the hyper-parameter settings. This portion of the memory is primarily used to optimize memory management and communication efficiency during the training process.

$\bullet$ \emph{Intermediate results and memory fragmentation.}
Besides the activations, there also exist intermediate results that will affect the peak memory consumption. Take the computation of the softmax function in Equation~\ding{201} as an example, the implementation of the Transformers library requires an additional $8BTV$ bytes of memory, as it needs to store two additional copies of the 32-bit input ($4BTV$ bytes each). Moreover, during the training process, memory fragmentation occurs due to the non-contiguous allocation and release of memory, typically leading to an additional 0.5GB to 1GB of memory consumption.

\subsubsection{{Memory Optimization Methods}}
Based on the aforementioned analysis, we will next introduce several typical methods for optimizing the memory usage for training LLMs. 

\paratitle{Gradient Checkpointing.}
Gradient checkpointing~\cite{Chen-arxiv-2016-training}, also known as activation recomputation, is a technique used to optimize memory usage during backpropagation. Specifically, the activations need to be retained during the forward pass. However, storing all activation values for each layer requires a significant amount of memory resources (detailed in Table~\ref{tab:activation}). To reduce the memory cost, gradient checkpointing retains only a subset of the activations during the forward pass and recomputes these values during the backward pass to save memory, albeit with additional computational overhead.
In implementation, gradient checkpointing typically involves storing the input of each Transformer layer and recomputing the corresponding activation values during backpropagation.

\paratitle{ZeRO.}
Zero redundancy optimizer (ZeRO)~\cite{Rajbhandari-IEEE-2020-ZeRO} technique, proposed by the DeepSpeed
library, focuses on alleviating the issue of memory redundancy in data parallelism. 
As mentioned in Section~\ref{subsub:scalable}, data parallelism requires each GPU to store the same copy of  the model states, resulting in a memory consumption of $16P$ bytes per GPU. 
A direct side effect of data parallelism is that it memory redundancy issues, since not all of the above data is necessary to be retained on each GPU.
To resolve it, the ZeRO technique aims to retain only a fraction of data on each GPU, while the rest data can be obtained from other GPUs when required.
Specifically, ZeRO provides three strategies, depending on how the three parts of the data are stored, namely optimizer state partitioning (ZeRO-1), gradient partitioning (ZeRO-2), and parameter partitioning (ZeRO-3).
Empirical results indicate that the first two strategies do not increase the communication overhead, and the third solution increases about 50\% communication overhead but saves memory proportional to the number of GPUs.
PyTorch has implemented a similar technique as ZeRO, called fully sharded data parallel (FSDP)~\cite{FairScale2021}.


\paratitle{Offload.}
In GPU-limited environments, DeepSpeed has proposed the offload technique~\cite{Ren-USENIX-2021-ZeRO}, which can significantly reduce the GPU memory required for training by offloading part of the model states and computational overhead to CPU memory. {Specifically, gradients and optimizer states would be offloaded to CPU memory, with only the model parameters kept on GPU.} The computationally intensive forward and backward propagation still need to be performed on GPU to ensure efficiency, while {parameter update, which requires relatively fewer computations,} are executed on CPU to reduce GPU memory overhead.
Furthermore, Infinity~\cite{Rajbhandari-sc-2021-zero} allows training models that exceed the GPU memory limits by utilizing high-speed disk storage (\eg NVMe).


\subsubsection{{Efficiency Optimization Methods}} \label{subsub:training-effi}
{In addition to memory-saving techniques, it is also crucial to maintain computational throughput as the model scales. 
}
In what follows, we will describe two representative efficiency optimization methods.

\paratitle{FlashAttention.}
FlashAttention~\cite{Dao-nips-2022-flashattention,Dao-2023-arxiv-flashattention2} is an optimization method for the attention mechanism that significantly reduces the memory transfer during attention computation. The core idea is to minimize the storage of intermediate results and directly obtain the final result. 
According to the attention computation equation $\operatorname{softmax}(\frac{\bm{Q}\bm{K}^{\intercal}}{\sqrt{D}})\bm{V}$, multiple intermediate results, such as $\bm{Q}\bm{K}^{\intercal}$ and the attention score matrix, need to be explicitly retained, leading to numerous memory read-write operations. 
FlashAttention uses specially designed methods, such as matrix partition and operator fusion, to keep intermediate results in the cache until the final result is obtained, thus reducing the amount of memory read and write operations.
Additionally, FlashAttention can effectively reduce the peak memory usage and activation memory consumption (Section~\ref{sec:training-effi}) during the LLM training and inference. By using FlashAttention, LLaMA-2~(7B) with a sequence length of 2,048 and a batch size of 8 requires only one-tenth of the computation time compared to the standard method.

\paratitle{Sequence Parallelism.}
Compared with the 3D parallelism introduced in Section~\ref{sec:training_settings}, sequence parallelism can be considered a fourth parallelism dimension in pre-training, particularly when handling long data sequences. The core idea is to partition the sequence across multiple devices for parallel computation. The primary challenge lies in minimizing communication across the devices during attention computation. DeepSpeed-Ulysses~\cite{Jacobs-arxiv-2023-deepspeed} partitions the sequence along the hidden dimension, allowing each device to receive a subset of the attention heads and compute attention for different heads in parallel. In comparison, Ring Attention~\cite{Liu-arxiv-2023-ring} partitions the sequence along the {length dimension}, where the query matrices on each device are in turn computed with the key and value matrices on other devices. Furthermore, Ring Attention is also compatible with FlashAttention and can be considered as its distributed extension.


\begin{table*}[!t]
\centering
\small
\caption{The computation, data transfer, and arithmetic intensity during the prefill stage. We use the asymptotic notation $O$ to denote the complexity of data transfer amount, where the constant factor of the complexity is related to the specific implementation method. Table source:~\cite{Chen-arxiv-2024-towards}.}
\label{tab:prefill-flops}
\begin{tabular}{llll}
\toprule
\textbf{Equations}                                                 & \textbf{Computation}            & \textbf{Data transfer}                  & \textbf{Arithmetic intensity}                                                             \\ \midrule
\ding{192} $\bm{Q}, \bm{K}, \bm{V} = \bm{X} \bm{W}^{Q,K,V}$   & $6 B T H^2$               & $O(B T H + H^2)$           & $O \left( \frac{1}{\frac{1}{H} + \frac{1}{B T}} \right)$                \\
\ding{193} $\bm{Q}, \bm{K} = \mathrm{RoPE}(\bm{Q}, \bm{K})$ & $6 B T H$                 & $O(B T H)$                 & $O(1)$                                                                  \\
\ding{194} $\bm{O} = \mathrm{Attn}(\bm{Q}, \bm{K}, \bm{V})$ & $4 B T^2 N D + 4 B T^2 N$ & $O(B T^2 N + B T N D)$     & $O \left( \frac{1 + \frac{1}{D}}{\frac{1}{D} + \frac{1}{T}} \right)$    \\
\ding{195} $\bm{X} = \bm{O} \bm{W}^O$                       & $2 B T H^2$               & $O(B T H + H^2)$           & $O \left( \frac{1}{\frac{1}{H} + \frac{1}{B T}} \right)$                \\
\ding{196} $\bm{X} = \mathrm{Add}\&\mathrm{Norm}(\bm{X})$   & $5 B T H$                 & $O(B T H + H)$             & $O \left( \frac{1}{1 + \frac{1}{B T}} \right)$                          \\
\ding{197} $\bm{G}, \bm{U} = \bm X [\bm{W}^G, \bm{W}^U]$        & $4 B T H H'$              & $O(B T H + B T H' + H H')$ & $O \left( \frac{1}{\frac{1}{H} + \frac{1}{H'} + \frac{1}{B T}} \right)$ \\
\ding{198} $\bm{D} = \mathrm{Swish}(\bm{G}) \cdot \bm{U}$    & $2 B T H'$                & $O(B T H')$                & $O(1)$                                                                  \\
\ding{199} $\bm{X} = \bm{D} \bm{W}^D$                       & $2 B T H H'$              & $O(B T H + B T H' + H H')$ & $O \left( \frac{1}{\frac{1}{H} + \frac{1}{H'} + \frac{1}{B T}} \right)$ \\
\ding{200} $\bm{X} = \mathrm{Add}\&\mathrm{Norm}(\bm{X})$   & $5 B T H$                 & $O(B T H + H)$             & $O \left( \frac{1}{1 + \frac{1}{B T}} \right)$                          \\ \bottomrule
\end{tabular}
\end{table*}

\begin{table*}[!t]
\centering
\small
\caption{The computation, data transfer, and arithmetic intensity during the decoding stage. Table source:~\cite{Chen-arxiv-2024-towards}.}
\label{tab:decoding-flops}
\begin{tabular}{llll}
\toprule
\textbf{Equations}                                                 & \textbf{Computation}            & \textbf{Data transfer}                  & \textbf{Arithmetic intensity}                                                             \\ \midrule
\ding{192} $\bm{q}, \bm{k}, \bm{v} = \bm{X} \bm{W}^{QKV}$                & $6 B H^2$             & $O(B H + H^2)$               & $O \left( \frac{1}{\frac{1}{H} + \frac{1}{B}} \right)$                   \\
\ding{193} $\bm{q}, \bm{k} = \mathrm{RoPE}(\bm{q}, \bm{k})$                   & $6 B H$               & $O(B H)$                     & $O(1)$                                                                   \\
\ding{194} $\bm{K}, \bm{V} = \mathrm{Cache}(\bm{k}, \bm{v})$        & -                     & $O(B T N D)$ or $O(B N D)$                    & -                                                                             \\
\ding{195} $\bm{o} = \mathrm{Attn}(\bm{q}, \bm{K}, \bm{V})$         & $4 B T N D + 4 B T N$ & $O(B T N + B T N D + B N D)$ & $O \left( \frac{1 + \frac{1}{D}}{1 + \frac{1}{D} + \frac{1}{T}} \right)$ \\
\ding{196} $\bm{X} = \bm{o} \bm{W}^O$                          & $2 B H^2$             & $O(B H + H^2)$               & $O \left( \frac{1}{\frac{1}{H} + \frac{1}{B}} \right)$                   \\
\ding{197} $\bm{X} = \mathrm{Add}\&\mathrm{Norm}(\bm{X})$ & $5 B H$               & $O(B H + H)$                 & $O \left( \frac{1}{1 + \frac{1}{B}} \right)$                             \\
\ding{198} $\bm{g}, \bm{u} = \bm{X} [\bm{W}^G, \bm{W}^U]$           & $4 B H H'$            & $O(B H + B H' + H H')$       & $O \left( \frac{1}{\frac{1}{H} + \frac{1}{H'} + \frac{1}{B}} \right)$    \\
\ding{199} $\bm{d} = \mathrm{Swish}(\bm{g}) \cdot \bm{u}$                 & $2 B H'$              & $O(B H')$                    & $O(1)$                                                                   \\
\ding{200} $\bm{X} = \bm{d} \bm{W}^D$                          & $2 B H H'$            & $O(B H + B H' + H H')$       & $O \left( \frac{1}{\frac{1}{H} + \frac{1}{H'} + \frac{1}{B}} \right)$    \\
\ding{201} $\bm{X} = \mathrm{Add}\&\mathrm{Norm}(\bm{X})$ & $5 B H$               & $O(B H + H)$                 & $O \left( \frac{1}{1 + \frac{1}{B}} \right)$                             \\ \bottomrule
\end{tabular}
\end{table*}

\subsection{Analysis and Optimization for {Model Inference}}
{In Section~\ref{sec-decoding}, we have introduced the basic decoding strategies for using LLMs. 
As inference efficiency is critically important for the application of LLMs, we next will quantitatively analyze the efficiency of the inference process and also present corresponding optimization methods. 
}

\subsubsection{Analysis of Inference Efficiency}

Overall, the inference process of LLMs can be divided into two stages for overhead analysis:  (1) the \emph{prefill} stage, which computes the states and caches the key-value tensors for the input sequence; and (2) the \emph{decoding} stage, which computes the states of the newly generated tokens, updates the key-value cache (KV cache, and continuously generate tokens in an auto-regressive way until the generation process is complete~\cite{Sheng-ICML-2023-FlexGen}.

\paratitle{Inference Efficiency Measurement.} 
To quantitatively analyze the inference efficiency, we next will introduce two widely-used metrics for measuring inference efficiency. 

$\bullet$ \emph{GPU performance metrics.} First, we introduce the \emph{compute capability} and \emph{memory bandwidth} to evaluate the efficiency of a certain GPU. 
The compute capability of a GPU refers to the number of floating-point operations (FLOP) that it can perform per second, measured in FLOP/s. The bandwidth of a GPU refers to the amount of memory read and write operations it can perform per second, measured in byte/s. The ratio of compute to bandwidth is known as the \emph{maximum arithmetic intensity} of the GPU, denoted as $I_{\text{max}}$, which is measured in FLOP/byte.
For example, the half-precision compute and bandwidth of the A100 GPU are 312 TFLOP/s and 2039GB/s, respectively. Correspondingly, its maximum arithmetic intensity is 142.51 FLOP/byte\footnote{\url{https://www.nvidia.com/en-us/data-center/a100/}}.

$\bullet$ \emph{Model efficiency metrics.}
Similarly, each operation (\eg matrix multiplication) of the model can be measured by two corresponding metrics: the \emph{computation amount} and the \emph{data transfer amount}. The former refers to the total number of floating-point operations, measured in FLOPs. The latter refers to the total amount of GPU memory read and write operations, measured in bytes. Analogous to the arithmetic intensity of a GPU, the \emph{arithmetic intensity} $I$ of a model operation (\eg matrix multiplication) can be defined as the ratio of computation to data transfer, with units of FLOP/byte. 

When the model's arithmetic intensity $I$ is less than the GPU's maximum arithmetic intensity $I_{\text{max}}$, it indicates that the maximum memory bandwidth of the GPU is lower than the speed required. Consequently, the model's efficiency will primarily be limited by memory bandwidth, and the operation is called  \emph{memory-bound}. 
Conversely, when $I$ exceeds $I_{\text{max}}$, it suggests that the GPU's maximum floating-point operation speed is lower than the speed required. In this case, the model's efficiency will mainly be constrained by the GPU's compute capability, and the operation is called \emph{compute-bound}.


\paratitle{Bottleneck Analysis.}
Based on the above analysis, we can obtain the arithmetic intensity for each operation during both the prefill and decoding stages, as shown in Tables~\ref{tab:prefill-flops} and~\ref{tab:decoding-flops}, thereby better identifying the bottleneck operations in the inference process.

$\bullet$ \emph{Prefill stage.}
In the following analysis, we will still take the LLaMA (7B) model in Table~\ref{tab:activation} as an example ($N=32, D=128, H=4096$) and assume a batch size of 8 and a sequence length of 1024 (\ie $B=8, T=1024$). Substituting these values into Table~\ref{tab:prefill-flops}, we can find that the arithmetic intensity for linear transformations (Equations~\ding{192}\ding{195}\ding{197}\ding{199}) is approximately 2730.67, for multi-head attention (Equation~\ding{194}) it is approximately 114.67, while the intensity for other operations (Equations~\ding{193}\ding{196}\ding{198}\ding{201}) is around 1. 
When using an A100 (80G) GPU with $I_{\text{max}}=142.51$, the arithmetic intensities of the linear transformations and multi-head attention operations are all above or close to the maximum value. Given that these operations occupy the majority of the computations during the prefill stage, we can conclude that prefill stage is actually compute-bound.

$\bullet$ \emph{Decoding stage.}
Similarly, substituting these values into the arithmetic intensity formulas in Table~\ref{tab:decoding-flops} for the decoding stage reveals that the arithmetic intensities of the linear transformations and multi-head attention are all below 8, which is much lower than the A100 GPU's maximum intensity 142.51. This indicates that the decoding stage is constrained by the GPU's data transfer speed (\ie memory-bound), a problem commonly referred to as the \emph{memory wall}. The analysis indicates that inefficiencies in LLM inference primarily occur during the decoding stage.

\subsubsection{System-level Optimization}
To mitigate the memory wall issue, an intuitive idea is to reduce the data transfer operations as possible, thereby enhancing the arithmetic intensity. In this part, we will introduce several system-level optimization methods to achieve the reduction in data transfer.

\paratitle{FlashAttention and Flash-Decoding.}
The FlashAttention method discussed in Section~\ref{subsub:training-effi} can also be applied at the prefill stage, as it reduces data transfer operations and effectively increases arithmetic intensity. However, this optimization technique is not directly applicable during the decoding stage, where only the {current query vector needs to be computed with the KV cache matrices}. To further optimize the decoding process,  Flash-Decoding~\cite{dao-web-2023-flash} has been proposed based on FlashAttention, particularly for long sequences, which shares a similar idea with sequence parallelism.   
Specifically, Flash-Decoding splits the KV cache into smaller chunks, {allowing the computation of the query vector with these chunks in parallel}, thereby improving the decoding efficiency.

\paratitle{PagedAttention.}
PagedAttention~\cite{vllm-pagedattention} focuses on optimizing {KV cache} and attention computation, significantly reducing data transfer operations in these two aspects. In KV cache concatenation, traditional methods often need to allocate new GPU memory for each concatenation, copying the original KV cache and the new hidden states into the newly allocated memory. This process leads to repeated memory read-write operations and substantial memory fragmentation. PagedAttention addresses this issue by introducing a memory paging management method, preallocating several blocks of memory for future KV caches, which can largely reduce the memory allocation operations during concatenation. Additionally, PagedAttention optimizes the attention computation by increasing the parallelism. It uses operator fusion to parallelize the {computation of the query vector with multiple KV cache chunk}, thereby enhancing the computational efficiency.

\paratitle{Batch Management Optimization.}
Batch management optimization aims to increase the batch size  during the decoding stage to enhance arithmetic intensity. A representative method is continuous batching, proposed by vLLM~\cite{vllm-pagedattention}. Unlike traditional fixed-length batch processing, this technique breaks down each request into a prefill iteration and several single-step {decoding iterations}, and continuous batching further employ heuristic algorithms to select requests for prefill or single-step decoding iteration. 
This fine-grained batching mechanism allows for handling more requests simultaneously, which is has the same effect as increasing the batch size. 
Furthermore, DeepSpeed-MII~\cite{Holmes-arxiv-2024-deepspeed} introduces Dynamic SplitFuse, which splits the prefill stage into multiple iterations and allows simultaneous prefill and decoding in one computation, resulting in larger batches and higher inference throughput.

\subsubsection{Algorithm-level Optimization}

In addition to system-level optimization methods, existing research work has proposed a series of improvements for autoregressive inference algorithms aimed at enhancing inference efficiency. This part introduces four typical inference optimization algorithms.

\paratitle{Speculative Decoding.}
Intuitively, the generation steps in language modeling have varied difficulty levels. For example, predicting the next word of ``\emph{The founder of Microsoft is}" may be more challenging than predicting the next word of ``\emph{The founder of Microsoft is Bill}". Even a small model may successfully predict the answer in this case. Based on this idea, \emph{speculative decoding}~\cite{Yaniv-ICML-2023-Fast,Chen-arxiv-2023-Accelerating} has been proposed to accelerate the inference speed. Specifically, it employs a relatively smaller yet more efficient model (such as an $n$-gram statistical model or a small pre-trained model) to autoregressively generate several tokens. Then, a larger model then verifies these tokens, determining {whether each token is the top-ranked prediction at the each generation step}. The small and large models iteratively repeat this process until decoding is complete. Speculative decoding can lead to a notable 2$\times$ to 3$\times$ speedup without compromising the generation quality.
Researchers further suggest several variants to improve the efficiency of this approach, such as a learning-based method to combine several small models~\cite{Miao-arxiv-2023-SpecInfer} and a stage-wise acceleration which employs a more smaller model to accelerate the small model first~\cite{Spector-2023-arxiv-Accelerating}.

\paratitle{Cascade Inference.}
{\emph{Cascade inference} optimizes the inference efficiency by addressing requests of varying difficulty with models of different scales.}
FrugalGPT~\cite{Chen-arxiv-2023-frugalgpt} introduces a series of models arranged by efficiency from high to low, sequentially processing a request through these models. A specially trained binary classification model then evaluates whether the generated result meets the task requirements. If the result is deemed reliable, subsequent models would be bypassed, thus improving the inference speed. This strategy can be applied to various open-source models and commercial APIs, allowing for the flexible adjustment the classification threshold to balance inference efficiency and generation quality according to specific needs. For reasoning tasks, researchers~\cite{Yue-arxiv-2023-large} further propose to utilize the self-consistency~\cite{Wang-arxiv-2022-Self-Consistency} of generated answers to evaluate the quality of the small model: the large model is employed for generation only when the small model's answers exhibit a low consistency. 


\paratitle{Non-autoregressive Decoding.}
Existing decoding methods predominantly adopt the autoregressive mechanism, generating tokens one by one, which is a primary reason for lower inference efficiency. Therefore,  \emph{non-autoregressive decoding}~\cite{Gu-iclr-2018-Non} has been proposed by generating  {all tokens} based on the input {at once}. However, the generation quality of this method still largely lags behind autoregressive methods. To improve the quality of the generated text, several studies attempt to combine both decoding methods, proposing semi-autoregressive decoding methods~\cite{Wang-emnlp-2018-Semi} that generate a group of tokens (\eg 3 to 10 tokens) at each step and use these tokens as input to generate the next group. However, existing mainstream  LLMs are pre-trained to predict the next token, making direct non- or semi-autoregressive generation infeasible. To address this, Medusa~\cite{Cai-arxiv-2024-Medusa} trains two additional prediction heads on the Vicuna model to predict the second and third tokens respectively, thereby achieving the generation of three tokens simultaneously. 
However, due to the decreased generation quality, these methods have been rarely used directly in practice, but are more often combined with other methods (\eg speculative decoding) to accelerate the inference process of LLMs. 
For instance, after Medusa generates three tokens in parallel, the original Vicuna model would still be   employed to verify the generation quality. 

\paratitle{Early Exit.}
It has been found that in multi-layer Transformer models, it may not be necessary to perform the computation through all layers to reliably predict the next token~\cite{Teerapittayanon-icpr-2016-branchynet}. Based on this idea, {several studies~\cite{Huang-iclr-2018-multi,Teerapittayanon-icpr-2016-branchynet}} have proposed improved generation methods based on \emph{early exit}. During model decoding, when  the conditions for early exit are satisfied, the model can directly use intermediate computation results from certain layers to generate tokens, thereby improving the inference efficiency. 
To determine the exit condition, prediction confidence~\cite{Huang-iclr-2018-multi} or the entropy~\cite{Teerapittayanon-icpr-2016-branchynet} of the next token's generation probability distribution can be used as reference measures. 
More recently, mixture-of-depths~\cite{raposo-arxiv-2024-mixture} has proposed to dynamically adjust the computation load of each layer. Similar to MoE networks, the mixture-of-depths method calculates a score for each layer's input via a routing network. If the score exceeds a preset threshold, the layer would be computed; otherwise, the layer would be skipped. Unlike traditional early exit mechanisms that skip all subsequent layers, the mixture-of-depths method selectively skips certain layers, which can adaptively utilize the characteristics of different layers during generation. 

\subsection{Model Compression}

Due to the huge number of model parameters, LLMs take a significant memory footprint for inference, making it very costly to be deployed in real-world applications~\cite{Wan-2024-Efficient-arXiv}. In this section, we focus on how to reduce the memory footprint of LLMs 
via technical approaches. In particular, we will primarily introduce the  model quantization approach, and also briefly 
discuss other model compression methods, \eg model pruning and distillation. 
\subsubsection{Quantization Methods}

There are generally two major model quantization approaches, namely \emph{quantization-aware training~(QAT)}~(requiring additional full model retraining) and \emph{post-training quantization~(PTQ)} (requires no model retraining). 
Compared with small-sized language models, two major differences need to be considered when designing or selecting quantization methods for LLMs. Firstly, LLMs consist of a huge number of parameters, and thus PTQ methods are more preferred due to a much lower computational cost than QAT methods. Secondly, LLMs exhibit very different activation patterns (\ie large outlier features), and it becomes more  difficult to quantize LLMs, especially  hidden activations. Next, we will briefly review several representative PTQ  methods\footnote{Since we mainly focus on discussing  quantization methods in the context of LLMs,  the line of  quantization work on small-sized language models (\eg BERT) has not been included in this survey. 
} for LLMs.   

\paratitle{Background for Quantization}.
In this part,  we present a general introduction of quantization techniques for neural networks. 
In neural network compression, quantization often refers to the mapping process from  floating-point numbers to  integers~\cite{Gholami-CoRR-2022-A}, especially the 8-bit integer quantization (\ie  \emph{INT8 quantization}). 
For neural network models, there are typically two kinds of data to be quantized, namely \emph{weights} (model parameters) and \emph{activations} (hidden activations), which are originally represented in floating-point numbers.  To illustrate the essential idea of model quantization,  we   introduce a simple yet popular quantization function:
$x_q = R(x/S) - Z$, which transforms a floating number $x$ into a quantized value $x_q$. In this function, 
$S$ and $Z$ denote the scaling factor (involving two  parameters $\alpha$ and $\beta$ that determine the clipping range) and zero-point factor (determining symmetric or asymmetric quantization),  respectively, and $R(\cdot)$ denotes the rounding operation that maps a scaled floating value to an approximate integer. 
As the reverse process,  \emph{dequantization}  recovers the original value from the quantized value accordingly: $\Tilde{x} = S\cdot (x_q + Z)$.  The  quantization error is calculated as the numerical difference between the original value $x$ and the recovered value $\Tilde{x}$.  
{The range parameters  $\alpha$ and $\beta$ have a large impact on the quantization  performance}, which often need to be  \emph{calibrated} according to real data distributions, in either a \emph{static} (offline) or \emph{dynamic} way (runtime). For more details, we refer to the readers to the excellent  survey~\cite{Gholami-CoRR-2022-A} about quantization methods on neural networks.  

\paratitle{Post-Training Quantization~(PTQ)}. We first introduce the PTQ methods for LLMs.

$\bullet$ \emph{Mixed-precision decomposition}. 
As found in \cite{Dettmers-arxiv-2022-LLM}, {extremely} large values would occur in hidden activations (called \emph{the emergence of outliers}) when the model size reaches  6.7B parameters or above. These outliers significantly influence the data distribution ranges of the hidden activations, making it challenging to conduct effective model quantization. 
\ignore{
Interestingly, these outliers are mainly distributed in  some specific  feature dimensions  at Transformer layers. 
Based on this finding, a vector-wise  quantization approach, called \emph{LLM.int8()}, has been proposed in \cite{Dettmers-arxiv-2022-LLM}, which separates  the feature dimensions with outliers and the rest dimensions  in  matrix multiplication. 
Then, the calculations for the two parts are performed with \emph{16-bit floating numbers} and \emph{8-bit integers}, respectively, so as to recover these outliers in a high precision. }
{To reduce the quantization error, a straightforward method is to separately process the outliers and the rest weight values. Specifically, LLM.int8()~\cite{Dettmers-arxiv-2022-LLM} has observed that these outliers are mainly distributed in certain feature dimensions at Transformer layers. Based on this finding, a vector-wise quantization approach is proposed to separate the outliers and the rest in matrix multiplication. 

$\bullet$ \emph{Salient weights protection}. For Transformer based language models, there often exists a subset of weight values that are more sensitive to quantization, {which are also referred to as~\emph{salient weights}~\cite{Lin-arXiv-2023-AWQ}.
}
{Unlike activation outliers, which occur dynamically during inference and may require complex runtime handling, weight outliers are static and can be pre-processed before model deployment.} By identifying and preserving these salient weights, the error associated with model quantization can be effectively reduced. In existing literature, various methods have been proposed to detect these salient weights. 
For instance, PB-LLM~\cite{Shang-CORR-2023-PBLLM} utilizes the magnitude of weights for finding critical weights, 
{SpQR~\cite{Tim-arxiv-2023-SpQR} categorizes the outliers in weights into small groups by investigating the structural patterns,} 
APTQ~\cite{Guan-CORR-2024-APTQ} employs the Hessian trace as a sensitivity metric, and OWQ~\cite{Lee-AAAI-2024-OWQ} selects the top sensitive columns based on both the Hessian matrix and weight perturbations.

$\bullet$ \emph{Fine-grained quantization}.
For Transformer models, weights and activations are usually represented in the form of tensors. A straightforward approach is to use coarse-grained   quantization parameters for the  whole tensor (\ie  per-tensor quantization)~\cite{Xiao-CoRR-2022-SmoothQuant}. However, it usually leads to inaccurate reconstruction results. 
Thus, fine-grained  methods are proposed to reduce the quantization error.  
ZeroQuant~\cite{Yao-NeurlPS-2022-ZeroQuant} adopts a  token-wise quantization approach with dynamic calibration for compressing activations. Whereas for weights (easier to be quantized), it uses a group-wise quantization. In practice, a group size of 128~\cite{Yao-NeurlPS-2022-ZeroQuant,Lin-arXiv-2023-AWQ} is commonly used for model quantization.  

$\bullet$ \emph{Balancing the quantization difficulty}.   
Considering that weights are easier to be quantized than activations, SmoothQuant~\cite{Xiao-CoRR-2022-SmoothQuant}  proposes to migrate the difficulty from  activations to weights. Specially, they incorporate a scaling transformation  to balance the difficulty between weights and activations in a linear layer: $\mathbf{Y} = (\mathbf{X}\text{diag}(\mathbf{s})^{-1}) \cdot (\text{diag}(\mathbf{s})\mathbf{W})$. 
By introducing an mathematically equivalent transformation, this formula  controls the quantization difficulty through the scaling factor $\mathbf{s}$. 
To set $\mathbf{s}$, it  incorporates a migration strength parameter $\alpha$ to balance the difficulties,  where each entry $s_j=\max(\mathbf{x}_j)^\alpha / \max(\mathbf{w}_j)^{(1-\alpha)}$ is determined by  the migration strength.  


{
$\bullet$ \emph{Layerwise quantization}. This approach finds optimal  quantized weights that minimize a layerwise reconstruction loss: {$\arg\min_{\widehat{\mathbf{W}}}\parallel \mathbf{W}\mathbf{X} -  \widehat{\mathbf{W}} \mathbf{X}\parallel_2^2$}.  
To efficiently optimize this objective, GPTQ~\cite{frantar-arxiv-2022-gptq} improves the  original optimal brain quantization~(OBQ)~\cite{Frantar-NeurIPS-2022-Optimal} method by fixing the quantization order of weights for all rows. 
Further, with specially designed   methods (\ie lazy batch-updates and Cholesky reformulation),  GPTQ  is feasible to quantize very large models (\eg 175B OPT)  in 3 or 4 bit precision.   
More recently, AWQ~\cite{Lin-arXiv-2023-AWQ} further simplifies the optimization form by incorporating activation-aware scaling for weights, which resembles the idea of  SmoothQuant~\cite{Xiao-CoRR-2022-SmoothQuant}: weights corresponding to outlier activations are more important to be precisely quantized. It does not directly optimize the reconstruction loss, but instead performs simple hyper-parameter search to achieve  the minimal  loss on calibration data. 
}

These strategies in the above methods can be jointly used to improve the quantization performance. 
{In order to achieve high-efficiency implementation, quantization methods also rely on hardware- or system-level support (\eg efficient GPU kernels or hardware-friendly group partition). 
}

\paratitle{Other Quantization Methods}. In the above, we mainly focus on PTQ methods, and next introduce two recent studies that explore efficient fine-tuning methods or QAT methods   for quanitizing LLMs.

$\bullet$ \emph{Efficient fine-tuning enhanced quantization.} 
For post-training quantization, direct low-bit  quantization (\eg INT4 quantization) often results in large performance degradation.  To overcome this challenge, QLoRA~\cite{Dettmers-CoRR-2023-QLoRA} 
incorporates additional small tunable  adapters (16-bit precision) into the quantized models,  to achieve an efficient, high-precision model fine-tuning.  
It combines the merits of LoRA~(See Section~\ref{sec-PEFT-methods}) and quantization methods. The experiment results show that 4-bit quantized  models can 
achieve the full 16-bit fine-tuning performance by QLoRA.

$\bullet$ \emph{Quantization-aware training~(QAT) for LLMs}. A recent study~\cite{liu-2023-arxiv-LLM-QAT} explores the effect of QAT methods by applying a data-free distillation method to compress the weights, activations as well as key-value cache. By conducting extensive experiments based on LLaMA, they show promising results with 4-bit quantization on both weights and key-value cache, but not on 4-bit activation quantization, which still needs more exploration. 

\label{sec:quantization_empirical}
\paratitle{{Empirical Analysis and Findings}.} Quantization has currently become a common technique to reduce the memory footprint and latency of LLMs in deployment. 
In particular, it is important to understand what level of precision (\eg INT8 or INT4) can be applied to quantize different parts of LLMs (\eg weights or activations), while retaining  a high accuracy. 
In this part, we first summarize the major findings about the quantization of LLMs in existing literature, and then present some empirical analysis with quantization experiments.  
\ignore{
\paratitle{Important Findings from Existing Work}. Recently, a very comprehensive evaluation~\cite{Yao-CoRR-2023-ZeroQuant-V2}  has been conducted about the impact of multiple factors (\eg model size and sensitivity) on the post-training quantization methods. Another  study~\cite{Dettmers-2022-arxiv-case} examines the scaling law of $k$-bit quantization in inference performance.   
{In addition to  the overall performance, the study~\cite{Liu-2023-arxiv-Do_emergent} specifically focuses on the potential impact of quantification on emergent capabilities, as well as the levels of performance that can be achieved across various levels of bit precision.}
Also,  prior  work (\eg LLM.int8()~\cite{Dettmers-CoRR-2022-LLM.int8}, GPTQ~\cite{frantar-arxiv-2022-gptq}, QLoRA~\cite{Dettmers-CoRR-2023-QLoRA}, and GLM~\cite{Zeng-arxiv-2022-GLM}) has also extensively examined the performance of quantization methods in various settings. Next, we summarize several important findings from these studies, which will be useful for those who may not want to delve into the technical details of quantization methods. }

$\bullet$ \emph{INT8 weight quantization can often yield very good results on LLMs, while the performance of lower precision weight quantization depends on specific  methods}~\cite{Yao-CoRR-2023-ZeroQuant-V2,Xiao-CoRR-2022-SmoothQuant,frantar-arxiv-2022-gptq,Lin-arXiv-2023-AWQ}.  In most cases, INT8 weight quantization can be effectively applied to reduce the memory footprint without performance degradation. 
While for INT4 (or INT3) weight quantization, existing methods rely on specific strategies to reduce the performance degradation, \eg layerwise method~\cite{frantar-arxiv-2022-gptq,Yao-NeurlPS-2022-ZeroQuant}, activation-aware scaling~\cite{Lin-arXiv-2023-AWQ} and low-rank adapter tuning~\cite{Dettmers-CoRR-2023-QLoRA}. Interestingly, LLMs seem to be less sensitive to low-bit weight quantization than small-sized language models~\cite{Yao-CoRR-2023-ZeroQuant-V2}.
In practice, with the same memory cost, it is suggested to use a larger language model with a lower quantization precision rather than a smaller language model with a higher quantization precision. For example, a 4-bit 60B LLM is demonstrated to have better performance than an 8-bit 30B LLM~\cite{Dettmers-2022-arxiv-case}. 
{Moreover, focusing on emergent capabilities, the study~\cite{Liu-2023-arxiv-Do_emergent} finds that in-context learning, step-by-step reasoning, and instruction following all seem to be seldom affected with 4-bit weight quantization. This result suggests that INT4 quantization exhibits a favorable trade-off in terms of both total bits and performance of emergent abilities.}

$\bullet$ \emph{Activations are more difficult to be quantized than weights}~\cite{Yao-CoRR-2023-ZeroQuant-V2,Dettmers-arxiv-2022-LLM,Xiao-CoRR-2022-SmoothQuant}. It has been found that large outliers would occur for Transformer language models having a size of 6.7B or above~\cite{Dettmers-arxiv-2022-LLM}. This issue has been one of the most fundamental difficulties to quantize LLMs.  
To overcome this issue, various methods, \eg mixed-precision decomposition~\cite{Dettmers-arxiv-2022-LLM}, fine-grained  quantization~\cite{wei-arxiv-2023-zero,Dettmers-arxiv-2022-LLM} and difficulty migration~\cite{Xiao-CoRR-2022-SmoothQuant},   can be applied to alleviate the influence of outlier values. 
Since large outliers mainly exist in the activations of LLMs, small language models are more resistant to activation quantization~\cite{Yao-CoRR-2023-ZeroQuant-V2,Liu-2023-arxiv-Do_emergent}.  
{In practice, high-quality INT8 activation quantization is still a difficult task, though several methods can attain satisfying results. }
Further, lower precision activation quantization has still not been successfully explored, even for QAT methods~\cite{liu-2023-arxiv-LLM-QAT}.  

{
$\bullet$ \emph{Efficient fine-tuning enhanced quantization is a good option to enhance the performance of quantized LLMs}~\cite{Dettmers-CoRR-2023-QLoRA,Hu-ICLR-2022-LoRA}.  
The benefits of efficient fine-tuning methods in quantization can be twofold. 
{Firstly, it can directly compensate for the performance degradation suffered from low-bit quantization. This can be achieved either by increasing the fitting capacity via updating high precision adapters~\cite{Yao-CoRR-2023-ZeroQuant-V2,Liu-2023-arxiv-Do_emergent,Xu-CORR-2023-qalora}, or {by finding a proper low-rank initizalization for LoRA fine-tuning~\cite{Li-ICLR-2024-loftq}.}
Secondly, it is flexible to support  task-specific or goal-specific fine-tuning of LLMs in a lightweight way~\cite{Dettmers-CoRR-2023-QLoRA}, \eg instruction tuning or chat-oriented tuning, by only tuning the small adapters. Overall, it makes a good trade-off between the effectiveness and training cost, which provides a promising approach to enhancing the performance of quantized LLMs.   
}

\paratitle{Empirical Analysis on Quantization Experiments}. 
To further help readers  understand the  impact of quantization on LLMs, we also conduct  a group of experiments to investigate the inference performance of quantized models here. 
Specifically, we focus on the fine-tuned LLaMA models~(\ie 7B and 13B) using popular SFT datasets, including FLAN-v2~\cite{Chung-arxiv-2022-Scaling}, Alpaca-52K~\cite{alpaca} and ShareGPT~\cite{ShareGPT}.
For evaluation, we utilize the same tasks  in Table~\ref{tab-instruction-tuning-res}, and  follow the quantization settings in the study~\cite{Liu-2023-arxiv-Do_emergent} examining the performance of quantized language models at three precision levels: 4-bit, 8-bit and 16-bit. The results are summarized in Table~\ref{tab-quantization-tuning-res}.
As can be observed from Table~\ref{tab-quantization-tuning-res}, the results obtained with 8-bit and 4-bit weight quantization are close to the performance of 16-bit models while significantly reducing memory consumption. 
In practice, it is recommended to first examine the performance of 4-bit weight quantization for LLMs if reducing memory usage is a critical consideration for deployment.

\begin{table*}[htb]
    \centering
    \caption{Evaluation results for quantized LLaMA models~(7B and 13B). {We employ existing model checkpoints provided by~\cite{Wang-arxiv-2023-How} for quantization experiments, which have been fine-tuned on FLAN-v2, Alpaca-52K, and ShareGPT, respectively.} Specifically, we report the performance with AlpacaFarm, MMLU, and BBH, as well as the memory usage of the loaded model~(Mem.).
    For quantization, we employ \emph{bitsandbytes} to quantize the 16-bit models to 8/4 bits by specifying the commands \texttt{load\_in\_8bit} and \texttt{load\_in\_4bit} when loading the weights. 
    It is worth noting that we select~\emph{text-davinci-003} as the baseline model for the AlpacaFarm dataset.}
    \label{tab-quantization-tuning-res}
\resizebox{2.0\columnwidth}{!}{
\begin{tabular}{llcccccccccccc}
\toprule
\multirow{2.5}{*}{\textbf{Models}}   & \multirow{2.5}{*}{\begin{tabular}[c]{@{}c@{}}\textbf{SFT Dataset}\end{tabular}} & \multicolumn{4}{c}{\textbf{16-bit}}& \multicolumn{4}{c}{\textbf{8-bit}} & \multicolumn{4}{c}{\textbf{4-bit}} \\ 
\cmidrule(r){3-6}\cmidrule(r){7-10}\cmidrule(r){11-14} & & AlpacaFarm & MMLU & BBH & Mem.$_{\rm{(GiB)}}$ & AlpacaFarm & MMLU & BBH & Mem.$_{\rm{(GiB)}}$& AlpacaFarm & MMLU & BBH & Mem.$_{\rm{(GiB)}}$\\
\midrule
LLaMA~(7B) 
& FLAN-v2 & 6.65 & 47.34 & 35.05 & 12.58 & 6.15 & 47.02 & 35.17 & 6.65 & 7.83 & 46.23 & 34.77 & 3.94 \\
 & Alpaca-52K & 32.55 & 40.87 & 33.66 & 12.58 & 33.60 & 39.98 & 34.38 & 6.65 & 29.57 & 39.24 & 32.80 & 3.94 \\
 & ShareGPT & 72.05 & 41.30 & 32.90 & 12.58 & 72.86 & 39.34 & 32.71 & 6.65 & 70.31 & 40.08 & 32.11 & 3.94 \\

 \midrule
 LLaMA~(13B) 
 & FLAN-v2 & 8.14 & 51.67 & 41.46 & 24.40 & 7.64 & 51.02 & 41.25 & 12.53 & 7.52 & 50.48 & 40.68 & 7.34 \\
 & Alpaca-52K & 33.60 & 47.63 & 36.10 & 24.40 & 31.43 & 47.04 & 35.98 & 12.53 & 30.87 & 46.20 & 36.16 & 7.34 \\
 & ShareGPT & 75.59 & 47.58 & 38.00 & 24.40 & 73.79 & 47.71 & 38.31 & 12.53 & 71.99 & 45.77 & 36.97 & 7.34 \\
\bottomrule
\end{tabular}
}
\end{table*}

{\subsubsection{Other Model Compression Methods} 

In addition to model quantization, we next introduce two other model compression methods for LLMs, namely model distillation and model pruning. Unlike model quantization, model distillation and pruning aim to simplify the model architecture, thereby reducing the total number of parameters.}

\paratitle{Distillation for LLMs.} 
{In general, \emph{model distillation} aims to transfer the capabilities from a capable model~(referred to as the~\emph{teacher model}) to a less capable model (referred to as the~\emph{student model}), thereby achieving the compression of the capable model. Based on whether the weights of {teacher models} are accessible, one can employ either the white-box approach or the black-box approach for LLM distillation. The white-box approach often employs the traditional knowledge distillation technique,  which incorporates additional loss functions~(\ie distillation loss) for aligning the outputs or intermediate states of the student model to those of the teacher model. Based on this approach, MINILLM~\cite{Gu-CORR-2023-knowledge} effectively distills the 13B LLaMA model down to a 7B model. The black-box approach~\cite{Hsieh-ACL-2023-distilling}, on the other hand, can only make use of the textual response of the teacher model for training the student model. These studies mainly focus on utilizing the generated responses for enhancing the key capabilities from the teacher {model~\cite{Taori-github-2023-Stanford, Luo-arxiv-2023-WizardMath}}, such as in-context learning and chain-of-thought prompting.}

\paratitle{Pruning for LLMs.} {The goal of model pruning is to minimize the number of parameters in a model while preserving its performance as much as possible. 
In general, model pruning methods can be categorized into two lines: structured pruning and unstructured pruning. Structured pruning aims to remove certain less important model components~(\eg neurons, channels, layers) that have minimal impact on performance. On the other hand, unstructured pruning mainly focuses on {removing individual weights or connections within a neural network model} without changing the model's main structure. As for LLMs, unstructured pruning can generally lead to higher compression rates. For instance, SparseGPT~\cite{Frantar-ICLR-2024-Sparsegpt} achieves 
{60\% unstructured sparsity 
for OPT-175B using unstructured pruning~(\ie 60\% of the elements in the weights are masked)}, and the pruned LLM still retains a relatively low perplexity. {With suitable strategies, structured pruning for LLMs can also  achieve promising model compression rate.} 
For instance, LLM-pruner~\cite{Ma-NIPS-2023-Llm-pruner} {selectively removes 20\% of the non-essential parameters from LLaMA~(7B) based on gradient information}, while maintaining 93.6\% performance of the original model. Furthermore, Sheared LLaMA~\cite{Xia-CORR-2023-Sheared} {introduces two techniques: targeted structured pruning and dynamic batch loading, which effectively prunes LLaMA-2~(7B) to a parameter size of 2.7B, while preserving 87.8\% of the original model's performance.}

\ignore{The core idea of model distillation involves introducing an additional loss function (referred to as distillation loss), which guides the student model to align its outputs to those of the teacher model.}

\subsubsection{Open-source Libraries} 

In this part, we briefly introduce the available open-source   libraries for memory-efficient deployment.

\paratitle{Quantization Libraries}. Next, we introduce three popular quantization libraries for LLMs, including:  

$\bullet$ \emph{Bitsandbytes}\footnote{https://github.com/TimDettmers/bitsandbytes} is developed based on the methods introduced in the papers of LLM.int8()~\cite{Dettmers-arxiv-2022-LLM} and 8-bit optimizers~\cite{Dettmers-ICLR-2022-8bit}. 
{It focuses on the quantization of both activations and weights for LLMs, including the support on 8-bit and 4-bit~(NF4,FP4) matrix multiplication for efficient inference, as well as an 8-bit optimizer for efficient training.}

$\bullet$ \emph{GPTQ-for-LLaMA}\footnote{https://github.com/qwopqwop200/GPTQ-for-LLaMa} is  developed specially for quantizing LLaMA models. It enables 4-bit quantization of LLaMA models of varied sizes based on the GPTQ algorithm~\cite{frantar-arxiv-2022-gptq}. Also, it provides a comparison with bitsandbytes in both memory and performance (PPL) on the project website. 


$\bullet$ \emph{AutoGPTQ}\footnote{https://github.com/PanQiWei/AutoGPTQ} is a quantization package developed based on the GPTQ algorithm~\cite{frantar-arxiv-2022-gptq}, which supports INT4 quantization for LLMs. 
It includes a number of quantized models in the library, and supports  LoRA by integrating with HuggingFace PEFT library.

$\bullet$ \emph{llama.cpp}\footnote{https://github.com/ggerganov/llama.cpp} makes it feasible to run  quantized LLaMA models on a MacBook device.
It supports INT4, INT5 and INT8 quantization, which is developed in efficient C/C++ implementation. It also supports a number of LLaMA based models, such as Alpaca and  Vicuna.

\paratitle{Other Libraries}. In addition, there are also libraries for supporting other model compression methods. 

$\bullet$ \emph{Torch-Pruning
}\footnote{https://github.com/VainF/Torch-Pruning} is a toolkit developed for general-purpose structural pruning, including the pruning for vision models, diffusion models and large language models. {It employs dependency graph for automatic structural pruning and supports several high-level pruners~(\eg MetaPruner and BNScalePruner).
} 

$\bullet$ \emph{LLM-Pruner}\footnote{https://github.com/horseee/LLM-Pruner} is designed specifically for the pruning of LLMs.  {It enables efficient gradient-based structral pruning for LLMs with minimal training samples and training time. Currently, it supports a number of LLMs, such as Baichuan, BLOOM, and LLaMA3.} 

\subsection{Retrieval-Augmented Generation}

When dealing with real-time information or specialized domain knowledge, LLMs may struggle to generate accurate outputs solely based on their internal knowledge. To address this issue,  \emph{retrieval-augmented generation}~(RAG) technique~\cite{ding2024survey, gao2023retrieval} has been proposed by incorporating external knowledge source for improving the model response. 
 This technique aims to retrieve relevant information from external sources (\eg the internet or domain-specific knowledge bases) using an information retrieval system, thereby providing LLMs with timely or domain-relevant context to reduce the factual errors in generated content. In the format, RAG can also be considered as a specific prompting strategy that integrates auxiliary information from external sources into the original prompt. Next, we will introduce the basic workflow of the retrieval-augmented generation technique and related optimization strategies.

\paratitle{Basic Workflow.}
Typically, the standard RAG procedure consists of three steps, including context retrieval, prompt construction, and response generation.

\textbullet~\emph{Context Retrieval.}
The retrieval step primarily focuses on finding relevant context information from existing information sources that are helpful for addressing the current information need. To achieve efficient retrieval, it is often necessary to build a search index over the collection of candidate documents and then use appropriate  methodologies for text retrieval. There are two  commonly used retrieval approaches: lexical-based retrieval~\cite{robertson2009probabilistic} using sparse vector representations and semantic retrieval methods using dense vector representations~\cite{Zhao-arxiv-2022-Dense}. The former tokenizes the documents and building an inverted index based on a vocabulary, followed by retrieving relevant documents using lexical matching. The latter maps documents to low-dimensional dense vectors and then constructs an efficient index of document vectors using approximate nearest neighbor search algorithms, ranking candidate documents based on the similarity of embeddings. Both methods can often perform well for large-scale document collection, which are widely used in existing RAG systems. 

\textbullet~\emph{Prompt Construction.}
After the retrieval stage returns the relevant documents, these documents need to be incorporated into the input prompt of the LLM along with the task description. The prompt should guide the model to utilize the retrieved information to complete the corresponding task. For example, a prompt could be, ``\emph{Please refer to the information contained in the following documents to complete the task}''.  Since the retrieved documents are typically lengthy, simply concatenating them into the prompt might lead to a poor utilization of the provided context due to the biased attention (\eg \emph{lost in the middle}~\cite{Liu-arxiv-2023-Lost}). 
To address this issue, existing approaches often introduce reranking models to select the most relevant documents from the retrieval results~\cite{wang2024rear}. Alternatively, information extraction or text compression techniques can be used to retain only the highly relevant information from the documents, thereby reducing the input context length~\cite{rau2024context, xu2024recomp}.

\textbullet~\emph{Response Generation.}
In this step, the constructed prompt is input into the LLM, enabling it to utilize the retrieved content to better accomplish the corresponding task. However, the retrieved documents may contain irrelevant information or even contradictory information to the true answer, which might affect the output generated by the LLM. To address this, the LLM can be further prompted to self-check the quality of the generated output and decide {whether to re-perform the retrieval based on the new outputs~\cite{shao2023enhancing}\ignore{啥意思？？如果质量不好，再次检索就能检索到好的吗？Kun: 是的，基于新的结果去检索}}, or it can perform a confidence assessment to determine whether the current task requires retrieval or the use of retrieved content~\cite{Jiang-2023-arxiv-Active}.

\paratitle{Improvement Strategies.}
In practice, factors such as the quality of retrieved documents, prompt design, and the generation method of  LLMs might impact the final performance of RAG.  
Next, we discuss how to enhance the RAG performance by summarizing existing improvement strategies. 

\textbullet~\emph{Retrieval method improvement.}
The incorporation of retrieval supplements the LLM with relevant contextual information, and the retrieval performance directly affects the quality of the final generated response~\cite{Ren-arxiv-2023-Investigating}. 
To design effective retrieval strategy, an important factor to consider is the text granularity. Intuitively, a coarser granularity (\eg document-level) may result in efficient retrieval but tend to incorporate substantial irrelevant information, while a finer granularity (\eg sentence-level) increases the proportion of relevant content in the retrieval results but can lead to higher retrieval latency. 
To balance relevance and latency, existing research work proposes using ``\emph{propositions}'' as the retrieval unit~\cite{chen2023dense}, corresponding to semantically complete and relatively independent text fragments, which can effectively reduce the recall of irrelevant information. In particular, they mainly use GPT-4 to synthesize instruction data for the extraction of proposition text, training a smaller model specifically to construct proposition text data~\cite{chen2023dense}.
Furthermore, to improve retrieval performance, methods such as query expansion and query rewriting can be utilized to optimize query formulation. Query expansion focuses on adding supplementary information to the original query, such as incorporating related entity information or providing detailed explanations of key information in the query~\cite{Wang-arxiv-2023-query2doc}, which helps strengthen relevance matching. However, traditional query expansion methods may disrupt the original semantics for complex queries. To address this issue, we can employ LLMs to decompose complex queries into several sub-queries, which are subsequently expanded individually, allowing for multi-path recall of related information~\cite{huang2023question}. As another query enhancment technique, query rewriting focuses on modifying the query content to highlight key information and eliminate potential ambiguities, facilitating the retrieval of related documents~\cite{he2016learning}. LLMs can be applied directly to query rewriting, transforming the original query into a more suitable form through well-designed prompts~\cite{liu2024query}. To reduce inference overhead, the query optimization capabilities of LLMs can also be transferred to smaller models through knowledge distillation~\cite{ye2023enhancing}.

{\textbullet~\emph{Retrieval results refinement.}}
\ignore{
(这一部分我建议重写，貌似和提示没啥关系，主要是对检索结果的refine，再看看如何写一下，另外，避免和retrieval optimization内容和重点重叠)}
{In addition to the initial retrieval methods, the refinement of retrieval results also plays an important role in RAG systems, since the retrieved documents may be not best suited for RAG systems, \eg LLMs might have difficulty in utilizing long contexts or be affected by irrelevant information in the retrieved documents.   
}
As a solution, the documents returned during the retrieval stage can be reranked according to their relevance to the input~\cite{jeong2024adaptive}, filtering out low-quality or irrelevant documents or placing less relevant documents in non-optimal positions within the prompt. 
Furthermore, both generation and reranking tasks~\cite{wang2024rear} can be jointly optimized to faciliate better utilize of context documents. 
Additionally, LLMs can be directly used for document re-ranking by designing specific prompts or using context examples to accomplish this task~\cite{sun-arxiv-2023-chatgpt}. 
In addition to document filtering or reranking, 
information extraction or automatic summarization techniques can be employed to refine the retrieved content by extracting more concise and query-relevant content from the retrieved documents. Furthermore, existing research has proposed token-level compression strategies~\cite{jiang2023llmlingua}, which select important tokens and remove unimportant parts from the candidate documents. 

\textbullet~\emph{Iterative retrieval enhancement.}
In some complex application scenarios,  a single retrieval procedure may not suffice for RAG systems. To address this issue, we can further use iterative retrieval augmentation and adaptive retrieval augmentation. 
Iterative retrieval augmentation aims to iteratively refine the initial query based on the model's generated results to achieve a comprehensive coverage of the required information. As it involves accumulating multiple rounds of retrieval information, the performance of RAG systems may be affected by redundant or conflicting information. To address this issue, stop mechanism has been introduced for retrieval iteration, 
using the LLM to evaluate the confidence of the current generation results to determine whether to continue the iteration process~\cite{Jiang-2023-arxiv-Active}. Additionally, for more complex scenarios, iterative retrieval can be combined with the LLM's own CoT reasoning capability. 
For example, intermediate results from the chain of thought can be used as the query input for the next round of retrieval, and after completing the retrieval process, the returned results can be integrated into the chain of thought. 
Building on the iterative retrieval augmentation method, adaptive retrieval augmentation further enhances the LLM's autonomous use of the retrieval mechanism~\cite{xu2024sayself}, thereby improving the overall framework's efficacy in using the retrieval systems. 
{In practical implementation, for the above two types of augmentation methods,  LLM first need to determine when to use the retriever and then utilize pre-set prompts to initiate query generation and retrieval result processing~\cite{asai2023self}.\ignore{(这句话的目的是什么？和前面的关系是？难道iterative的就不用这个方法了吗？Kun: 都用，已改)}}

\textbullet~\emph{RAG-enhanced training.}
In addition to the improvement strategies mentioned above, specialized training tasks can be designed to further enhance the LLM's ability to utilize the retrieved content, including both instruction tuning and pre-training tasks. 
By constructing instruction data focused on retrieval context utilization~\cite{luo-arxiv-2023-sail},  instruction tuning can improve the LLM's ability to utilize relevant retrieval information. When curating the instruction data, {it is essential to consider two important issues: positional bias and irrelevant information within the input context\ignore{需要重写，这也不是两个issue啊，特别是第二点; Kun: 已改}}. Specifically, relevant documents can be placed at different positions within the prompt, which can enhance the model's attention to relevant content in various positions and prevent the model from neglecting certain positions~\cite{Liu-arxiv-2023-Lost}. Additionally,  irrelevant information can be added to the instructions data, so as to improving the model's ability to resist interference from such information~\cite{lin-arxiv-2023-ra}.
In addition, special training tasks can be introduced during the pre-training stage to further enhance the LLM's retrieval and generation capabilities~\cite{guu-PMLR-2020-retrieval, Lewis-NeurIPS-2020-Retrieval}. Existing work mainly constructs unsupervised pre-training data aimed at retrieval augmentation. A common data construction method uses portions of the original document as queries and then trains the model to reconstruct the remaining content of the original document based on the retrieval results~\cite{lee-ACL-2019-latent}. 


\subsection{Hallucination}
Hallucination, which refers to the phenomenon that LLMs generate content inconsistent with factual information, has become a significant issue that greatly affects the task performance of LLMs~\cite{Li-arxiv-2024-dawn}.
In this section, we focus on discussing the topic of LLM hallucination, first introducing the definition and source of hallucination and then summarizing the detection and mitigation methods.

\subsubsection{Definition of Hallucination}
Early research typically defines hallucinations based on the relationship between a model's output and the given input~\cite{Ji-ACM-2023-survey}. In this manner, hallucinations are categorized into \emph{intrinsic hallucinations} where the model's output does not match the input text and \emph{extrinsic hallucinations} where the model's output cannot be verified against the input. However, in real-world scenarios, user inputs often do not contain reference documents, and thus existing work mainly focuses on open-domain factual hallucinations, where the model-generated content does not align with or cannot be verified by existing world knowledge~\cite{Zhang-CoRR-2023-Siren,Li-arxiv-2024-dawn}. According to a recent study~\cite{Li-arxiv-2024-dawn}, factual hallucinations can be further categorized into the following types:

\textbullet~\emph{Entity-error hallucination.} This type of hallucination refers to LLMs generating text containing incorrect entities, such as names of people, dates, locations, or objects that contradict world knowledge.

\textbullet~\emph{Relation-error hallucination.} This type of hallucination involves LLMs generating incorrect relationships between entities, such as inaccurate quantitative or chronological connections.

\textbullet~\emph{Incompleteness hallucination.} LLMs may produce incomplete outputs, especially when generating lengthy or list-based responses. This hallucination arises when LLMs are asked about aggregated facts and they fail to reserve the factual completeness.

\textbullet~\emph{Outdatedness hallucination.} This type of hallucination occurs when LLMs generate information that was accurate at a past time but is no longer correct at present. This issue typically arises due to that most LLMs were trained on time-limited corpora.

\textbullet~\emph{Overclaim hallucination.} This type of hallucination refers to cases where the statement expressed in the generated text of LLMs is beyond
the scale of factual knowledge.

\textbullet~\emph{Unverifiability hallucination.} This hallucination refers to cases where the information produced by LLMs cannot be verified against existing information sources, making it difficult to assess its accuracy.


\subsubsection{Source of Hallucination}
In this part, we will discuss the potential factors that might lead to hallucination for LLMs.


\paratitle{Training Data.} The quality of training data significantly impacts the model's output and is a primary source of hallucinations. Further, the distribution of training data also plays a key role in shaping the behaviors of LLMs. We next introduce the effect of training data on hallucinations from these two aspects. 

\textbullet~\emph{Data quality.}
In practice, the pre-training dataset is typically constructed by collecting diverse data from various sources. 
While increasing pre-training data can lead to improved model performance, low-quality data can severely damage the generation performance of large models. On the one hand, pre-training data may contain erroneous information, and the goal of training large models is to imitate and memorize the training data as possible. If inaccurate information frequently appears in the training data, the model may memorize and directly copy this content during generation, leading to the phenomenon known as ``\emph{imitative falsehoods}''~\cite{Lin-ACL-2022-TruthfulQA}. On the other hand, pre-training data may contain biased content and the subjective views of its creators. Such biased content can severely affect the model's learning of world knowledge, possibly leading to inappropriate representations.

\textbullet~\emph{Data distribution.} The distribution of pre-training data also significantly affects the model's behavior. Firstly, regarding the recency factor, LLMs are typically trained on data from a limited period. As world knowledge continuously evolves, the model's stored knowledge can become outdated, thereby likely leading to fabrications or outdated information when addressing questions beyond its knowledge scope. In terms of data composition, pre-training data may lack domain-specific knowledge, which would affect the model performance on tasks requiring specialized knowledge, such as medical or legal issues, and it will also result in significant hallucinations. Additionally, recent studies show that when addressing questions involving long-tail knowledge that appears infrequently in the training corpus, models are more likely to generate inaccurate content~\cite{Li-arxiv-2024-dawn}.


\paratitle{Training Methods.}
The training process of LLMs typically includes two major stages: pre-training and post-training. Inappropriate training methods across the two stages are also likely to result in the hallucination behaviors of LLMs. 

\textbullet~\emph{Pre-training.}
Currently, the pre-training stage primarily employs the next token prediction method for model training. {Recent studies~\cite{Liu-arxiv-2023-Lost}} indicate that under the autoregressive training method, the model's attention distribution tends to decay as the sequence length increases. This would prevent LLMs from effectively modeling long-range dependencies, potentially resulting in inference errors or hallucinations. Additionally, the teacher-forcing strategy is commonly used during the training of large models.  In this approach, the correct tokens from the previous steps are used to predict the next token instead of the model output. However, during model inference, the model can only use its own generated content for subsequent predictions. This discrepancy between the training and generation phases leads to ``\emph{exposure bias}''~\cite{Bengio-nips-2015-scheduled}, which may in turn cause hallucination issues.

\textbullet~\emph{Post-training.} 
During the instruction-tuning process, existing works typically employ knowledge distillation to improve the model's instruction-following ability. This involves using high-performance models (such as GPT-4) to generate large-scale instruction data and then fine-tuning weaker models with this data. However, these synthesized data may contain hallucinated content, which might lead to more hallucinations for the trained model. Additionally, during the human alignment process, existing training methods may also cause hallucination issues. Some research work has revealed that LLMs may cater to human responses for earning higher rewards, likely resulting in answers that do not align with factual knowledge~\cite{Sharma-arxiv-2023-towards}.

\paratitle{Response Generation.} Given the input prompt, LLMs employ decoding strategies (\eg top-$k$ sampling in {Section~\ref{sec-decoding}}) for generating the response. 
In this process, the prompt formulation and the decoding strategies potentially affect the generation behaviors of LLMs. 

\textbullet~\emph{Prompt design.}
Prompting has become the primary way for using LLMs to solve downstream tasks. However, inappropriate prompt design can cause the model to overlook or misunderstand important information, leading to incorrect or irrelevant content~\cite{Li-arxiv-2024-dawn}. Recent studies have shown that the readability, format, and concreteness of user instructions would impact the model's output~\cite{Rawte-arxiv-2023-exploring}. For instance, the use of complex words or long phrases in the prompt reduces the readability, which makes LLMs more difficult to understand the real intentions of user instruction, thereby increasing the chance of hallucination. Additionally, non-standard expressions or abstract concepts can also exacerbate hallucinations.

\textbullet~\emph{Decoding strategy.}
To improve the diversity of the generated content, multiple random sampling strategies are introduced (\eg beam search, top-$p$ sampling). However, increasing diversity also brings a higher likelihood of generating hallucinated content. For example, increasing the temperature $t$ (Equation~\ref{eqn:temperature}) will result in a more uniform token probability distribution,  which potentially leads to more hallucinations, since low-frequency yet irrelevant words would be assigned a higher probability for generation in this setting.

\subsubsection{Hallucination Detection} 
To effectively detect the hallucinated content, existing work mainly adopts three approaches, namely model-based, uncertainty-based and tool-based methods.


\paratitle{Model-based Methods.} 
Due to the powerful language capabilities and rich world knowledge, existing work extensively adopts powerful LLMs to detect hallucinations from the model-generated text. In this approach, hallucination detection can be considered as a normal text task that requires prompt formulation. To facilitate the research in  this line, 
HaluEval~\cite{Li-arxiv-2023-HaluEval} introduces a comprehensive dataset of model-generated and human-annotated hallucinated samples to evaluate how well LLMs can identify such instances, and they empirically show specific prompting strategies such as CoT can effectively improve the model's accuracy in detecting hallucinations. Furthermore, research work proposes to decompose the hallucination detection into two subtasks: first, extract factual statements, and then assess whether each statement is hallucinated or not~\cite{Li-arxiv-2024-dawn, Dhuliawala-arxiv-2023-chain}. 

\paratitle{Uncertainty-based  Methods.}
Recent studies suggest that the occurrence of hallucinations in LLMs may be related to the uncertainty of their outputs~\cite{Manakul-emnlp-2023-selfcheckgpt}. Based on such assumptions, a series of works propose detecting hallucinations by assessing the uncertainty of model-generated content. Some research work focuses on the internal features of LLMs, such as token probability and logits. For key concepts in the generated text, a lower token probability indicates a higher uncertainty, which represents an increased likelihood of hallucination~\cite{Varshney-arxiv-2023-stitch}. Other research efforts evaluate the uncertainty by examining the consistency of the models' responses. For instance, SelfCheckGPT~\cite{Manakul-emnlp-2023-selfcheckgpt} lets LLMs answer the same questions multiple times to judge whether the generated answers are consistent or not. Another alternative way requires LLMs to reconstruct the input questions based on the responses and then check the consistency between the generated and original questions~\cite{Yehuda-arxiv-2023-search}.

\paratitle{Tool-based Methods.} LLMs can detect hallucinations by calling external tools to verify the model-generated content. Typically, the model's output contains various segments of factual knowledge, which can be broken down into fine-grained factual statements. FActScore~\cite{Min-arxiv-2023-FActScore} refers to knowledge sources like search engines to verify these statements. FacTool~\cite{Chern-arxiv-2023-FacTool} further proposes to use a series of external verification tools such as calculators and code interpreters to check different types of text. In addition, HaluAgent~\cite{Chen-arxiv-2024-haluagent} proposes an agent framework to employ smaller open-source models for hallucination detection. With the assistance of tools like search engines and calculators, HaluAgent enables 7B-size models to achieve comparable performance as GPT-4 in hallucination detection. 

\subsubsection{{Hallucination Mitigation}} 
In practice, it is essential to effectively mitigate the hallucination behaviors of LLMs, to provide accurate and helpful responses. In this part, we will introduce several widely-used approaches for alleviating the hallucination, including human alignment, retrieval-augmented generation and improved decoding strategy. 


\paratitle{Human Alignment.}
Hallucination mitigation is closely related to the \emph{honest} criterion in  ``3H'' standards for human alignment, and various alignment methods like RLHF can be adopted to mitigate the model hallucination. HaluEval 2.0~\cite{Li-arxiv-2024-dawn} proposes to first collect hallucinated and non-hallucinated responses to train a reward model, and then fine-tune the LLM with the reward model’s feedback using RL algorithms. However, recent research shows that human preference data may lead LLMs to exhibit sycophantic behavior~\cite{Sharma-iclr-2024-Sycophancy}, where models prioritize catering to human demands over maintaining truthfulness. Some work proposes to refine the annotation process of preference data, such as by aggregating multiple human preferences to improve feedback quality~\cite{Sharma-iclr-2024-Sycophancy} or fine-tuning LLMs on prompts where the truthfulness of a claim is independent of the user’s opinion~\cite{Wei-arxiv-2023-sycophancy}.

\paratitle{Retrieval-Augmented Generation.}
Providing LLMs with highly reliable external knowledge as context can help reduce hallucinations. RARR~\cite{Gao-2023-ACL-RARR} first generates multiple questions about the generated text,  then retrieves web pages from Google Search as evidence, and finally, an editing model is employed if any disagreement is detected between the evidence and the generated text. LLM-Augmenter~\cite{Peng-arxiv-2023-Check} further expands the knowledge source to local databases, devising an agent framework to retrieve, consolidate, and generate feedback to the LLM for the final answer. Other research explores placing the retrieval process at different positions relative to the generation process. Verify-and-Edit~\cite{Zhao-acl-2023-VE} proposes to perform the retrieval procedure after the generation process, allowing the original answer to be edited based on the retrieved documents. Furthermore, to help LLMs better handle complex tasks, IRCoT~\cite{Trivedi-arxiv-2022-Interleaving} interleaves the knowledge retrieval process with CoT generation, where the retrieved documents guide the LLM in generating additional reasoning steps and CoT sentences assist in retrieving more relevant and diverse documents.

\paratitle{Improved Decoding Strategy.}
In addition to the above methods, hallucinations can also be mitigated by using improved decoding strategies.  
Typically, the internal states or knowledge of LLMs themselves can be exploited to reduce the hallucinations. DoLa~\cite{Chuang-arxiv-2023-DoLa} proposes that the lower layers of LLMs tend to assign higher probabilities to syntactically plausible words, while higher layers encode more factual knowledge. Therefore, DoLa devises a contrastive decoding strategy by subtracting the lower logits from the last layer's logits and using the results for next-token prediction. ITI~\cite{Li-2023-nips-ITI} finds that specific attention heads show high linear probing accuracy and regards their activation as truth-correlated directions. During inference, certain heads' activations would be shifted along these pivot directions. Some other work introduces external knowledge sources to aid the decoding process. CAD~\cite{Shi-arxiv-2023-Trusting} provides LLMs with extra context about the query, and then contrasts the output probabilities by those without using context, {thereby adjusting the influence of the model's prior knowledge. \ignore{有点夸张了，不可能覆写吧？}} KCTS~\cite{Choi-arxiv-23-KCTs} applies an auxiliary knowledge classifier on top of the LLM to detect hallucinations, and uses its knowledge faithfulness score to reweight the token distribution.

\subsection{Complex Reasoning}\label{sec-long-cot}
In this section, we introduce a new reasoning paradigm for LLMs aimed at solving complex tasks by allocating more time to \emph{thinking} before responding to a problem, \ie conducting complex reasoning.
Specially, we focus on long chain-of-thought~(CoT) reasoning\footnote{The phrase ``\emph{long CoT}'' may not be conceptually precise since the model's thought process could be tree- or graph-structured rather than strictly linear. We use this terminology in line with OpenAI's introduction of the o1 model, which generally refers to extended thought processes for complex reasoning.}, which is the mainstream approach taken by recent large reaonsing models, such as OpenAI's o-series models.
We will begin by providing an overview of long CoT reasoning, then introduce the construction of long CoT data and the corresponding training methods, and finally discuss more general test-time scaling methods.

\subsubsection{Overview and Analysis}\label{subsec-long-cot-intro}
Generally, long CoT reasoning is a method to search for solutions within the natural language space, as reflected in the output responses of LLMs. This approach is akin to the \emph{slow thinking mode} of the human brain~\cite{kahneman-2011-thinking-book}, which takes significantly more time to think through difficult problems compared to the \emph{fast thinking mode} used for simpler ones. This subsection will first qualitatively analyze the  reasoning patterns and then briefly discuss the main advantages of this reasoning mode. 

\begin{figure}[h]
  \centering
  \includegraphics[width=\linewidth]{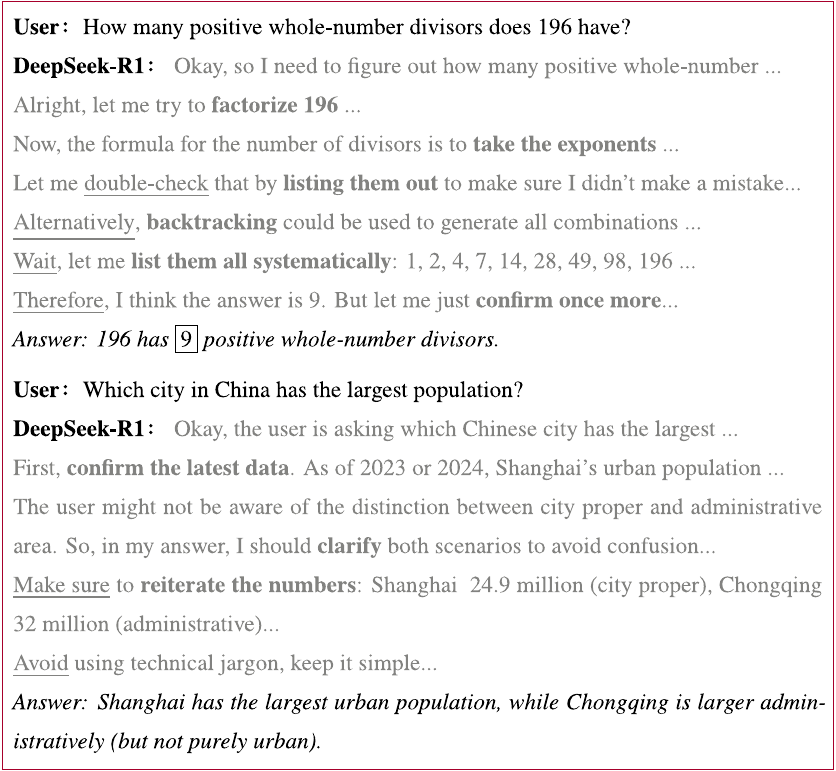}
  \caption{Examples of long CoT reasoning from DeepSeek-R1 (accessed on January 25, 2025). Grey fonts denote the thought part of the model output, and italic fonts denote the final answer.}
  \label{example-long-cot}
\end{figure}

\paratitle{Reasoning Patterns Analysis.} 
As demonstrated in Example~\ref{example-long-cot}, existing long CoT reasoning models typically generate \emph{an extended thought process} (in grey) before arriving at \emph{the final answer} (in italic). It is crucial to understand how this thought process is conducted and the types of reasoning patterns generated by LLMs during problem-solving. To provide an intuitive understanding of this reasoning process, we present two examples from the DeepSeek-R1 model.

In the first example, we present a mathematical problem to the model, and the corresponding long CoT can be observed in the reasoning portion of the response. The thought process here is informal and flexible, while showcasing a systematic exploration of the solution within the natural language space. 
Concretely, the model follows a complete reasoning process consisting of action steps like ``\emph{factorize 196}'' and``\emph{take the exponents}''. 
Notably, the thought process naturally includes trigger keywords like ``\emph{double check}'' and ``\emph{wait}'', which invoke the corresponding verification or reflection actions.
In the second example, we ask the model which Chinese city has the largest population. Interestingly, it exhibits similar thought patterns, even though the question could be addressed in a more compact and straightforward manner. The model generates a comprehensive reasoning process with actions such as ``\emph{confirm the latest data}'' and ``\emph{clarify}'', with trigger keywords like ``\emph{make sure}'' and ``\emph{avoid}''.  

To gain a more comprehensive understanding of this complex reasoning mode, some research has further analyzed the reasoning patterns exhibited in the o1 model~\cite{wu-2024-comparative-arxiv}. These studies, based on empirical investigation, have identified several key reasoning patterns, including systematic analysis, method reuse, divide-and-conquer, self-refinement, context identification, and constraint emphasis. Additionally, the use of these reasoning patterns varies across different tasks, significantly enhancing cognitive processes compared to standard CoT reasoning.

\paratitle{Reasoning Advantages.} 
Unlike standard CoT reasoning, this approach does not enforce a linear reasoning chain. Instead, it integrates various reasoning actions and strategies, such as reflection and backtracking, into a single response. Overall, it has two major advantages compared to the standard CoT method or direct prompting methods.

Firstly, due to the autoregressive nature, the standard generation paradigm of LLMs is a ``\emph{one-time}'' reasoning process. This means that if the generated solution contains obvious mistakes, or even if LLMs are aware of other promising solutions, there are no opportunities for refinement or verification. This issue becomes more pronounced in complex reasoning tasks, where the search space is much larger, preventing LLMs from fully exploring it \cite{zhong-2024-evaluation-arxiv}. In contrast, long CoT reasoning mitigates this problem by allowing the model to autonomously check and revise its attempts, thus enabling more effective reasoning.

Secondly, this text-based reasoning process can, in principle, emulate various search algorithms that rely on  more complex search structures. For example, to represent a tree-structured search space, one might employ a textual process that combines forward exploration with backward revisits, incorporating necessary reflection and verification steps along the way. Consequently, long CoT reasoning can replicate the effects of previously introduced methods like tree-of-thought (ToT) and graph-of-thought (GoT). However, this capability is not inherently present in the LLM; it emerges in a manner similar to the standard CoT ability, developing through appropriate training (see Section~\ref{subsec-long-cot-training}).

Overall, long CoT represents a significant different reasoning mode compared to the standard CoT method, facilitating search algorithms within the natural language space of LLMs.
It emphasizes how to navigate correct paths through a trial-and-error approach, typically incorporating critical reasoning actions such as planning, evaluation, reflection, and exploration. In contrast, short CoT data typically presents a direct solution process in which all reasoning steps are expected to be correct.

\subsubsection{Construction of Long CoT Data}\label{subsec-long-cot-data}

To guide LLMs in producing long-form reasoning followed by solutions, it is crucial to curate high-quality long CoT data for warming up or training the models.
While human annotators can construct extended CoT data, this process is costly and requires professional expertise for challenging problems, making it difficult to scale.
Consequently, existing studies often develop various methods for automatically constructing long CoT data, such as distillation from more advanced models, search based data synthesis, and multi-agent collaboration, which are detailed below.

\paratitle{Long CoT Data Distillation.}
Benefiting from the openness of o1-like LLMs endowed with powerful reasoning capabilities, the leading approach to curating high-quality long CoT data involves using open models or APIs for data synthesis. The basic idea is to first construct a set of prompts (\ie problems) and then feed them into the teacher model to collect the corresponding long CoT response data. 
Specifically, STILL-2~\cite{Min-arxiv-2024-Imitate} utilizes two slow-thinking systems, \ie DeepSeek-R1-Lite-Preview~\cite{DeepSeek-arxiv-2024-Deepseek} and QwQ-32B-preview~\cite{Qwen-arxiv-2024-Qwq} for distillation to construct a dataset of long-form thought processes. A key finding is that length distribution is a critical factor in determining the quality of long CoT data. They suggest that length directly reflects the difficulty of prompt problems, with mathematical problems being particularly important to collect, as they often involve extensive thought processes in their solutions. The research shows that even a small amount of carefully curated long CoT data can effectively activate the slow-thinking mode in LLMs. Furthermore, this effect is corroborated by the work on DeepSeek-R1~\cite{Deepseek-arxiv-2025-DeepSeek}, which demonstrates that training with distilled data from DeepSeek-R1 consistently enhances the performance of multiple Qwen and Llama models.

\paratitle{Search based Data Synthesis.}
Search algorithms like Monte Carlo Tree Search (MCTS) \cite{Silver-nat-2017-Mastering} have been widely applied to synthesizing long-form reasoning  data. As a representative technique, MCTS integrates the principles of tree exploration and random simulation to estimate potential outcomes of actions, making it particularly effective for decision-making tasks with large action spaces. 
In complex problem-solving, MCTS decomposes the process into multi-step generation, with each node at a specific tree layer representing a step in the solution~\cite{jiang-2024-arxiv-still1}. At each step, a LLM, serving as the policy model, samples several candidate nodes, each generating a one-step CoT. MCTS extensively uses rollouts to automatically assign a Q-value to each intermediate step based on its contribution: steps potentially leading to more trajectories that correctly solve the problem receive higher Q-values.
After iterating through multiple steps to successfully address the problem, the complete reasoning trajectories from the root node to the terminal node can be viewed as long-form CoT data, where intermediate nodes represent either correct reasoning steps or trial-and-error attempts.

\paratitle{Multi-Agent Collaboration.}
Beyond relying on a single model, an alternative approach to generating long CoT data is to construct a multi-agent framework~\cite{Liang-arxiv-2023-Encouraging} in which several models collaborate or debate to produce long-form reasoning data. 

The multi-agent framework for synthesizing long-form CoT data typically involves the coordination of multiple autonomous agents, each specializing in distinct roles or functions. These agents work together using iterative reflection and strategic debate to enhance the reasoning process. 
Within this framework, one agent might initiate a chain of thought by presenting an initial hypothesis or argument, while others critique and challenge these ideas through logical examination and counter-arguments. This process encourages deep reflection by prompting agents to reconsider assumptions, address potential biases, and refine conclusions through continuous discourse. 
In this context, reflection involves not only reconsidering past decisions but also assessing whether each agent's contribution is grounded in logical consistency. Additionally, the debate mechanism incorporates alternative perspectives and counterarguments into the reasoning process, resulting in more robust and nuanced outcomes for complex decision-making tasks. By combining these cognitive processes, the framework fosters an environment where complex problems can be tackled collaboratively, with diverse viewpoints contributing to more comprehensive solutions.

\subsubsection{Training Methods}\label{subsec-long-cot-training}
To elicit and enhance long CoT reasoning capabilities, the existing literature extensively explore two methods: long CoT instruction tuning and scaling reinforcement learning (RL) training. We will describe each approach in detail below.

\paratitle{Long CoT Instruction Tuning.}
As discussed in Section~\ref{subsec-long-cot-intro}, long-form thought processes require models to engage in extended reasoning before responding. To develop this reasoning capability, we can instruction-tune LLMs using carefully curated long CoT data. The core concept is to train LLMs to ``\emph{imitate}'' the demonstrated behaviors presented in the long CoT data.

In general, this fine-tuning method aims to achieve two key objectives: \emph{format adherence} (\ie following a long CoT  format) and \emph{ability elicitation} (\ie  activating the complex reasoning mode). Specifically, format adherence requires the model to produce outputs consisting of two sequential parts—thought and solution—while ability elicitation activates the model's inherent capacities for executing appropriate long-form thought processes. It has been shown that both objectives can be effectively achieved through supervised fine-tuning: a small amount of high-quality long CoT data can suffice to elicit the long CoT reasoning capabilities of LLMs. For instance, by fine-tuning Qwen2.5 (32B) on just 3.9K distilled long CoT data, STILL-2 \cite{Min-arxiv-2024-Imitate} achieved performance comparable to industry counterparts such as o1-preview and QwQ in mathematical problem-solving. This effectiveness is largely because strong LLMs inherently possess various specific reasoning abilities (\eg reflection and backtracking). Instruction tuning with long CoT data further enhances these innate abilities, comprehensively integrating and extending their utilization, which enables the model to manage more complex reasoning processes. 

An interesting finding is that this reasoning capability appears to generalize well across different domains. For example, when trained exclusively on mathematical data, it can lead to significant improvements in other disciplines, such as physics and chemistry~\cite{Min-arxiv-2024-Imitate}. 
This is primarily because long CoT reasoning is inherently a reasoning mode rather than a specific ability tied to any particular domain. This can be seen in the example shown in Example~\ref{example-long-cot}, where the query, ``\emph{Which city in China has the largest population?}'', is answered through a  complex thought process, despite being solvable in a more straightforward manner.  
Moreover, this capability can be naturally extended to multimodal LLMs, as these models are typically built on the backbone of language models~\cite{Du-arxiv-2025-Virgo}.

Furthermore, this training approach can be naturally enhanced by other supervised training strategies~\cite{Min-arxiv-2024-Imitate}, such as rejection sampling and directional preference optimization. In general, one can begin by warming up a LLM through instruction tuning with long CoT instruction data and then use the model itself to generate rollout samples as training data.  
These enhancements can have a certain effect, particularly when the amount of warmup instruction data is limited. However, their impact tends to diminish when sufficient long CoT instruction data is available, especially if the quality of self-generated samples is not superior to that of the demonstration data~\cite{Min-arxiv-2024-Imitate}. These findings suggest that this advanced capability of a model may quickly reach a performance ceiling when trained through supervised fine-tuning, due to the inherent limitations typical of imitation learning (for further discussion, see Section~\ref{sec-longcot-discussion}). 

Another downside of this fine-tuning method is its tendency to default to long CoT reasoning mode even for simpler problems (See Example~\ref{example-long-cot}). To better manage reasoning behavior, it is essential to explore systematic approaches that integrate both long CoT reasoning mode and standard response mode.

\paratitle{Scaling RL Training.}
Although OpenAI has not disclosed technical details about the o-series models, training methods have been published through initiatives that implement long chain-of-thought (CoT) reasoning systems, such as DeepSeek-R1~\cite{Deepseek-arxiv-2025-DeepSeek} and Kimi-K1.5~\cite{Kimi-arxiv-2025-Kimi}, which have demonstrated performance comparable to o1. The technical methods employed converge on the approach of scaling RL training to enhance the complex reasoning capabilities of LLMs. In the following part, we introduce the detailed RL method through three components: the policy model, the reward model, and the RL training algorithm. 

$\bullet$ \emph{ Policy model}. The policy model refers to the LLM that needs to be enhanced by the complex reasoning capacities.
Typically, it should be warmed up through supervised fine-tuning with long CoT data, as outlined in the aforementioned method. The main purpose of this warm-up is to activate the long CoT reasoning mode, enabling the policy model to conduct appropriate explorations using a long-form thought process. It is also recommended that the policy model possesses strong foundational capabilities, as this is crucial for eliciting high-reward actions in a more efficient way. An interesting attempt taken by DeepSeek-R1-Zero is to omit the supervised fine-tuning step. Instead, it leverages its strong instruction-following capacity to  adhere to the response format and reasoning mode, guiding the model to generate formatted responses comprising two parts: thought and answer. This method uses a format reward to reinforce the correct reasoning mode.

$\bullet$ \emph{ Reward model}. To effectively guide the policy model, it is necessary to set an appropriate reward model in RL algorithms. As discussed in Section~\ref{sec-alignment}, RLHF employs a specially trained reward model to instruct the learning of the policy model. However, this approach has become less effective for long CoT reasoning, given the difficulty of training reliable reward models to assess the quality of long CoT reasoning processes. Consequently, existing approaches typically employ a verifiable reward model primarily built on reference answers (\eg mathematical problems) or test samples (\eg coding problems). Typically, the mathematical domain serves as the major source of training data, where problems with specific answers are selected. The ground-truth answer is used to derive the reward scores, such as 1 for a correct solution and 0 for an incorrect solution. 
This might seem counterintuitive: how can a complex reasoning system be effectively developed using such a simple reward model? The explanation lies in the essence of RL: unlike supervised fine-tuning, it encourages the autonomous explorations of models through simple yet appropriate incentives. In this way, the complex reasoning capability can be well internalized within the model. In addition to the accuracy reward, other simple rewards can be considered, including completeness, avoidance of excessively long texts, and other formatting issues like repetition. 
OpenAI has proposed the \emph{reinforcement fine-tuning (ReFT)}~\cite{OpenAI-openai-2024-Reinforcement} approach for tuning the o-series models to build domain-specific models, which also uses a simple accuracy reward to guide the training. One limitation of this reward model is that it can only utilize problems with definite and concise answers for training. More general task data, such as summarization, cannot be directly used for training. In such cases, incorporating a trainable reward model becomes necessary. However, as we have discussed, once this reasoning mode is elicited in specific domains, it can naturally generalize well across different domains.

$\bullet$ \emph{ RL Training}. After configuring the policy and reward models, suitable RL algorithms are selected to train the policy model~\cite{zeng2024scaling, chen-2025-still3}. In Section~\ref{sec-alignment}, we provide a detailed implementation of the PPO algorithm, which can be applied directly for training such models. Nonetheless, PPO requires the maintenance and updating of an additional value model, which leads to high training costs, especially when scaling RL training. As a result, existing approaches~\cite{Deepseek-arxiv-2025-DeepSeek} often prefer more simplified RL algorithms, such as GRPO~\cite{Shao-arxiv-2024-Deepseekmath} and RLOO~\cite{Kool-ICLR-2019-Buy}, which use heuristic methods to eliminate the need of a value model.
These algorithms typically exhibit higher efficiency and demonstrate strong training performance, especially in long CoT reasoning. A critical factor to monitor during RL training is the response length of the reasoning models, as a longer average response length often corresponds to enhanced reasoning capabilities. Therefore, it is important to track the trends in average response lengths. With appropriate training, the model should show progressively longer response lengths, accompanied by simultaneous performance improvements.
In fact, response length is directly connected to the \emph{test-time scaling law} demonstrated by OpenAI\footnote{\url{https://openai.com/index/learning-to-reason-with-llms/}}. This law suggests that as more output tokens are generated, a model's reasoning performance can improve substantially. Nonetheless, achieving stable and effective RL training is challenging and necessitates consideration of various factors, such as the selection of query problems (\eg choosing problems that are challenging yet solvable by the model), the updating of the reference model (\eg continually updating it as training progresses), and the enhancement of exploration strategies (\eg sampling more responses with higher temperature settings).

\subsubsection{Extended Discussion}\label{sec-longcot-discussion}

In the preceding discussions, we have introduced the long CoT reasoning in technical detail. Actually, it can be considered a specific approach to achieve \emph{test-time scaling} (\aka inference-time scaling), which is the focus of this subsection.  

From a broader perspective, test-time scaling encompasses various approaches that enhance model performance by increasing the outputs or computations from LLMs. In this way, many methods can be considered test-time scaling techniques.
For example, Self-Consistency~\cite{Wang-arxiv-2022-Self-Consistency} generates multiple responses and then aggregates the solutions using majority vote, resulting in higher inference costs due to the increased number of rollouts. Additionally, planning techniques (Section~\ref{subsec-planning}) and their agentic instantiations (Section~\ref{sec:llm_based_agent}) can also be considered test-time scaling approaches, as they involve prompting LLMs multiple times and utilizing tools or memory components. 
Therefore, the essence of test-time scaling is to trade additional inference costs for performance gains. Unlike previous approaches, long CoT reasoning directly searches for solutions within the natural language space, notably within a single response.

When comparing different test-time scaling methods, two critical factors require careful examination: \emph{token efficiency} (the performance improvement per token cost) and \emph{performance ceiling} (the maximum attainable performance).
Research has shown that scaling test-time computation can effectively enhance model performance~\cite{Charlie-arxiv-2024-Scaling,Deepseek-arxiv-2025-DeepSeek} through the use of simple aggregation methods or specially trained models, though token efficiency may vary. Overall, scaling RL training tends to exhibit higher token efficiency compared to existing test-time scaling methods~\cite{Deepseek-arxiv-2025-DeepSeek}.
Additionally, both heuristic methods and supervised fine-tuning often exhibit a relatively limited performance ceiling that cannot be substantially elevated once scaling reaches a certain level~\cite{Min-arxiv-2024-Imitate,Charlie-arxiv-2024-Scaling}. In contrast, scaling RL training can lead to continuous performance improvements in reasoning models as training time increases. For example, DeepSeek-R1-Zero demonstrates a consistent upward trend in performance even after more than 8,000 training steps~\cite{Deepseek-arxiv-2025-DeepSeek}.

These scaling effects are crucial for solving complex tasks. Notably, a potential advantage of long CoT reasoning models is that they make it feasible to develop expert-level models in specialized domains or for specific tasks, which could significantly impact the advancement of scientific research challenges. 
Moreover, as inference methods and hardware techniques improve, the deployment and use cost of these models will be significantly reduced, enhancing the contribution of these highly intelligent models to real-world applications.
Additionally, addressing security issues in long CoT reasoning models is crucial. Given their unique reasoning mode, specialized alignment strategies should be developed to ensure safer use of these models.

\section{Conclusion and Future Directions}
\label{sec-con}
In this survey, we have reviewed the recent progress of large language models~(LLMs), and introduced the key concepts, findings, and techniques for understanding and utilizing LLMs. We focus on the large-sized models (\ie having a size larger than 10B) while excluding the contents of early pre-trained language models (\eg BERT and GPT-2) that  have been well covered in the existing literature.  
In particular, our survey has discussed four important aspects of LLMs, \ie pre-training, adaptation, utilization, and evaluation. For each aspect, we highlight the techniques or findings that are key to the success of LLMs.
Furthermore, we also summarize the available resources for developing LLMs and discuss important implementation guidelines for reproducing LLMs. 
This survey tries to cover the most recent literature about LLMs and provides a good reference resource on this topic for both researchers and engineers.
      
Next, we summarize the discussions of this survey, and introduce the challenges and future directions for LLMs, in the following aspects. 

\paratitle{Basics and Principles.} 
Instead of training on specific task goals, LLMs learn from unsupervised pre-training on large-scale text data. This is quite different from previous multi-task learning approaches, which aim to extend the training tasks as possible to achieve sufficient generalization.  
Thus, it is essential to reveal the basic principles  or elements that establish the foundation of the abilities of LLMs.  
Although the basic idea of language models is intuitive, it is still challenging to formally explain why LLMs trained by  simple language modeling objectives (\eg next token prediction) can become capable of solving various real-world tasks. 
To investigate this problem, a promising approach is to study the  
capacity learning (or selection) mechanism based on unsupervised pre-training, since 
  the model capacity of LLMs strongly depends on  pre-training data.  
In addition, \emph{scaling}   plays an important role in improving the capacity of LLMs~\cite{Brown-NeurIPS-2020-Language,Wei-arxiv-2022-Emergent,Rae-arxiv-2021-Scaling}, and it is very useful to conduct more theoretical analysis about how the behaviors of large models relate to those of small models, \eg what behaviors of large models can be inferred from small models and what can't be predicted indeed.  
Another research direction is to explore more deep analysis on model generalization for LLMs, since increasing concerns have been raised about whether LLMs can generalize beyond the knowledge encoded by pre-training data. Furthermore,
data contamination has become a severe issue for fairly assessing the performance of LLMs~\cite{zhou-arxiv-2023-dont}, and thus 
setting appropriate evaluation protocol will be the basis to investigate and analyze the model capacity of LLMs. 

\paratitle{Model Architecture.} Due to the scalability and effectiveness, 
Transformer 
has become the de facto architecture for building LLMs. %
Various strategies have been proposed to improve the performance of this architecture, such as  neural network configuration  and scalable parallel  training (see discussions in Section~\ref{sec:configuration}).   
However, Transformer still suffers from  high training  costs and slow inference rates. More efforts~\cite{peng-2023-arxiv-rwkv,sun-2023-arxiv-retnet} are still in need to develop improved model architectures  for large-scale pre-training. 
Specially, system-level or hardware-level optimization (\eg FlashAttention~\cite{Dao-2023-arxiv-flashattention2}) is worth more  exploration to improve the efficiency of Transformer architectures. 
In addition, as an important basic  capacity, existing LLMs typically maintain a long context window. For example, the most recent GPT-4 Turbo enables a long context of  128K tokens, and Claude 2.1 also supports the input up to 200K tokens. Although many efforts have been made to enhance  the long context modeling ability of LLMs~\cite{su-online-2023-Rerope,Press-ICLR-2022-Train}, the resulting models still can't well process the information in the context window~\cite{Liu-arxiv-2023-Lost}. To address this issue, specific architecture adaptations or algorithms might be needed to enhance the modeling and utilization  of long context information.  Another worrying concern is that existing work mostly focuses on training LLMs with decoder-only Transformers. Despite the effectiveness, it severely limits the more wide, diverse explorations on alternative model architectures. 


\paratitle{Model Training.} 
For pre-training, it is essential to establish a data-centric infrastructure and training procedure for LLM optimization, which can effectively support a systematic process of data collection, data cleaning, data mixture, and data  curriculum. Furthermore, it also calls for more flexible mechanisms of hardware support or resource schedule, so as to better organize and utilize the resources in a computing cluster. 
In  practice, it is very  challenging to pre-train capable LLMs, due to the huge compute consumption and the sensitivity to data quality and  training tricks~\cite{Zeng-arxiv-2022-GLM,Scao-arxiv-2022-BLOOM}. 
Thus, it becomes particularly important to develop systemic, economical pre-training approaches for optimizing LLMs, \eg predictable scaling~\cite{OpenAI-OpenAI-2023-GPT-4} and proxy model training~\cite{Xie-arxiv-2023-doremi}.  
More training recipes or principles should be investigated and shared to reduce the potential  risk of degradation or failure in large-scale model optimization.  
Although increasingly more model checkpoints and cleaned datasets have been released, there still lacks  reproducible work on pre-training data preparation (\eg detailed cleaning strategies) and data scheduling (\eg data mixture and curriculum).   
Since it is very costly to pre-train a LLM from scratch, it is  important to design suitable mechanisms for continually pre-training or fine-tuning the LLM based on publicly available model checkpoints (\eg LLaMA~\cite{Touvron-arxiv-2023-LLaMA} and Flan-T5~\cite{Chung-arxiv-2022-Scaling}).    
For this purpose, a number of technical issues have to be resolved, \eg  catastrophic forgetting and task specialization.   
{Furthermore, it is also useful to develop effective tuning strategies that effectively inject or edit specific knowledge~\cite{Yao-arxiv-2023-Editing}, \eg correcting the outdated facts.}


\paratitle{Model Utilization.} 
Based on the natural language interface, \emph{prompting} has become the prominent approach for using LLMs to solving various tasks. 
By combining task descriptions and demonstration examples into prompts, in-context learning~(ICL) endows LLMs with the ability to perform well on new tasks, even outperforming full-data fine-tuned models in some cases. 
To enhance the ability of complex reasoning, advanced prompting techniques have been proposed,  exemplified by the chain-of-thought~(CoT) strategy, which includes the intermediate reasoning steps into prompts.
Furthermore, planning is a promising approach for solving  complex tasks, which iteratively  invokes LLMs by leveraging tool use capacities. Despite these efforts, several basic  problems related to prompting are still  under-explored:   why a good prompt can elicit the correct answer but a bad prompt cannot,  how to reveal the working principles of advanced prompting methods (\eg ICL and  CoT) and  further improve these existing approaches, and how to efficiently find the effective prompts for  LLMs on  specific tasks.     
\ignore{
However,  existing prompting approaches still have several deficiencies described as follows. Firstly,
it involves considerable human efforts in the design of prompts. It would be quite useful to automatically generate effective prompts for solving various tasks.
 Secondly, some complex tasks (\eg formal proof and numerical computation) require specific knowledge or logic rules, which may not be well expressed  in natural language or demonstrated by examples.  
 Thus, it is important to develop more informative, flexible task formatting methods for prompts\footnote{It seems that an alternative approach to this issue is to invoke external tools, \eg the plugins for ChatGPT, when the task is difficult to solve via text generation.}.
Thirdly, existing prompting strategies mainly focus on single-turn performance. 
 It is useful to develop interactive prompting  mechanisms (\eg through natural language conversations) for  solving complex tasks, which have been demonstrated to be very useful by ChatGPT. %
}
Furthermore, from a practical perspective, it has become a fundamental challenge to reduce the inference cost of LLMs, especially in large-scale deployment. 
Another popular research direction is  retrieval-augmented generation, where retrieved contexts from supporting sources are included into prompts for task solving. It has been shown that retrieval augmentation can extend the knowledge boundary and improve the question answering capacity~\cite{Ren-arxiv-2023-Investigating}, but may suffer from the effectiveness of long context utilization by LLMs~\cite{Liu-arxiv-2023-Lost}. 

\paratitle{Safety and Alignment.}
Despite the capacities, LLMs are faced with great safety challenges in practical use.
As a fundamental issue of probabilistic modeling nature, LLMs exhibit a tendency to generate hallucinations~\cite{Bang-arxiv-2023-A}, referring to texts that seem plausible but may be factually incorrect~\cite{OpenAI-OpenAI-2023-GPT-4}.  
%
What is worse, 
LLMs might be elicited by intentional instructions to produce harmful, biased, or toxic texts for malicious systems, leading to the potential risks of misuse~\cite{Brown-NeurIPS-2020-Language,Ouyang-arxiv-2022-Training}. 
To have a detailed  discussion of the safety issues of LLMs (\eg privacy, overreliance, disinformation, and influence operations), the readers can refer to the {GPT-3/4 technical reports~\cite{OpenAI-OpenAI-2023-GPT-4,Brown-NeurIPS-2020-Language}.  %
As the major technical approach to averting these issues, alignment methods (\eg RLHF)~\cite{Ouyang-arxiv-2022-Training,Glaese-arxiv-2022-Improving} have been widely used by leveraging human feedback for developing well-aligned LLMs.
However, RLHF heavily relies on high-quality human feedback data from professional labelers, which is costly and time-consuming to recruit qualified human annotators. 
Therefore, it is necessary to improve the RLHF framework for reducing the efforts of human labelers and seek a more efficient annotation approach with guaranteed data quality,  \eg LLMs can be employed to assist the labeling work. 
Furthermore, it is also suggested to develop simplified optimization algorithms for alignment~\cite{Rafailov-arxiv-2023-Direct,Guo-arxiv-2023-Beyond}, to reduce   the training difficulty and unstability of RLHF. 
As another practical approach, red teaming~\cite{Ganguli-arxiv-2022-Red,Perez-EMNLP-2022-Red} has been adopted for improving the model safety of LLMs, which utilizes the collected adversarial prompts to refine the LLMs (\ie  avoiding the attacks from red teaming). 
In addition, 
privacy  concerns are also important to consider when fine-tuning LLMs with domain-specific data, and thus federated based learning~\cite{Kuang-2023-arxiv-FederatedScope} can be useful in privacy-restricted scenarios.   

\paratitle{Application and Ecosystem.}
As LLMs have shown strong capacities  in solving various tasks, they can be applied in a broad range of real-world applications (\ie following task-specific natural language instructions).   %
As a remarkable progress, ChatGPT has potentially changed the way how humans access information, which has been additionally integrated in  the release of \emph{New Bing}. Generally, in the near future, it can be foreseen that LLMs would have a significant impact on  information-seeking techniques, including both search engines and recommender systems.
Furthermore, LLMs make it possible to develop more intelligent systems (\eg autonomous AI agents) to tackle various complex tasks in real-world scenarios.  
Specially, Assistants API has been launched by OpenAI (featured by  instructions, knowledge and tool use), enabling rapid development of agent-like assistants within the applications.  
This wave of technical innovation would  lead to an  ecosystem of LLM-empowered applications (\eg OpenAI’s GPT Store), which has a close connection  with human life.   
Lastly, the rise of LLMs sheds light on the exploration of artificial general intelligence~(AGI). It is promising to develop more smart AI systems than ever. However, in this development process, AI safety should be one of the primary concerns, \ie making AI lead to good for humanity but not bad~\cite{OpenAI-blog-2023-Planning}.

\section*{\textsc{Coda}}
It is not an easy job to write this long survey and update its content with timely work. First of all, we would like to sincerely thank the support from the readers and our team members. We work very hard on this survey, and hope that it can present a comprehensive, timely reference for LLMs. 

\paratitle{Survey Writing}. This survey was planned during a   discussion meeting held by our research team, and we aimed to summarize the recent advances of large language models as a highly readable report for our team members. The first draft was finished on March 13, 2023, in which our team members tried their best to include the related studies about LLMs in a relatively  objective, comprehensive way. 
Then, we have extensively revised the writing and contents in several passes.  
Due to the space limit, we can only include a fraction of existing  LLMs in Figure~\ref{fig:llms_timeline} and Table~\ref{tab:resource_model}, by setting the selection criterion.
However, we set a more relaxed criterion for model selection on our GitHub page (\url{https://github.com/RUCAIBox/LLMSurvey}), which will be regularly maintained. 
We release the initial version on March 31, 2023,  the major revision on June 29, 2023, and second   version on September 10, 2023, and this latest version (major revision) on November 23,  2023. 

\paratitle{Seeking for Advice}. Despite all our efforts, this survey is still far from perfect: we are likely to miss important references or topics, and might also have non-rigorous expressions or discussions. 
We will continuously update this survey, and improve the quality as much as we can. 
For us, survey writing is also a learning process for LLMs by ourselves. 
For readers with constructive suggestions to improve this survey, you are welcome to leave comments on the GitHub page of our survey or directly email our authors. 
We will make  revisions following the received comments or suggestions in a future version, and acknowledge the readers who have contributed constructive suggestions in our survey.

\paratitle{Update log}. In this part, we regularly  maintain an update log for the submissions of this survey to arXiv: 
\begin{itemize}
\item First release on March 31, 2023: the initial  version. 
\item Update on April 9, 2023: add the affiliation information, revise Figure~\ref{fig:llms_timeline} and Table~\ref{tab:resource_model} and clarify the corresponding selection criterion for LLMs, improve the writing,  and correct  some minor errors. 
\item Update on April 11, 2023: correct  the errors for library resources.  
\item Update on April 12, 2023: revise  Figure~\ref{fig:llms_timeline} and Table~\ref{tab:resource_model}, and clarify the release date of LLMs.
\item Update on April 16, 2023: add a new Section~\ref{sec-GPT-series} about the technical evolution of GPT-series models.
\item Update on April 24, 2023: add the discussion about scaling laws and add some explanations about the model sizes for emergent abilities (Section~\ref{sec-background}); add an illustrative figure for the attention patterns for different architectures in Figure~\ref{fig:architectures}, and add the detailed formulas in Table~\ref{tab:detailed_configuration}.
\item Update on April 25, 2023: revise some copy errors in figures and tables. 
\item Update on April 27, 2023: add efficient tuning in Section~\ref{sec-PEFT}.
\item Update on April 28, 2023: revise  Section~\ref{sec-PEFT}.
\item Update on May 7, 2023: revise Table~\ref{tab:resource_model}, Table~\ref{tab:corpora}, and some minor points.
\item {Update on June 29, 2023 (major revision): }
\begin{itemize}
\item Section~\ref{sec:introduction}: add Figure~\ref{fig:paper_number} for the trends of published LLM papers in arXiv;
\item  Section~\ref{sec-overview}: add  Figure~\ref{fig:openai} for GPT's evolution and the corresponding discussion;
\item  Section~\ref{sec-resource}: add Figure~\ref{fig:llama_family} for LLaMA family and the corresponding discussion;
\item  Section~\ref{sec-adaptation}: add latest discussion about the synthetic data formatting of instruction tuning in Section~\ref{sec-instruction-formatted},  the empirical analysis for instruction tuning in Section~\ref{instruction-results},   parameter-efficient model adaptation in Section~\ref{sec-PEFT} and memory-efficient adaptation in Section~\ref{sec-PEFT};
\item Section~\ref{sec-utilization}: add latest discussion about the underlying mechanism of ICL~\ref{sec-ICL-mechanism},  planning for complex task solving in Section~\ref{subsec-planning};
\item Section~\ref{sec-evaluation}: update Table~\ref{tab:dataset} for representative datasets for evaluating advanced abilities of LLMs, and empirical ability evaluation in Section~\ref{sec-empirical};
\item  Section~\ref{subsec:promptdesign}: add prompt design;
\item  Section~\ref{sec-application}: add the discussions on applications of LLMs in finance and scientific research domains;
\end{itemize}
\item {Update on September 10, 2023 (major revision): }
\begin{itemize}
\item 
{Claim the copyrights of the figures and tables in this paper.}
\item 
{Add latest LLMs, techniques and their descriptions in Section~\ref{sec-resource}, Section~\ref{sec-pretraining}, Section~\ref{sec-adaptation}, Section~\ref{sec-utilization} and Section~\ref{sec-evaluation};}
\item {Section~\ref{sec-pretraining}: add latest discussion about the decoding strategy in Section~\ref{sec-decoding};}
\item {Section~\ref{sec-adaptation}: add latest discussion about the practical tricks for instruction tuning in Section~\ref{sec-ituning-strategy}, the empirical analysis on LLaMA (13B) for instruction tuning in Section~\ref{instruction-results}, practical strategies for RLHF in Section~\ref{sub:RLHF},  alignment without RLHF in Section~\ref{sec-alignment-withoutRL} and remarks on SFT and RLHF in Section~\ref{sec-remarks-SFTRL};}
\item {Section~\ref{sec-utilization}: update the content about the planning for complex task solving in Section~\ref{subsec-planning};}
\item {Section~\ref{sec-evaluation}: add  discussions about evaluation approaches in Section~\ref{subsec-evaapp}, Table~\ref{tab-category-evaluation} for the category of existing evaluation work, and update empirical ability evaluation in Section~\ref{sec-empirical} and the results on Table~\ref{tab-experimental-res};}
\item  {Section~\ref{subsec:promptdesign}: add new prompt examples in Table~\ref{tab-tips};}
\end{itemize}

\item {Update on November 23, 2023 (major revision): }
\begin{itemize}
    \item 
    {
Section~\ref{sec:introduction}: add Figure~\ref{fig:task_solvers} for the evolution process of four generations of language models;}
    \item 
    {Section~\ref{sec-overview}: add more discussion about scaling laws and how emergent abilities relate to scaling laws;}
    \item {Section~\ref{sec-resource}: add latest LLMs in Figure~\ref{fig:llms_timeline} and Table~\ref{tab:resource_model}, latest APIs in Section~\ref{sec:apis_for_llms}, commonly used datasets for instruction tuning and alignment tuning in Section~\ref{sec:commonly_used_fituning}, and several libraries in Section~\ref{sec:library};}
    \item {Section~\ref{sec-pretraining}: add latest discussion about the data scheduling, including data mixtures and data curriculum in Section~\ref{sec:data_scheduling}; add summary of data preparation in Section~\ref{sec:data_prepare_sug}; add discussion about modeling long context in Section~\ref{sec:long_context}; add discussion about decoding efficiency issues and add latest decoding strategies in Section~\ref{sec-decoding};}
    \item {Section~\ref{sec-adaptation}: add latest discussion about instance construction and tuning strategies in Section~\ref{sec-instruction}; add latest discussion about process-supervised RLHF in Section~\ref{sub:RLHF}, and the empirical study on quantized LLaMA models (7B and 13B) in Section~\ref{sec:quantization_empirical};}
    \item {Section~\ref{sec-utilization}: add latest discussion about prompt optimization in Section~\ref{sec:prompt_opt}, and update the content about chain-of-thought prompting in Section~\ref{subsec-cot};}
    \item {Section~\ref{sec-application}: add latest discussion about LLM for research directions in Section~\ref{sec:llm4community};}
    \item {Section~\ref{sec-con}: revise the content in the several aspects.}
\end{itemize}

\item {Update on September 25, 2024: }
\begin{itemize}
    \item {Section~\ref{sec-resource}: reorganize the content of ``public available model checkpoints'' into multiple series; add the latest LLMs in Figure~\ref{fig:llms_timeline}.}
    \item {Section~\ref{sec-pretraining}: add LLM-based data filtering and selection methods in Section~\ref{sec:data_pre_processing}; update Section~\ref{sec:archs}, ``Emergent Architectures'' to include more discussions about SSM-based architectures; add Table~\ref{tab-new-architectures} to compare parallelism and complexity of different architectures.}
    \item  {Section~\ref{sec-adaptation}: add latest discussion about instruction quality improvement and instruction selection in Section~\ref{sec-instruction-formatted}; add latest discussion about practical strategies for RLHF and process-supervised RLHF in Section~\ref{sub:RLHF}; update the content about supervised alignment tuning in Section~\ref{sec-alignment-withoutRL}.}
    \item {Section~\ref{sec-utilization}: add latest papers about discrete prompt optimization in Section~\ref{sec:prompt_opt}.}
    \item {Section~\ref{sec-advanced-topics}: add latest discussion about advanced topics, including long context modeling, LLM-based agent, analysis and optimization for training and inference, model inference, model compression, retrieval-augmented generation, and hallucination.}
\end{itemize}

\item {Update on October 12, 2024: }
\begin{itemize}
    \item Section~\ref{sec-KGLLM}: correct the errors.
\end{itemize}

\item {Update on March 11, 2025: }
\begin{itemize}
    \item Section~\ref{sec-long-cot}: add latest papers about long CoT reasoning, including the analysis of reasoning patterns and advantages, construction of long CoT data (\ie distillation, search-based, and multi-agent collaboration), and training methods (\ie instruction tuning and reinforcement learning).  
\end{itemize}

\item {Update on February 27, 2026: }
\begin{itemize}
    \item Figure~\ref{fig:llms_timeline}: update with new LLMs.
\end{itemize}
\end{itemize}


\paratitle{Clarifications on Experiments}. In this version, we have included a number experiments on instruction-tuning (Table~\ref{tab-instruction-tuning-res}), overall ability evaluation (Table~\ref{tab-experimental-res}), and prompt engineering (Table~\ref{tab-instructions}). Due to the limit of computational resources, our experiments are not complete, limited to small-sized models or a few comparisons. Despite that, we feel that it might be  meaningful to share the partial results to the public. We will try to include the missing results of larger models or more comparisons in the future versions. \textbf{We also call for support of computing power for conducting more comprehensive experiments.}  

\paratitle{Chinese Book}. We also released  a Chinese book based on this survey article, at the link: \url{https://llmbook-zh.github.io}. This book is in the publication process.  

\ifCLASSOPTIONcompsoc
  \section*{Acknowledgments}
\else
  \section*{Acknowledgment}
\fi

The authors would like to thank Yankai Lin and Yutao Zhu for proofreading this paper. 
Since the first release of this paper, we have received a number of valuable comments from the readers. 
We sincerely thank the readers who have written to us with constructive suggestions and comments: Tyler Suard, Damai Dai, Liang Ding,  Stella Biderman,  Kevin Gray,  Jay Alammar, Yubo Feng,  Mark Holmstrom, Xingdong Liu, Il-Seok Oh, Yiting Liu,  Shaojun Wang,  Gaoyan Ou,  Todd Morrill, Hao Liu,  Zhenyu Zhang, and Xinlin Zhuang.
\\

Since the v11 version (June 29, 2023),  we have been adding a large number of experiments and prompt practices. These new contents are completed by a number of volunteers in our team. Here, we add a special part to thank all the students who have worked very hard on this part (also including the ones on our author list). 

\paratitle{Contribution on Experiments.} 
We would like to sincerely thank the following people for their hard work involved in experiments shown in Table~\ref{tab-experimental-res}.

$\bullet$ Xiaoxue Cheng: implement the experiments for evaluation on Language Generation and HaluEval tasks.

$\bullet$ Yuhao Wang: implement the experiments for evaluation on interaction with environment tasks.

$\bullet$ Bowen Zheng: implement the experiments for evaluation on tool manipulation tasks.

\paratitle{Contribution on Tips.}
We list the following guys for their contributions on the corresponding numbers of provided tips for designing prompts in Table~\ref{tab-tips}.

$\bullet$ Xiaolei Wang: T3, O3

$\bullet$ Beichen Zhang: D2, D5

$\bullet$ Zhipeng Chen: D3, D4

$\bullet$ Junjie Zhang: D6

$\bullet$ Bowen Zheng: D7

$\bullet$ Zican Dong: D8

$\bullet$ Xinyu Tang: C2

$\bullet$ Yifan Du: T4

$\bullet$ Tianyi Tang: O6, O7, D9

$\bullet$ Yupeng Hou: O8, C3

$\bullet$ Salvatore Raieli: C4

\ifCLASSOPTIONcaptionsoff
  \newpage
\fi

\bibliographystyle{IEEEtran}
\bibliography{newbib}

\end{document}